\documentclass[acmtog, authorversion, nonacm]{acmart}

\usepackage{booktabs} %

\usepackage{hhline}
\usepackage[ruled]{algorithm2e} %
\usepackage{subcaption}
\usepackage{pifont}
\usepackage{arydshln}
\usepackage{multirow}
\usepackage{multicol}
\usepackage{xcolor}
\usepackage{cleveref}
\usepackage{xspace}
\usepackage{rotating}
\usepackage{xhfill}

\usepackage[toc,page,header]{appendix}

\makeatletter

\crefname{section}{Sec.}{Secs.}
\Crefname{section}{Section}{Sections}
\Crefname{table}{Table}{Tables}
\crefname{table}{Tab.}{Tabs.}

\definecolor{mydarkblue}{rgb}{0,0.08,1}
\definecolor{mydarkgreen}{rgb}{0.02,0.6,0.02}
\definecolor{mydarkorange}{rgb}{0.40,0.2,0.02}

\makeatletter
\DeclareRobustCommand\onedot{\futurelet\@let@token\@onedot}
\def\@onedot{\ifx\@let@token.\else.\null\fi\xspace}

\def\etal{\emph{et al}\onedot}

\def\etal{\emph{et al}\onedot}

\newif\ifdraft
\draftfalse

\def\linenull#1{%
\begin{tikzpicture}
\draw (-5.5,0) -- (-1,0);
\end{tikzpicture}}

\definecolor{mygreen}{RGB}{112,173,71}
\definecolor{myblue}{RGB}{91,155,213}
\definecolor{myorange}{RGB}{237,125,49}
\definecolor{myred}{RGB}{255,22,67}
\definecolor{honey}{RGB}{201,89,95}
\definecolor{amber}{RGB}{180,148,41}
\definecolor{darkblue}{RGB}{84,96,126}

\definecolor{darkgreen}{HTML}{008000}
\definecolor{jazzberryjam}{rgb}{0.65, 0.04, 0.37}
\definecolor{lightgreen}{HTML}{82cf8e}
\definecolor{lightblue}{HTML}{618fab}
\definecolor{lightpurple}{HTML}{ac85cc}

\title{p\textcolor{lightgreen}{\hspace{0.2mm}\raisebox{0.2mm}{\scalebox{0.75}{$\pmb{\bigcirc}$}}}ps: Photo-Inspired Diffusion \textcolor{lightpurple}{\raisebox{0.2mm}{\scalebox{1.2}{$\pmb{\bigcirc}$}}}perators}

\author{Elad Richardson}
\affiliation{%
 \institution{Tel Aviv University}
 \country{Israel}
}
\author{Yuval Alaluf}
\affiliation{%
 \institution{Tel Aviv University}
 \country{Israel}
}
\author{Ali Mahdavi-Amiri}
\affiliation{%
 \institution{Simon Fraser University}
 \country{Canada}
}
\author{Daniel Cohen-Or}
\affiliation{%
 \institution{Tel Aviv University}
 \country{Israel}
}

\setcopyright{none}

\makeatletter
\let\@authorsaddresses\@empty
\makeatother

\begin{document}

\begin{abstract}
Text-guided image generation enables the creation of visual content from textual descriptions. 
However, certain visual concepts cannot be effectively conveyed through language alone.  This has sparked a renewed interest in utilizing the CLIP image embedding space for more visually-oriented tasks through methods such as IP-Adapter. Interestingly, the CLIP image embedding space has been shown to be semantically meaningful, where linear operations within this space yield semantically meaningful results. Yet, the specific meaning of these operations can vary unpredictably across different images.
To harness this potential, we introduce \textit{pOps}, a framework that trains specific semantic operators directly on CLIP image embeddings. 
Each \textit{pOps} operator is built upon a pretrained Diffusion Prior model. 
While the Diffusion Prior model was originally trained to map between text embeddings and image embeddings, we demonstrate that it can be tuned to accommodate new input conditions, resulting in a diffusion operator.
Working directly over image embeddings not only improves our ability to learn semantic operations but also allows us to directly use a textual CLIP loss as an additional supervision when needed.
We show that \textit{pOps} can be used to learn a variety of photo-inspired operators with distinct semantic meanings, highlighting the semantic diversity and potential of our proposed approach.
Code and models are available via our project page: \url{https://popspaper.github.io/pOps/}.

\end{abstract}

\begin{teaserfigure}
\centering
\includegraphics[width=\textwidth]{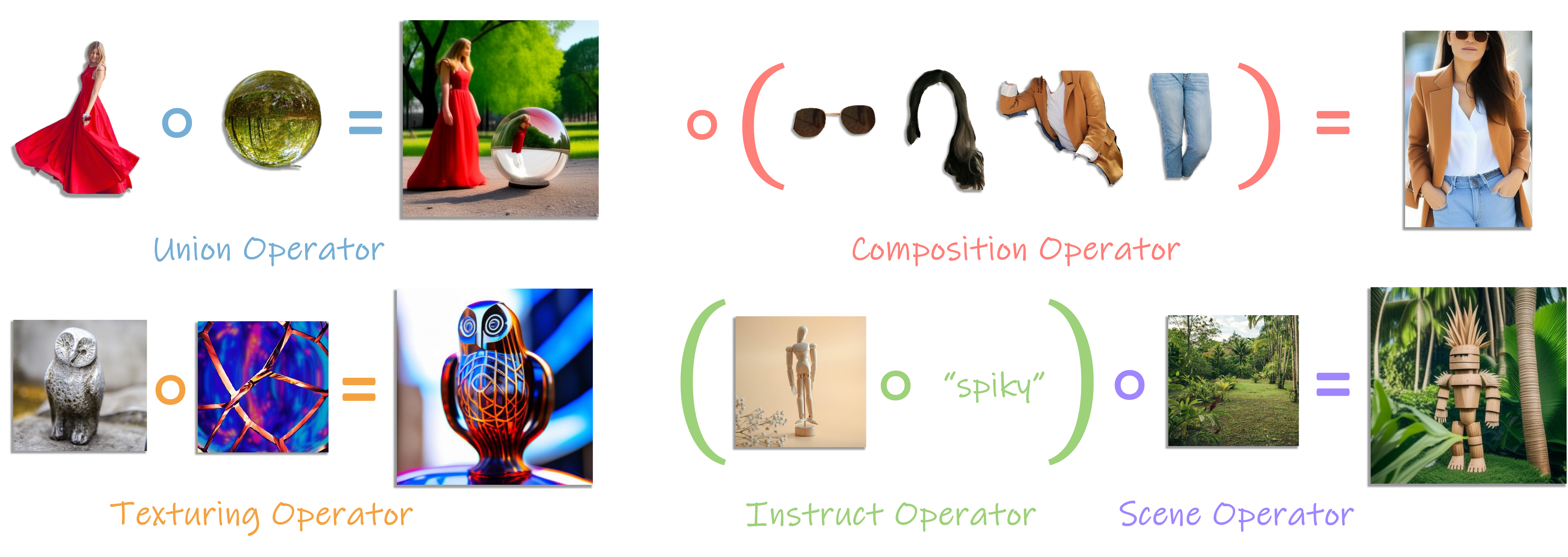}  \\[-0.45cm]
\caption{Different operators trained using \textit{pOps}. Our method learns operators that are applied directly in the image embedding space, resulting in a variety of semantic operations that can then be realized as images using an image diffusion model.}
\label{fig:teaser}
\end{teaserfigure}

\maketitle

\section{Introduction}
Operators are often among the first concepts we learn in mathematics.
They offer an intuitive means to describe complex concepts and equations, accompanying us from basic arithmetic operations to advanced mathematics.
In the field of visual content generation, text has emerged as the de facto interface for describing and generating complex concepts. However, attaining precise control over the generated content through language is challenging, often requiring extensive prompt engineering. Drawing inspiration from the intuitiveness of operators and classical generation approaches such as Constructive Solid Geometry~\cite{foley1996constructive}, we propose an operator-based generation mechanism built on top of the CLIP~\cite{radford2021learning} image embedding space. 

\begin{figure}[b]
    \centering
    \includegraphics[width=0.45\textwidth]{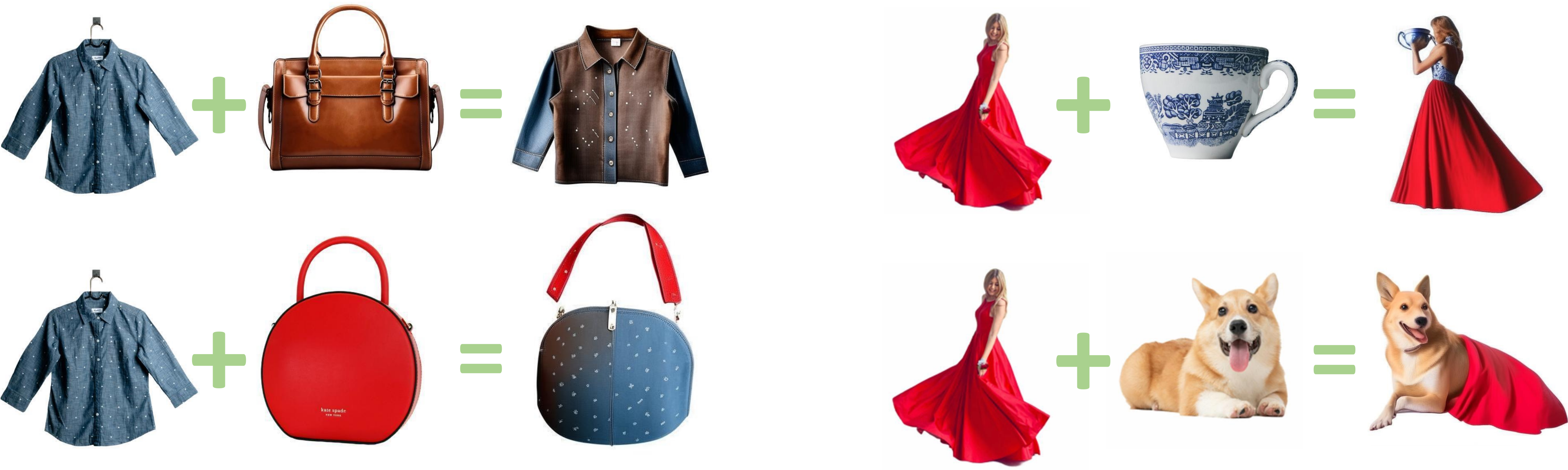}  \\[-0.25cm] 
    \caption{
    \textbf{Averaging in latent space.} Given two images we encode them to the CLIP embedding space, average their representations, and pass the result as a condition to an image diffusion model to generate an image.
    As shown, averaging in latent space has semantic meaning even with no training but the meaning can change unexpectedly and is not controllable. 
    }
    \label{fig:avg_vis}
\end{figure}

Interestingly, as observed by Ramesh~\etal~\shortcite{ramesh2022hierarchical}, the CLIP image embedding space is already semantically meaningful, where linear operations within this subspace yield semantically meaningful embedding representations. As illustrated in~\Cref{fig:avg_vis}, these operations correspond to manipulations of generated images, such as compositions or the merging of concepts. However, being a vector space, users lack direct control over the exact operations performed over embeddings residing within this space. 
Motivated by this observation, we propose \textit{pOps}, a general framework for training specific operators within the CLIP~\cite{radford2021learning} image embedding space, with each operator reflecting a unique semantic operation. 
Importantly, all \textit{pOps} operators share the same architecture, differing only in the training data and objective. As shall be demonstrated, this unified framework allows one to compose different semantic manipulations, providing much-needed control and flexibility over the image embeddings used to guide the generation process.

We represent these manipulations using the Diffusion Prior model, introduced in DALL-E 2~\cite{ramesh2022hierarchical}.
We show that the Diffusion Prior, originally trained to map text embeddings into image embeddings, can be naturally extended and fine-tuned to accommodate other conditions.
In its original training scheme, the Diffusion Prior was trained to denoise image embeddings based on either text conditions or null inputs. Intuitively, the prior needed to learn not only the properties of its input conditions but also the characteristics of a broad target domain and the relation between the two.
Subsequently, when fine-tuning the model over a new condition, the model can now leverage its prior understanding of the image domain, thereby focusing on relearning the condition-specific aspect of the mapping. 
In fact, we show that even when fine-tuning a subset of the prior model layers, the model can still operate over new input conditions. This observation also aligns with existing literature on text-to-image diffusion models, where introducing new controls such as image embeddings (IP-Adapter~\cite{ye2023ip-adapter}) or spatial controls (ControlNet~\cite{zhang2023adding}) can be achieved with a relatively short fine-tuning performed over a pretrained model.

To illustrate the flexibility of \textit{pOps}, we design several operators, highlighting different potential semantic applications, including:
\begin{enumerate}
    \item \textit{The Union Operator}. Given two image embeddings representing scenes with one or multiple objects, combine the objects appearing in the scenes.
    \item \textit{The Texturing Operator}. Given an image embedding of an object and an image embedding of a texture exemplar, paint the object with the provided texture.
    \item \textit{The Scene Operator}. Given an image embedding of an object and an image embedding representing a scene layout, generate an image placing the object within a semantically similar scene.
    \item \textit{The Instruct Operator}. Given an image embedding of an object and a single-word adjective, apply the adjective to the image embedding, altering its characteristics accordingly.
    \item \textit{The Composition Operator}. Given a set of object parts (e.g., articles of clothing), create a scene composing the objects together (e.g., a complete outfit). 
\end{enumerate}

For each operator, we independently fine-tune the Diffusion Prior model on the corresponding task to generate the desired image embedding representation. Observe that some operators (e.g., texturing and union) can be trained by defining a paired dataset of image embeddings. However, in some instances, defining a paired dataset is impractical. As such, we show how one can train operators using supervision realized by a textual CLIP loss, eliminating the need for direct image supervision. 

\begin{figure}
    \centering
    \includegraphics[width=0.46\textwidth]{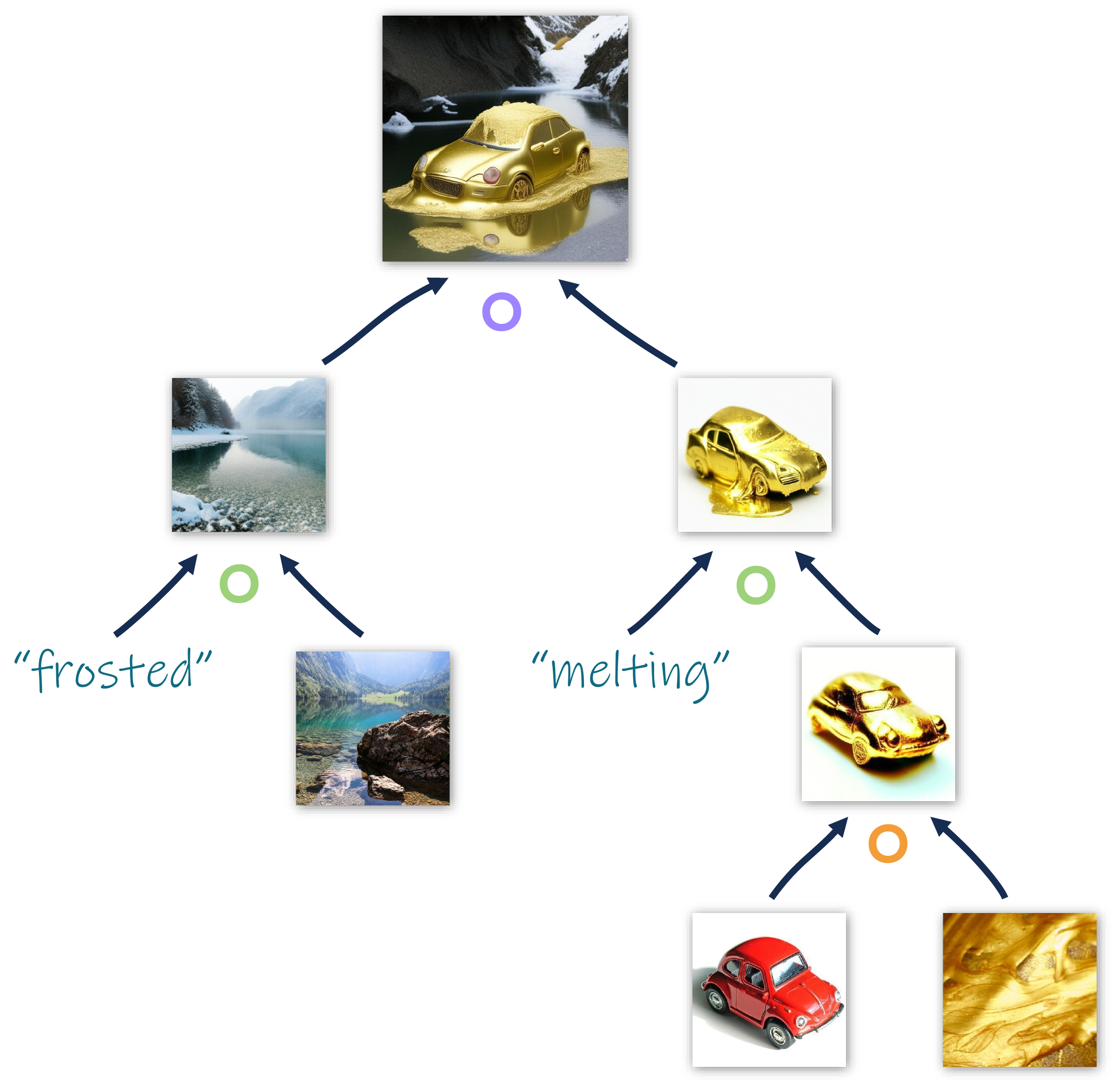} \\[-0.2cm]
    \caption{
    \textit{pOps} operators can be composed into \textit{generative trees}, each node specifying a different operator applied in the CLIP image embedding space. \\[-1cm]
    }
    \label{fig:composition_tree}
\end{figure}

Finally, given a set of trained \textit{pOps} operators, we can also compose them together to form more complex semantic operations, creating a new generation paradigm. Rather than providing all conditions simultaneously and generating the output in a single shot, we can carefully design each element in the CLIP embedding space and compose them together into a \textit{generative tree}. 
This allows users to design a more granular generation process wherein objects are first generated independently, manipulated individually, and finally merged together into a single embedding. This final embedding can then be \textit{``rendered''} into a corresponding image using a pretrained image denoising network. This methodology aligns well with traditional generation processes in computer graphics, such as Constructive Solid Geometry~\cite{foley1996constructive}, which builds upon an iterative, tree-like modeling approach, as illustrated in~\Cref{fig:composition_tree}. 

\section{Related Work}

\paragraph{\textbf{Text-to-Image Generation}}
Recent advancements in large-scale generative models~\cite{po2023state,yin2024survey} have quickly revolutionized content creation, particularly in the domain of visual content generation. Notably, the progress in large-scale diffusion models~\cite{ramesh2022hierarchical,nichol2021glide,balaji2023ediffi,kandinsky2,ding2022cogview2,saharia2022photorealistic,rombach2021highresolution} has resulted in unprecedented quality, diversity, fidelity. However, these models primarily rely on a free-form text prompt as guidance, often requiring extensive prompt engineering to reach the desired result~\cite{witteveen2022investigating,wang2022diffusiondb,liu2022design,marcus2022very}. As a result, many have explored new avenues for providing users with more precise control over the generative process. This control is often realized through spatial conditions~\cite{avrahami2023spatext,bar2023multidiffusion,li2023gligen,lhhuang2023composer,zhang2023adding,voynov2023sketch,dahary2024yourself}, including but not limited to segmentation masks, bounding boxes, and depth maps. 
While effective for defining structure, these methods still lack the ability to control the style and appearance of the generated image.

\vspace{-0.2cm}
\paragraph{\textbf{Image-Conditioned Generation}}
To address the limitations of text representations, some approaches aim to integrate image embeddings directly into pretrained denoising networks, most commonly through cross-attention layers. For instance, T2I-Adapter~\cite{mou2023t2iadapter} controls the global style of generated images by appending image features extracted from a CLIP image encoder to the text embeddings. 
Similarly, Uni-ControlNet~\cite{zhao2024uni} introduces an adapter tasked with projecting CLIP image embeddings to the text embedding space to achieve global control over the generated image.
Most relevant to our work, IP-Adapter~\cite{ye2023ip-adapter} employs a decoupled cross-attention mechanism and an Image Prompt Adapter to project image features into a pretrained text-to-image diffusion model. 
While all of these methods allow conditioning on image embeddings, manipulating the embeddings themselves is challenging, as they are fed into the network as-is. As a result, it remains difficult to precisely control the actual effect of this condition.

\vspace{-0.2cm}
\paragraph{\textbf{Diffusion Prior Model}}
In Ramesh~\etal~\shortcite{ramesh2022hierarchical}, the authors introduce the \textit{Diffusion Prior} model, tasked with mapping an input text embedding to a corresponding image embedding in the CLIP~\cite{radford2021learning} embedding space. This image embedding is then used to condition the generative model to generate the corresponding image. 
This mechanism allows them to not only use existing image embeddings as a condition but also generate such inputs using a separate generative process. Originally the authors demonstrated that leveraging the Diffusion Prior leads to improved image diversity while supporting image variations, interpolation, and editing.  
Since then it has been shown that the prior mechanism can also be adopted for a wide range of generative tasks, including creative image generation~\cite{richardson2023conceptlab}, text-to-video generation~\cite{singer2023makeavideo,esser2023structure}, and 3D generation~\cite{xu2023dream3d,mohammad2022clip}.

\vspace{-0.2cm}
\paragraph{\textbf{Operators and Composable Generation}}
In the context of few-shot learning, Alfassy~\etal~\shortcite{alfassy2019laso} demonstrate how to construct a new feature vector such that its semantic content aligns with the output of a set operation applied over a set of input vectors (e.g., intersection and union). 
This technique was shown to assist in few-shot discriminative settings as a form of augmentation in the feature space. In the generative domain, Composable-Diffusion~\cite{liu2022compositional} proposed using conjunction and negation operators to compose text prompts and better control the generation process. Concept Algebra has also been shown to be feasible in existing text-to-image models by leveraging their learned representations~\cite{gandikota2023concept,brack2024sega} or using a small exemplar dataset~\cite{wang2024concept,motamed2023lego}.

While composite generation remains an under-researched task, it has become common in the generative community to use tools such as ComfyUI and WebUI to compose different methods into a single generative scheme. 
In a sense, this can be viewed as a hierarchical generative process where each model serves as an operator with a dedicated task (e.g. a try-on operator (\cite{choi2024improving,xu2024ootdiffusion}), a texturing operator (\cite{cheng2024zest}), a stylization operator~\cite{wang2024instantstyle}). 
While this aligns with the inspiration behind our work, these operators are typically applied as an afterthought in the image domain, whereas we focus on manipulations in the semantic image embedding domain.

\vspace{-0.1cm}
\paragraph{\textbf{Inspired Generation}}
Human creativity has been heavily studied in the context of computer graphics, with many exploring whether computers can be used to aid the creative design process~\cite{hertzmann2018can,elhoseiny2019creativity,kantosalo2014isolation,wang2024can,Oppenlaender_2022,esling2020creativity}.
At the core of the creative design process lies the ability to draw upon past knowledge to inspire the creation of novel ideas~\cite{bonnardel2005towards,wilkenfeld2001similarity}. Crucially, this process involves associating past ideas to produce original concepts rather than simply mimicking prior work~\cite{brown2008guiding,rook2011creativity}. This is often achieved through the use of exemplars, drawing inspiration from their shape, color, or function. 

Recently, Vinker~\etal~\shortcite{vinker2023concept} utilized a VLM to decompose a visual concept into different visual aspects, organized in a hierarchical tree structure. In doing so, they demonstrate how novel concepts and creative ideas can be discovered from a single original concept. Building on this, Lee~\etal~\shortcite{lee2024languageinformed} learn concept representation into disentangled language-informed axes such as category, color, and material, enabling novel concept compositions using the disentangled sub-concepts. Finally, Ng~\etal~\shortcite{ng2023dreamcreature} extract localized sub-concepts (e.g., body parts) in an unsupervised manner that can be used to create hybrid concepts by merging the learned sub-concepts. 

In this work, we focus on composing different aspects of visual concepts to inspire the generation of new visual content. This idea also draws inspiration from Constructive Solid Geometry (CSG)~\cite{foley1996constructive}, which combines geometric primitives via a set of boolean operators to form complex objects.

\section{Preliminaries}

\paragraph{\textbf{Diffusion Prior.}}
Text-to-image diffusion models are typically trained using a conditioning vector $c$, which is derived from a pretrained CLIP~\cite{radford2021learning} text encoder based on a user-provided text prompt $p$.
Ramesh~\etal~\shortcite{ramesh2022hierarchical} propose a two-stage approach to the text-to-image generative process. Firstly, they train a Diffusion Prior model to map a given text embedding to a corresponding image embedding. Subsequently, the predicted image embedding is fed into a denoising diffusion probabilistic model (DDPM)~\cite{ho2020denoising} to generate an image.

The training process of this two-step framework resembles that of standard text-conditioned diffusion models. First, a DDPM is trained following the standard diffusion objective and aims to minimize:
\begin{equation}~\label{eq:ldm}
    \mathcal{L} = \mathbb{E}_{z,y,\varepsilon,t} \left [ || \varepsilon - \varepsilon_\theta(z_t, t, c) ||_2^2 \right ].
\end{equation}
Here, the denoising network $\varepsilon_\theta$ is tasked with removing the noise $\varepsilon$ added to the latent code $z_t$ at timestep $t$, given the conditioning vector $c$, where $c$ is now an image embedding.

Next, the Diffusion Prior model, $P_\theta$, is trained to predict a denoised image embedding $e$ from a noised image embedding $e_t$ at timestep $t$, given a text prompt $y$, by minimizing the objective given by: 
\begin{equation}\label{eq:prior}
    \mathcal{L}_{prior} = \mathbb{E}_{e,y,t} \left [ || e - P_\theta (e_t, t, y) ||_2^2 \right ].
\end{equation}

\begin{figure*}
    \centering
    {
    \includegraphics[width=0.925\textwidth]{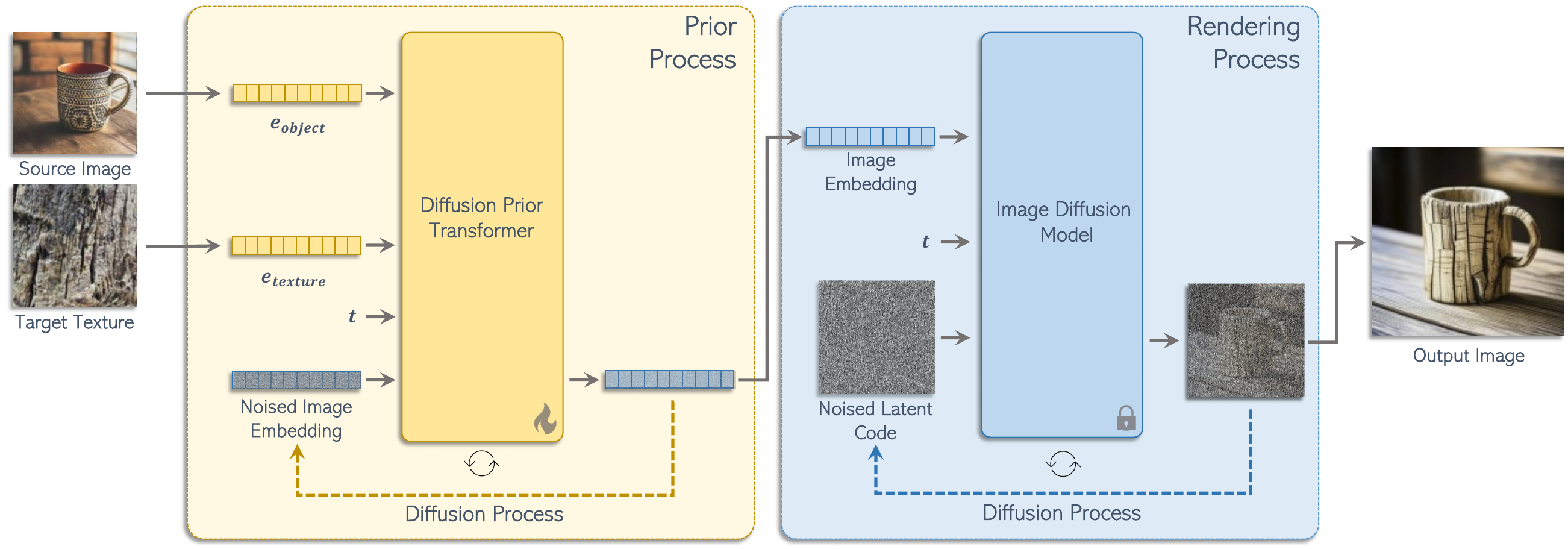}
    } \\[-0.25cm]
    \caption{
    \textbf{pOps Overview for the Texturing Operator}.
    Given an image representing our source object and an image representing our target texture, we first encode both images into the CLIP embedding space, resulting in embeddings $e_{object}$ and $e_{texture}$, respectively. 
    To train our Diffusion Prior model on the specific semantic task (shown in \textcolor{amber}{yellow}), we perform optimization as follows.
    At each timestep $t$, we pass the two image embeddings, an encoding of $t$, and a noised image embedding to our Diffusion Prior model. The model is tasked with outputting a denoised image embedding that matches the target embedding $e_{target}$.
    Following training, we can pair our trained Diffusion Prior model with a pretrained, fixed image diffusion model. The learned image embedding serves as a conditioning to the diffusion model to effectively ``render'' the corresponding image (illustrated in \textcolor{blue}{blue}). \\[-0.5cm]
    }
    \label{fig:method}
\end{figure*}

In this work, we explore how the Diffusion Prior can be adapted to operate over image embeddings rather than the standard text embeddings. In doing so, we present a versatile framework capable of mapping various user inputs to their corresponding image embeddings. 

\section{The pOps Framework}
Here, we demonstrate how \textit{pOps} can be utilized to realize a variety of semantic operators.
While all the \textit{pOps} operators share the same architecture, they differ in terms of input conditions and corresponding training objectives.

\subsection{Binary Image Operators}
We begin with binary operators that are conditioned on two provided image embeddings and produce a single image embedding that aligns with the desired task. An overview is provided in~\Cref{fig:method}.

\subsubsection{Architecture and Training}
Following Ramesh~\etal~\shortcite{ramesh2022hierarchical}, we divide the generation process into two stages. First, an image embedding is generated utilizing a dedicated transformer model. This image embedding then serves as a condition for the image diffusion model to generate the desired image.
Since we work directly over image embeddings, training is required only for the prior, while the diffusion image model, acting as a ``renderer'', remains fixed.

\begin{figure}
    \centering
    \setlength{\belowcaptionskip}{-4pt}
    \setlength{\tabcolsep}{0pt}
    \renewcommand{\arraystretch}{1}
    \newcommand{\pl}{0.2}
    {\small
    \begin{tabular}{ccc@{\hskip 2pt}ccc}

    \includegraphics[width=0.075\textwidth]{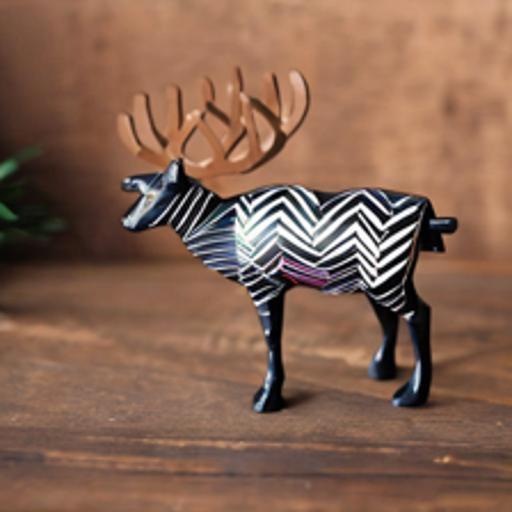} & \includegraphics[width=0.075\textwidth]{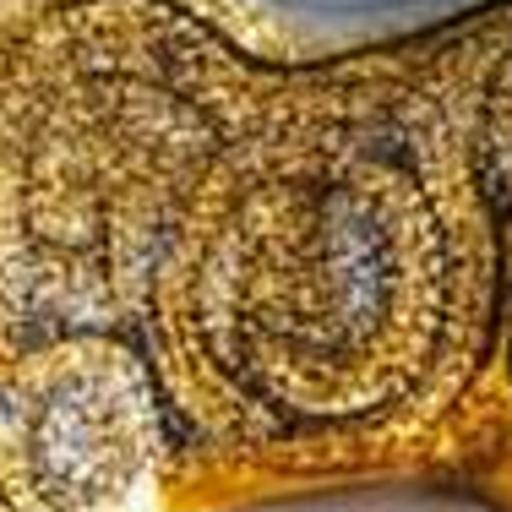} & 
    \includegraphics[width=0.075\textwidth]{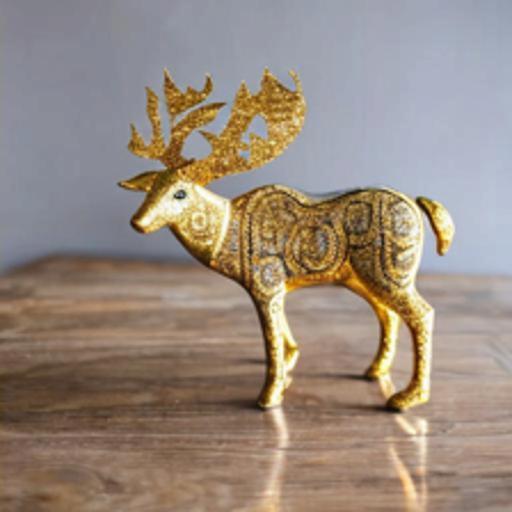}  &
    \includegraphics[width=0.075\textwidth]{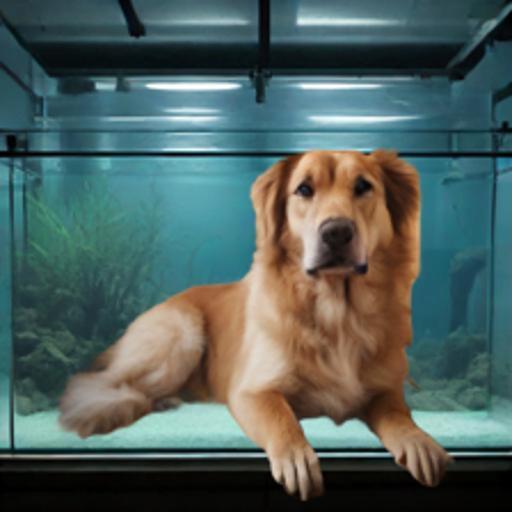} & \includegraphics[width=0.075\textwidth]{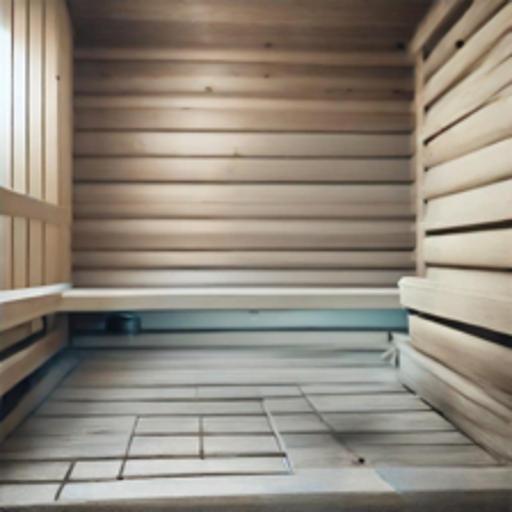} & 
    \includegraphics[width=0.075\textwidth]{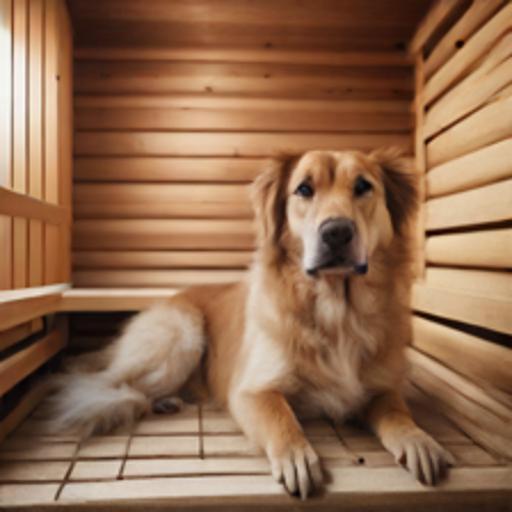} \\
        
    $I_a$ & $I_b$ & $I_{target}$ & $I_a$ & $I_b$ & $I_{target}$ \\[-0.3cm]
        
    \end{tabular}
    }
    \caption{Generated paired data for various \textit{pOps} operators. During training, the images are encoded to embeddings $e_a$, $e_b$. and $e_{target}$, respectively. \\[-0.4cm]}
    \label{fig:data_small}
\end{figure}

For our binary operators, the learnable task is defined using a paired dataset of input conditions, ($I_a$, $I_b$), and a corresponding target image $I_{target}$, see~\Cref{fig:data_small}. These pairs represent the semantic mapping we aim to learn. As we operate in the image embedding space, we first encode all images using a pretrained CLIP image encoder~\cite{radford2021learning}, $E_{im}(\cdot)$, resulting in corresponding embeddings $e_a$, $e_b$, and $e_{target}$.
We note that the original prior model received $77$ input tokens, representing the $77$ text tokens extracted from the pretrained CLIP text encoder. Here, we repurpose these inputs, placing our two embeddings $e_a$ and $e_b$ at the start and filling the remaining entries with zero embeddings. 
As shall be demonstrated, reusing the original entries of the prior model allows us to adapt the number of image embeddings that we pass to the diffusion prior model to match each operator.
These embeddings are followed by an encoding of the timestep $t$ and the noised image embedding we aim to denoise. 
The predicted output of the prior model is taken from the token output associated with the input noised image embedding, \textcolor{amber}{yellow} highlighted section of~\Cref{fig:method}.

During training, at each optimization step, we randomly sample a timestep $t$ and add a corresponding noise to $e_{target}$, resulting in the noisy image embedding $e_{target}^t$. We then train our prior model using the standard denoising objective:
\begin{equation}\label{eq:prior}
    \mathcal{L}_{prior} = \mathbb{E}_{e_{target},y,t} \left [ || e_{target} - P_\theta (e_{target}^t, t, e_a, e_b) ||_2^2 \right ].
\end{equation}
Thus, our model learns to denoise $e_{target}^t$ while taking into account the conditional embeddings $e_a$ and $e_b$. 
During inference, we perform $25$ denoising steps, starting from random noise, with an additional classifier-free guidance term where we drop the $e_a$ and $e_b$ inputs.

\subsubsection{Data Generation}
When trying to solve a specific image-to-image task, it is common to incorporate task-specific modules into the architecture, such as a dedicated depth estimation model applied to the input image or a background extraction model to isolate the object of interest.
Instead, in \textit{pOps}, we adopt a unified architecture for all our binary operators. Our model implicitly learns to manipulate the image embeddings based on the desired task. This is achieved by generating data that simulates our target task, leveraging the powerful vision and vision-language models released in recent years. 
Below, we outline the data generation process for the various binary operators considered in this work, with additional details and generated samples provided in~\Cref{sec:additional_details}.

\begin{figure}
    \centering
    {
    \includegraphics[width=0.465\textwidth]{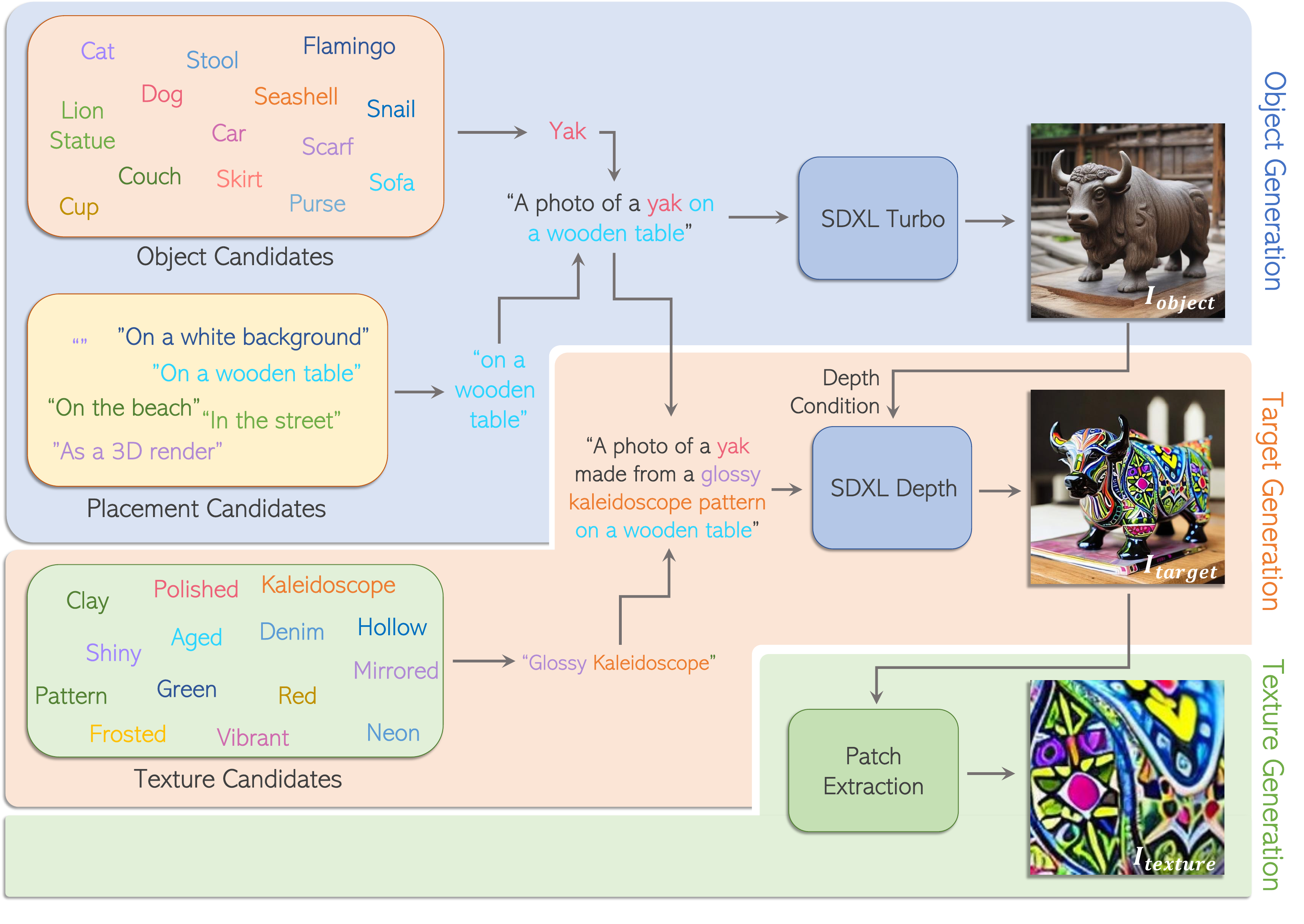}
    } \\[-0.4cm]
    \caption{
    \textbf{Data Generation Scheme}. An example scheme for our data generation, illustrated over our texturing operator. \\[-0.8cm]
    }
    \label{fig:data_generation_scheme}
\end{figure}

\paragraph{\textbf{Texturing}}
In the texturing operator, our input image embeddings consist of $e_{object}$, the embedding of the object to be textured, and $e_{texture}$, representing the desired texture. Our goal is to generate a target embedding $e_{target}$, depicting an image of $e_{object}$ textured with $e_{texture}$. 
The data generation protocol used to create our paired texturing dataset is illustrated in~\Cref{fig:data_generation_scheme}.

We begin by generating an object using SDXL-Turbo~\cite{sauer2024fast}. The resulting image embedding then serves as $e_{object}$ used during training. 
Next, we compile a set of attributes associated with textures and randomly sample a subset of these properties, composing them into a descriptive sentence. 
We then generate an image using a depth-conditioned Stable Diffusion model, conditioned on the depth of the generated object image and the composed text prompt. This process results in an image of our original object with a new texture, which we utilize to generate the embedding $e_{target}$.
Finally, to generate $e_{texture}$, we automatically extract a small patch from within the target image and define it as our texture exemplar. 

It is important to highlight that the texture is directly extracted from the target image. This encourages specificity, as a textual prompt can generate a range of plausible textures, whereas here, we condition the model on a \textit{specific} texture. Furthermore, achieving a complete match between the target and object images is not necessary. For instance, there can be variations in the background between the two, as long as they remain \textit{semantically} consistent.

\vspace{-0.1cm}
\paragraph{\textbf{Scene}}
In our scene operator, we receive two input embeddings: $e_{object}$, representing our object of interest, and $e_{back}$, denoting a target scene background for placing the object. As in texturing, we initially generate an image of our object using SDXL-Turbo, which corresponds to $e_{target}$. Next, we employ a background removal model~\cite{bria_br_model} to isolate our object from the generated image. The segmented object is then positioned either on a white background or within a newly generated background, which is encoded into the $e_{object}$ embedding.
Lastly, we utilize a Stable Diffusion inpainting model to produce an image containing only the original background, which we encode to $e_{back}$. 
In essence, during the data generation phase, we decompose the target into separate representations of its object and background. Through this process, \textit{pOps} can learn how to effectively compose the two elements back together.

\vspace{-0.1cm}
\paragraph{\textbf{Union}}
In our union operator, we receive two image embeddings representing two objects, denoted as $e_a$ and $e_b$, with the aim of generating an image embedding that plausibly incorporates both objects. To construct the union dataset, we build on the intuition that separating objects from existing scenes is typically easier than integrating them together.
Therefore, we first construct a dataset of images containing pairs of objects by randomly selecting two object classes and generating an image containing both objects using SDXL-Turbo (e.g., ``a cat and a banana''). This resulting image is then encoded to define $e_{target}$. Next, we employ a grounded detection method, OWLv2~\cite{minderer2024scaling}, to extract each object of interest as an individual crop, generating $e_a$ and $e_b$, respectively.
The \textit{pOps} operator is then tasked with composing these part embeddings back into a single image combining both parts.

\vspace{-0.1cm}
\subsection{Multi-Image Compositions}
While binary operators cover a wide range of tasks and can be combined in a tree-like structure to execute more complex operations, some operators can benefit from considering all inputs simultaneously. 
To illustrate this, we explore a specific composition operator that takes a set of embeddings, each representing a distinct clothing item, and combines them into a single representation of a person wearing those clothes.
To train such an operator, we extend the input sequence to accommodate the set of clothing items, setting a fixed input index for each clothing type.
This again leverages the original design of the prior, which was tailored to process a sequence of $77$ input text tokens.
For training, we utilize the ATR dataset~\cite{liang2015deep}, developed for human parsing. We encode the given complete image as our target embedding $e_{target}$ and decompose the clothing items using the segmentation masks annotated in the dataset to form our input sequence.
The training scheme itself is identical to the binary operators, utilizing~\Cref{eq:prior} to train the prior model on our composition task.

\begin{figure*}
    \centering
    \setlength{\tabcolsep}{0.5pt}
    {\small

    \begin{tabular}{c c c c @{\hspace{0.4cm}} c c c @{\hspace{0.4cm}} c c c }
        \raisebox{0.035\linewidth}{\rotatebox[origin=t]{90}{\begin{tabular}{c@{}c@{}c@{}c@{}} Texturing \end{tabular}}} &
        \includegraphics[width=0.1025\textwidth]{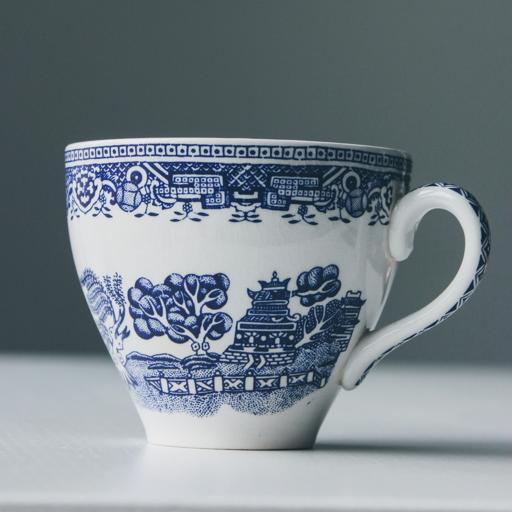} &
        \includegraphics[width=0.1025\textwidth]{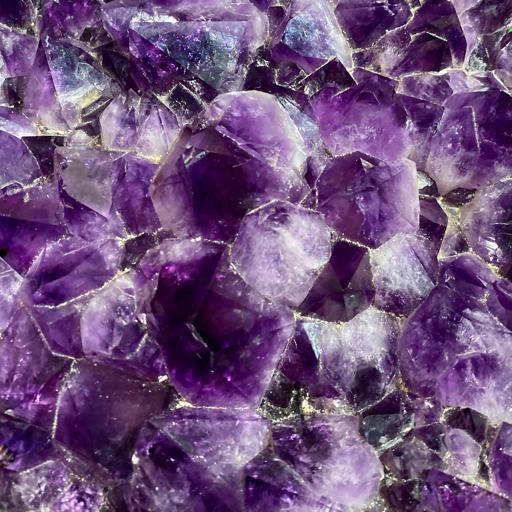} &
        \includegraphics[width=0.1025\textwidth]{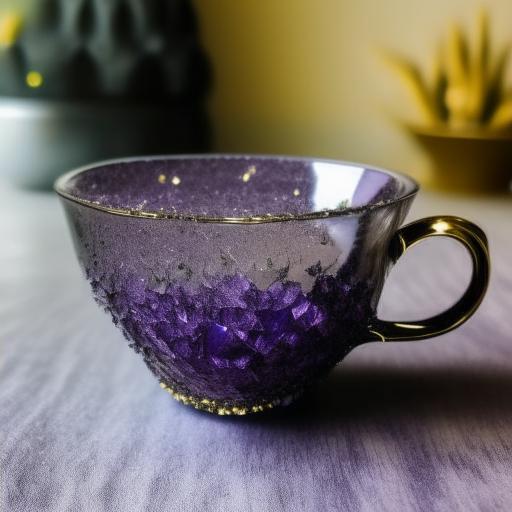} &
        \includegraphics[width=0.1025\textwidth]{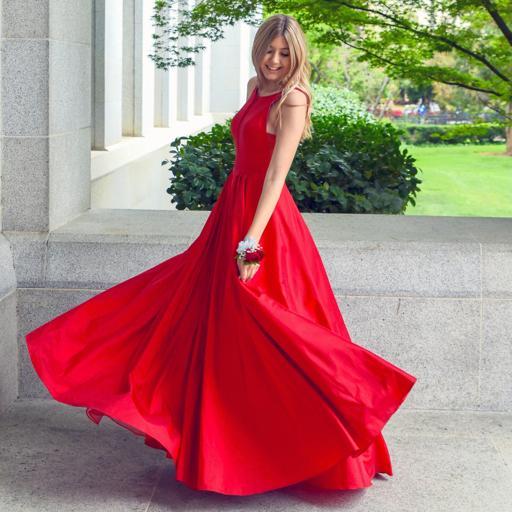} &
        \includegraphics[width=0.1025\textwidth]{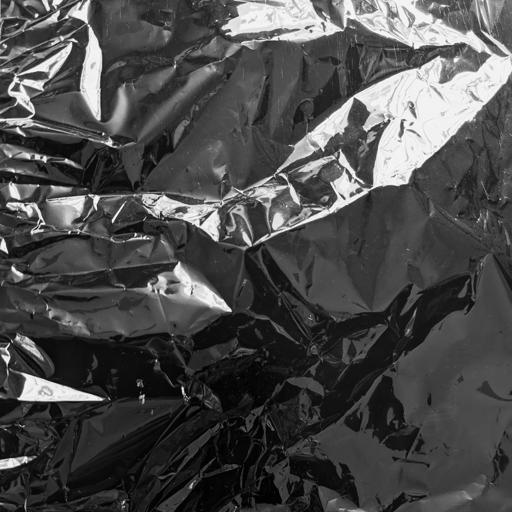} &
        \includegraphics[width=0.1025\textwidth]{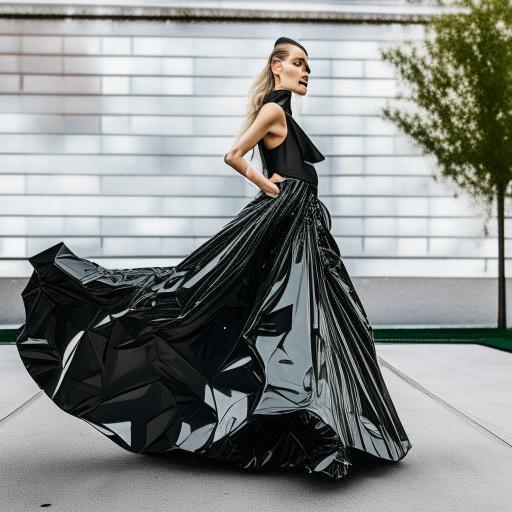} &
        \includegraphics[width=0.1025\textwidth]{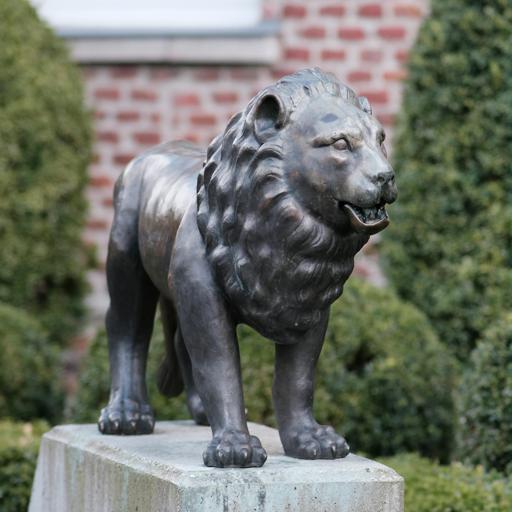} &
        \includegraphics[width=0.1025\textwidth]{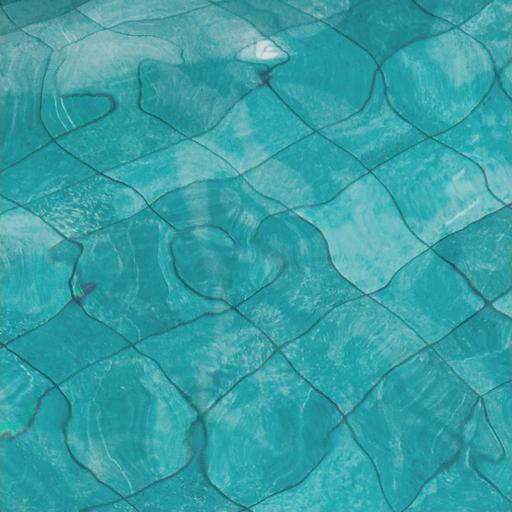} &
        \includegraphics[width=0.1025\textwidth]{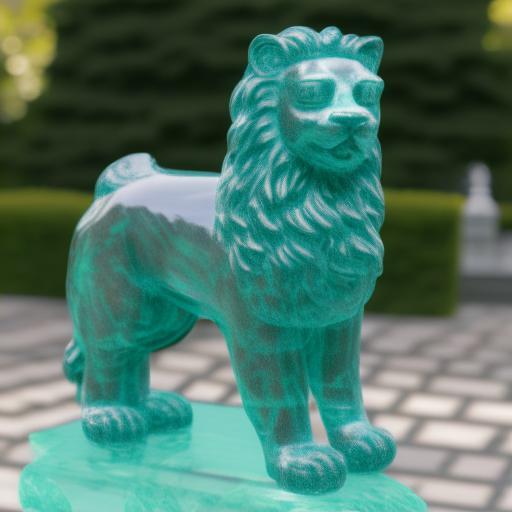} 
        
        \vspace{-0.1cm}    
        \\ 
    
        & Object & Texture & Result & Object & Texture & Result & Object & Texture & Result \\ \\[-0.3cm]

        \raisebox{0.035\linewidth}{\rotatebox[origin=t]{90}{\begin{tabular}{c@{}c@{}c@{}c@{}} Scene \end{tabular}}} &
        \includegraphics[width=0.1025\textwidth]{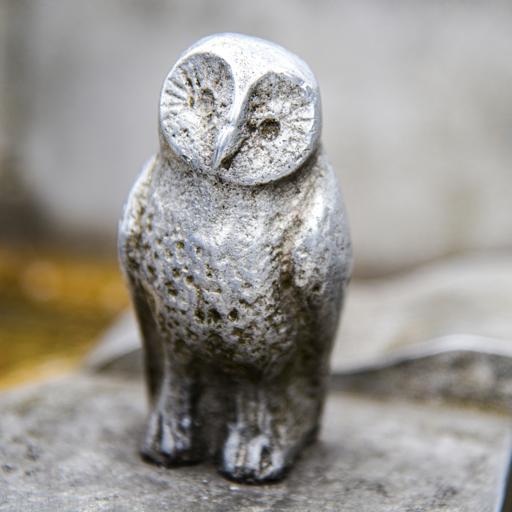} &
        \includegraphics[width=0.1025\textwidth]{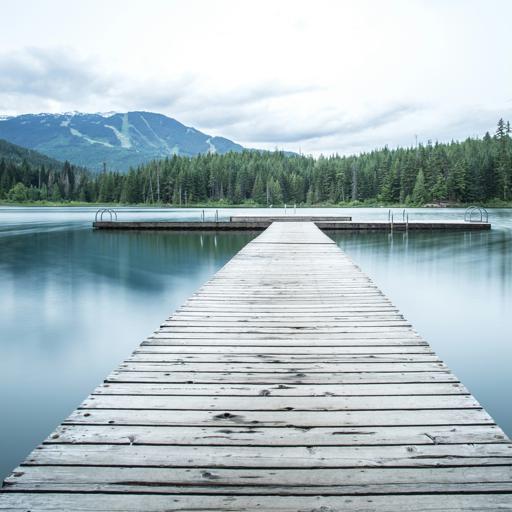} &
        \includegraphics[width=0.1025\textwidth]{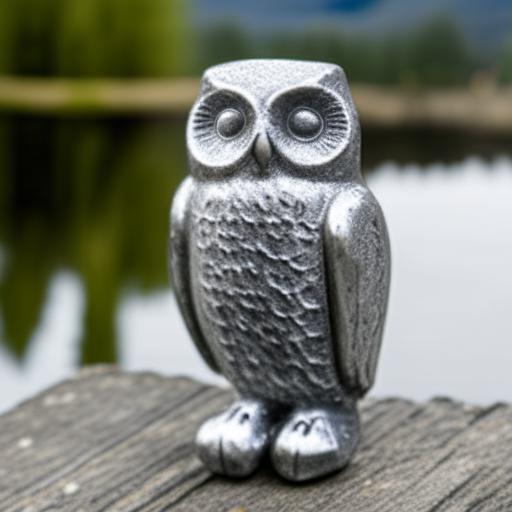} &
        \includegraphics[width=0.1025\textwidth]{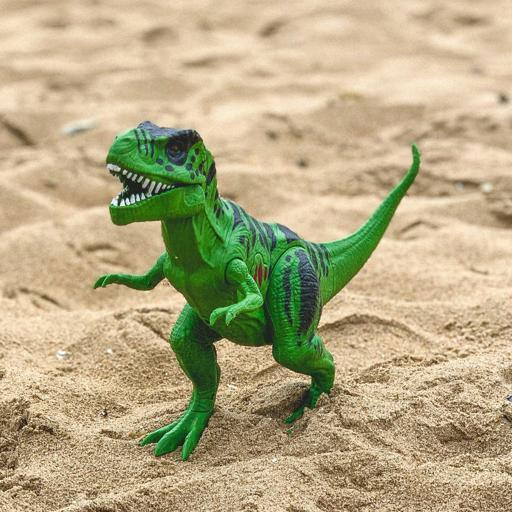} &
        \includegraphics[width=0.1025\textwidth]{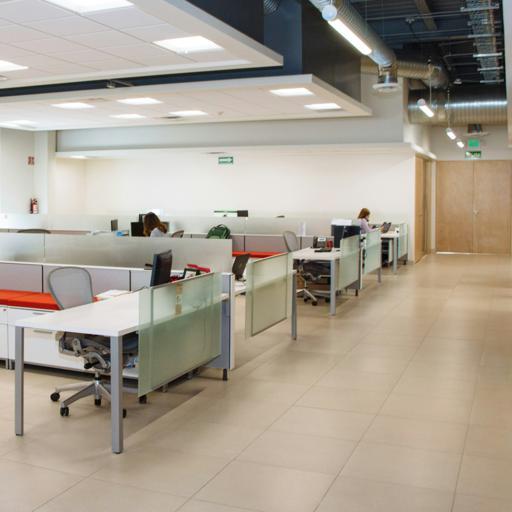} &
        \includegraphics[width=0.1025\textwidth]{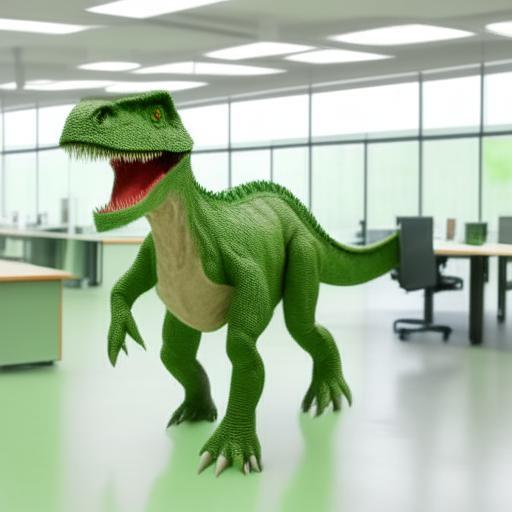} &

        \includegraphics[width=0.1025\textwidth]{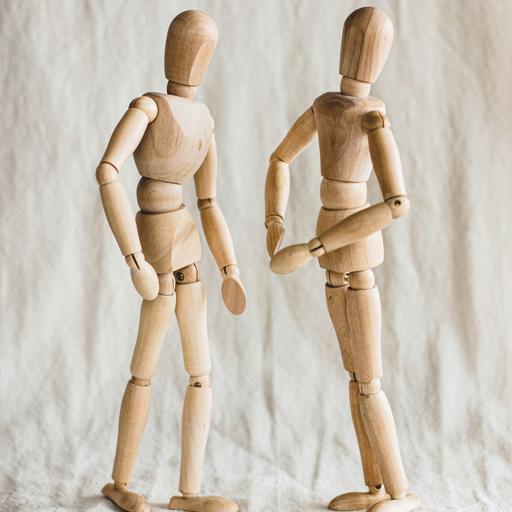} &
        \includegraphics[width=0.1025\textwidth]{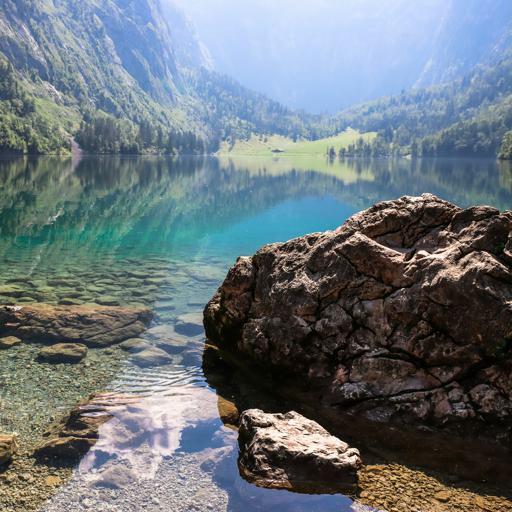} &
        \includegraphics[width=0.1025\textwidth]{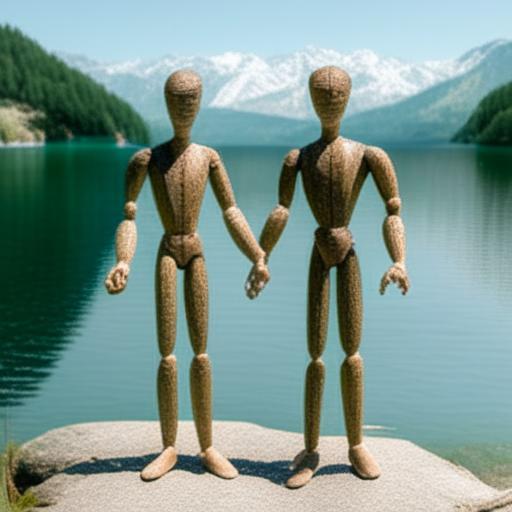} 
        \vspace{-0.1cm}    
        \\

        & Object & Background & Output &  Object & Background & Output & Object & Background & Output \\ \\[-0.3cm]

        \raisebox{0.035\linewidth}{\rotatebox[origin=t]{90}{\begin{tabular}{c@{}c@{}c@{}c@{}} Union \end{tabular}}} &
        
        \includegraphics[width=0.1025\textwidth]{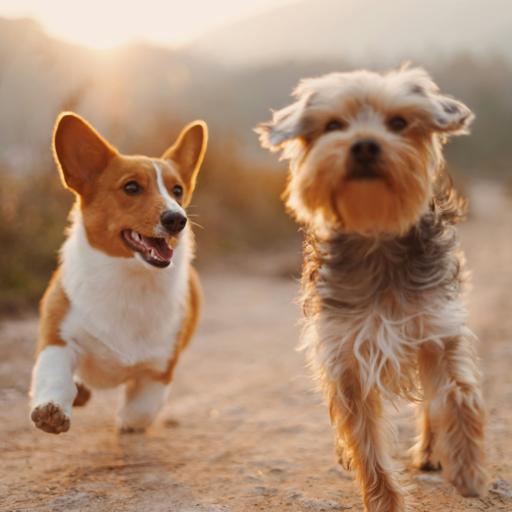} &
        \includegraphics[width=0.1025\textwidth]{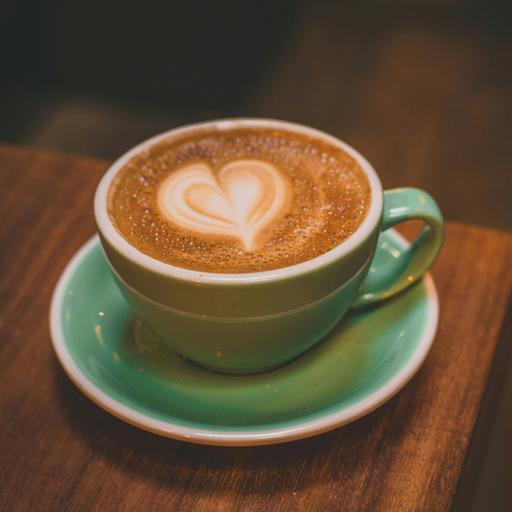} &

        \includegraphics[width=0.1025\textwidth]{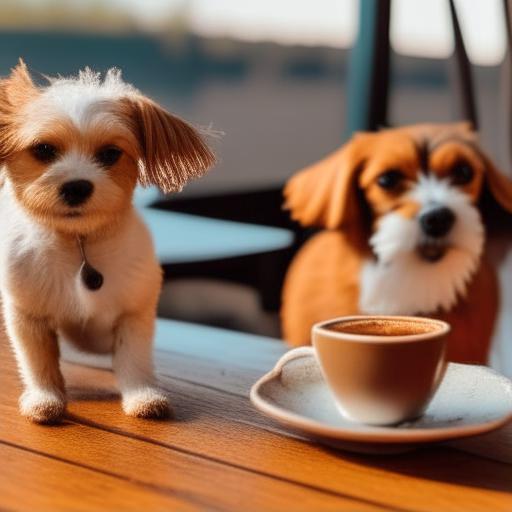} &
        
        \includegraphics[width=0.1025\textwidth]{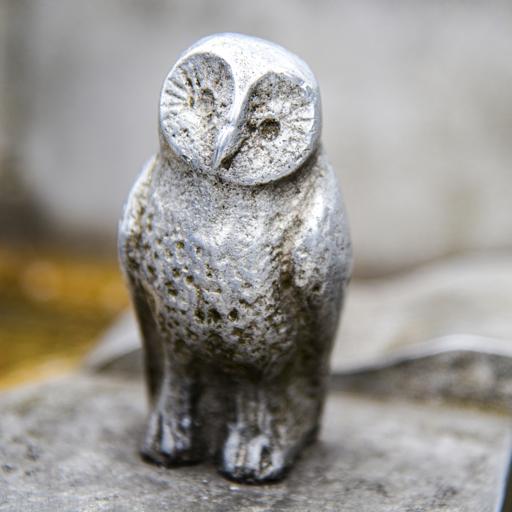} &
        \includegraphics[width=0.1025\textwidth]{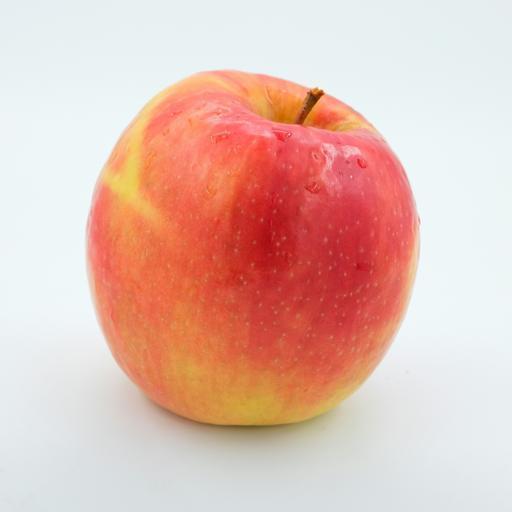} &

        \includegraphics[width=0.1025\textwidth]{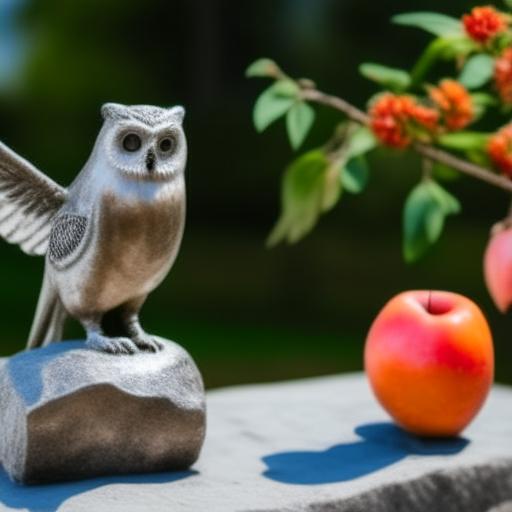} &

        \includegraphics[width=0.1025\textwidth]{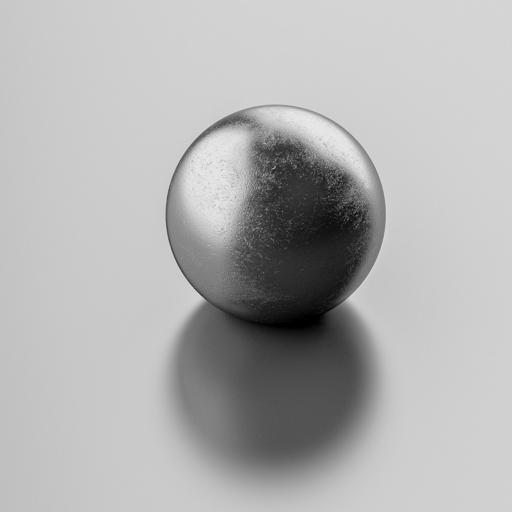} &
        \includegraphics[width=0.1025\textwidth]{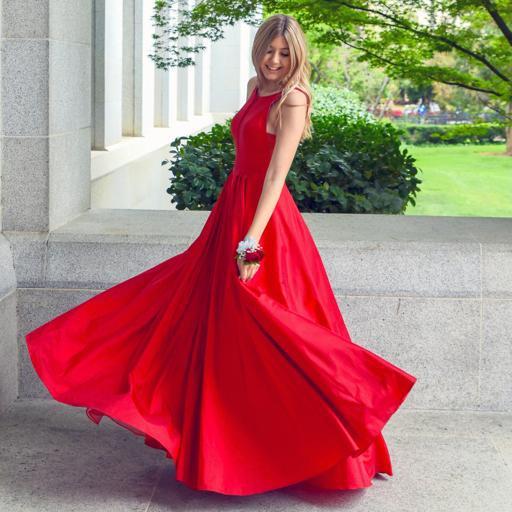} &
        \includegraphics[width=0.1025\textwidth]{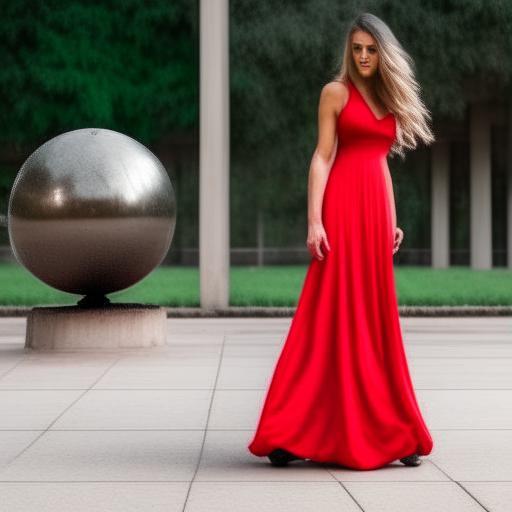} 
        \vspace{-0.1cm}    \\

    \end{tabular}
    }
    \vspace{-0.3cm}    
    \caption{Results obtained with our binary \textit{pOps} operators. Notice that while images are visualized, all operations are applied within the embedding space. 
    }
    \label{fig:binary_results}
\end{figure*}

\vspace{-0.15cm}
\subsection{The Instruct Operator}
All the operators discussed so far have assumed a paired dataset with a well-defined target embedding. However, operating in the CLIP embedding space presents interesting opportunities to easily apply additional losses within this space.
In particular, we explore a binary operator that takes as input a CLIP \textbf{image} embedding of an object, denoted as $e_{object}$, and a CLIP \textbf{text} embedding of a target adjective, labeled as $e_{instruct}$ (e.g., ``spiky'', ``hairy'', ``melting''). With these inputs, the prior model is tasked with generating an embedding $e_{object}$ corresponding to an image portraying the adjective described in $e_{instruct}$.

\setlength{\abovedisplayskip}{4pt}
\setlength{\belowdisplayskip}{4pt}

Since both the image embedding and text embedding reside in a shared CLIP space of the same dimensionality, we can easily feed both into our transformer. 
To train our task, we introduce an additional loss objective that evaluates the CLIP similarity between the generated image and the embedding $e_{text}$ of the prompt combining the target adjective and object class (e.g., ``a spiky dog''). Formally, our new loss objective is given by:
\begin{equation}\label{eq:prior}
    \mathcal{L} = \mathcal{L}_{prior} + \lambda \langle e_{text}, P_\theta (e_c^t, t, e_{object}, e_{instruct}) \rangle,
\end{equation}
where $P_\theta (e_c^t, t, e_{object}, e_{instruct})$ is the generated embedding.

\section{Experiments}
We now turn to validate the effectiveness of \textit{pOps} through a comprehensive set of evaluations. Additional details, along with a large gallery of results, are available in~\Cref{sec:additional_details,sec:additional_results,sec:additional_comparisons}.

\begin{figure}
    \centering
    \setlength{\tabcolsep}{0.5pt}
    \addtolength{\belowcaptionskip}{-5pt}
    {\small
    \begin{tabular}{c c c @{\hspace{0.2cm}} c c}

        \includegraphics[width=0.09\textwidth]{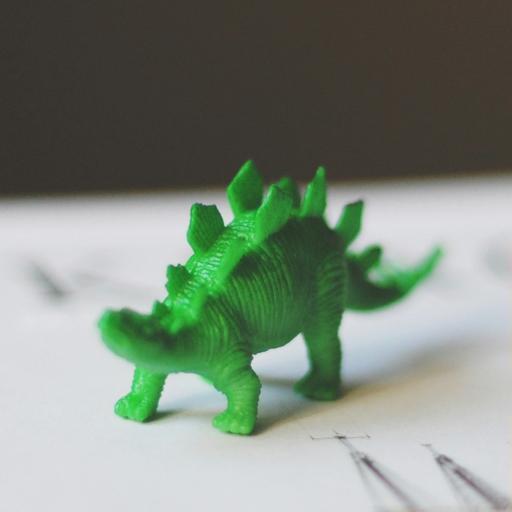} &
        \includegraphics[width=0.09\textwidth]{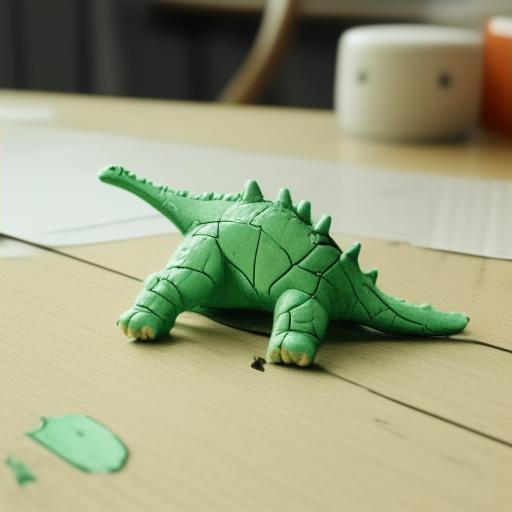} &
        \includegraphics[width=0.09\textwidth]{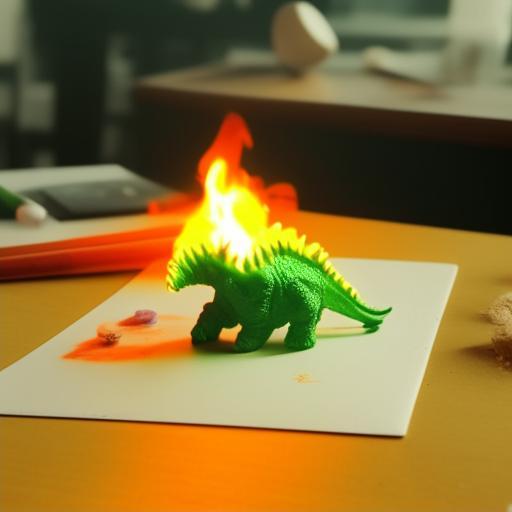} &
        \includegraphics[width=0.09\textwidth]{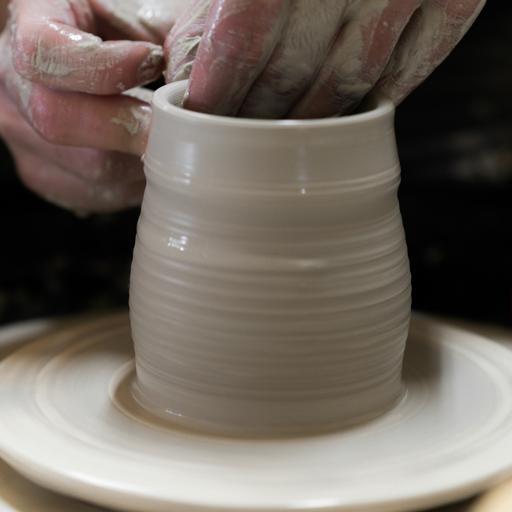} &
        \includegraphics[width=0.09\textwidth]{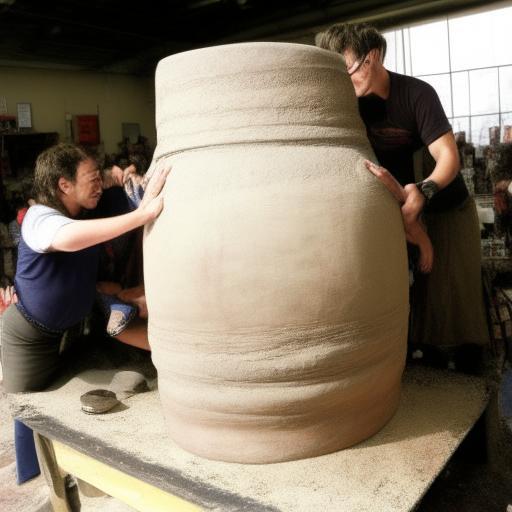} \\ 
        Input & ``cracked''   & ``burning''  & Input & ``enormous'' \\

        \includegraphics[width=0.09\textwidth]{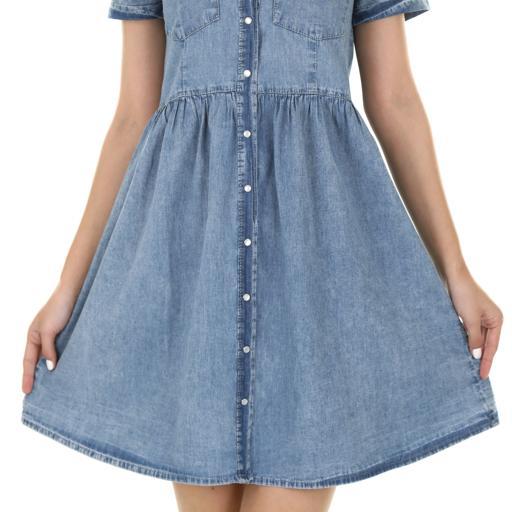} &
        \includegraphics[width=0.09\textwidth]{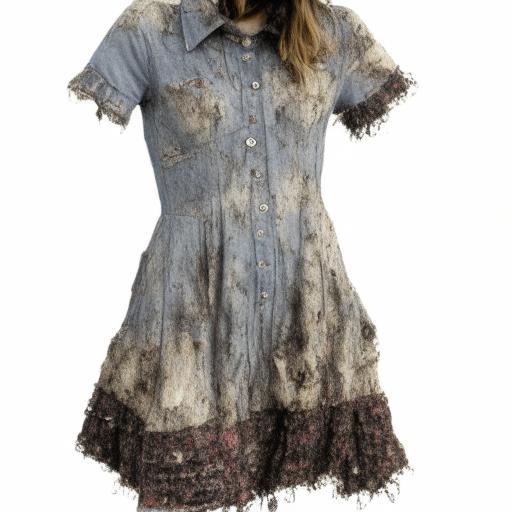} &
        \includegraphics[width=0.09\textwidth]{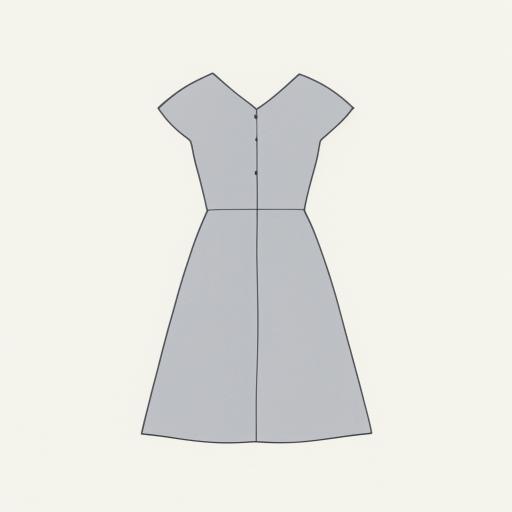} &
        \includegraphics[width=0.09\textwidth]{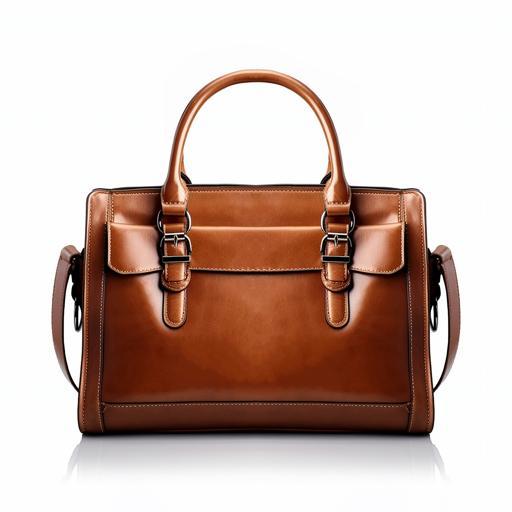} &
        \includegraphics[width=0.09\textwidth]{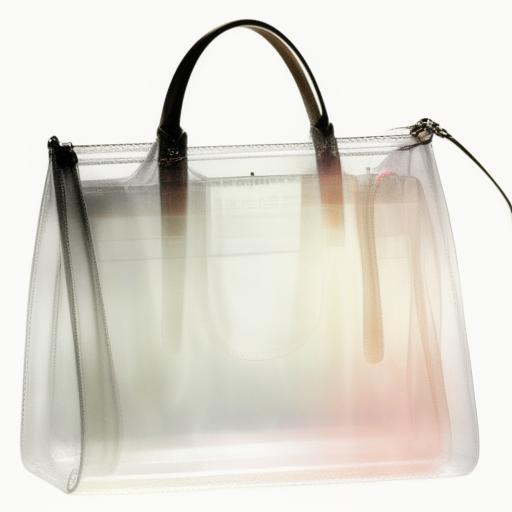} \\
        Input & ``rotten'' & ``minimalistic'' & Input & ``translucent'' \\

        \includegraphics[width=0.09\textwidth]{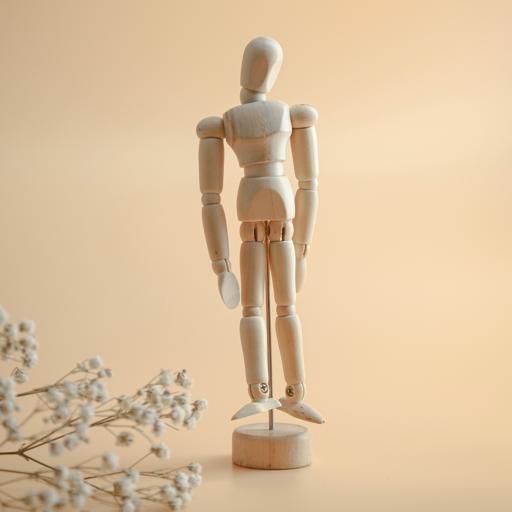} &
        \includegraphics[width=0.09\textwidth]{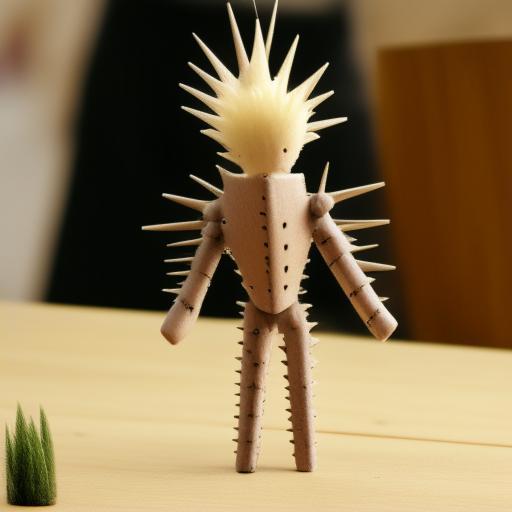} &
        \includegraphics[width=0.09\textwidth]{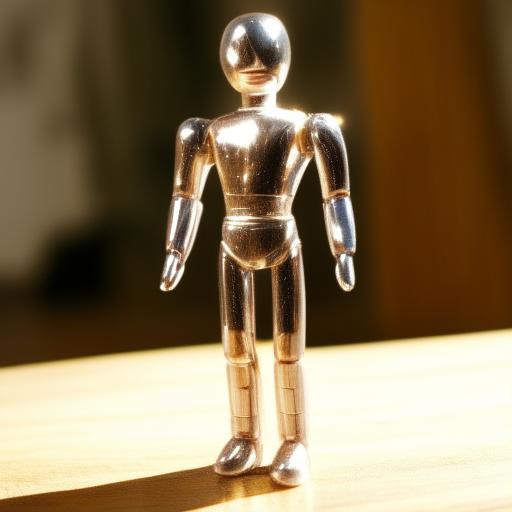} &
        \includegraphics[width=0.09\textwidth]{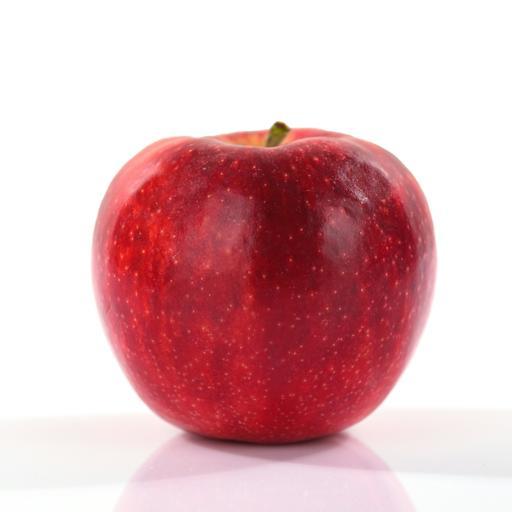} &
        \includegraphics[width=0.09\textwidth]{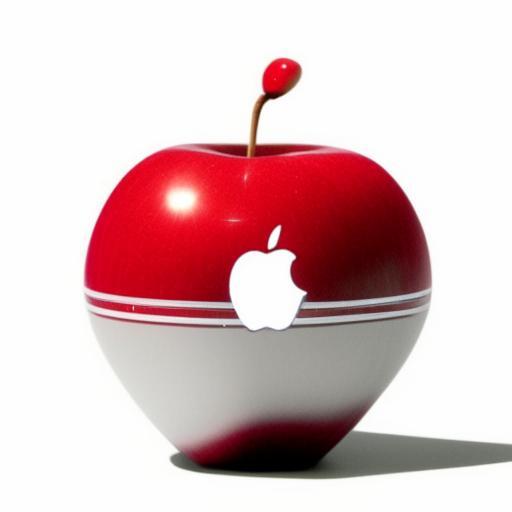}  \\

        Input & ``spiky'' & ``shiny'' & Input & ``futuristic'' \\

    \end{tabular}
    }
    \caption{Instruct operator results obtained by \textit{pOps}.
    }
    \label{fig:instruct_operator_small}
\end{figure}

\begin{figure}
    \centering
    \setlength{\tabcolsep}{0.5pt}
    \renewcommand{\arraystretch}{0.2}
    {\small
    \begin{tabular}{
        c 
    }

    \includegraphics[width=0.49\textwidth]{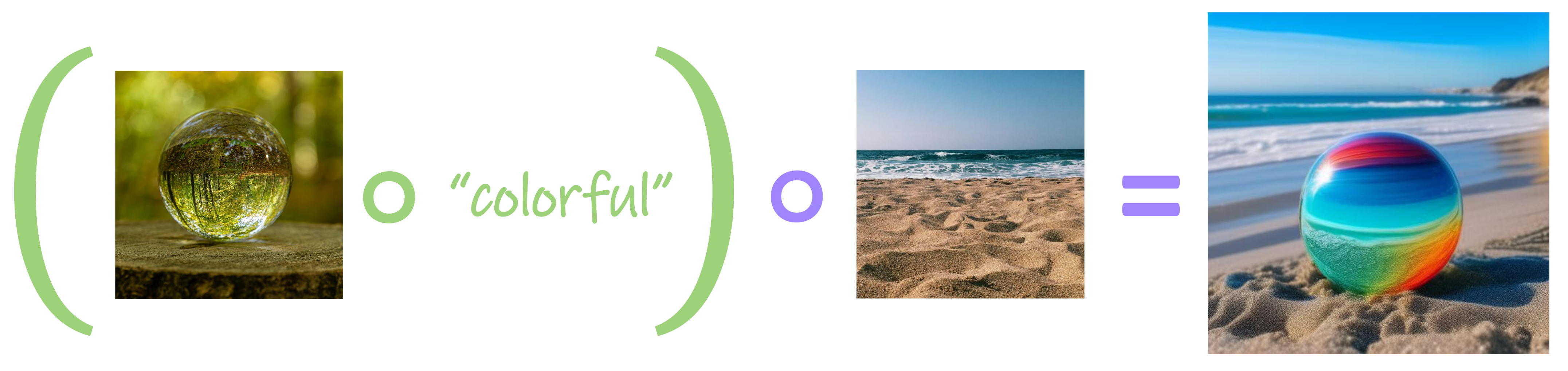} \\
    \includegraphics[width=0.49\textwidth]{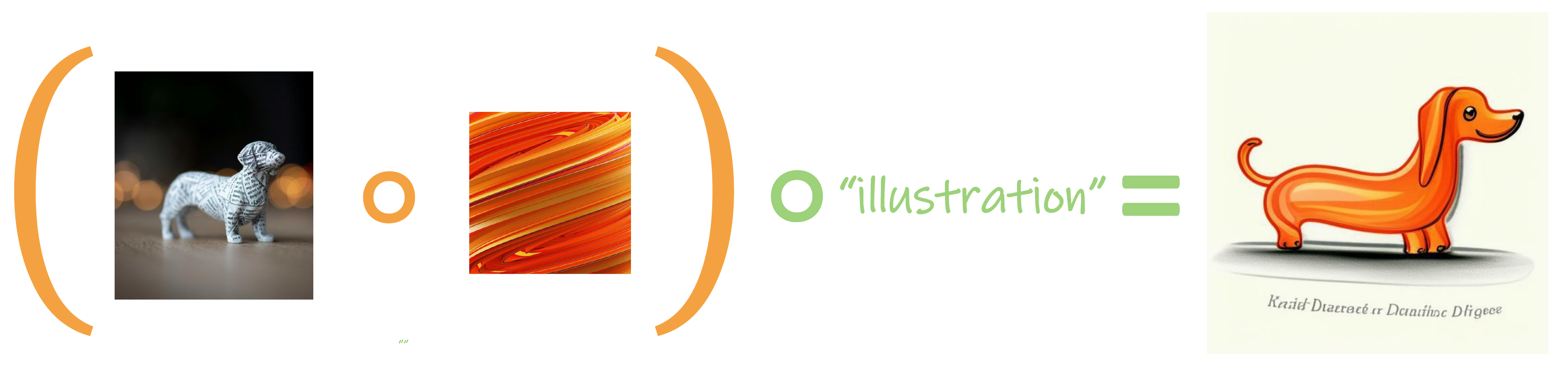} \\
    \includegraphics[width=0.49\textwidth]{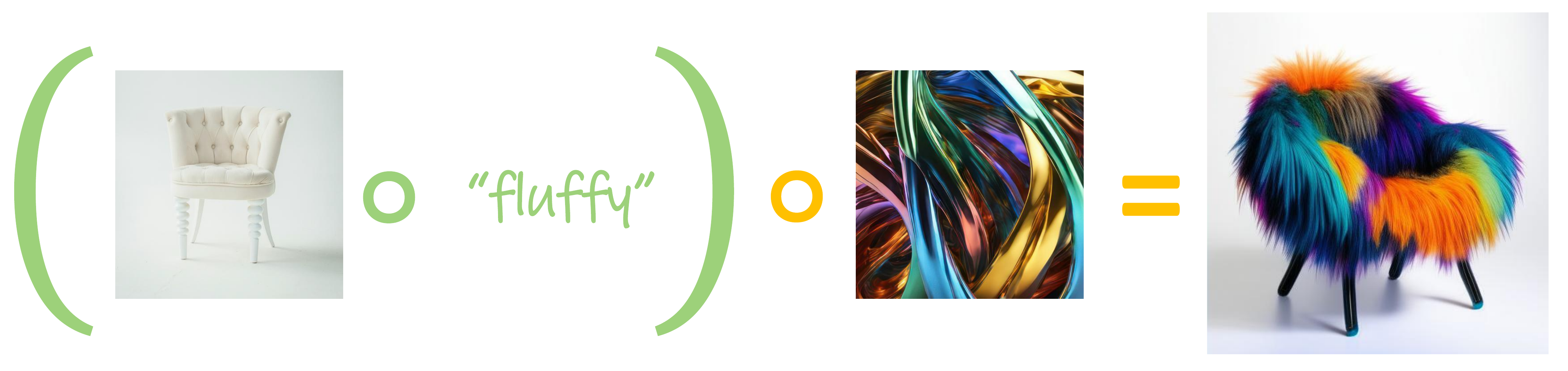} \\

    \end{tabular}
    }
    \caption{Multi-operator compositions obtained by our \textit{pOps} method. \\[-0.1cm]}
    \label{fig:compositions_small}
\end{figure}

\begin{figure}
    \centering
    \setlength{\tabcolsep}{3pt}
    \addtolength{\belowcaptionskip}{-8pt}
    {\small
    \begin{tabular}{c @{\hspace{4pt}} c}

        \includegraphics[height=0.09\textwidth]{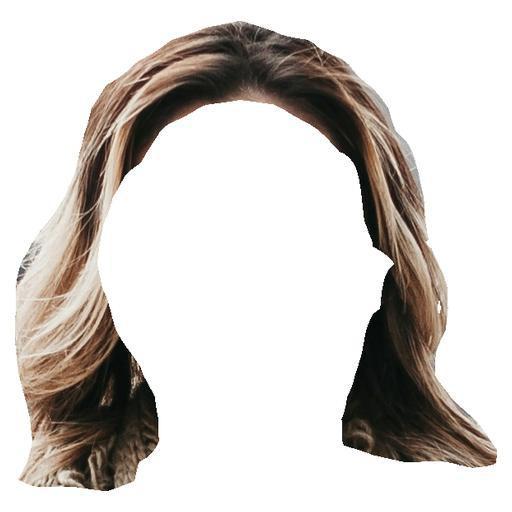} 
        \includegraphics[height=0.1\textwidth]{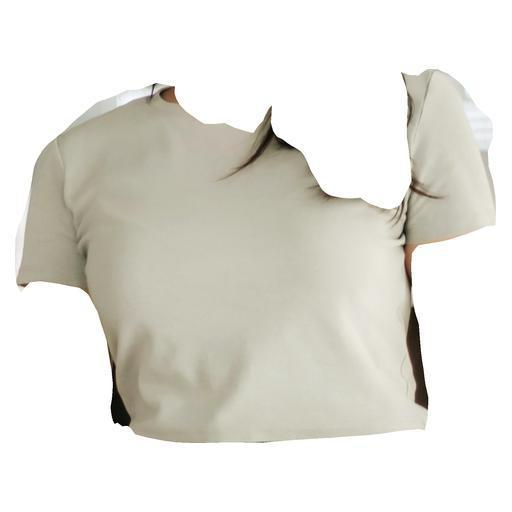} 
        \includegraphics[height=0.09\textwidth]{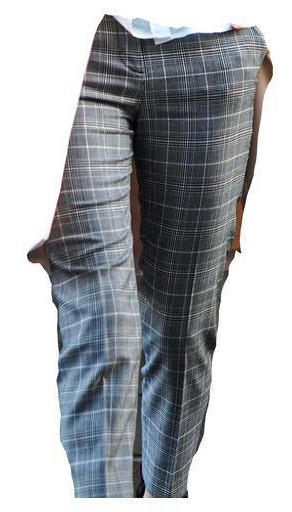} &
        \includegraphics[height=0.1\textwidth]{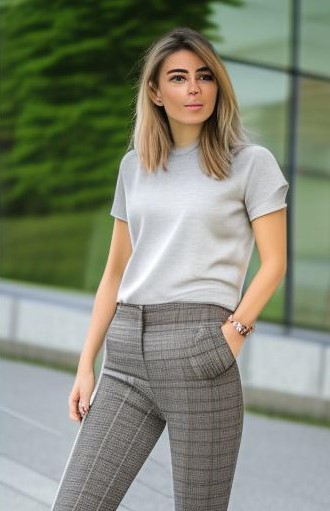} \\
        \includegraphics[height=0.1\textwidth]{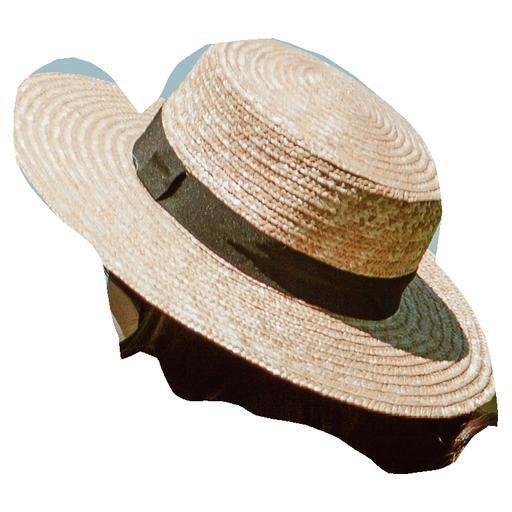} 
        \includegraphics[height=0.1\textwidth]{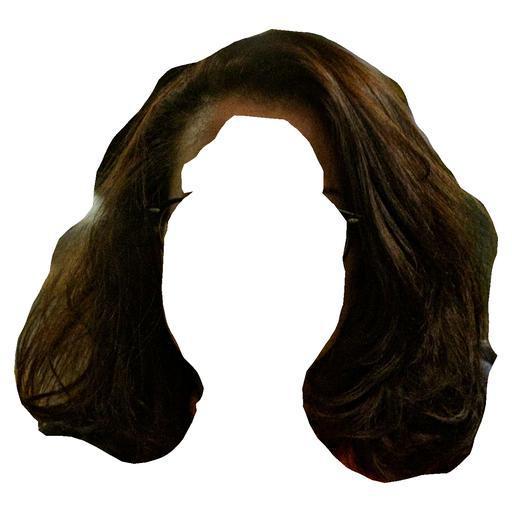} 
        \includegraphics[height=0.1\textwidth]{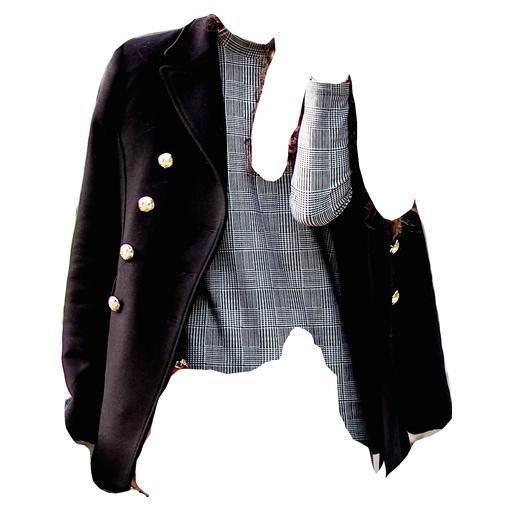} 
        \includegraphics[height=0.1\textwidth]{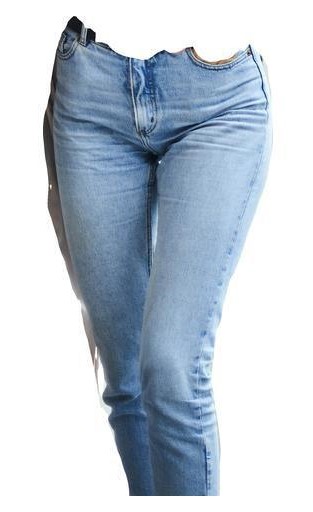} &
        \includegraphics[height=0.1\textwidth]{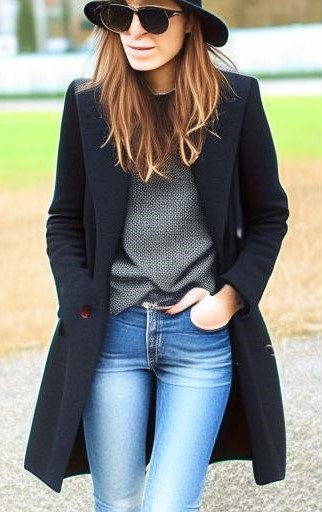} \\

         Inputs & Result \\

    \end{tabular}
    }
    \caption{Multi-image compose operator results obtained by pOps.}
    \label{fig:clothes_compose_small}
\end{figure}

\paragraph{\textbf{Operator Results.}}
Results for our binary operators are provided in~\Cref{fig:binary_results}, where each operator effectively realizes a specific and consistent semantic operation. Given that we operate within the CLIP embedding space, the operators focus on preserving the semantic nature of the inputs while being agnostic to the structure or placement of the objects. 
Next, we present results for our instruct operator in~\Cref{fig:instruct_operator_small}. Given a single descriptive word, our operator successfully generates a plausible output incorporating both the adjective and input object. 
As shown in~\Cref{fig:compositions_small}, our operators can also be combined into generative equations representing more complex semantic operations. These operations are applied directly in the image embedding space, where only the final embedding is ``rendered'' into a corresponding image.
Finally, \Cref{fig:clothes_compose_small} shows a multi-input example where \textit{pOps} was trained to take a sequence of embeddings corresponding to articles of clothing and output an embedding that represents the complete outfit. 
Additional results for all operators are available in~\Cref{sec:additional_results}.

\begin{figure}
    \centering
    \setlength{\tabcolsep}{0.5pt}
 {\small
    \begin{tabular}{c c c  @{\hspace{0.3cm}} c c c}
         
         \includegraphics[height=0.084\textwidth]{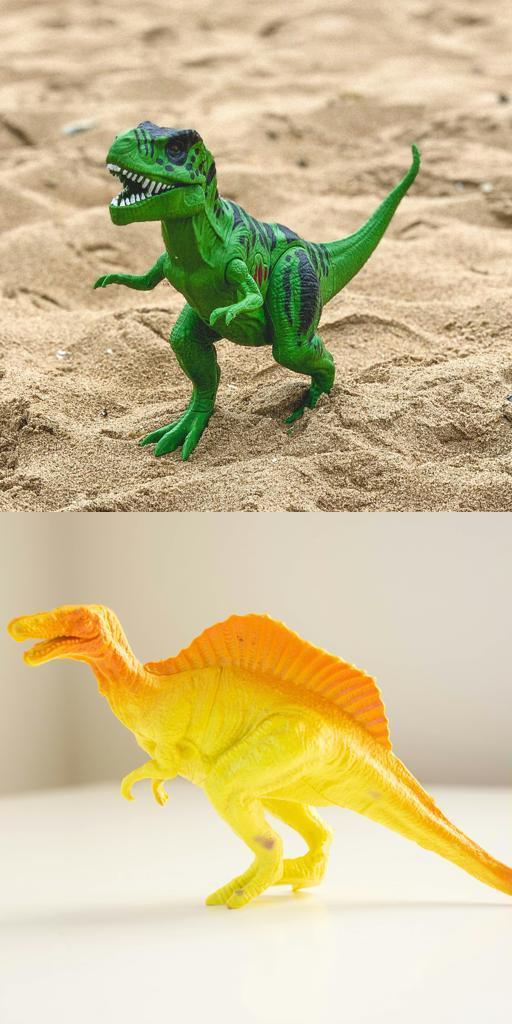} &
        \includegraphics[height=0.084\textwidth]{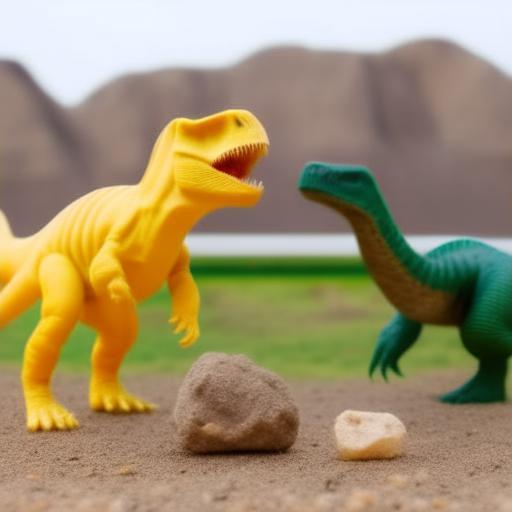} &
        \includegraphics[height=0.084\textwidth]{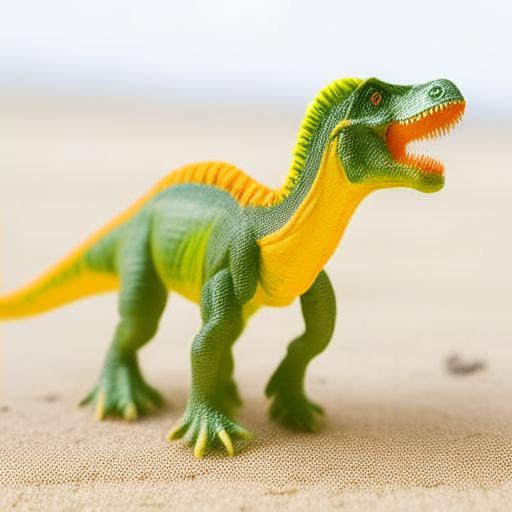} &
         \includegraphics[height=0.084\textwidth]{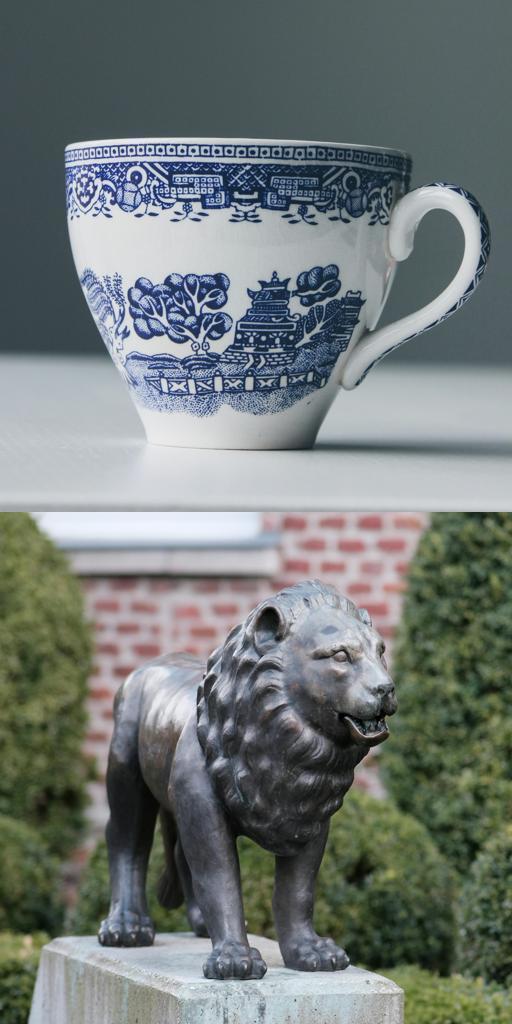} &
        \includegraphics[height=0.084\textwidth]{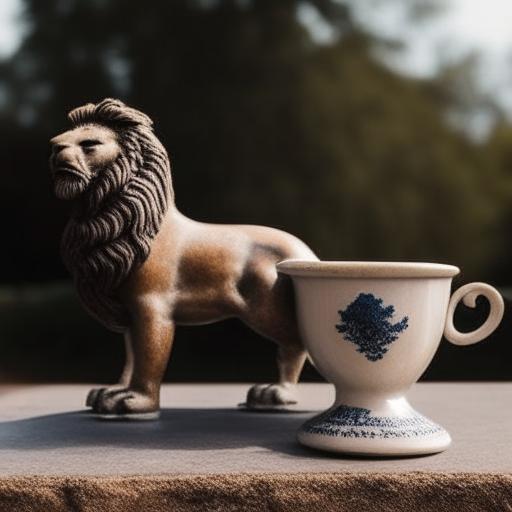} &
        \includegraphics[height=0.084\textwidth]{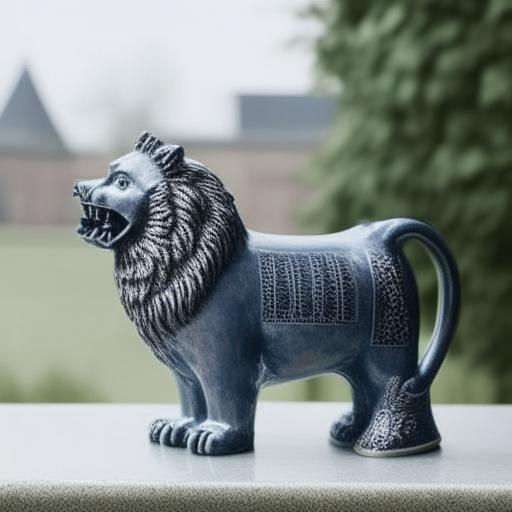} \\
        Input & Union & Average & Input & Union & Average \\
        \includegraphics[height=0.084\textwidth]{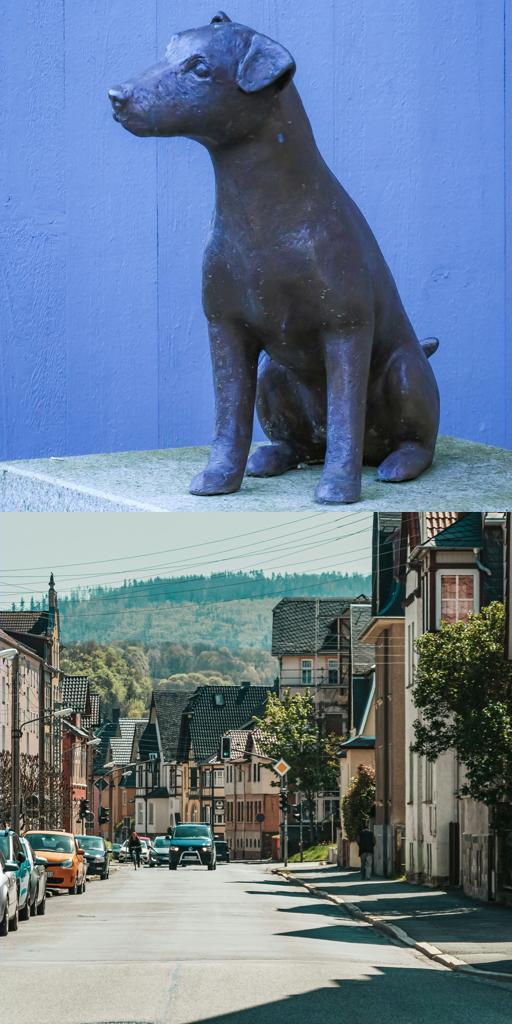} &
        \includegraphics[height=0.084\textwidth]{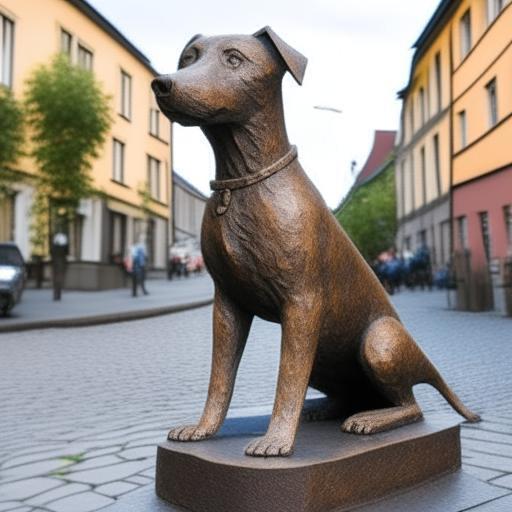} &
        \includegraphics[height=0.084\textwidth]{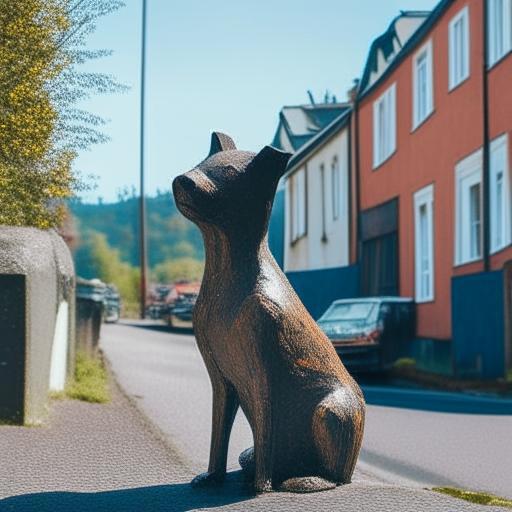} &
         \includegraphics[height=0.084\textwidth]{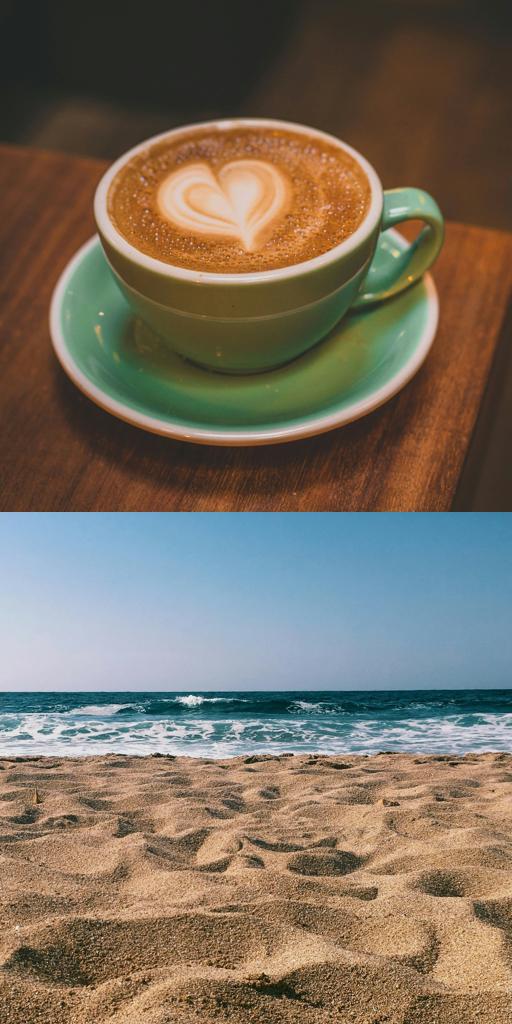} &
        \includegraphics[height=0.084\textwidth]{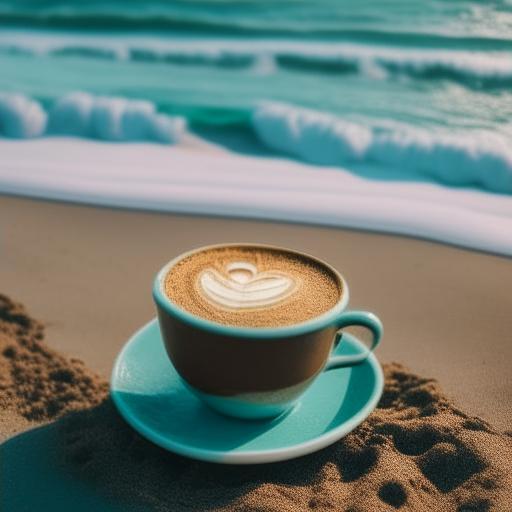} &
        \includegraphics[height=0.084\textwidth]{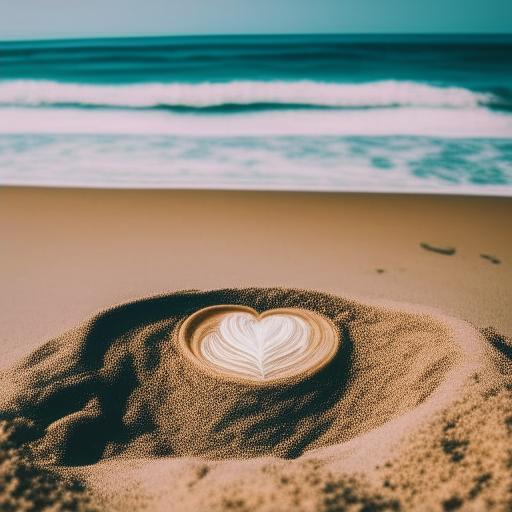} \\
        Input & Scene & Average & Input & Scene & Average 

    \end{tabular}
 }
    \caption{Qualitative comparison of \textit{pOps} to latent averaging.}
    \label{fig:average_small}
    
\end{figure}

\begin{figure}
    \centering
    \setlength{\tabcolsep}{0.5pt}
    {\small
    \begin{tabular}{c c@{\hspace{0.25cm}} c c c c c}

        \includegraphics[width=0.09\textwidth]{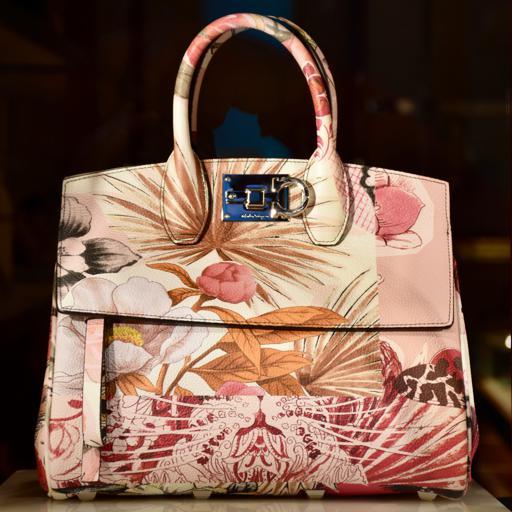} &
        \includegraphics[width=0.09\textwidth]{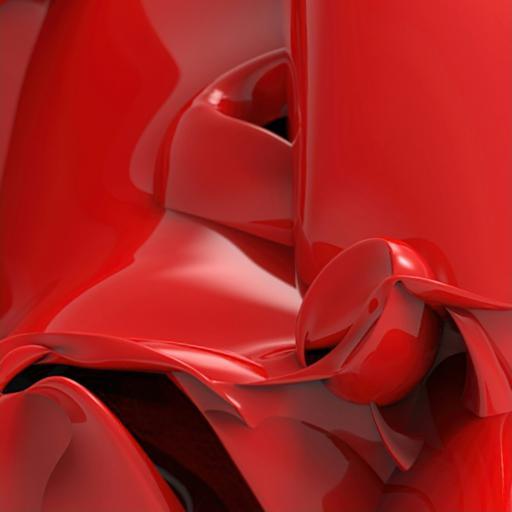} &
        \includegraphics[width=0.09\textwidth]{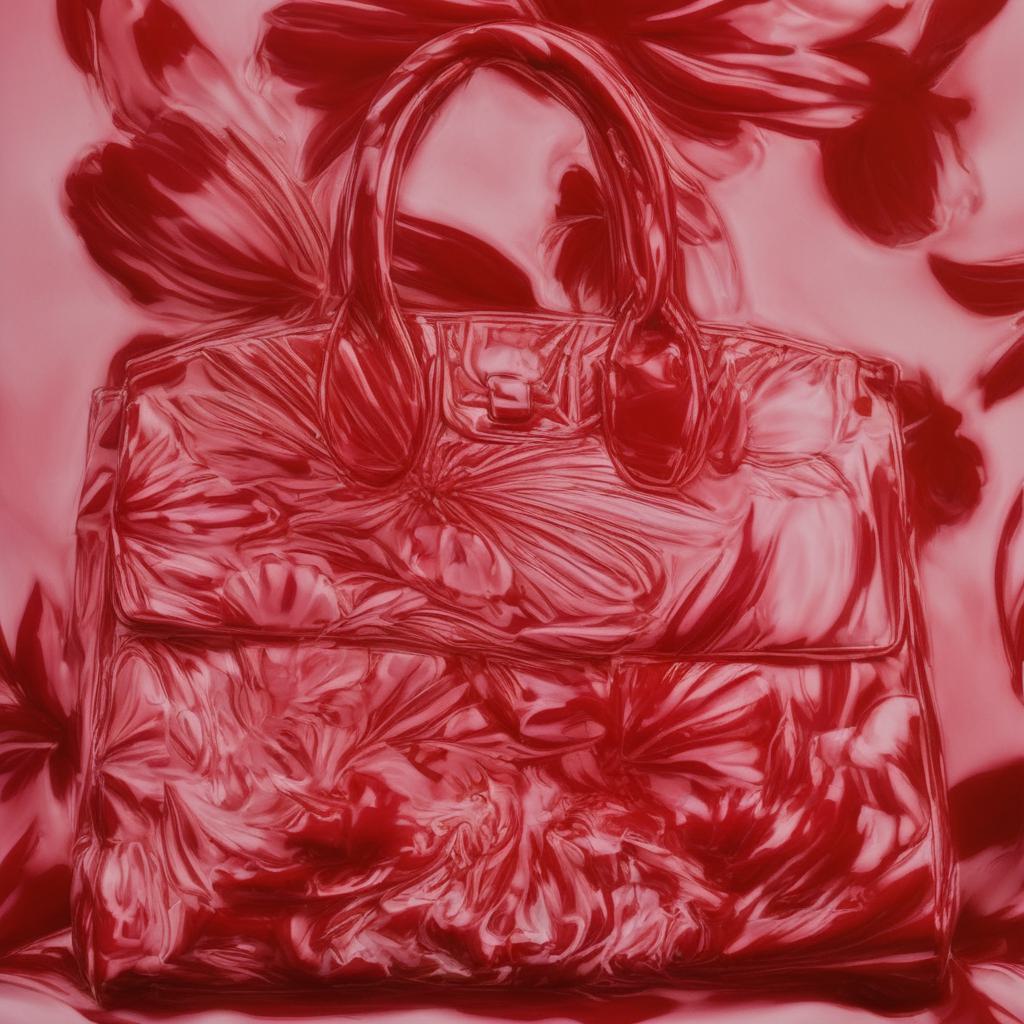} &
        \includegraphics[width=0.09\textwidth]{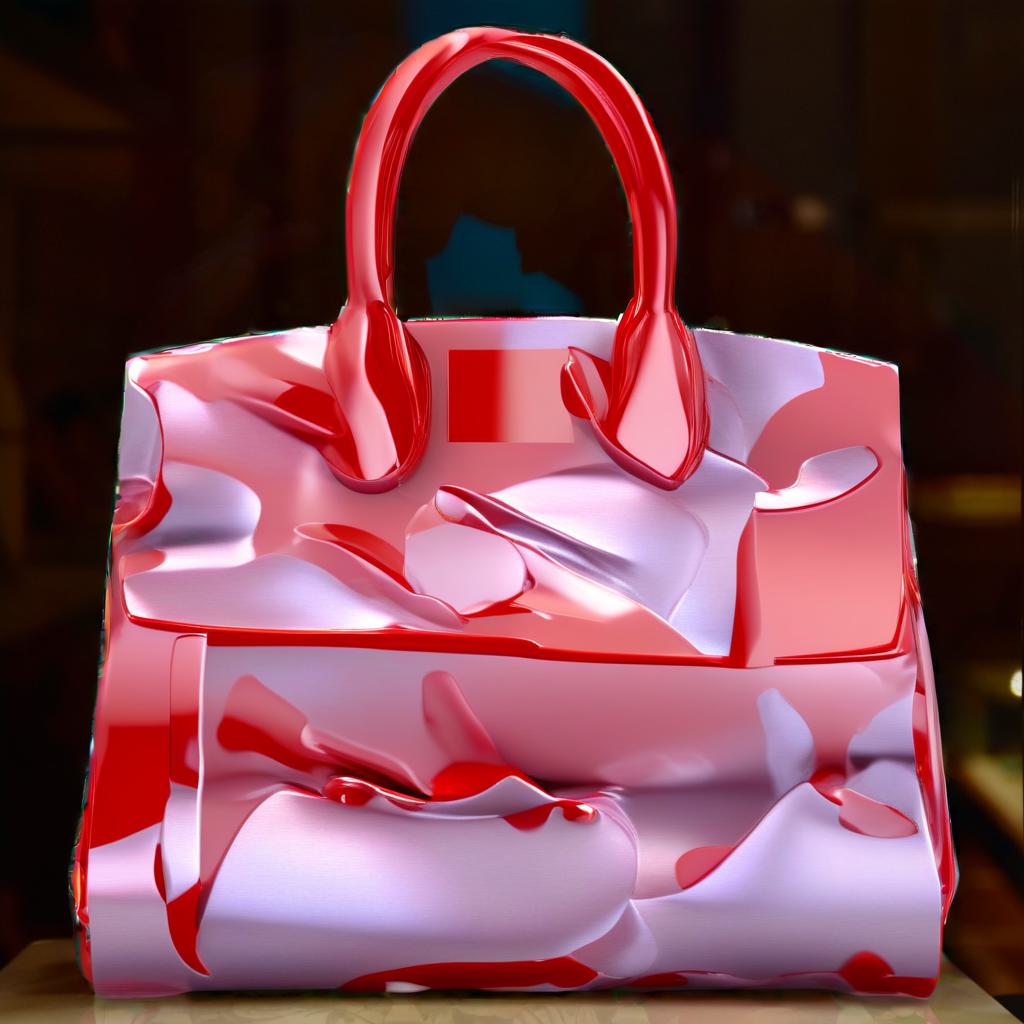} &
        \includegraphics[width=0.09\textwidth]{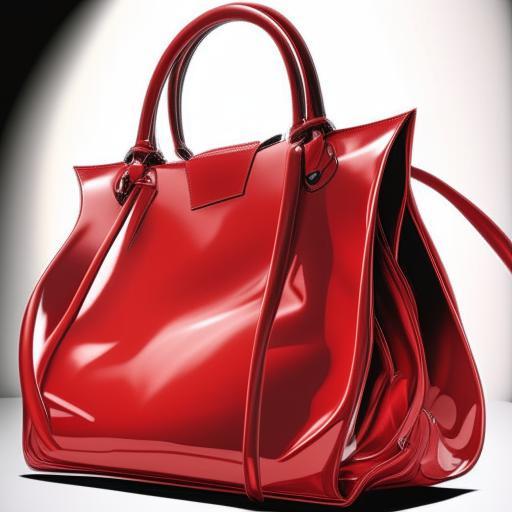} \\

        \includegraphics[width=0.09\textwidth]{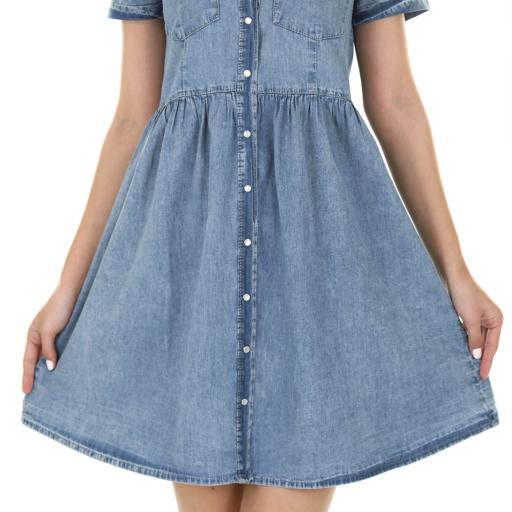} &
        \includegraphics[width=0.09\textwidth]{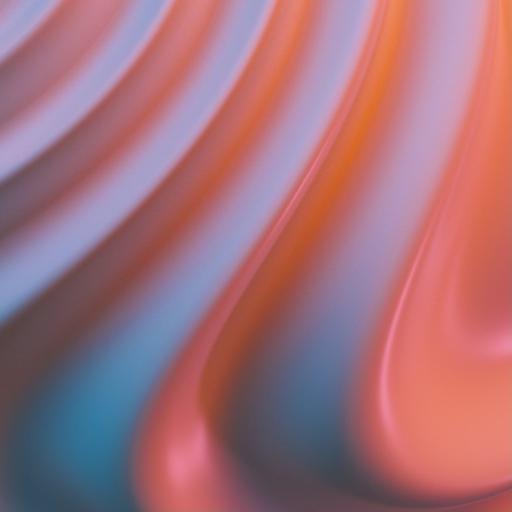} &
        \includegraphics[width=0.09\textwidth]{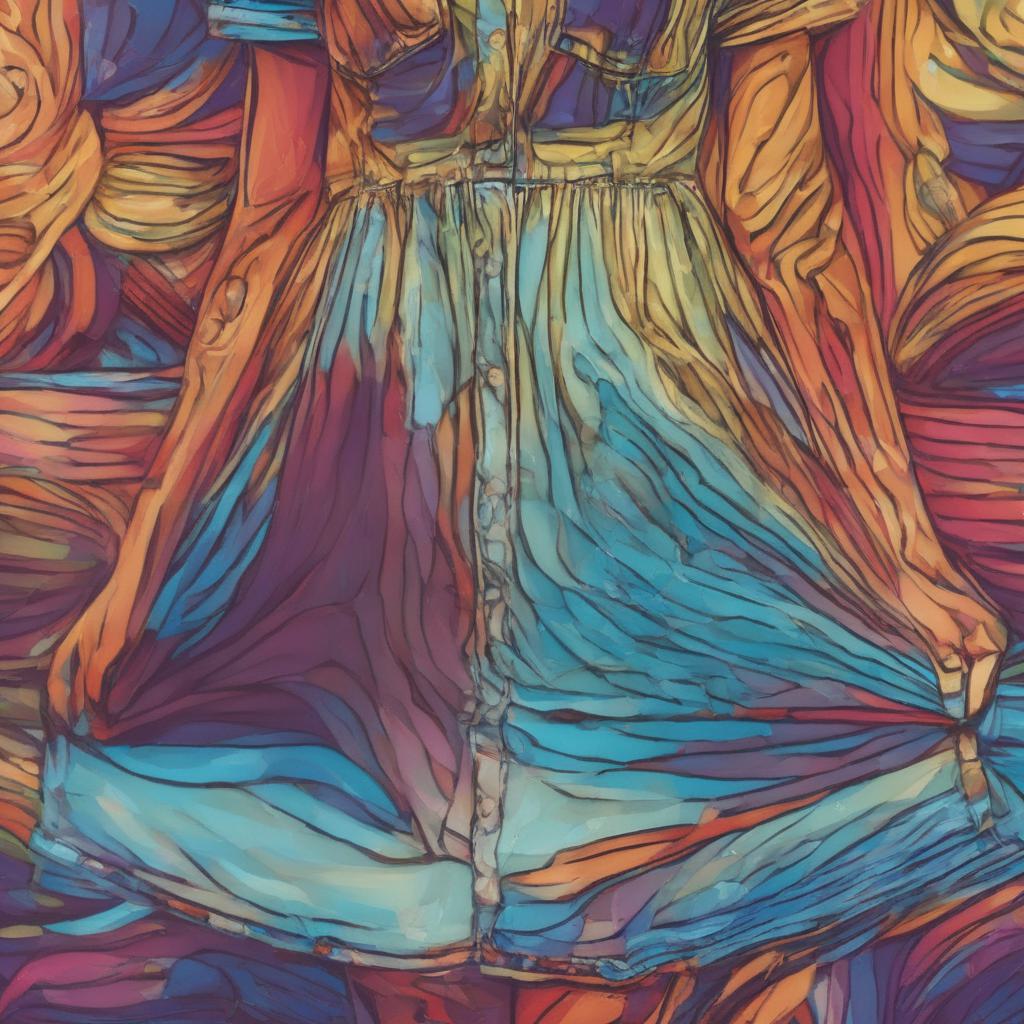} &
        \includegraphics[width=0.09\textwidth]{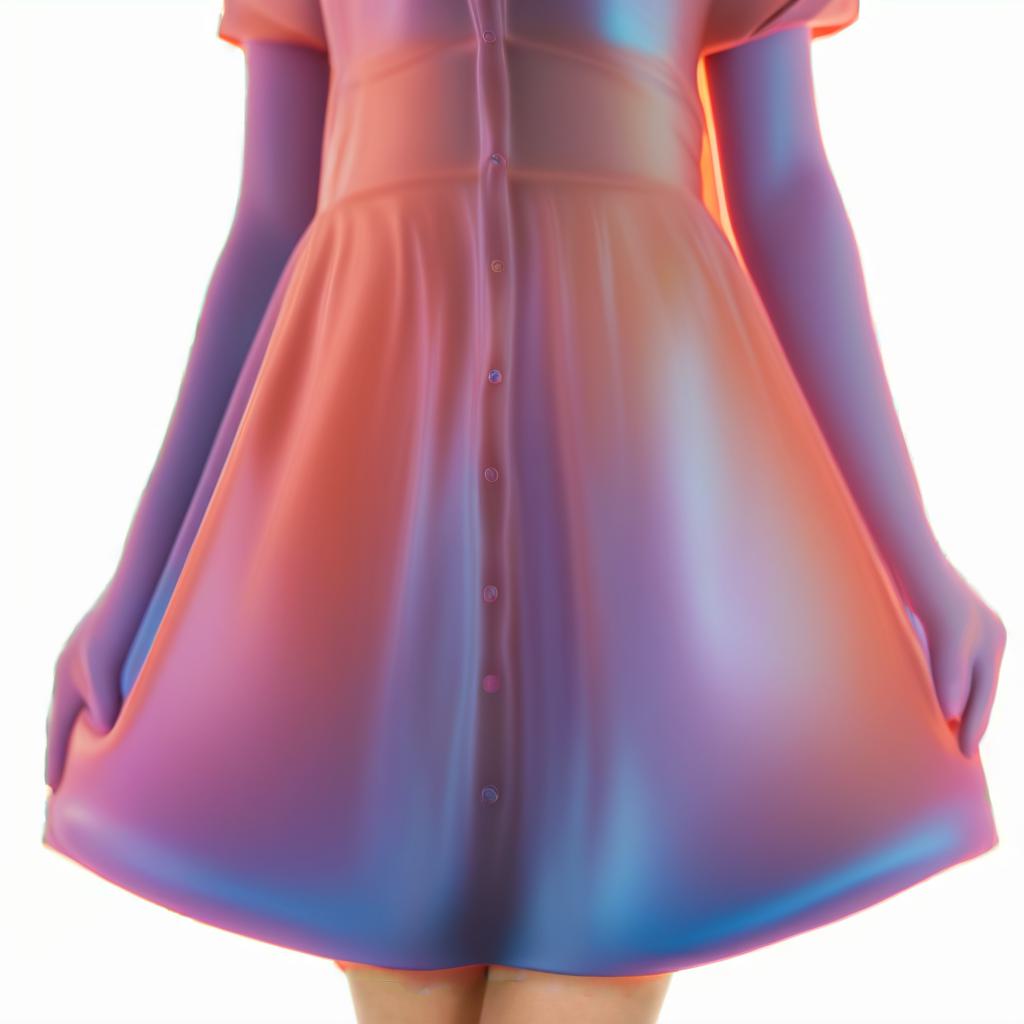} &
        \includegraphics[width=0.09\textwidth]{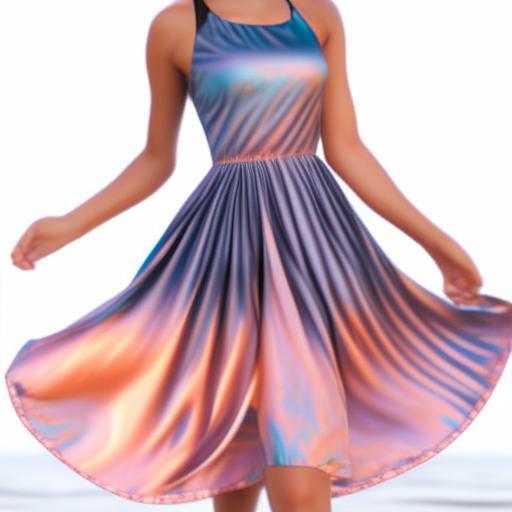} \\

        Object & Texture & VSP &  ZeST & pOps \\[-0.3cm]

    \end{tabular}
    }
    \caption{Qualitative comparison for the \textit{pOps} texturing operator. \\[-0.1cm]}
    \label{fig:texturing_comparisons_small}
\end{figure}

\paragraph{\textbf{Qualitative Comparisons.}}
Next, we evaluate our union and scene operators in comparison to latent space averaging. As can be seen in~\Cref{fig:average_small}, \textit{pOps} applies a consistent operation to the provided inputs, whereas averaging yields outputs with varying semantic meanings. This observation aligns with our expectations that the CLIP embedding space is well-suited for semantic operations but is inconsistent when used na\"{\i}vely.
We proceed to evaluate our texturing and instruct operators by comparing them to relevant literature. In~\Cref{fig:texturing_comparisons_small}, we compare our texturing operator to Visual Style Prompting~\cite{jeong2024visual} and ZeST~\cite{cheng2024zest}. 

\begin{figure}
    \centering
    \setlength{\tabcolsep}{0.5pt}
    {
    \begin{tabular}{c c c c c c c}

        \raisebox{0.1in}{\rotatebox{90}{ Input }} &
        \includegraphics[width=0.0875\textwidth]{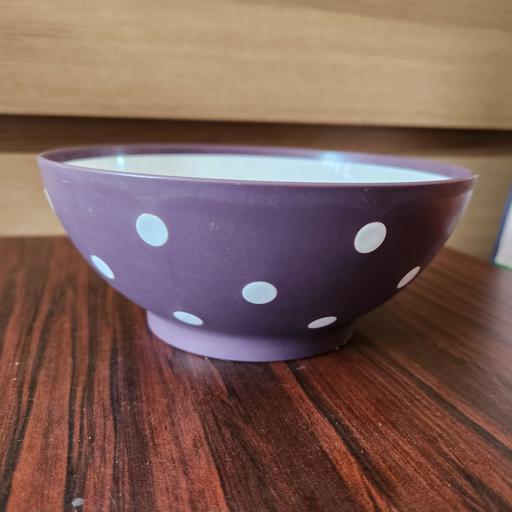} &
        \includegraphics[width=0.0875\textwidth]{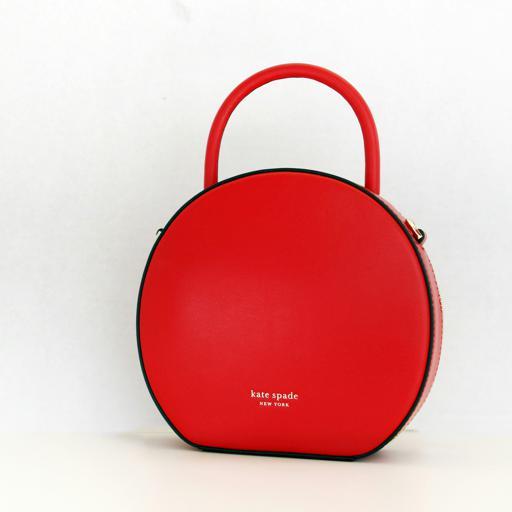} &
        \includegraphics[width=0.0875\textwidth]{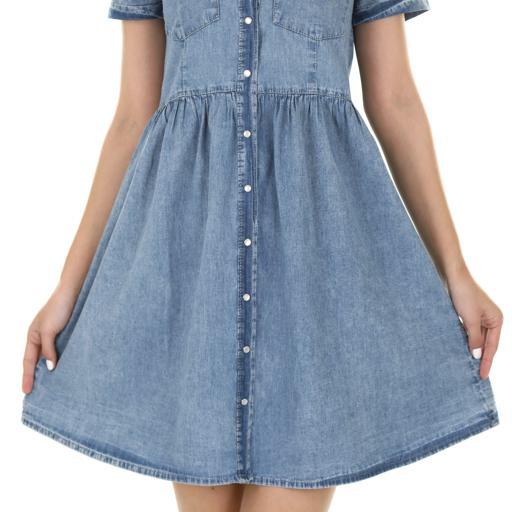} &
        \includegraphics[width=0.0875\textwidth]{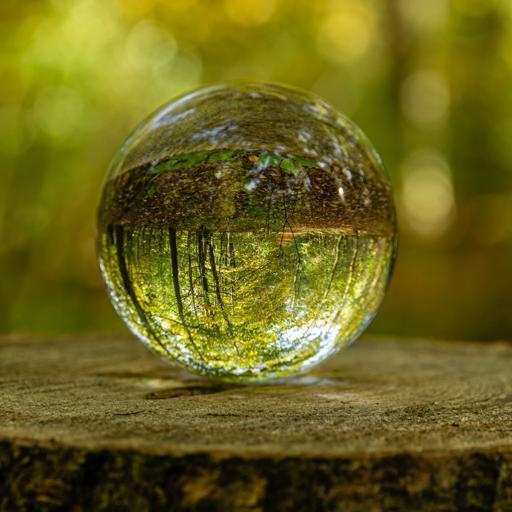} &
        \includegraphics[width=0.0875\textwidth]{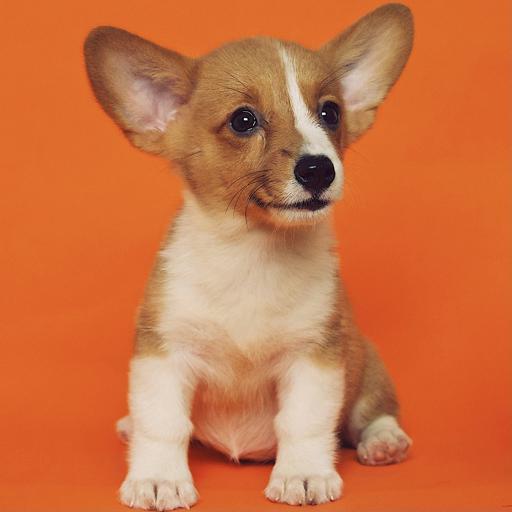} &
        \\

        & 
        ``melting'' & ``shattered'' & ``burning'' & ``many'' & ``muddy'' \\

        \raisebox{-0.03in}{\rotatebox{90}{ InstructP2P }} &
        \includegraphics[width=0.0875\textwidth]{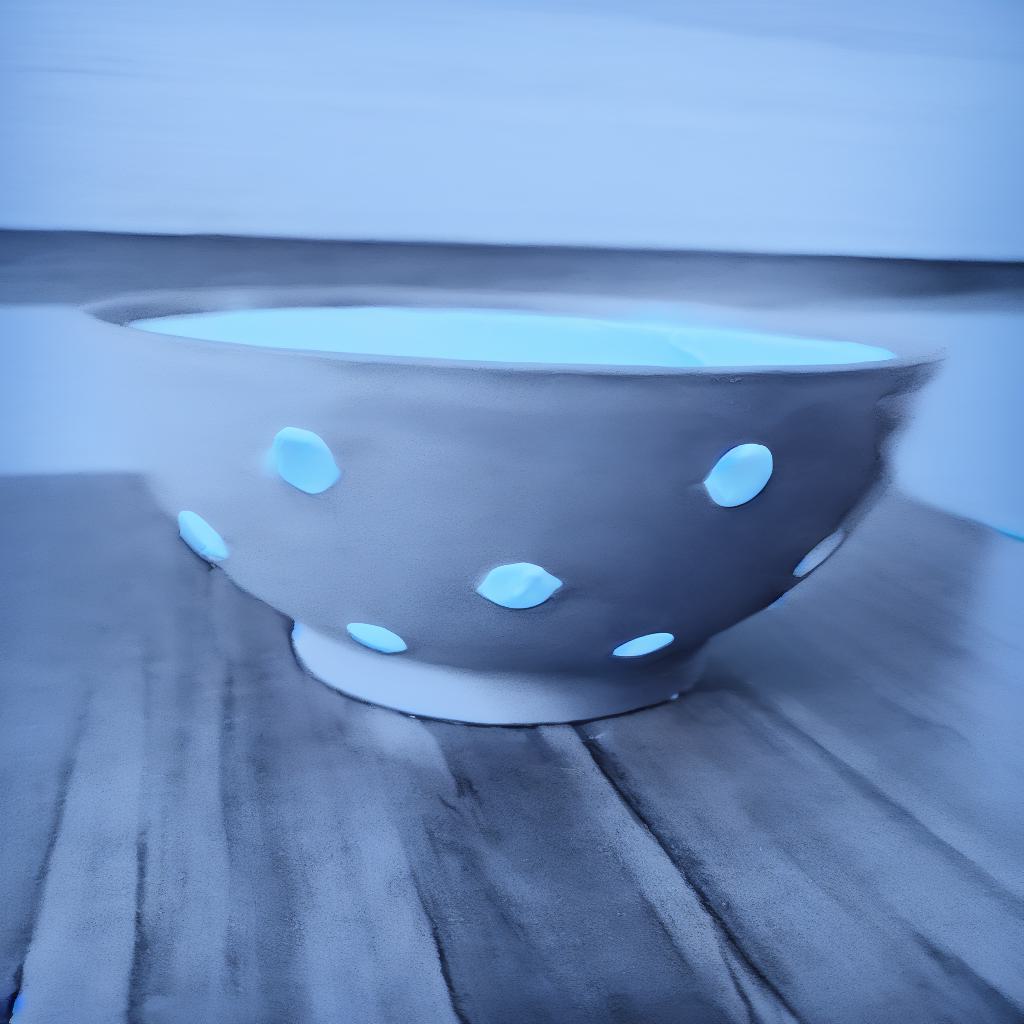} &
        \includegraphics[width=0.0875\textwidth]{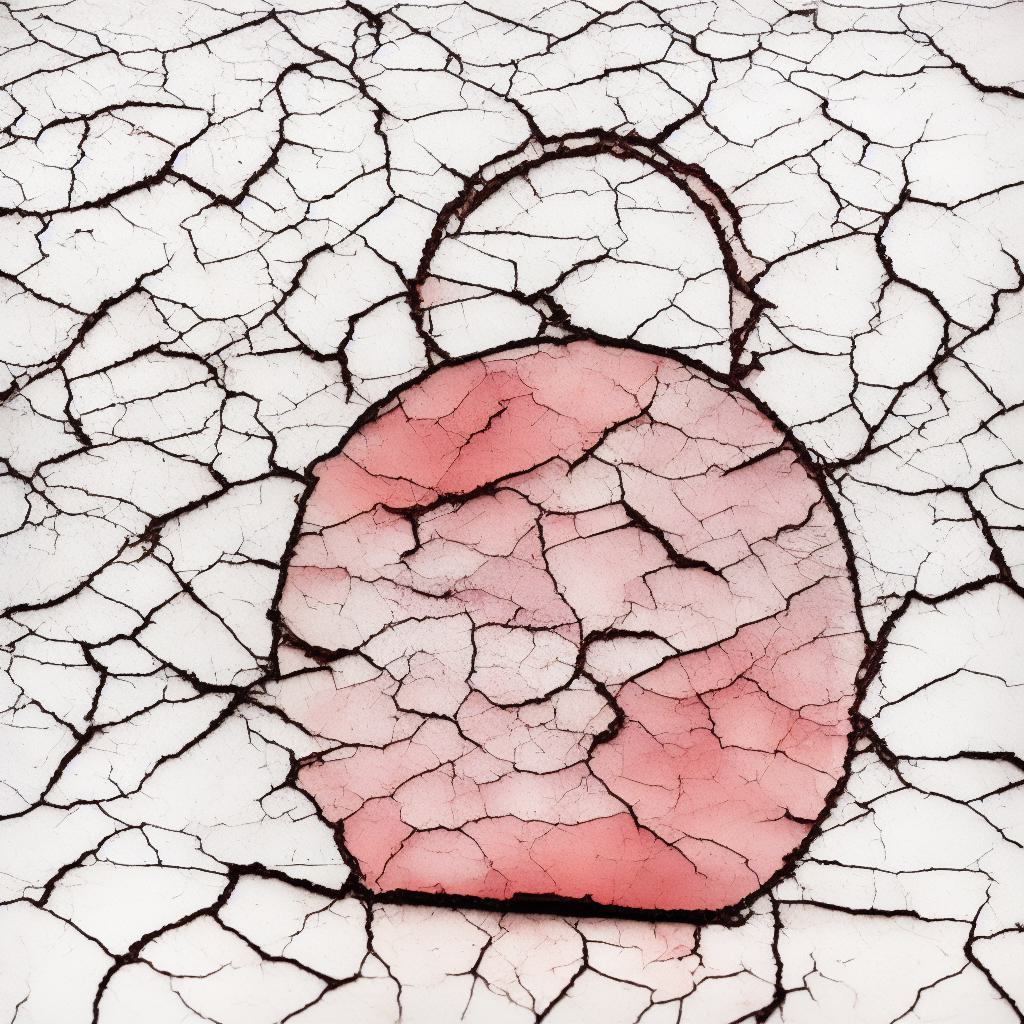} &
        \includegraphics[width=0.0875\textwidth]{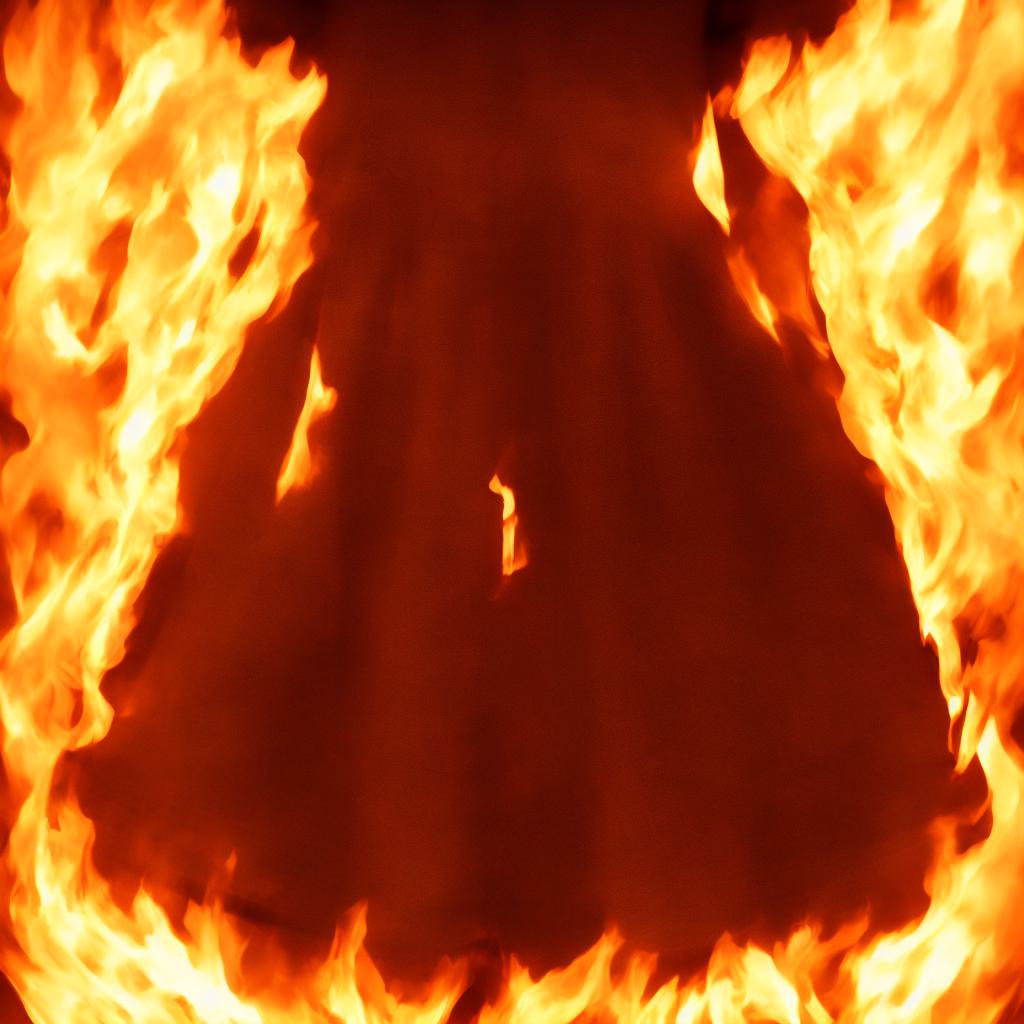} &
        \includegraphics[width=0.0875\textwidth]{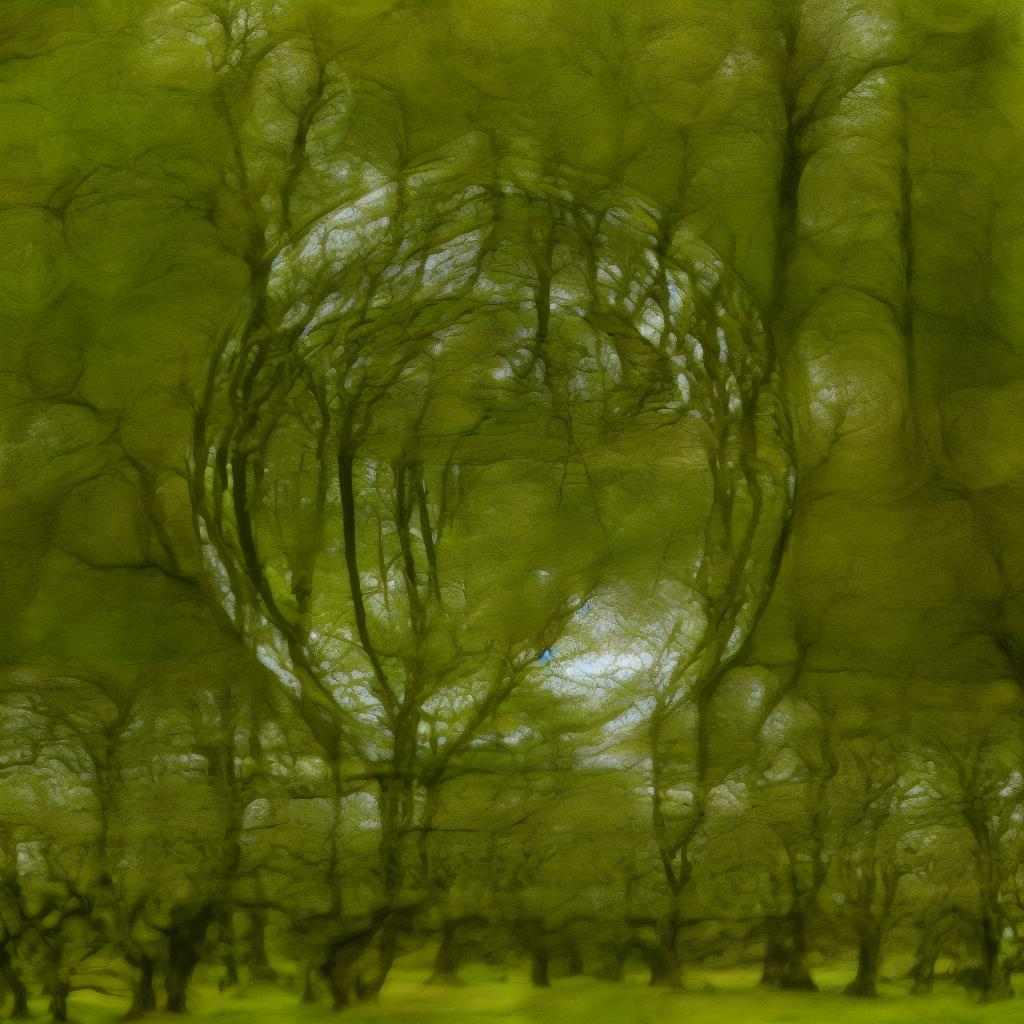} &
        \includegraphics[width=0.0875\textwidth]{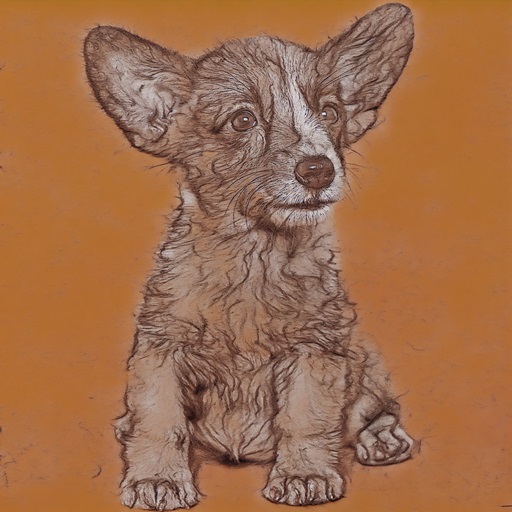} &
        \\

        \raisebox{-0.015in}{\rotatebox{90}{ IP-Adapter }} &
        \includegraphics[width=0.0875\textwidth]{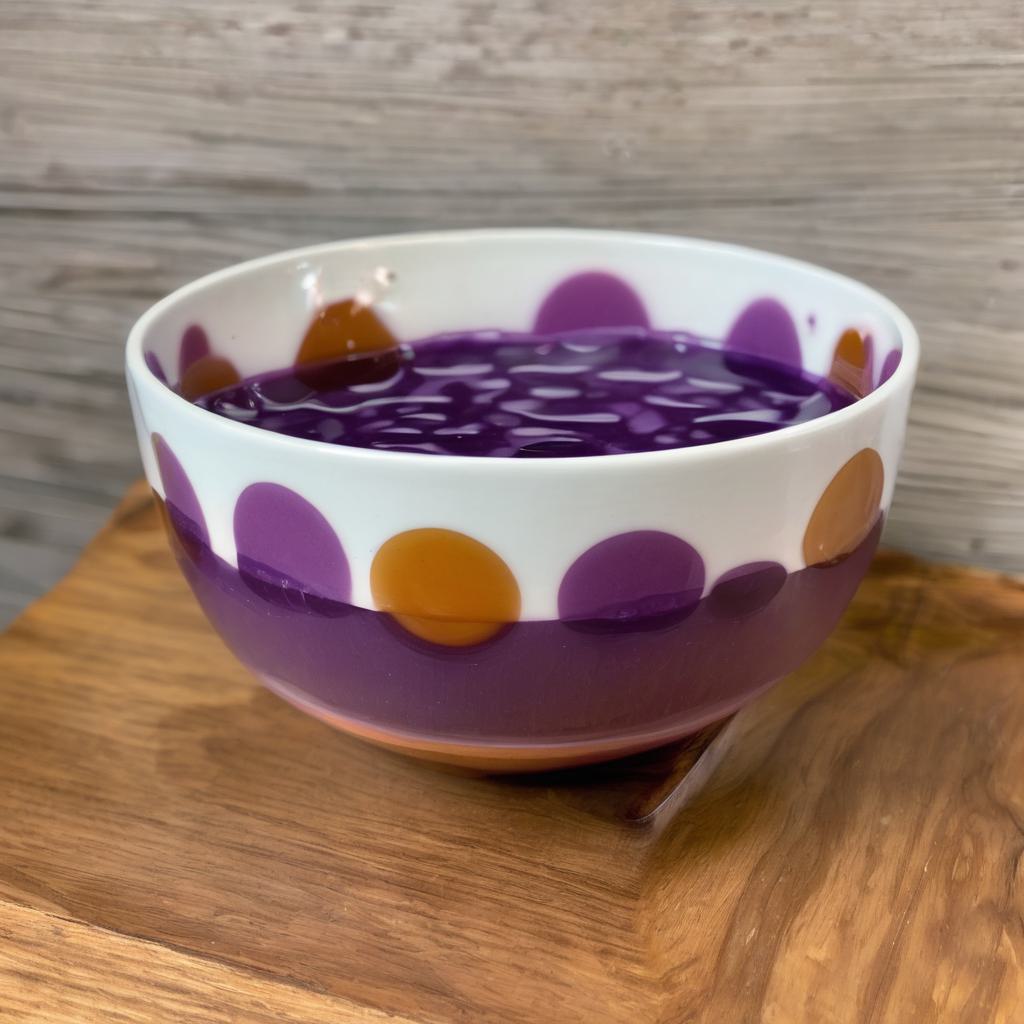} &
        \includegraphics[width=0.0875\textwidth]{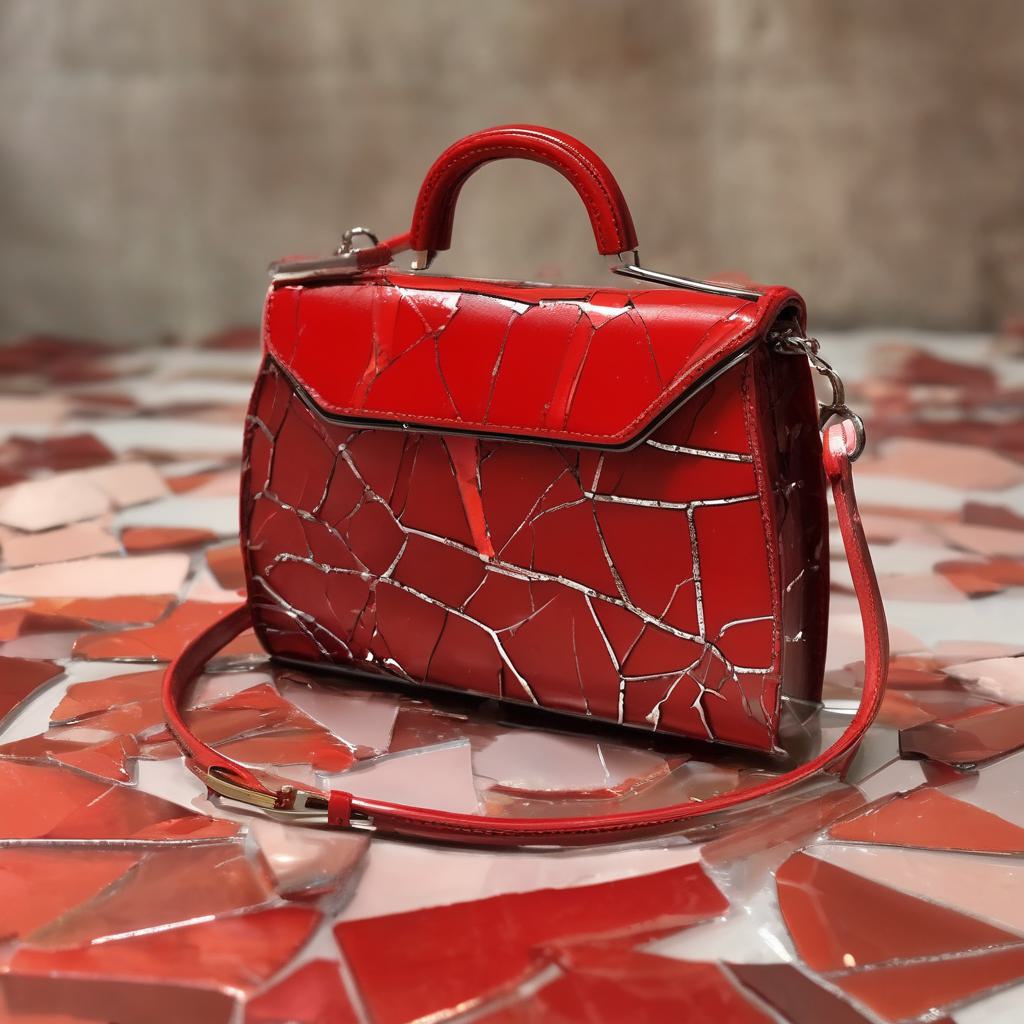} &
        \includegraphics[width=0.0875\textwidth]{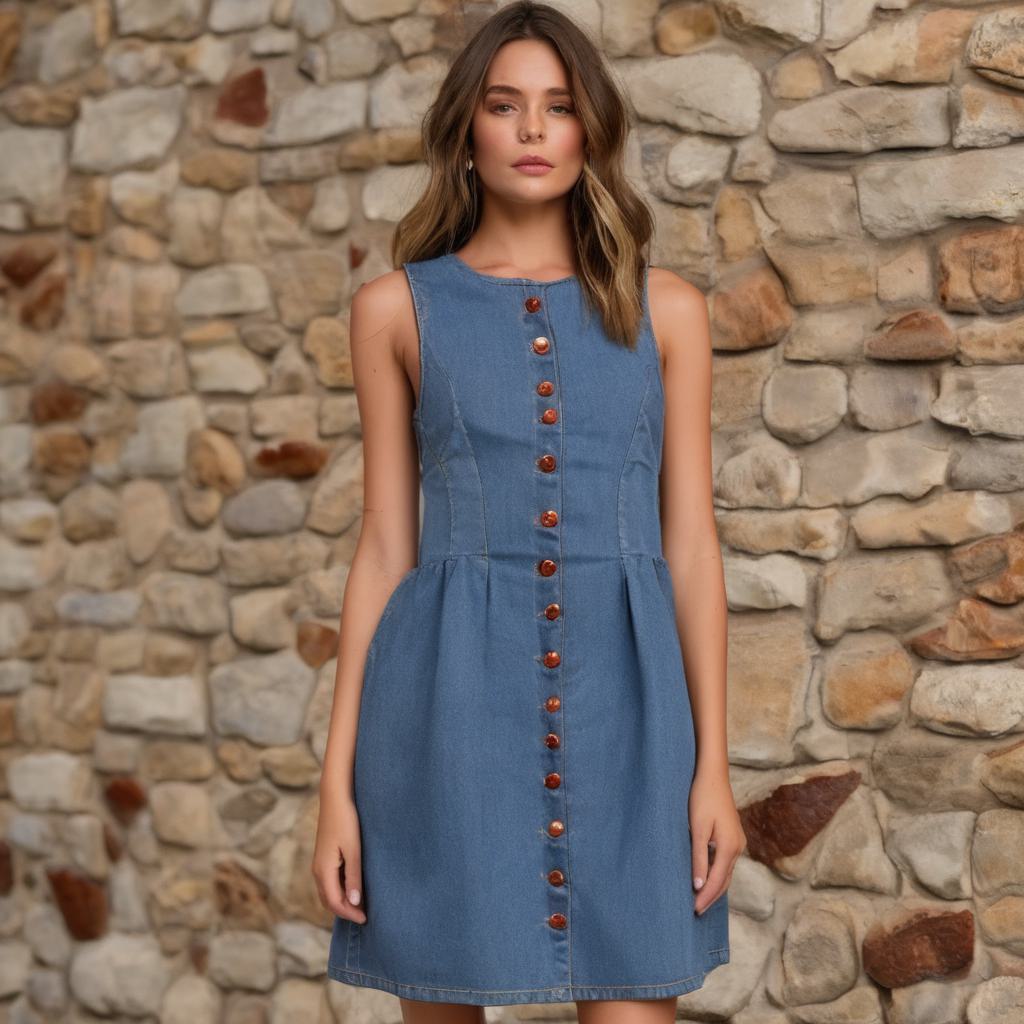} &
        \includegraphics[width=0.0875\textwidth]{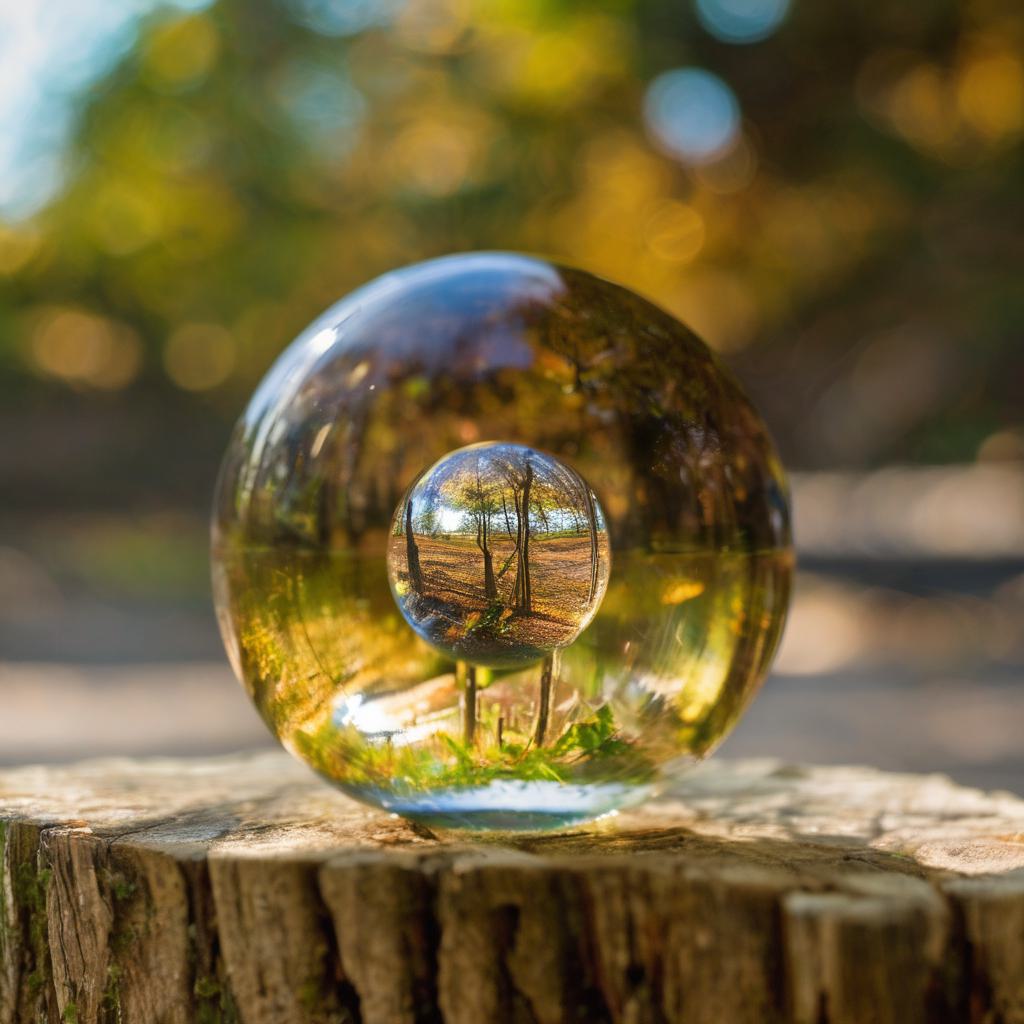} &
        \includegraphics[width=0.0875\textwidth]{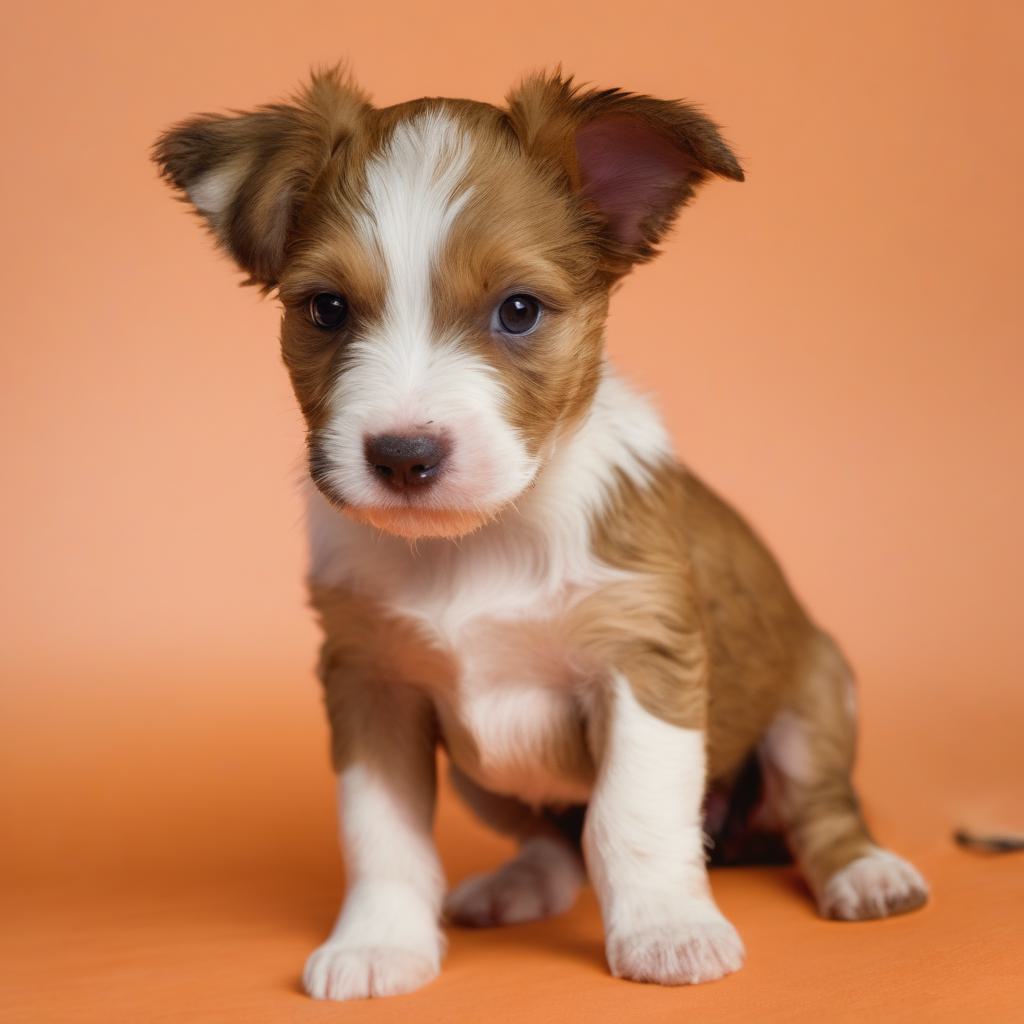} &
        \\

        \raisebox{0.125in}{\rotatebox{90}{ \textit{pOps} }} &
        \includegraphics[width=0.0875\textwidth]{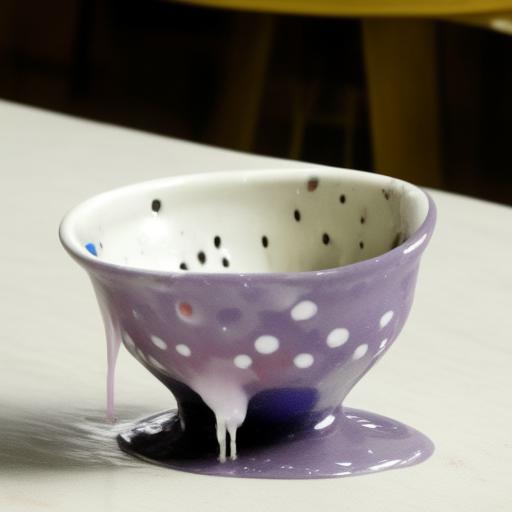} &
        \includegraphics[width=0.0875\textwidth]{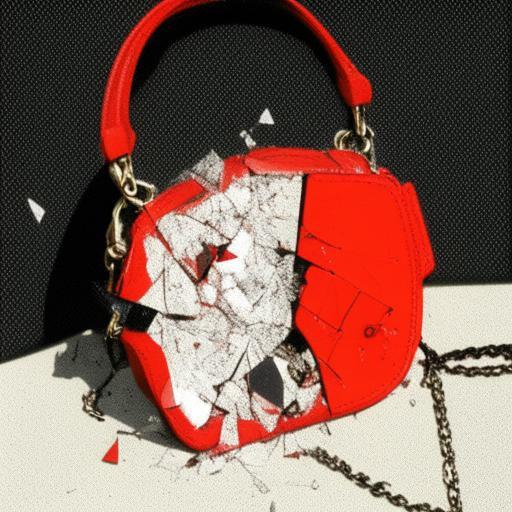} &
        \includegraphics[width=0.0875\textwidth]{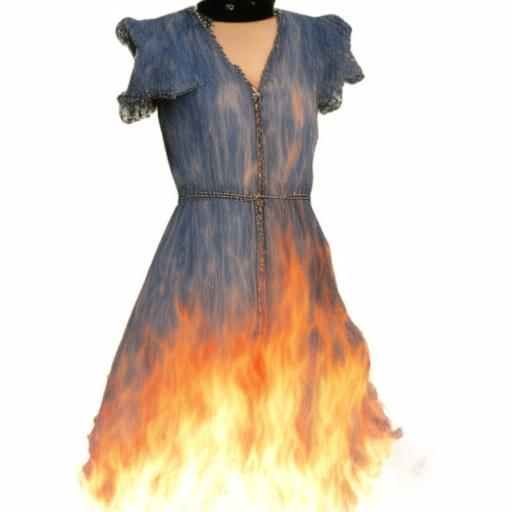} &
        \includegraphics[width=0.0875\textwidth]{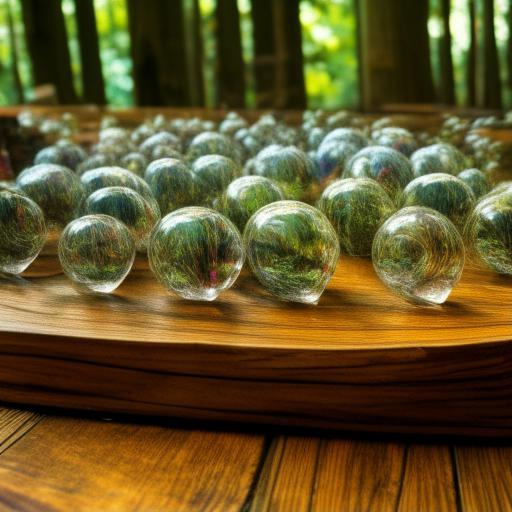} &
        \includegraphics[width=0.0875\textwidth]{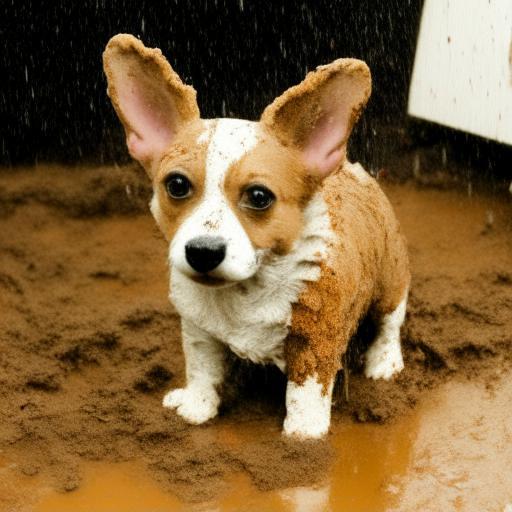} \\

    \end{tabular} 

    }
    \vspace{-0.3cm}
    \caption{Qualitative comparison for the instruct operator to existing approaches: InstructPix2Pix~\cite{brooks2022instructpix2pix} \& IP-Adapter~\cite{ye2023ip-adapter}. \\[-0.2cm]}
    \label{fig:instruct_comparisons_small}
\end{figure}

Similarly, in~\Cref{fig:instruct_comparisons_small}, we compare our instruct operator to InstructPix2Pix~\cite{brooks2022instructpix2pix} and IP-Adapter~\cite{ye2023ip-adapter}. 
Note that \textit{pOps} has seen the instructions during training, but without direct supervision that was used in InstructPix2Pix. Comparisons to additional baselines can be found in~\Cref{sec:additional_comparisons}.

\paragraph{\textbf{Quantitative Comparisons.}}
We conduct two forms of quantitative evaluation to validate the effectiveness of our approach. In~\Cref{tb:instruct_quantitative}, we utilize image and text similarity metrics to compare our instruct operator to InstructPix2Pix and IP-Adapter. One can see that our method attains higher image similarity than IP-Adapter with a scale of $0.1$ while still retaining high text similarity values.
Next, we perform a user study for the instruct and texturing tasks alongside their alternatives. The results in~\Cref{tb:user_study} demonstrate that \textit{pOps} compares favorably to the recent state-of-the-art in both tasks.

\vspace{-0.1cm}
\paragraph{\textbf{Analysis.}}
As discussed, the image generation process in \textit{pOps} is independent of the trainable operator itself. 
Therefore, we have the flexibility to employ any compatible image generation model that can be conditioned on our CLIP image embeddings.
While our model of choice was Kandinsky 2~\cite{kandinsky2}, in~\Cref{fig:ip_adapter_small} we show that our method is also compatible with IP-Adapter without requiring any modification or tuning. 
This compatibility enables us to leverage a diverse range of models supported by IP-Adapter, including a depth-conditioned ControlNet.

\begin{table}
\small
\centering
\setlength{\tabcolsep}{2.5pt}
\caption{Quantitative Comparison for the Instruct Operator. Image similarity is computed with DreamSim~\cite{fu2023dreamsim} and text similarity with CLIP ViT-L/14. Results are averaged across $52$ objects and $65$ adjectives.\\[-0.65cm]} 
\begin{tabular}{l c c c c} 
    \toprule
    Method & Image Similarity $\uparrow$ & Text Similarity $\uparrow$ & BERT Similarity $\uparrow$ \\
    \midrule
    InstructPix2Pix  & $0.455$ & $0.237$ & $0.424$ \\
    IP-Adapter (0.5) & $0.826$ & $0.211$ & $0.544$ \\
    IP-Adapter (0.1) & $0.584$ & $0.219$ & $0.531$ \\
    \midrule
    pOps             & $0.6607$ & $0.236$ & $0.437$ \\
    \bottomrule
\end{tabular}
\vspace{-0.1cm}
\label{tb:instruct_quantitative}
\end{table}

\begin{table}
\small
\centering
\setlength{\tabcolsep}{2.5pt}
\caption{User study results for the instruct and texturing operators.
\\[-0.3cm]} 
\begin{tabular}{l c c c c} 
    \toprule
    \multicolumn{5}{c}{Instruct} \\
    \midrule
    Metric & InstructP2P & IP-Adapter & IP-Adapter (0.1) & \textit{pOps} \\
    \midrule
    Percent Preferred $\uparrow$ & $23.81\%$ & $3.18\%$ & $12.70\%$ & $\textbf{60.31\%}$ \\
    Average Rating $\uparrow$    & $1.65$    & $1.95$   & $2.75$    & $\textbf{3.49}$    \\
    \toprule
    \multicolumn{5}{c}{Texturing} \\
    \midrule
    Metric & IP-Adapter & VSP & ZeST & \textit{pOps} \\
    \midrule
    Percent Preferred $\uparrow$ & $3.57\%$ & $1.79\%$ & $37.50\%$ & $\textbf{57.14\%}$ \\
    Average Rating $\uparrow$    & $1.45$   & $2.21$   & $3.66$    & $\textbf{3.98}$    \\
    \bottomrule
\end{tabular}
\label{tb:user_study}
\end{table}

Finally, since \textit{pOps} employs a diffusion model to generate image embeddings, we can sample different seeds for the same input conditions. Interestingly, in~\Cref{fig:null_small}, we demonstrate that when providing only a single input to the texturing operator, the model can sample diverse and plausible results based on the given input. Additional examples for both analyses are provided in~\Cref{sec:additional_results}.

\begin{figure}
    \centering
    \setlength{\tabcolsep}{0.5pt}
    \addtolength{\belowcaptionskip}{-5pt}
    {
    \begin{tabular}{c c @{\hspace{0.2cm}} c c c}

        \includegraphics[width=0.0875\textwidth]{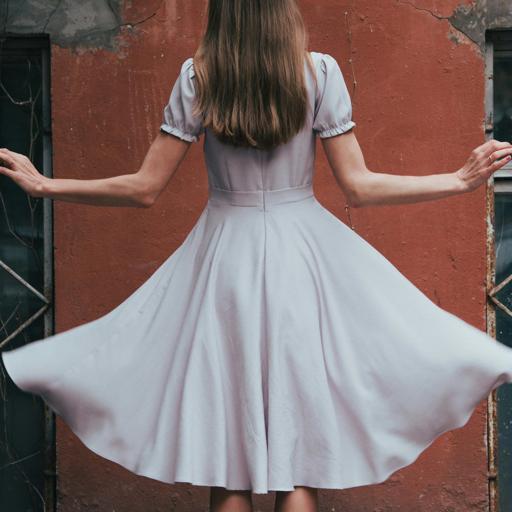} &
        \includegraphics[width=0.0875\textwidth]{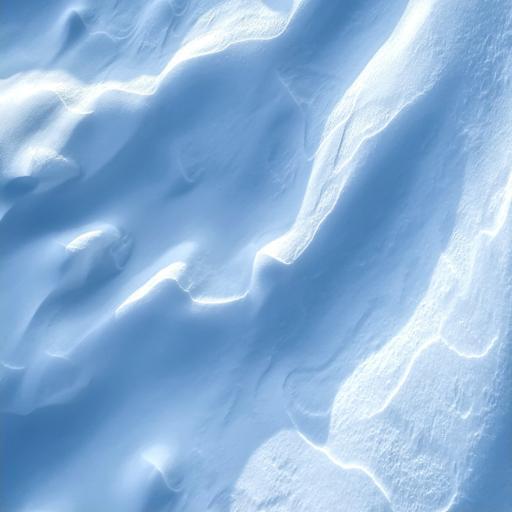} &
        \includegraphics[width=0.0875\textwidth]{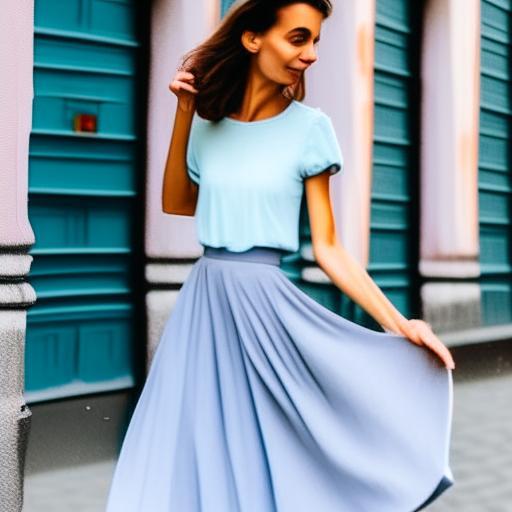} &
        \includegraphics[width=0.0875\textwidth]{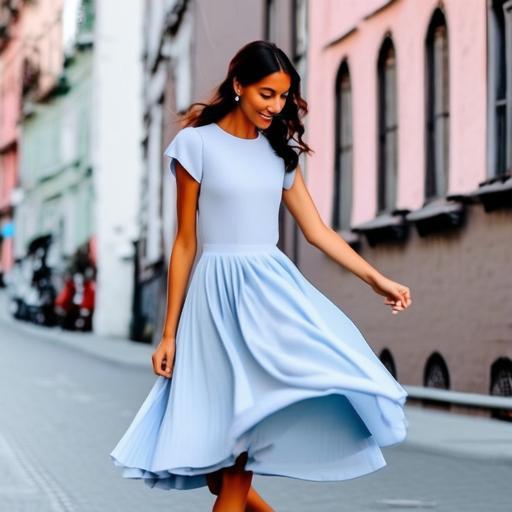} &
        \includegraphics[width=0.0875\textwidth]{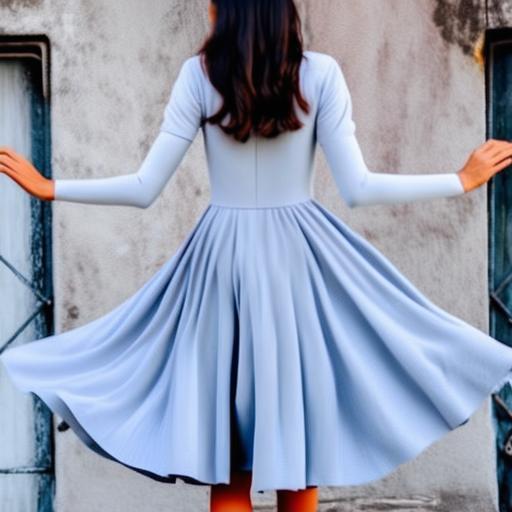} \\

        \includegraphics[width=0.0875\textwidth]{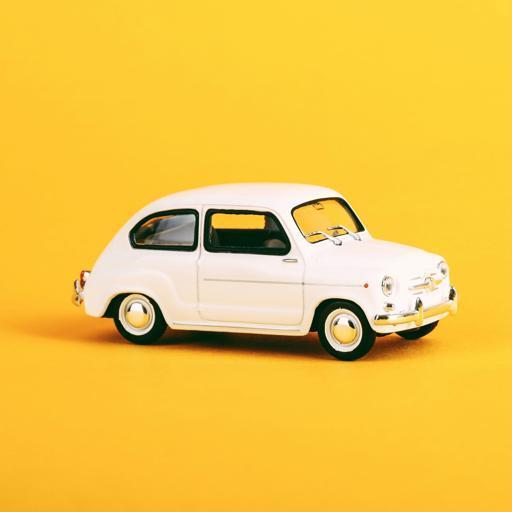} &
        \includegraphics[width=0.0875\textwidth]{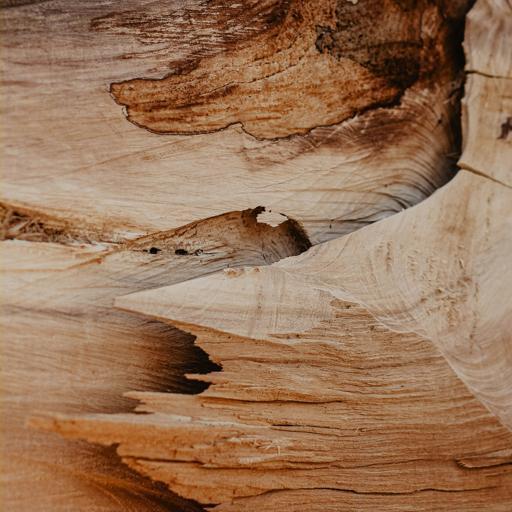} &
        \includegraphics[width=0.0875\textwidth]{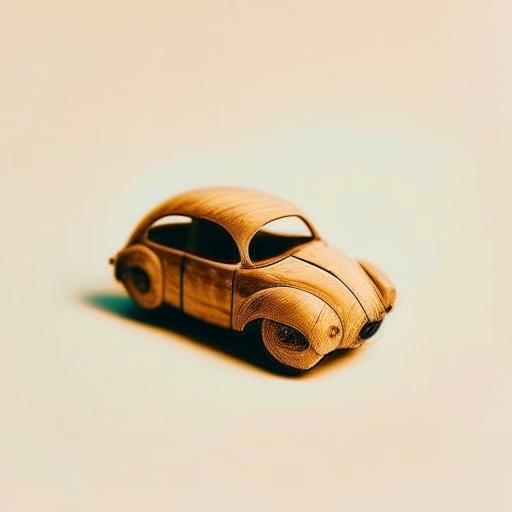} &
        \includegraphics[width=0.0875\textwidth]{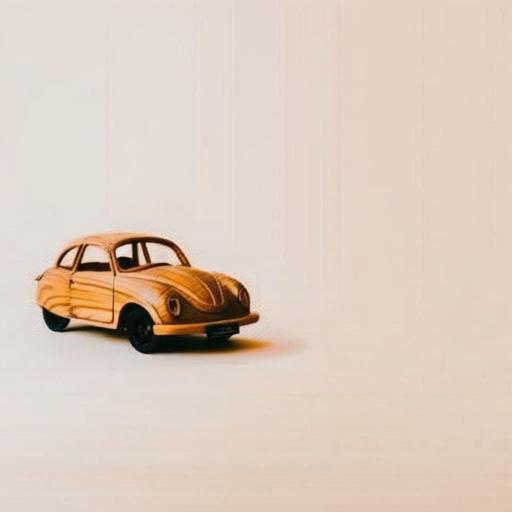} &
        \includegraphics[width=0.0875\textwidth]{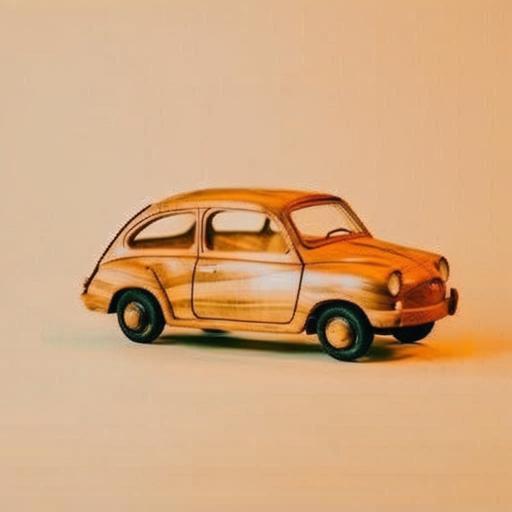} \\

        \includegraphics[width=0.0875\textwidth]{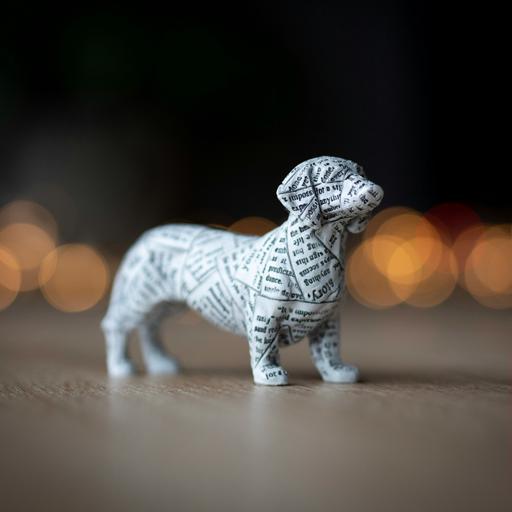} &
        \includegraphics[width=0.0875\textwidth]{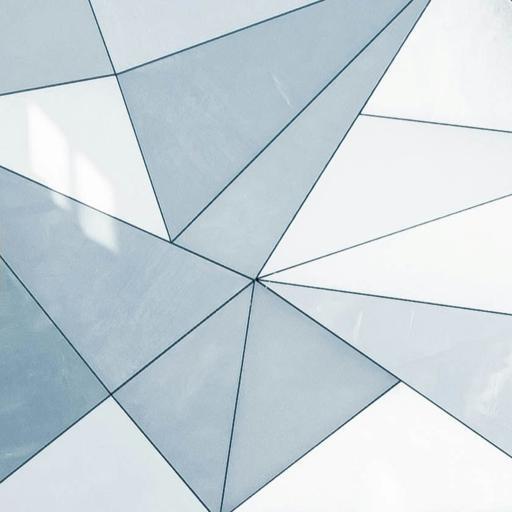} &
        \includegraphics[width=0.0875\textwidth]{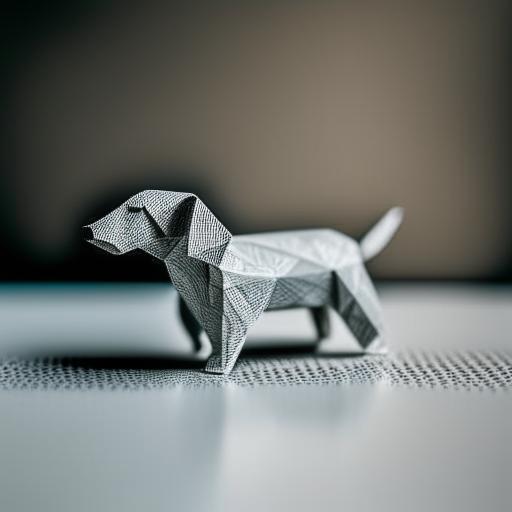} &
        \includegraphics[width=0.0875\textwidth]{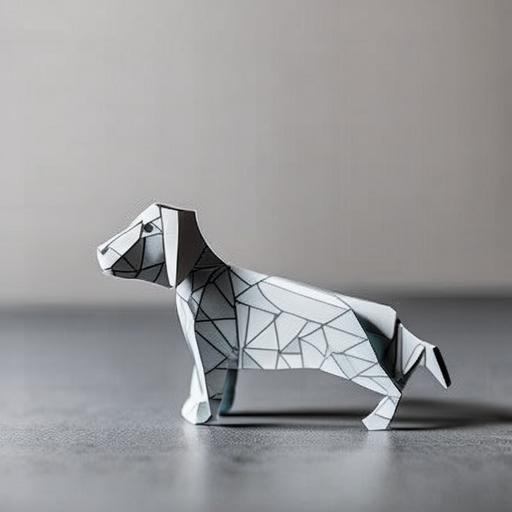} &
        \includegraphics[width=0.0875\textwidth]{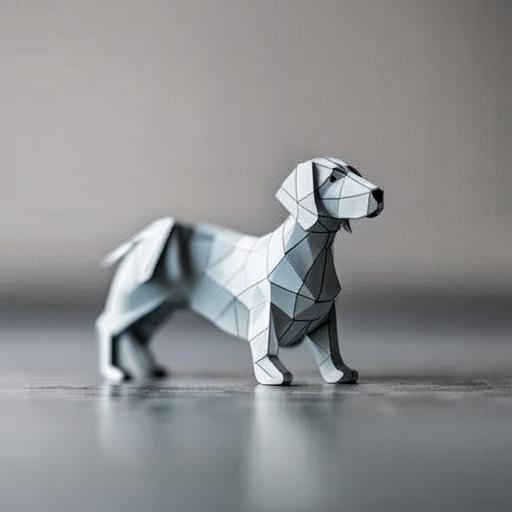} \\

        Object & Texture & Kandinsky & IP-Adapter & +Depth

    \end{tabular}
    }
    \vspace{-0.3cm}
    \caption{\textbf{Different Renderers}. \textit{pOps} outputs can be directly fed to either Kandinsky or IP-Adapter and incorporated alongside spatial conditions 
    }
    \label{fig:ip_adapter_small}
\end{figure}

\begin{figure}
    \centering
    \setlength{\tabcolsep}{0.5pt}
    \addtolength{\belowcaptionskip}{-5pt}
    {
    \begin{tabular}{c @{\hspace{0.2cm}} c c c c}

        \includegraphics[width=0.0875\textwidth]{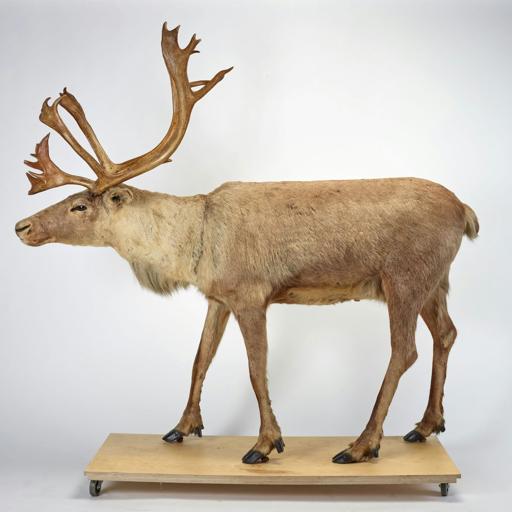} &
        \includegraphics[width=0.0875\textwidth]{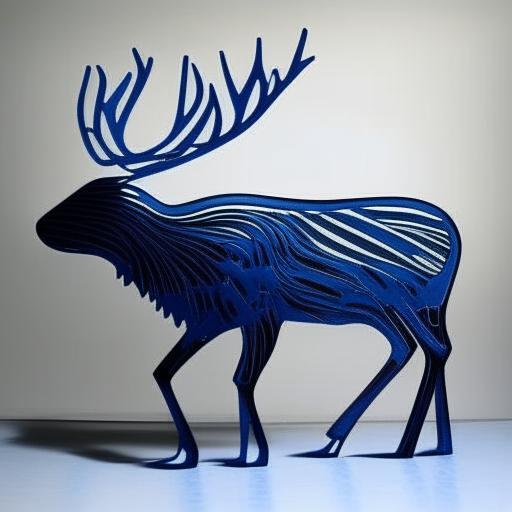} &
        \includegraphics[width=0.0875\textwidth]{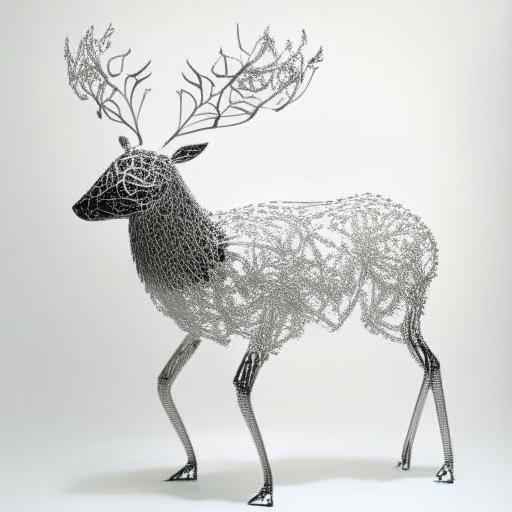} &
        \includegraphics[width=0.0875\textwidth]{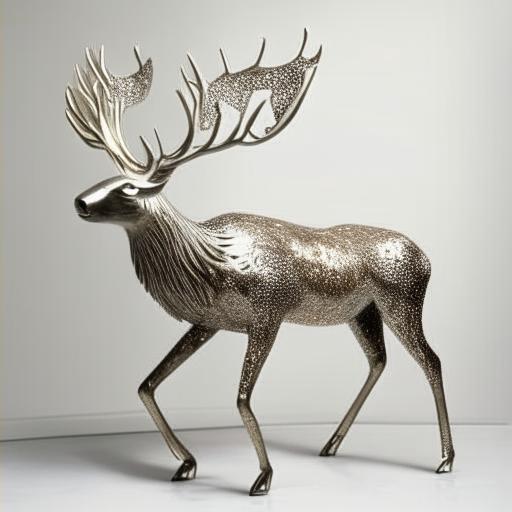} &
        \includegraphics[width=0.0875\textwidth]{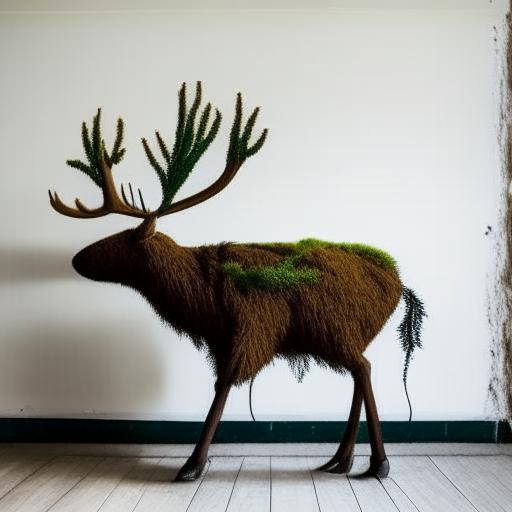} \\

        \includegraphics[width=0.0875\textwidth]{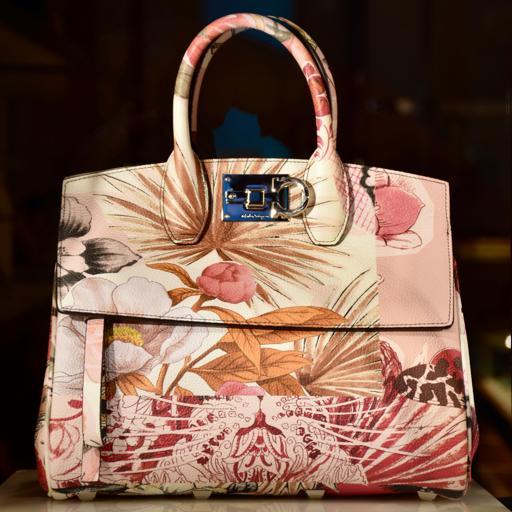} &
        \includegraphics[width=0.0875\textwidth]{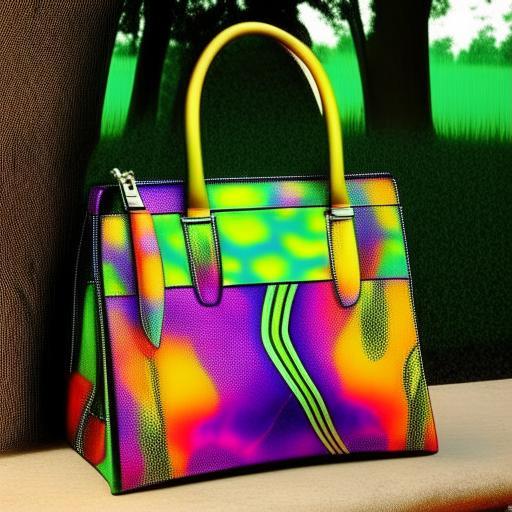} &
        \includegraphics[width=0.0875\textwidth]{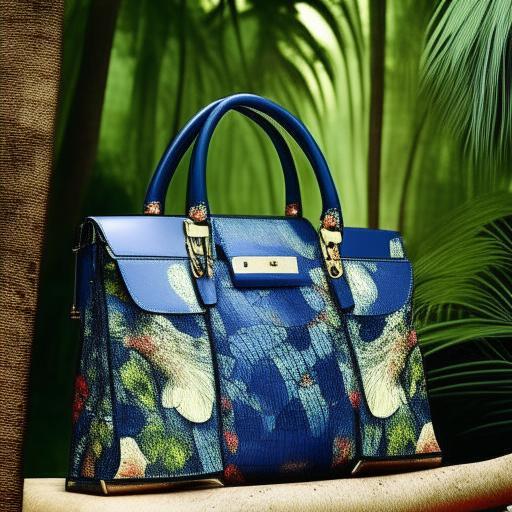} &
        \includegraphics[width=0.0875\textwidth]{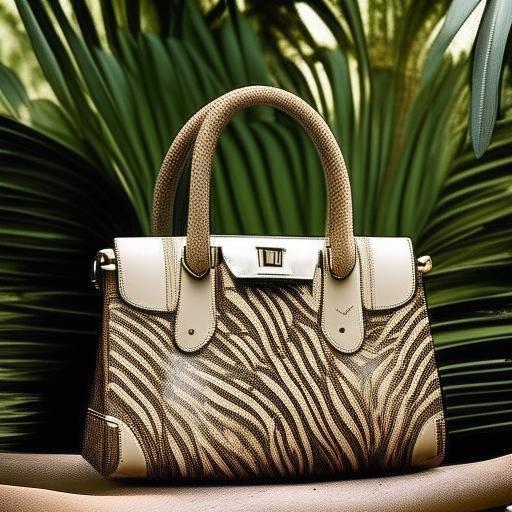} &
        \includegraphics[width=0.0875\textwidth]{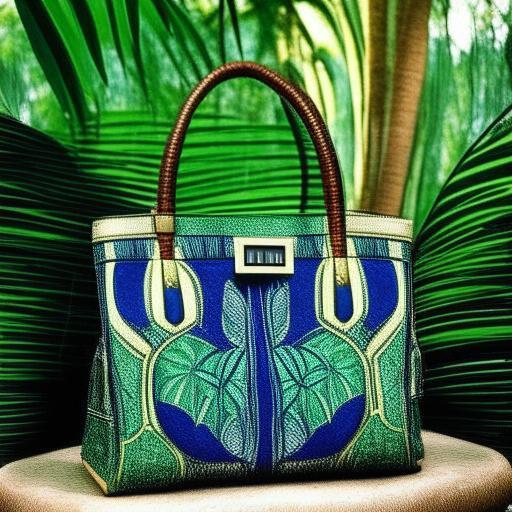} \\

        Object & \multicolumn{4}{c}{Sampled Textures} \\

        \includegraphics[width=0.0875\textwidth]{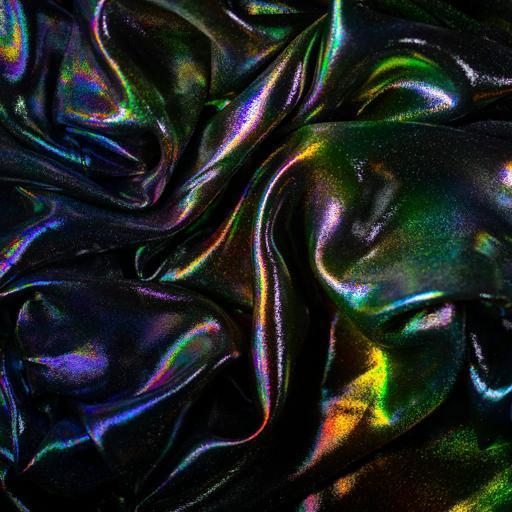} &
        \includegraphics[width=0.0875\textwidth]{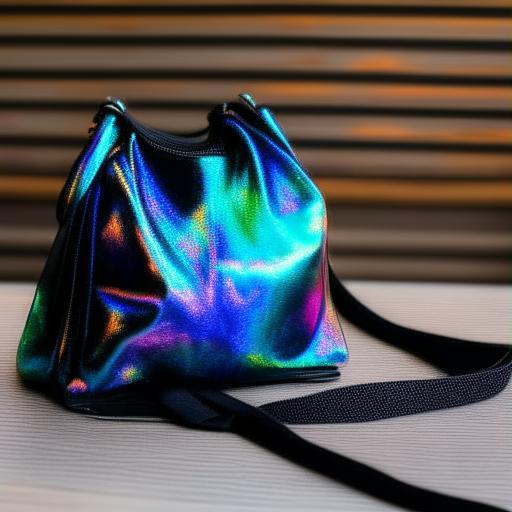} &
        \includegraphics[width=0.0875\textwidth]{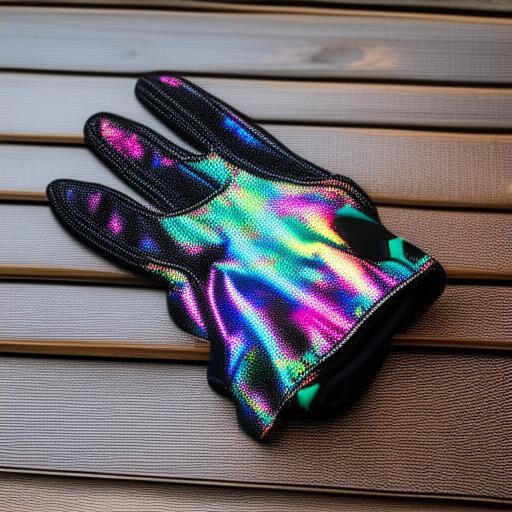} &
        \includegraphics[width=0.0875\textwidth]{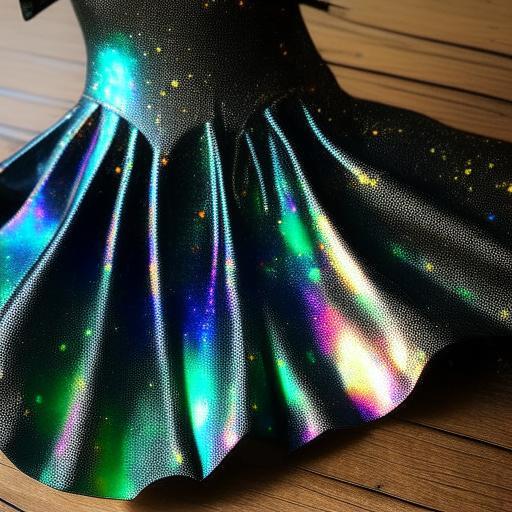} &
        \includegraphics[width=0.0875\textwidth]{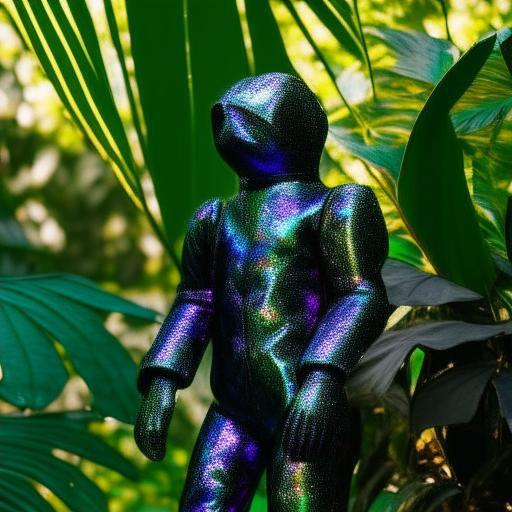} \\

        \includegraphics[width=0.0875\textwidth]{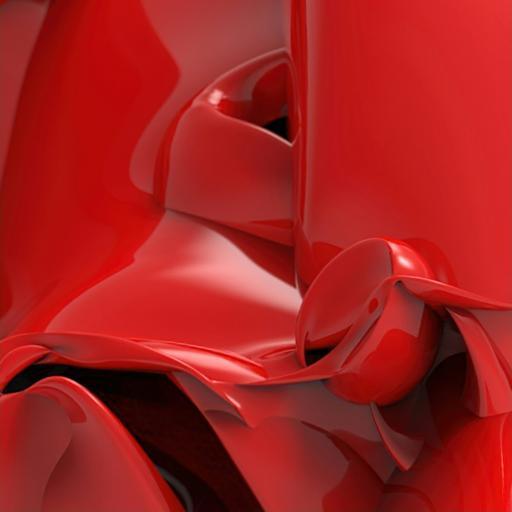} &
        \includegraphics[width=0.0875\textwidth]{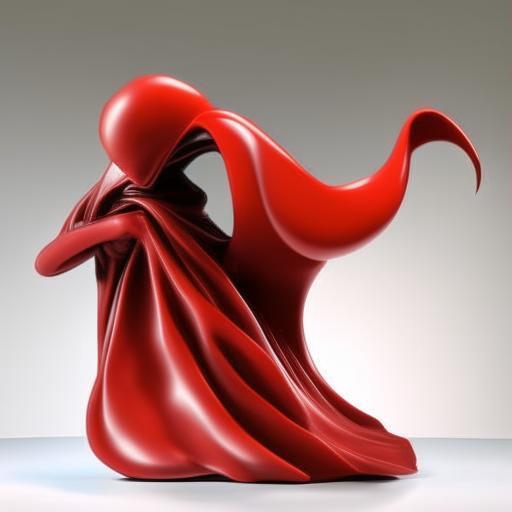} &
        \includegraphics[width=0.0875\textwidth]{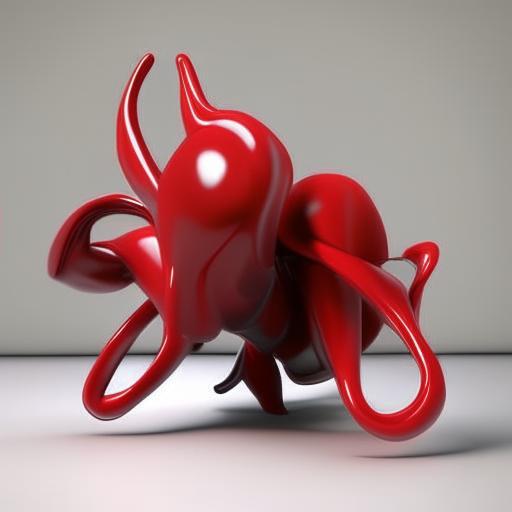} &
        \includegraphics[width=0.0875\textwidth]{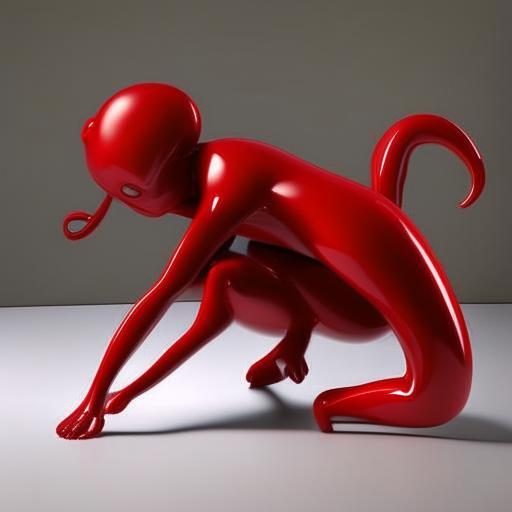} &
        \includegraphics[width=0.0875\textwidth]{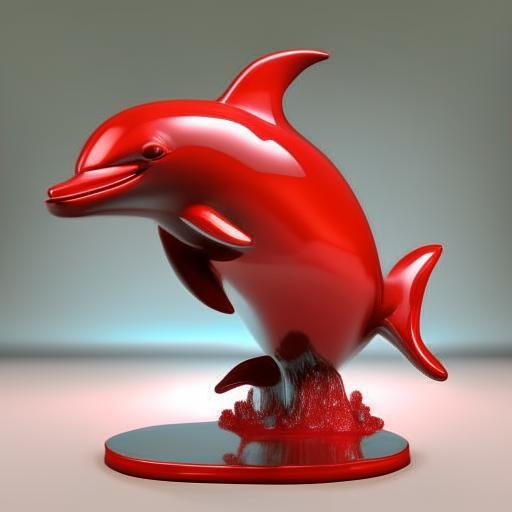} \\

        Texture & \multicolumn{4}{c}{Sampled Objects} \\[-0.35cm]

    \end{tabular}
    }
    \caption{\textbf{Sampling from missing inputs}. Given only an object or a texture, the \textit{pOps} texturing operator can successfully sample diverse textured objects.
    }
    \label{fig:null_small}
\end{figure}

\section{Limitations}
\begin{figure}
    \centering
    \setlength{\tabcolsep}{0.5pt}
    \renewcommand{\arraystretch}{1}
    \newcommand{\pl}{0.2}
    {\small
    \begin{tabular}{cc  @{\hspace{0.4cm}} ccc}

    \multicolumn{2}{c}{Reconstruction Results} & \multicolumn{3}{c}{Operator Results} \\

    \includegraphics[width=0.09\textwidth]{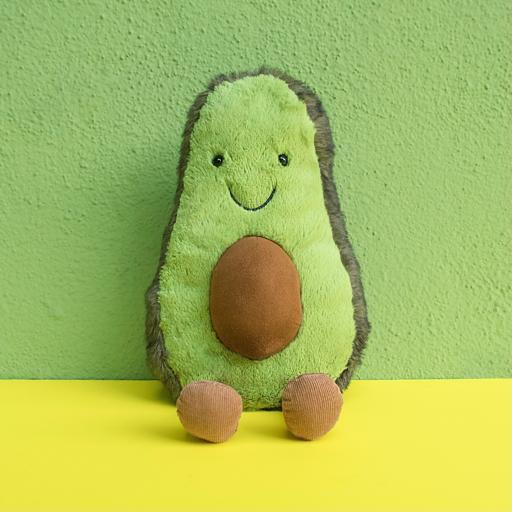} & 
    \includegraphics[width=0.09\textwidth]{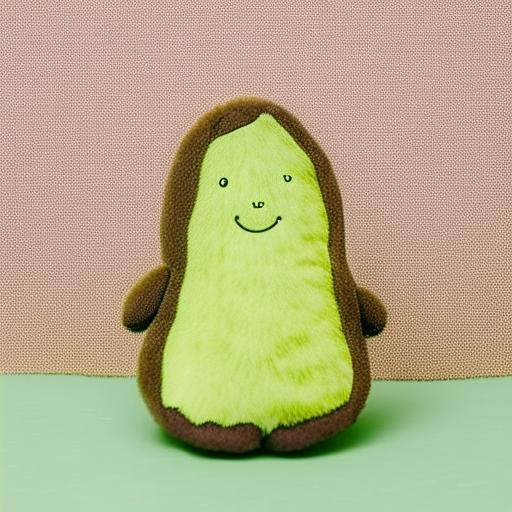} & 
    \includegraphics[width=0.09\textwidth]{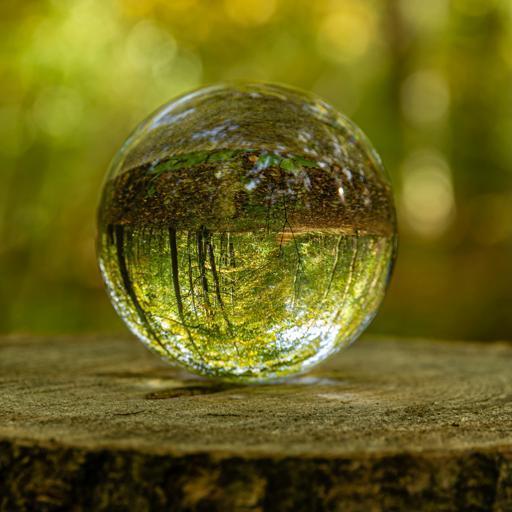} & 
    \includegraphics[width=0.09\textwidth]{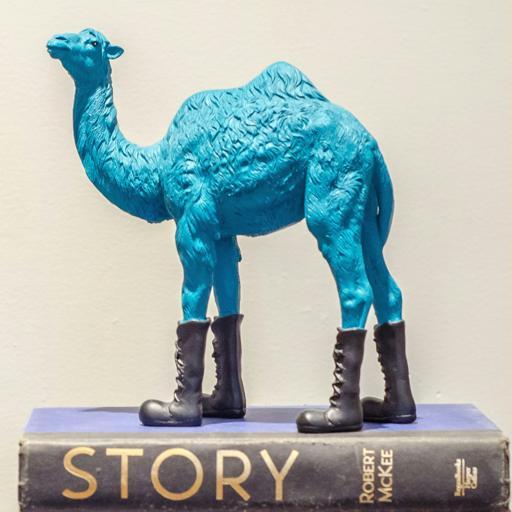} & 
    \includegraphics[width=0.09\textwidth]{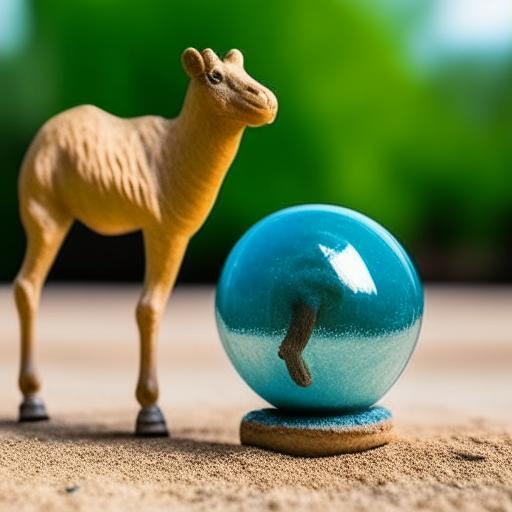} 
     \\

    \includegraphics[width=0.09\textwidth]{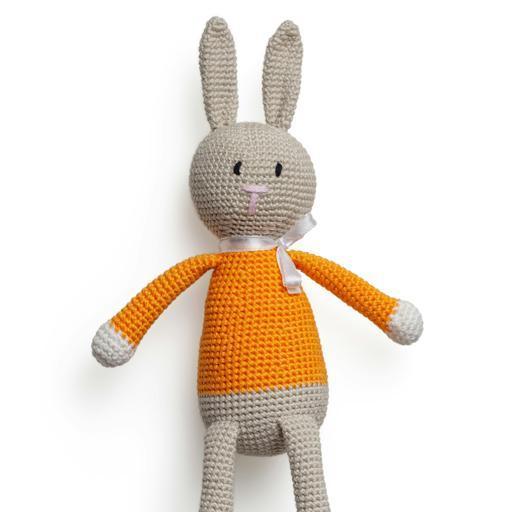} & 
    \includegraphics[width=0.09\textwidth]{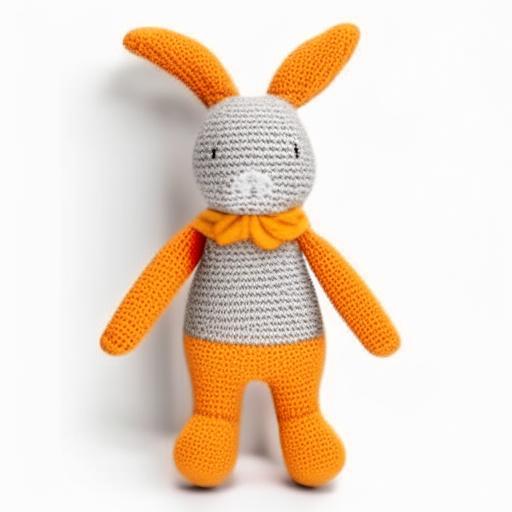} & 
        \includegraphics[width=0.09\textwidth]{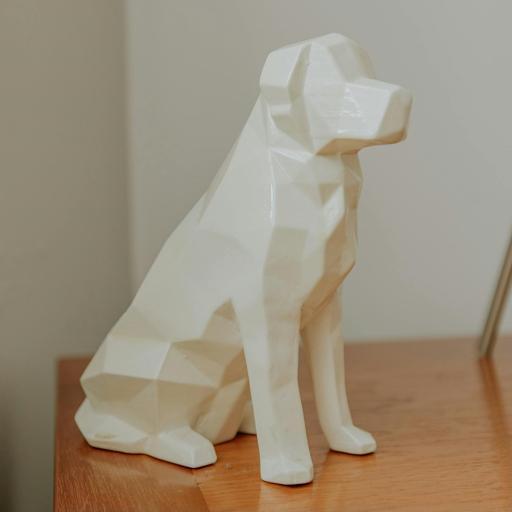} & 
    \includegraphics[width=0.09\textwidth]{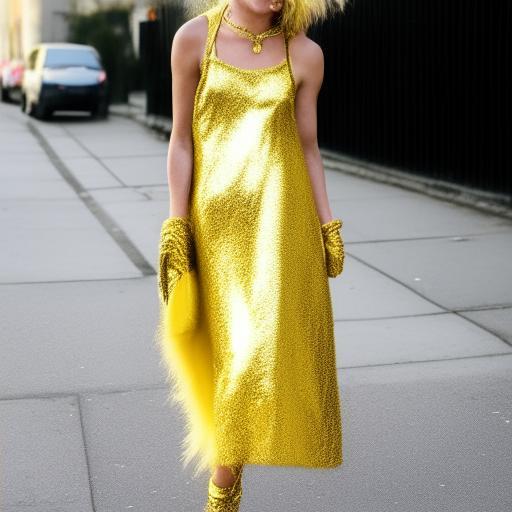} & 
    \includegraphics[width=0.09\textwidth]{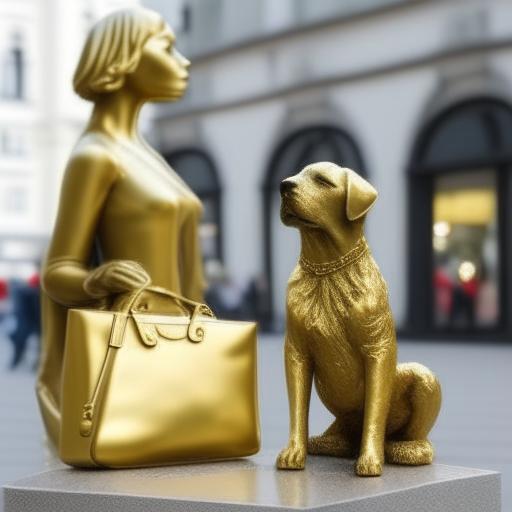} \\

    \includegraphics[width=0.09\textwidth]{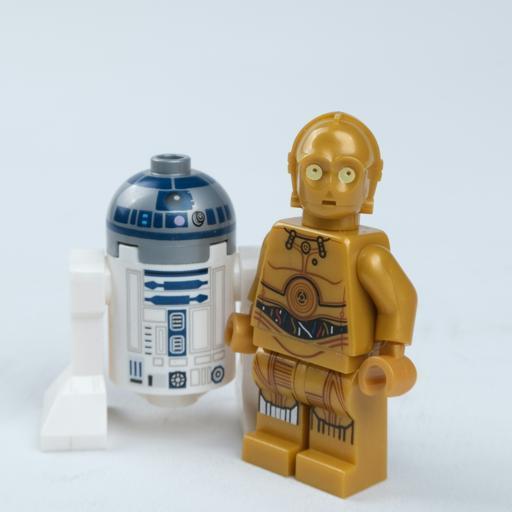} & 
    \includegraphics[width=0.09\textwidth]{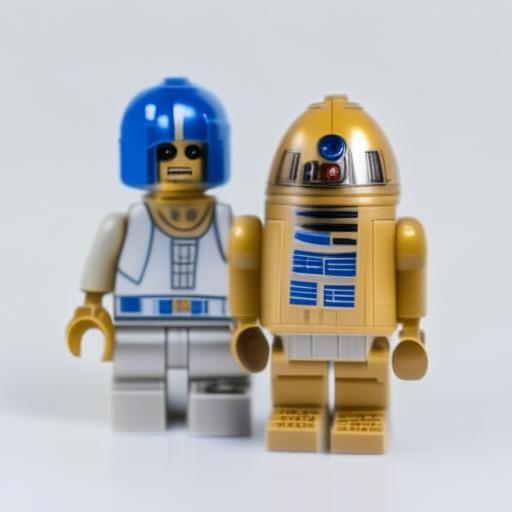} & 
    \includegraphics[width=0.09\textwidth]{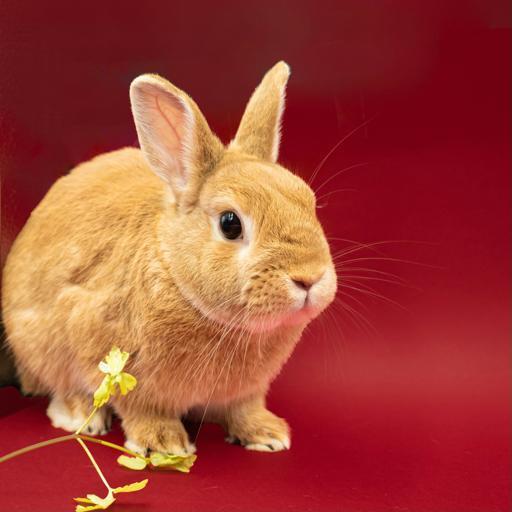} & 
    \includegraphics[width=0.09\textwidth]{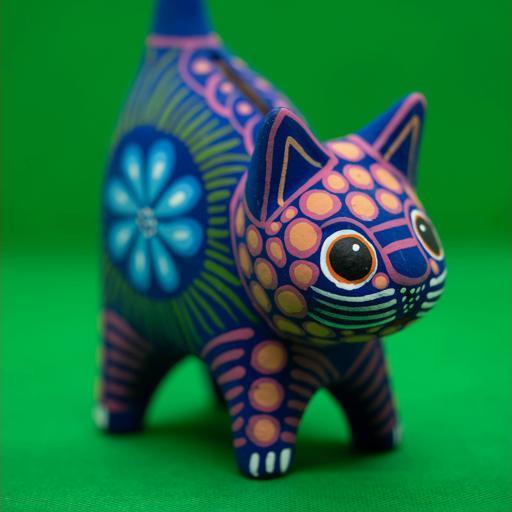} & 
    \includegraphics[width=0.09\textwidth]{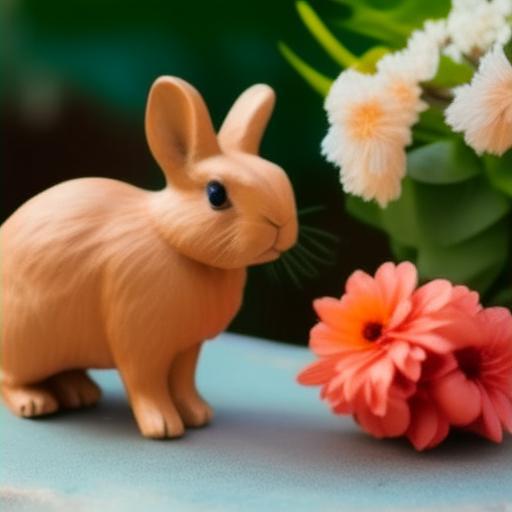} \\

    \includegraphics[width=0.09\textwidth]{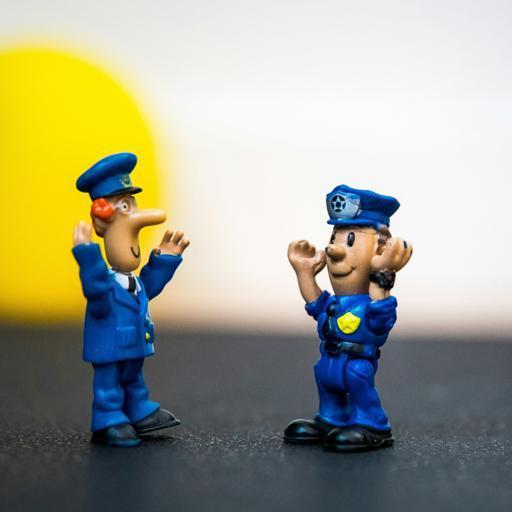} & 
    \includegraphics[width=0.09\textwidth]{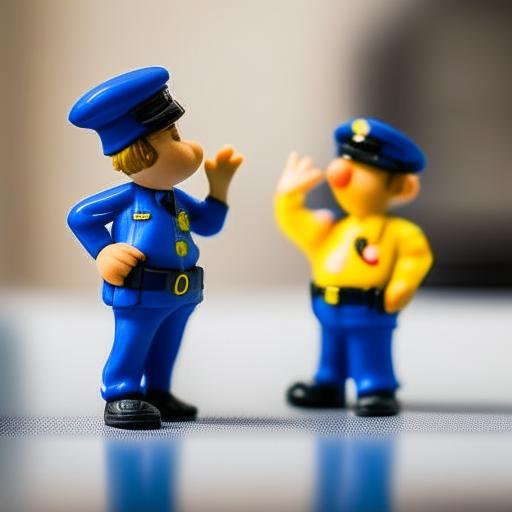} & 
    \includegraphics[width=0.09\textwidth]{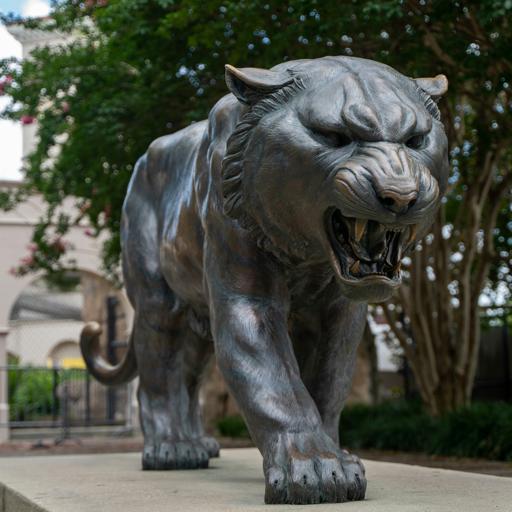} & 
    \includegraphics[width=0.09\textwidth]{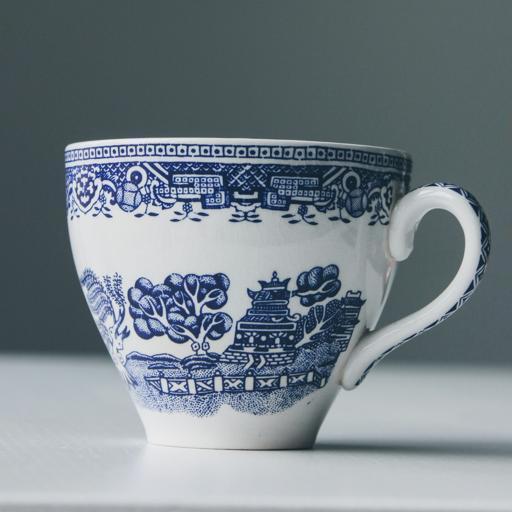} & 
    \includegraphics[width=0.09\textwidth]{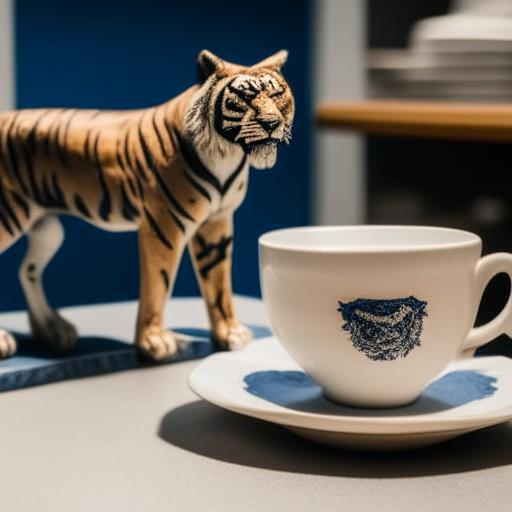}\\
    Input & Reconstruct & Input A & Input B & Union Result

    \end{tabular}
    }
    \caption{\textbf{Limitations of \textit{pOps}.} On the left we show reconstructions achieved by directly embedding an image into CLIP and reconstructing it with Kandinsky2, highlighting the limitations of the embedding space. On the right, we show failure cases for our union operator, where attribute leakage is visible or where the operator struggles with preserving both objects.}\label{fig:limitations}
\end{figure}

While our experiments highlight the potential of \textit{pOps} for semantic control, it is important to also discuss the limitations of our approach. 
First, there are inherent limitations when operating within the CLIP domain. As previously discussed in Ramesh~\etal~\shortcite{ramesh2022hierarchical}, the semantic embedding fails to preserve some visual attributes.
In~\Cref{fig:limitations} we visualize these limitations by viewing direct reconstructions of images when passing them through the CLIP embedding space. Although the embedding space effectively encodes the objects semantically, it struggles with encoding their distinct visual appearance compared to optimization-based personalization methods. As shown, CLIP also struggles with binding two different visual attributes to two distinct objects. This was most evident in our results for the union operation where the ``rendered'' result may leak colors between the two objects, struggling with maintaining the distinct appearance of each one.

Additionally, \textit{pOps} tunes each operator independently, where it might be more beneficial to train a single diffusion model capable of realizing all of our different operators together or alternatively do only a low-rank adaptation~\cite{hu2021lora} when training an operator.
Finally, all \textit{pOps} operators were trained on a single GPU for a few days. This leads us to believe that further computational scaling could potentially improve performance even within the limitations of the CLIP space and current architecture.

\section{Conclusions}
In this work, we have introduced \textit{pOps}, a framework designed for training semantic operations directly on CLIP image embeddings. \textit{pOps} offers a new take on image generation, providing users with specific forms of semantic control over image embeddings that can then be joined together to form the desired concept.
Our method builds upon both generated datasets that represent the task at hand and can also be supervised directly using a CLIP-based objective. 
We believe that \textit{pOps} opens up new possibilities for training a wide variety of operators within the CLIP space and other semantic spaces. These new operators can then be composed with one another to create even more creative possibilities along the generation process.

\begin{acks}
We would like to thank Rinon Gal, Yael Vinker, and Or Patashnik for their discussions and valuable input which helped improve this work. We would also like to thank Andrey Voynov for his early feedback on this work. This work was supported by the Israel Science Foundation under Grant No. 2366/16 and Grant No. 2492/20.
\end{acks}


\clearpage

\clearpage

\appendix
\appendixpage
\setlength{\abovedisplayskip}{5pt}
\setlength{\belowdisplayskip}{5pt}

\vspace*{0.2cm}
\section{Additional Details}~\label{sec:additional_details}

\vspace*{-0.5cm}
\subsection{Implementation Details}

\paragraph{\textbf{Models and Architectures.}}
In this work, we use the CLIP ViT-bigG-14-laion2B-39B-b160k model~\cite{radford2021learning,dosovitskiy2021an} for our embedding space, implemented using the Transformers library~\cite{wolf-etal-2020-transformers}. 
The architecture of our Diffusion Prior model follows the same architecture as used in Kandinsky 2~\cite{kandinsky2}.
For our diffusion models, we show results over both the Kandinsky 2.2 model~\cite{kandinsky2} and IP-Adapter~\cite{ye2023ip-adapter}, both of which support this specific CLIP model.

\paragraph{\textbf{Training Scheme.}}
We train all models using a batch size of $1$ over a single GPU. The models are trained using the AdamW optimizer~\cite{loshchilov2018decoupled} with a constant learning rate of $1e-5$. Each operator is trained for approximately $500,000$ training steps when trained from scratch. However, we found, empirically, that fine-tuning the model from an existing operator rather than the original Diffusion Prior model speeds up convergence. Unless otherwise noted, we train all the layers of the Diffusion Prior model.

\subsection{Data Generation}

\begin{figure}
    \centering
    \setlength{\belowcaptionskip}{-4pt}
    \setlength{\tabcolsep}{0pt}
    \renewcommand{\arraystretch}{1}
    \newcommand{\pl}{0.2}
    {\small
    \vspace*{0.5cm}
    \begin{tabular}{ccc@{\hskip 2pt}ccc}

    \multicolumn{6}{c}{Texturing} \\
    \includegraphics[width=0.075\textwidth]{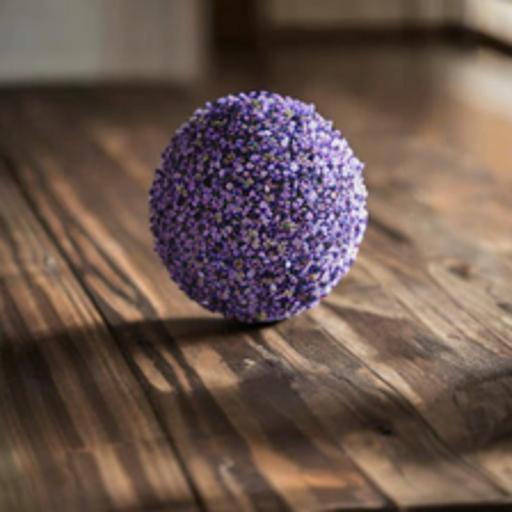} & \includegraphics[width=0.075\textwidth]{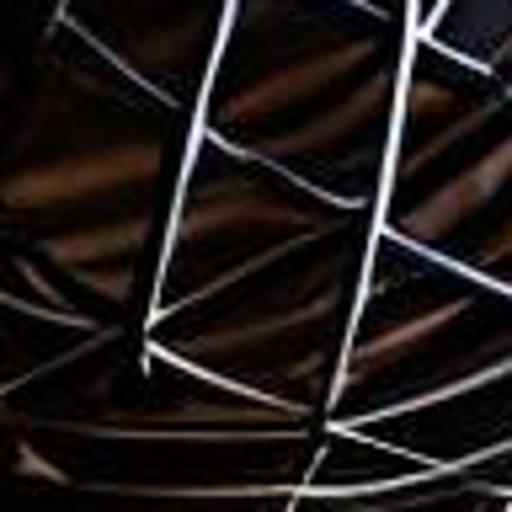} & 
    \includegraphics[width=0.075\textwidth]{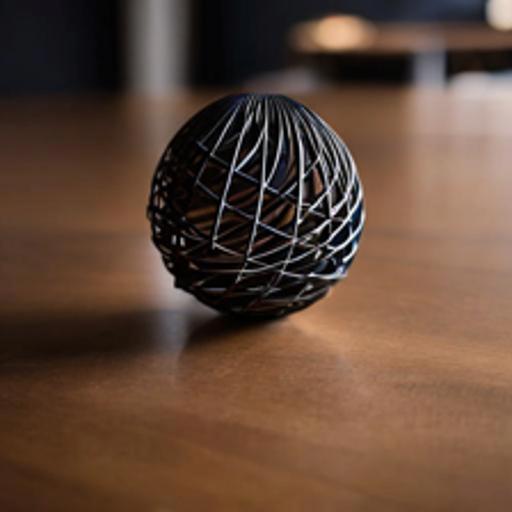} &
    \includegraphics[width=0.075\textwidth]{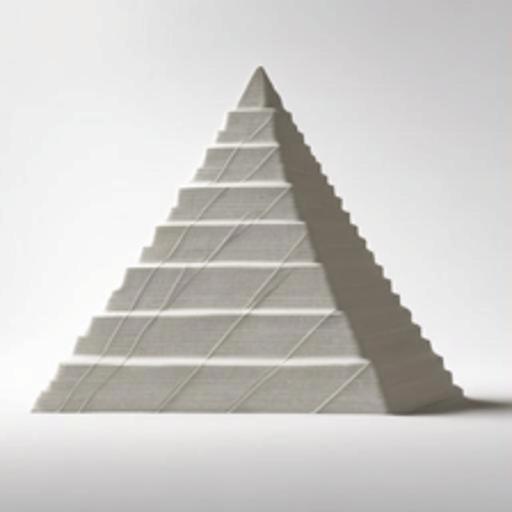} & \includegraphics[width=0.075\textwidth]{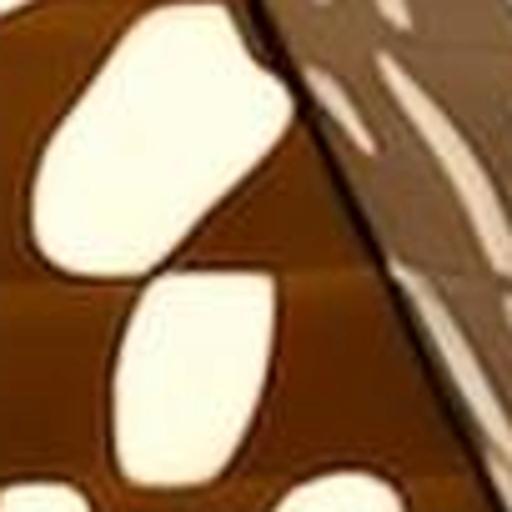} & 
    \includegraphics[width=0.075\textwidth]{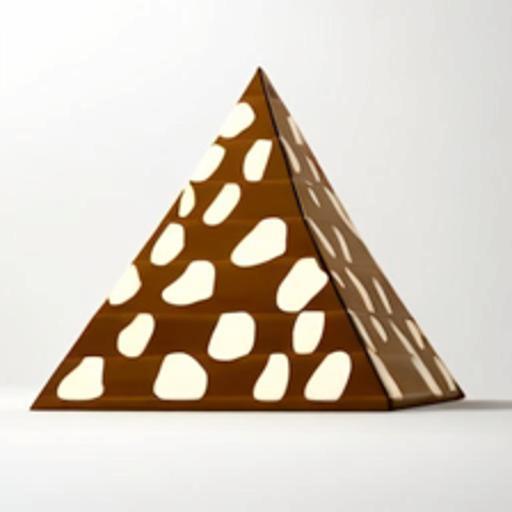} \\
    \includegraphics[width=0.075\textwidth]{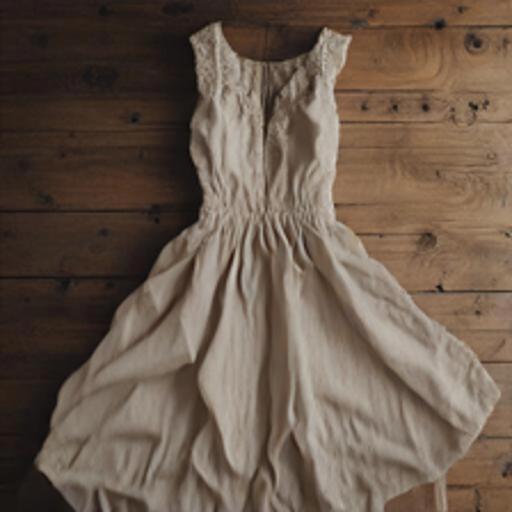} & \includegraphics[width=0.075\textwidth]{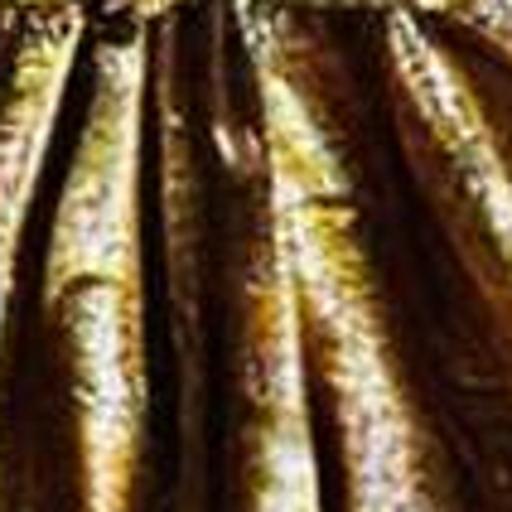} & 
    \includegraphics[width=0.075\textwidth]{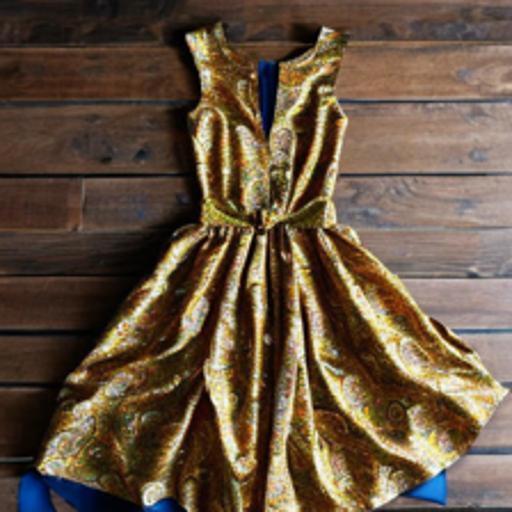} &
    \includegraphics[width=0.075\textwidth]{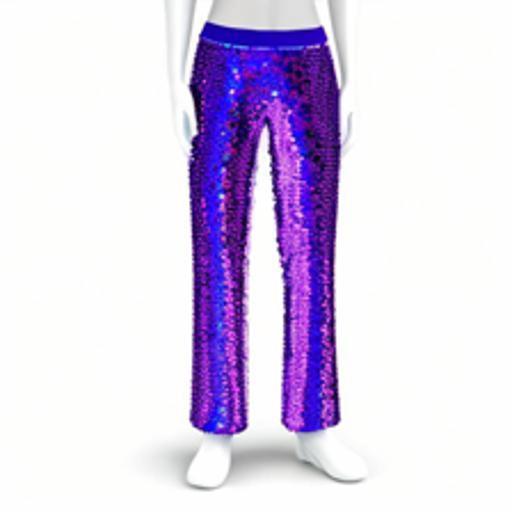} & \includegraphics[width=0.075\textwidth]{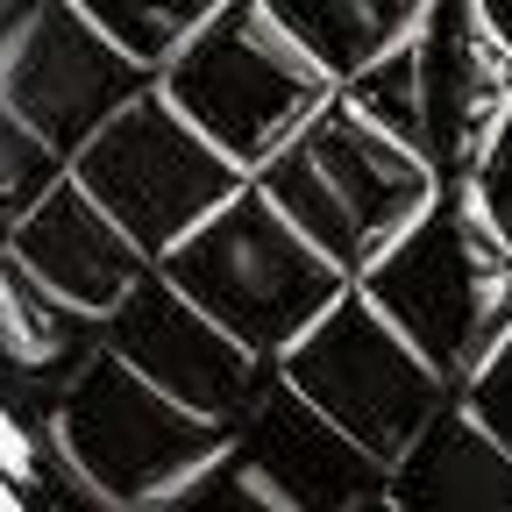} & 
    \includegraphics[width=0.075\textwidth]{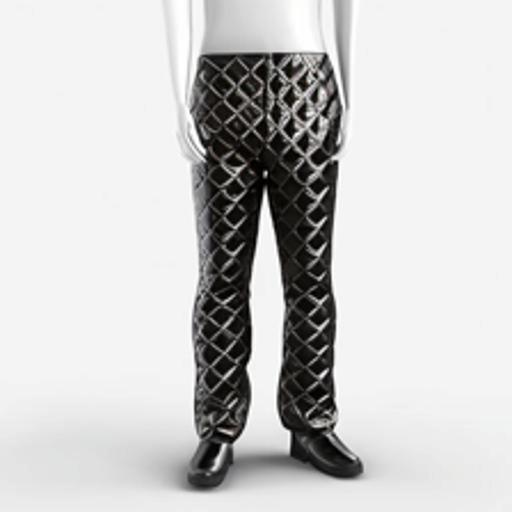} \\
    
    \\
    
    \multicolumn{6}{c}{Scene} \\
    
    \includegraphics[width=0.075\textwidth]{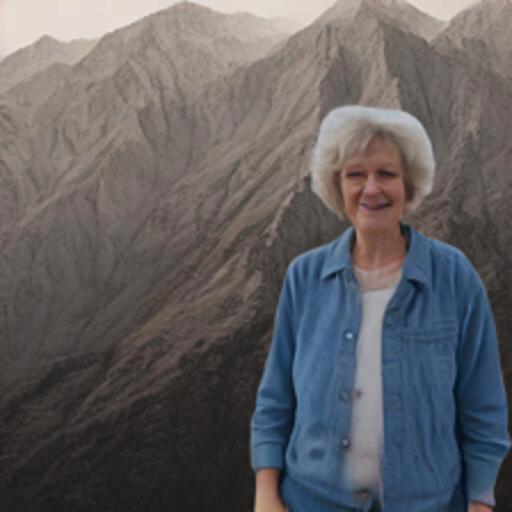} & \includegraphics[width=0.075\textwidth]{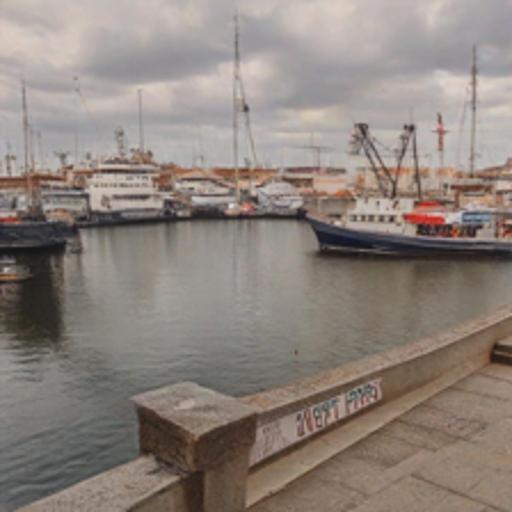} & 
    \includegraphics[width=0.075\textwidth]{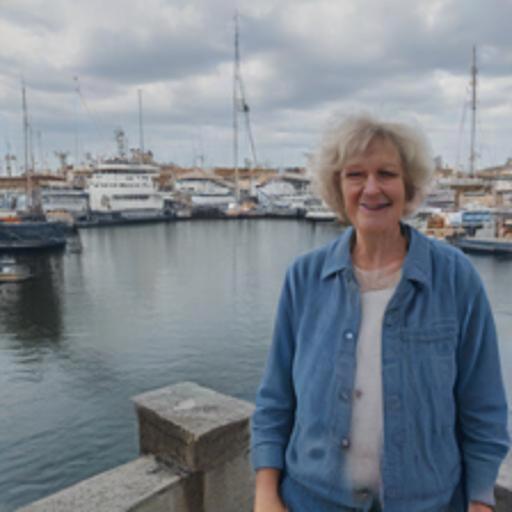} &
    \includegraphics[width=0.075\textwidth]{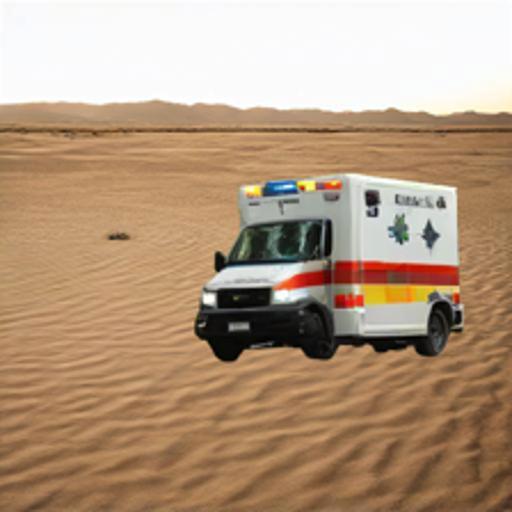} & \includegraphics[width=0.075\textwidth]{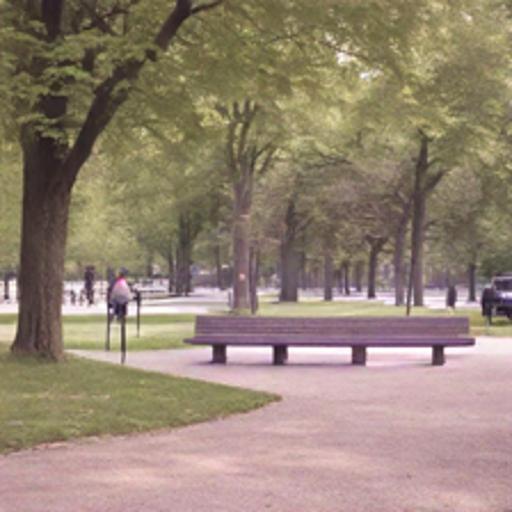} & 
    \includegraphics[width=0.075\textwidth]{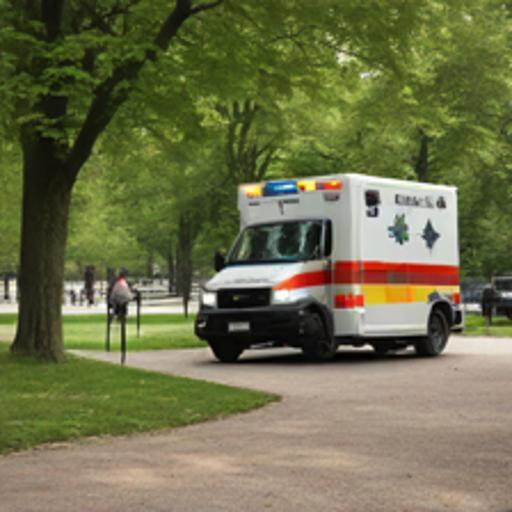}\\

    \includegraphics[width=0.075\textwidth]{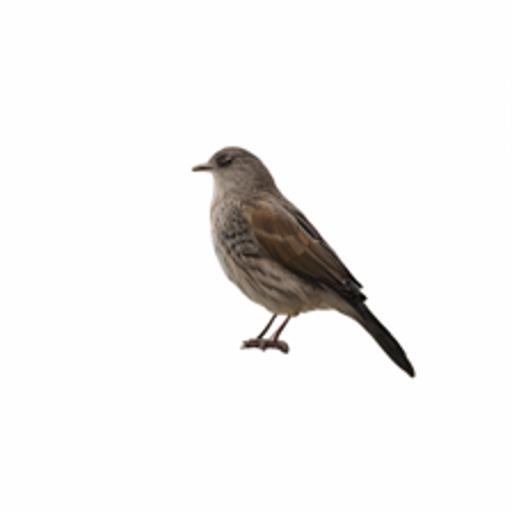} & \includegraphics[width=0.075\textwidth]{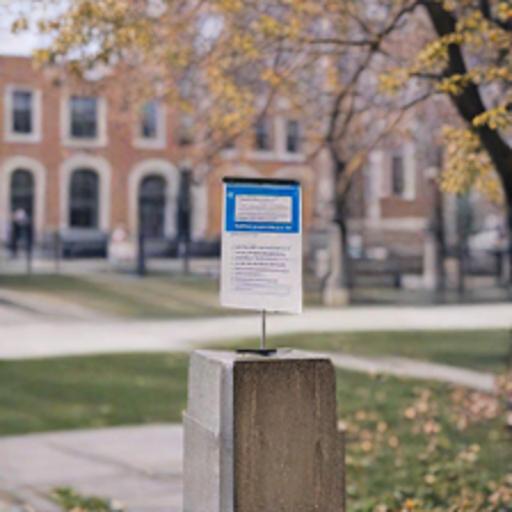} & 
    \includegraphics[width=0.075\textwidth]{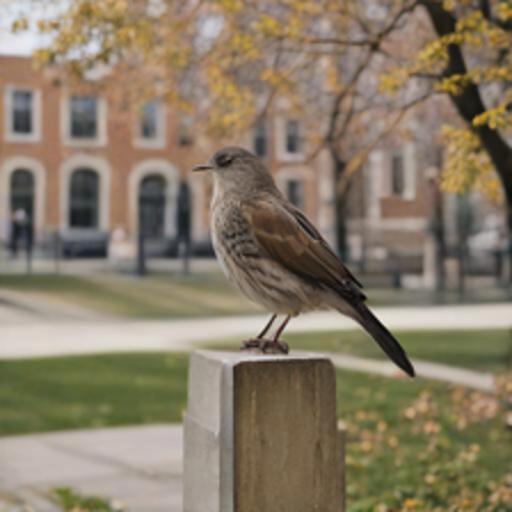} &
    \includegraphics[width=0.075\textwidth]{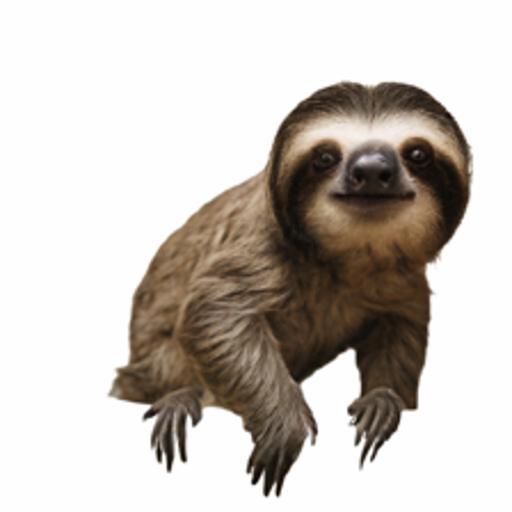} & \includegraphics[width=0.075\textwidth]{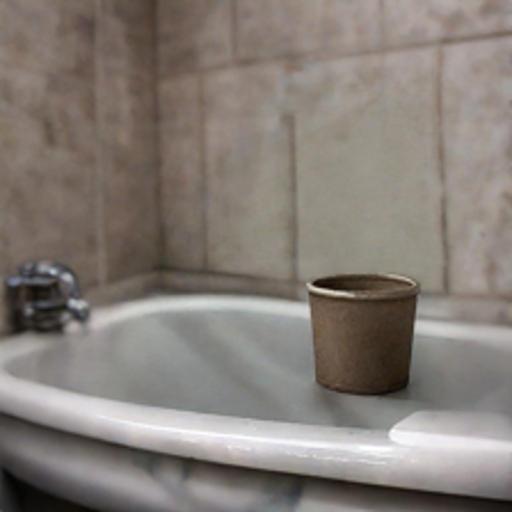} & 
    \includegraphics[width=0.075\textwidth]{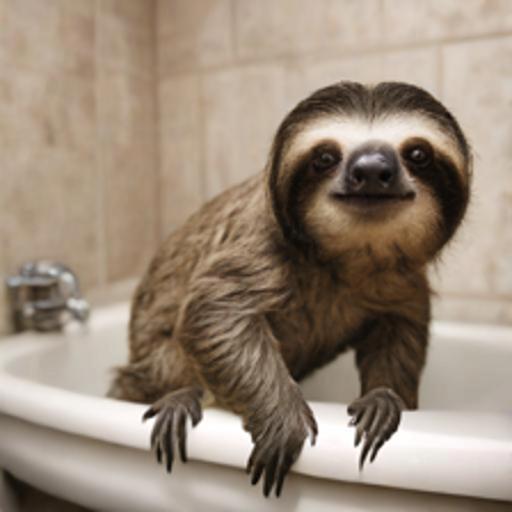}\\

    \\

    \multicolumn{6}{c}{Union} \\
    \includegraphics[width=0.075\textwidth]{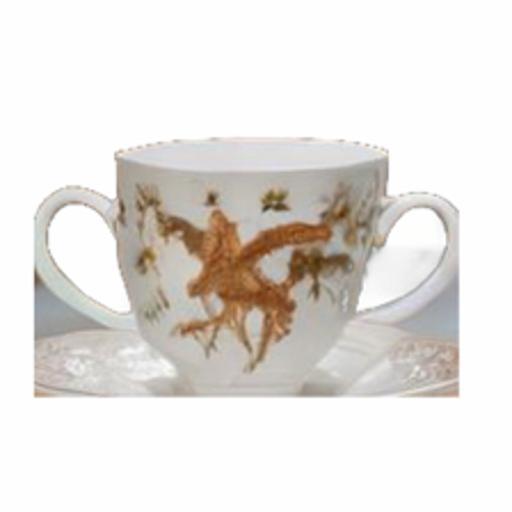} & \includegraphics[width=0.075\textwidth]{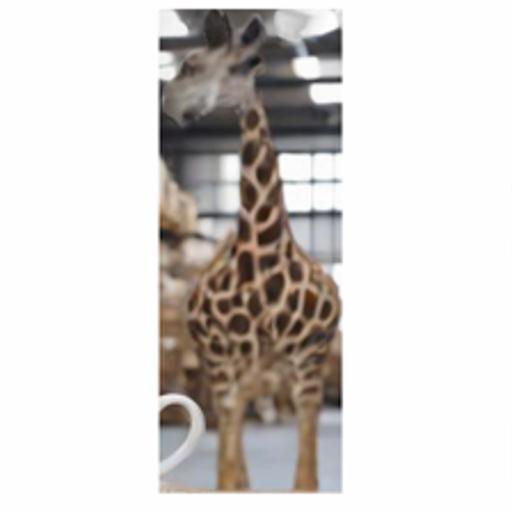} & 
    \includegraphics[width=0.075\textwidth]{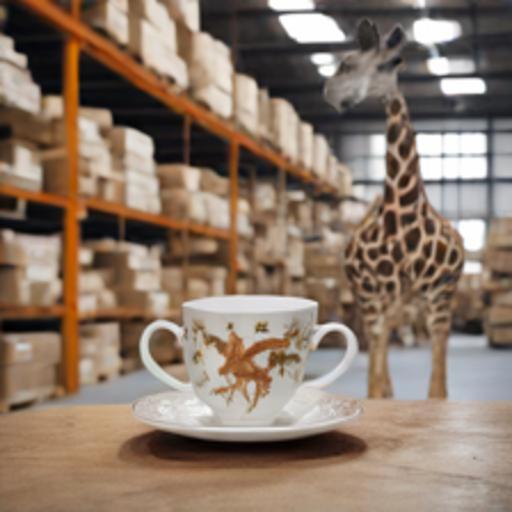} &
    \includegraphics[width=0.075\textwidth]{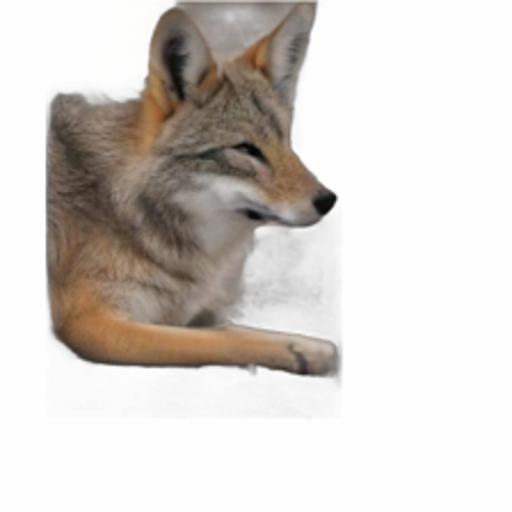} & \includegraphics[width=0.075\textwidth]{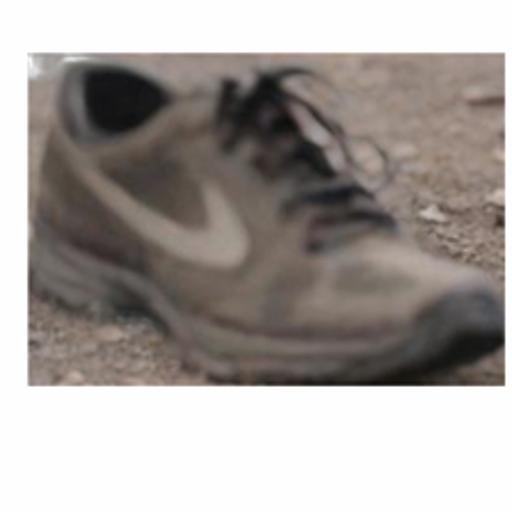} & 
    \includegraphics[width=0.075\textwidth]{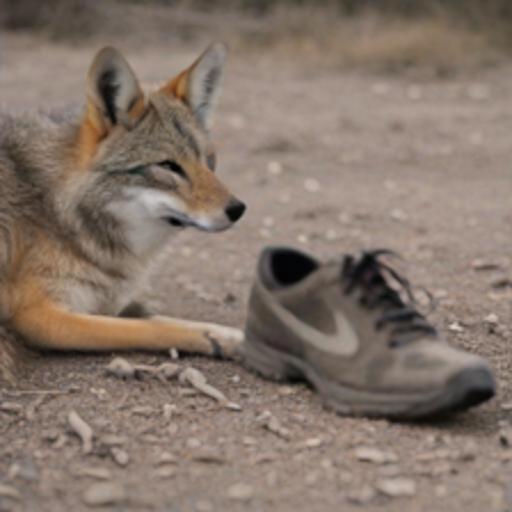}\\
    \includegraphics[width=0.075\textwidth]{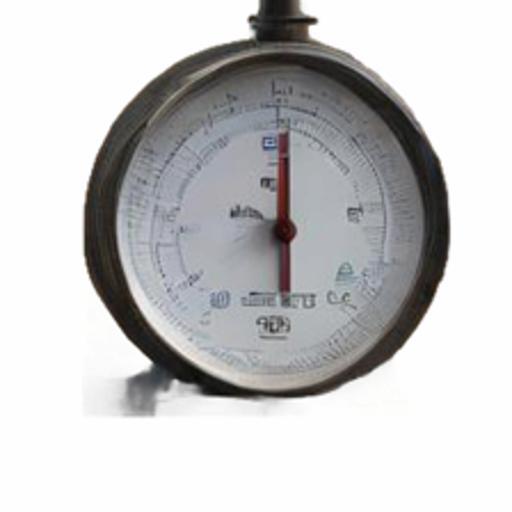} & \includegraphics[width=0.075\textwidth]{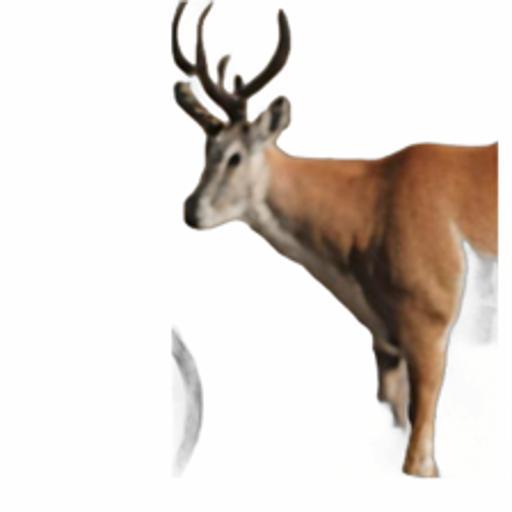} & 
    \includegraphics[width=0.075\textwidth]{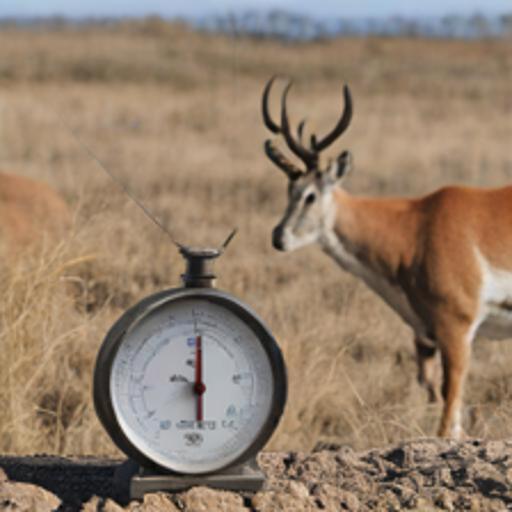} &
    \includegraphics[width=0.075\textwidth]{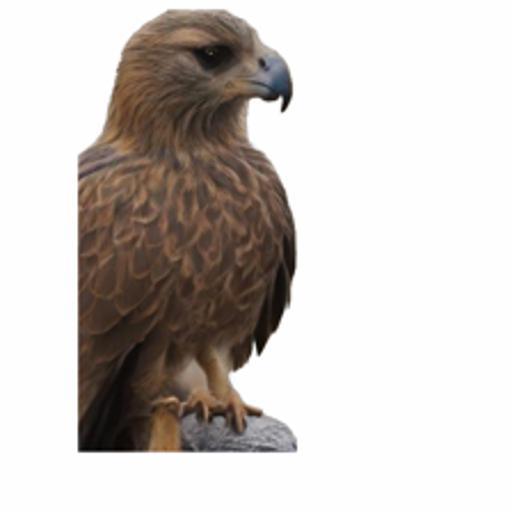} & \includegraphics[width=0.075\textwidth]{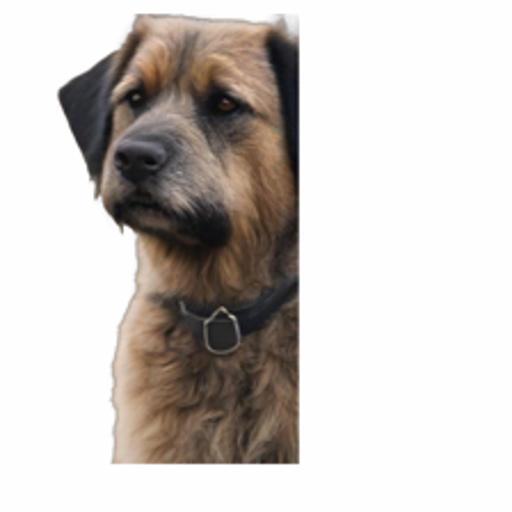} & 
    \includegraphics[width=0.075\textwidth]{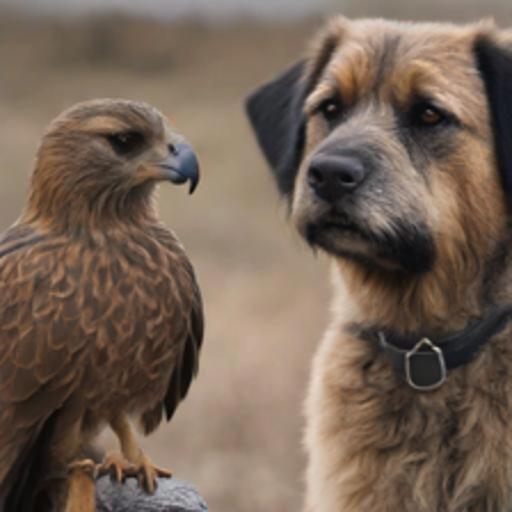} \\
    $I_a$ & $I_b$ & $I_{target}$ & $I_a$ & $I_b$ & $I_{target}$ \\[-0.3cm]

    \end{tabular}
    }
    \caption{Generated paired data for various \textit{pOps} operators. During training, the images are encoded to embeddings $e_a$, $e_b$. and $e_{target}$, respectively. \\[-0.4cm]}
    \label{fig:data}
\end{figure}

In the main paper, we discuss the process used for generating data for each operator. Below, we provide additional details. Samples of the generated data are illustrated in~\Cref{fig:data}. Unless otherwise noted, for each operator, we generate approximately $50,000$ samples.

\paragraph{\textbf{Texturing.}}
The data generation scheme for our texturing operator is illustrated in Figure 5 of the main paper. We consider $290$ object candidates across various categories such as geometric objects, animals, statues, and other miscellaneous common objects. We additionally consider $24$ different object placement candidates and $310$ texture attributes. For generating the object images, we use SDXL-Turbo~\cite{sauer2024fast} and use prompts of the form ``A photo of a \textcolor{blue}{<object>} \textcolor{red}{<placement>}.''. 

To generate the target image $I_{target}$, we then sample between one and five texture attributes and generate an image using a depth-conditioned Stable Diffusion 2.0~\cite{rombach2021highresolution} model using prompts of the form ``A photo of a \textcolor{blue}{<object>} made from \textcolor{darkgreen}{<texture 1, texture 2, ...>} \textcolor{red}{<placement>}.'' The generation process is conditioned on the depth map extracted from $I_{object}$. 

Finally, we are left to extract the patch representing our input texture image $I_{texture}$. To this end, we first detect the object in the generated image using an OWLv2~\cite{minderer2023scaling} model with the prompt ``A \textcolor{blue}{<object>}''. We then select a small patch from within the output bounding box and use this as our texture image.

\paragraph{\textbf{Scene.}}
For our scene operator, as noted in the main paper, $I_{object}$ is created either by pasting the segmented object either on a white background or a newly generated background. For the newly generated background, we compose a set of $208$ possible backgrounds such as ``On the beach'', ``On the farm'', ``In the castle'', etc. For our inpainting model, used to create $I_{back}$, we employ the SD-XL Inpainting 0.1 model using the mask extracted from our object.

\paragraph{\textbf{Union.}}
For generating our union dataset, we consider $20,000$ different objects, taken from the raw classes list from Open Images~\cite{OpenImages}.

\paragraph{\textbf{Instruct.}}
Here, we sample our images from a set of $20,000$ possible classes, as above, and a list of $60$ possible adjectives.

\paragraph{\textbf{Composition.}}
As noted in the main paper, for our composition operator, we use the ATR dataset~\cite{liang2015deep} for training. In total, we use $17,000$ images for training, comprising $12$ different clothing categories.

\section{Evaluation Setup}

\subsection{Baseline Methods}

\paragraph{\textbf{Texturing.}}
For evaluating our texturing operator, we consider four alternative methods: (1) Cross-Image Attention~\cite{alaluf2023cross}, (2) IP-Adapter~\cite{ye2023ip-adapter}, (3) Visual Style Prompting~\cite{jeong2024visual}, and (4) ZeST~\cite{cheng2024zest}. 

For all methods, we use their official implementation and default hyperparameters. For IP-Adapter~\cite{ye2023ip-adapter}, we consider IP-Adapter trained over Stable Diffusion 1.5~\cite{rombach2021highresolution} which uses OpenCLIP-ViT-H-14 for extracting the conditioning image embeddings.

\paragraph{\textbf{Instruct.}}
For the instruct operator, we consider three approaches: (1) IP-Adapter~\cite{ye2023ip-adapter}, (2) InstructPix2Pix~\cite{brooks2022instructpix2pix}, and (3) NeTI~\cite{alaluf2023neural}. 

For IP-Adapter, we consider two variants. First, we use the IP-Adapter Plus variant trained over Stable Diffusion 1.5 using a scale of $0.5$, where we pass the adjective as the guiding text prompt. 
However, we attained better results when using the more recent IP-Adapter for SDXL 1.0 which is conditioned on image embeddings extracted from OpenCLIP-ViT-H-14 (ip-adapter-plus\_sdxl\_vit-h). We found that to achieve meaningful semantic modifications, a low scale factor of $0.1$ was needed. However, when doing so, the resulting images generated by IP-Adapter no longer resembled the original images. As such, we captioned the original images using BLIP-2~\cite{li2023blip} and passed the image caption along with the desired adjective to IP-Adapter as the guiding text prompt. We found that this allowed for better alignment with the adjective (thanks to the low scale) while better preserving the original image (thanks to the image caption).

Finally, we compare \textit{pOps} to NeTI, an optimization-based personalization method (see~\Cref{fig:instruct_comparisons_neti}). We follow the default hyperparameters and train a new concept using the image of the object. The best results were achieved when training for $250$ optimization steps, as additional training led to overfitting the original image. At inference, we generated images using prompts of the form ``A photo of a <adjective> $S_*$''. When needed, we manually modified the prompts to ensure that they were grammatically correct.

\subsection{Quantitative Evaluations}

Below we provide details regarding the evaluation data and protocol reported in the main paper.

\paragraph{\textbf{Texturing.}}
To quantitatively evaluate performance on the texturing task, we consider $52$ images of objects spanning various categories including animals, statues, food items, accessories, and more. For each object, we paint the object using $16$ different texture patches, resulting in $832$ object-texture combinations. For each of the considered methods, we utilized three different random seeds, which gave $2,496$ total results.

As no standard metric exists for evaluating the quality of the texturing, we perform a perceptual user study. We consider two types of questions: (1) top preference and (2) rating. More specifically, users were first shown the results of the four methods side-by-side and asked to choose the result they most preferred while taking into account both how the original object was preserved and how the target texture was applied. Next, users were asked to rate the result of each method on a scale of $1$ to $5$, with $5$ being the best, on how well the original object was preserved and the texture was applied to it. Each user was shown $7$ questions for each of the two types.

\paragraph{\textbf{Instruct.}}
To evaluate our instruct operator, we similarly construct an evaluation set. Here, we consider the same $52$ objects as above and construct a set of $65$ adjectives. We then modify each of the $52$ objects with each adjective, resulting in $3,380$ combinations. As above, each method is applied using three different seeds, resulting in $10,140$ generated images. 

For our evaluation metric, we first consider the standard CLIPScore~\cite{hessel2021clipscore} and measure CLIP-space similarities. Specifically, we first compute the image similarity between the generated images and the original image. Next, we calculate the CLIP-space similarity between the embeddings of the generated images and the embedding of text prompts of the form ``A <adjective> photo''. Finally, we consider an additional text-based similarity metric. Here, we first manually create a short caption of the target object (e.g., ``A lion statue'', ``A dress''). We then caption the generated images using BLIP-2~\cite{li2023blip}. We then compute a sentence similarity measure~\cite{devlin2018bert,reimers-2019-sentence-bert}, computing the average cosine similarity between sentence embeddings extracted from the generated caption and captions of the form ``A photo of a <adjective> <caption>.'' This metric was designed to better capture the ability of the methods to integrate the desired adjective while preserving the original object class.

\section{Additional Comparisons}~\label{sec:additional_comparisons}
\begin{figure*}
    \centering
    \setlength{\tabcolsep}{0.5pt}
    \renewcommand{\arraystretch}{0.5}
    {\small
\begin{tabular}{c c c c@{\hspace{0.4cm}} c c c c@{\hspace{0.4cm}}}

        \includegraphics[width=0.105\textwidth]{figures/union_comparisons/images/alvan-nee-T-0EW-SEbsE-unsplash_pablo-merchan-montes-_Tw4vCs9C-8-unsplash/object_1.jpg} &
        \includegraphics[width=0.105\textwidth]{figures/union_comparisons/images/alvan-nee-T-0EW-SEbsE-unsplash_pablo-merchan-montes-_Tw4vCs9C-8-unsplash/object_2.jpg} &
        \includegraphics[width=0.105\textwidth]{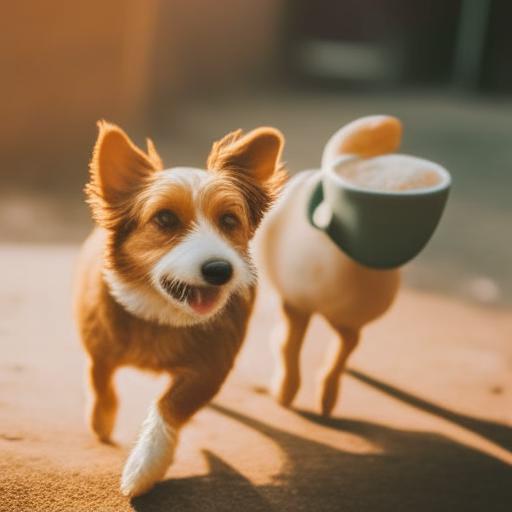} &
        \includegraphics[width=0.105\textwidth]{figures/union_comparisons/images/alvan-nee-T-0EW-SEbsE-unsplash_pablo-merchan-montes-_Tw4vCs9C-8-unsplash/ours.jpg} &

        \includegraphics[width=0.105\textwidth]{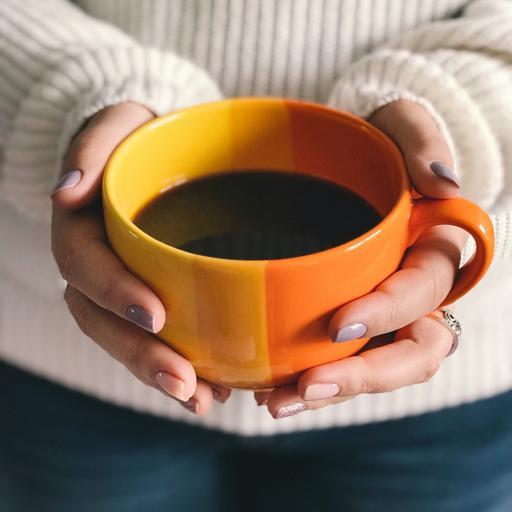} &
        \includegraphics[width=0.105\textwidth]{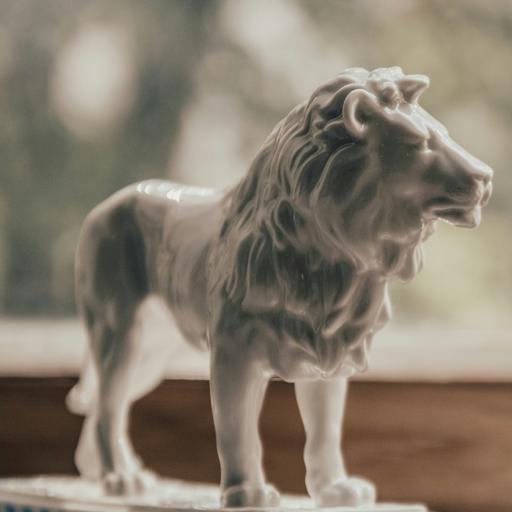} &
        \includegraphics[width=0.105\textwidth]{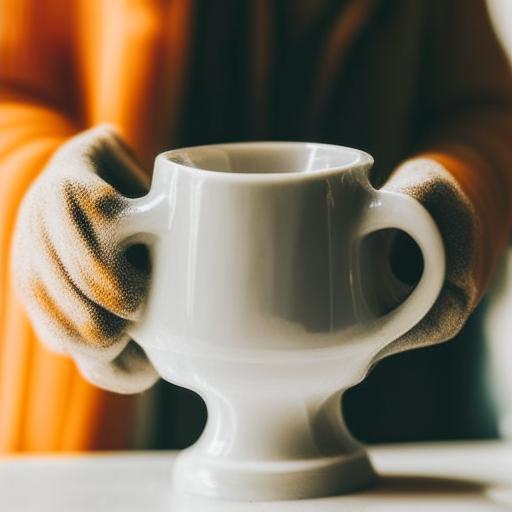} &
        \includegraphics[width=0.105\textwidth]{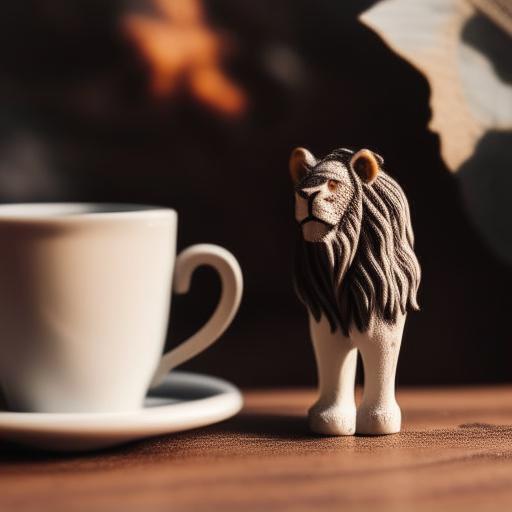} \\

        Object A & Object B & Average & Ours & Object A & Object B & Average & Ours \\ \\
\end{tabular}
    }
    \caption{Qualitative comparison of our \textit{pOps} union operator compared to results obtained by averaging over the CLIP image embdedings.}
    \label{fig:union_mean_comparisons}
\end{figure*}

\begin{figure*}
    \centering
    \setlength{\tabcolsep}{0.3pt}
    \renewcommand{\arraystretch}{0.5}
    {\small
    \begin{tabular}{c c c c@{\hspace{0.4cm}} c c c c@{\hspace{0.4cm}}}

        \includegraphics[width=0.105\textwidth]{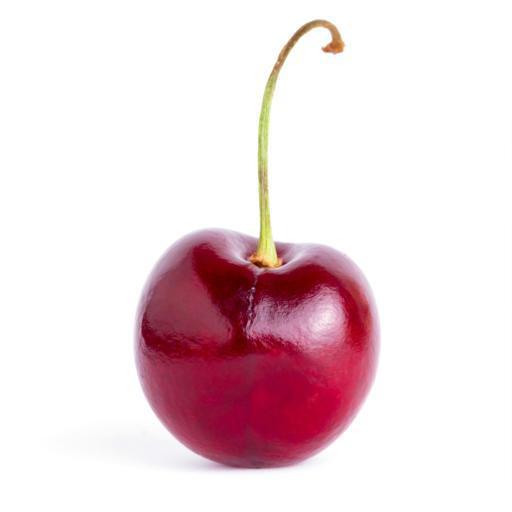} &
        \includegraphics[width=0.105\textwidth]{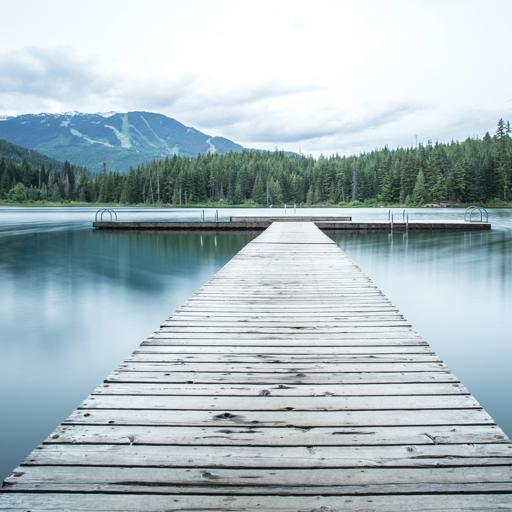} &
        \includegraphics[width=0.105\textwidth]{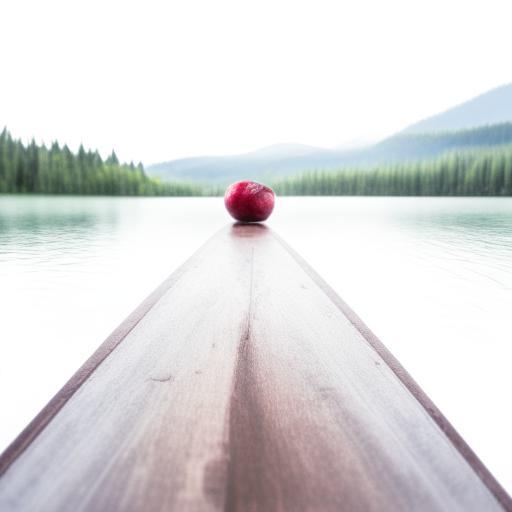} &
        \includegraphics[width=0.105\textwidth]{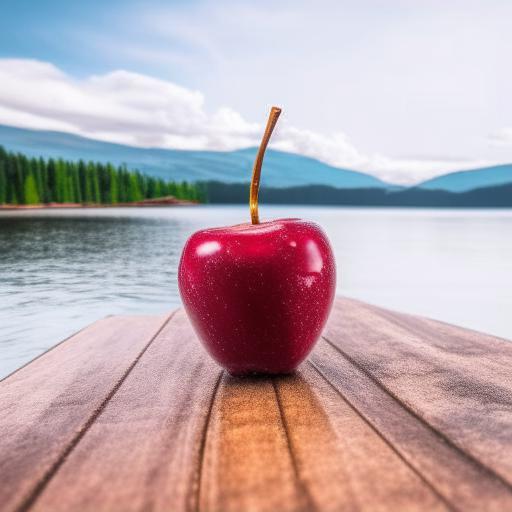} &

        \includegraphics[width=0.105\textwidth]{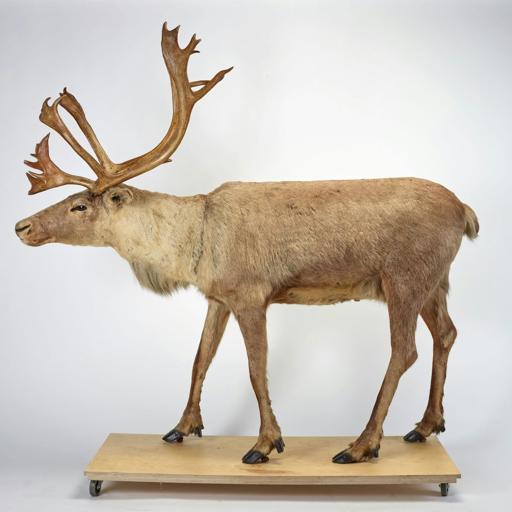} &
        \includegraphics[width=0.105\textwidth]{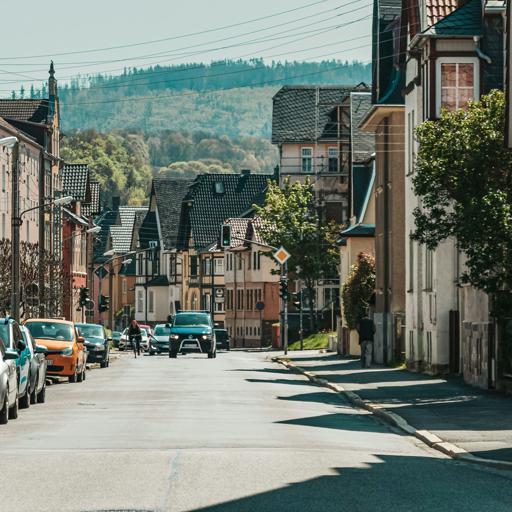} &
        \includegraphics[width=0.105\textwidth]{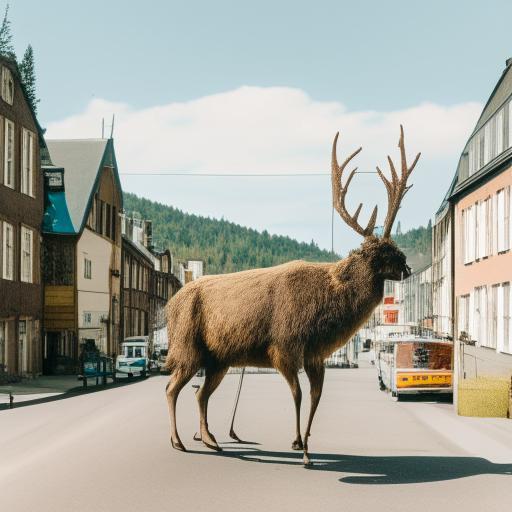} &
        \includegraphics[width=0.105\textwidth]{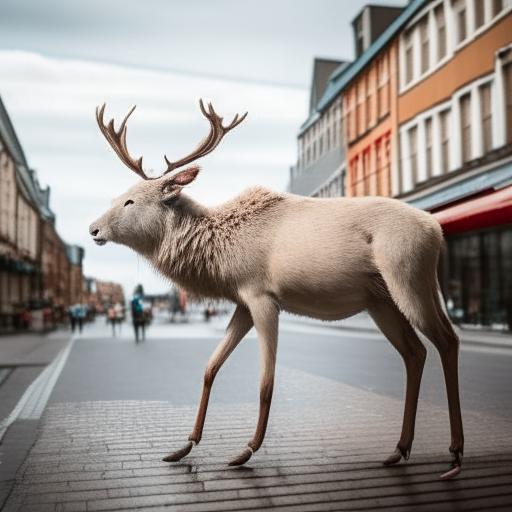} \\

        Object & Scene & Average & Ours & Object & Scene & Average & Ours \\ \\
\end{tabular}
    }
    \caption{Qualitative comparison of our \textit{pOps} scene operator compared to results obtained by averaging over the CLIP image embdedings.}
    \label{fig:scene_mean_comparisons}
\end{figure*}

We provide additional qualitative comparisons, as follows:
\begin{enumerate}
    \item First, in~\Cref{fig:texture_mean_comparisons,fig:scene_mean_comparisons,fig:union_mean_comparisons}, we provide additional comparisons over our binary operators (union, scene, and texturing), comparing our \textit{pOps} results with those obtained from a simple latent averaging within the CLIP embedding space.
    \item In~\Cref{fig:supp_texturing_comparisons}, we provide additional qualitative comparisons to alternative texturing approaches: Cross-Image Attention~\cite{alaluf2023cross}, IP-Adapter~\cite{ye2023ip-adapter}, Visual Style Prompting~\cite{jeong2024visual}, and ZeST~\cite{cheng2024zest}.
    \item In~\Cref{fig:supp_instruct_comparisons}, we show additional qualitative comparisons over our instruct operator, comparing \textit{pOps} to two alternative approaches: InstructPix2Pix~\cite{brooks2022instructpix2pix} and IP-Adapter~\cite{ye2023ip-adapter}.
    \item Finally, in~\Cref{fig:instruct_comparisons_neti}, we compare our instruct operator to an additional optimization-based personalization approach, NeTI~\cite{alaluf2023neural}.
\end{enumerate}

\section{Additional Results}~\label{sec:additional_results}
Finally, in the below Figures, we provide additional results:
\begin{enumerate}
    \item In~\Cref{fig:null_object} and~\Cref{fig:null}, we show additional results obtained by our texturing Diffusion Prior model when using null inputs for the object input and texture input, respectively.
    \item In~\Cref{fig:texturing_operator_1,fig:texturing_operator_2,fig:texturing_ours_multi_textures}, we provide additional texturing results.
    \item In~\Cref{fig:union_operator_1}, we provide additional union results.
    \item In~\Cref{fig:scene_operator_results_1,fig:scene_operator_results_2,fig:scene_operator_results_3}, we show additional scene operator results.
    \item In~\Cref{fig:instruct_operator_1}, we provide additional instruct operator results.
    \item In~\Cref{fig:clothes_compose_1,fig:clothes_compose_2}, we show additional multi-image clothing composition results obtained with \textit{pOps}.
    \item In~\Cref{fig:ip_adapter_supp}, we show results obtained with both Kandinsky~\cite{kandinsky2} and IP-Adapter~\cite{ye2023ip-adapter} as renderers as well as results obtained with IP-Adapter alongside ControlNet~\cite{zhang2023adding} with depth-conditioning.
    \item Finally, in~\Cref{fig:instruct_scene_compositions,fig:instruct_texturing_compositions,fig:texturing_instruct_compositions}, we provide examples of operator compositions, combining our scene, instruct, and texturing \textit{pOps} operators.
\end{enumerate}

\clearpage
\newpage

\begin{figure*}
    \centering
    \setlength{\tabcolsep}{0.5pt}
    \addtolength{\belowcaptionskip}{-10pt}
    \renewcommand{\arraystretch}{0.5}
    {\small
    \begin{tabular}{c c c c@{\hspace{0.4cm}} c c c c@{\hspace{0.4cm}}}

        \\ \\ \\ \\ \\ \\
        \\ \\
\end{tabular}
    }
    \caption{Qualitative comparison of our \textit{pOps} texturing operator compared to results obtained by averaging over the CLIP image embdedings.}
    \label{fig:texture_mean_comparisons}
\end{figure*}

\begin{figure*}
    \centering
    \setlength{\tabcolsep}{0.5pt}
    \addtolength{\belowcaptionskip}{-10pt}
    {
    \begin{tabular}{c c@{\hspace{0.4cm}} c c c c c}

        \\ \\

        \includegraphics[width=0.11\textwidth]{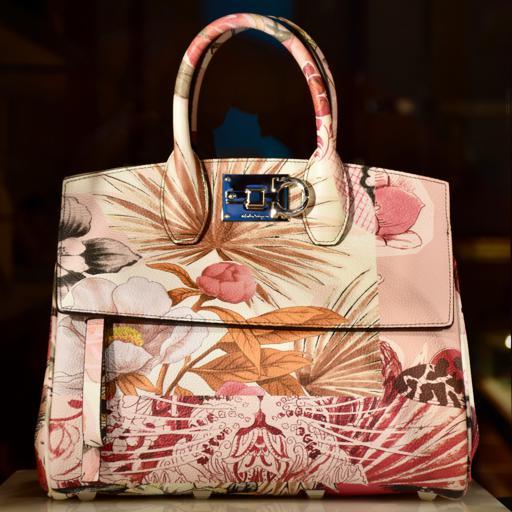} &
        \includegraphics[width=0.11\textwidth]{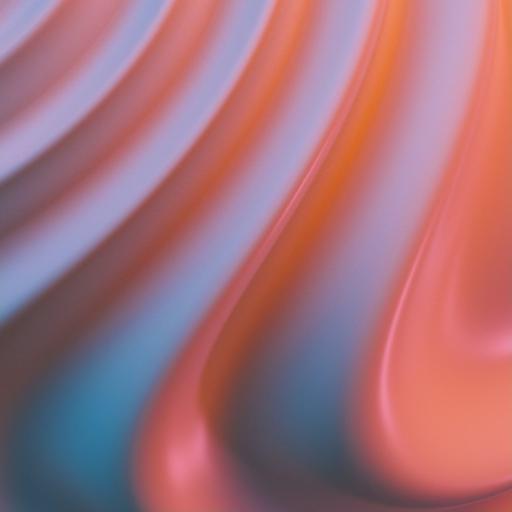} &
        \includegraphics[width=0.11\textwidth]{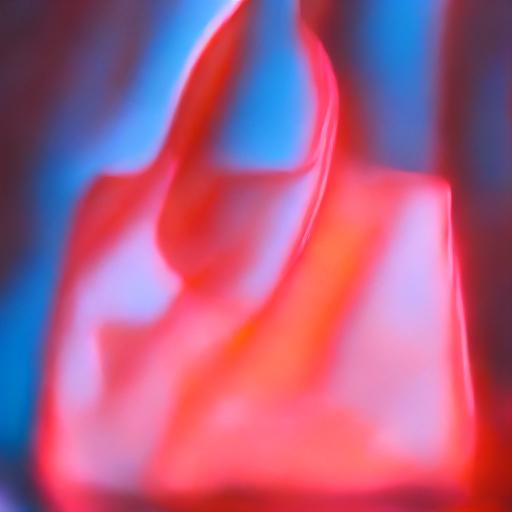} &
        \includegraphics[width=0.11\textwidth]{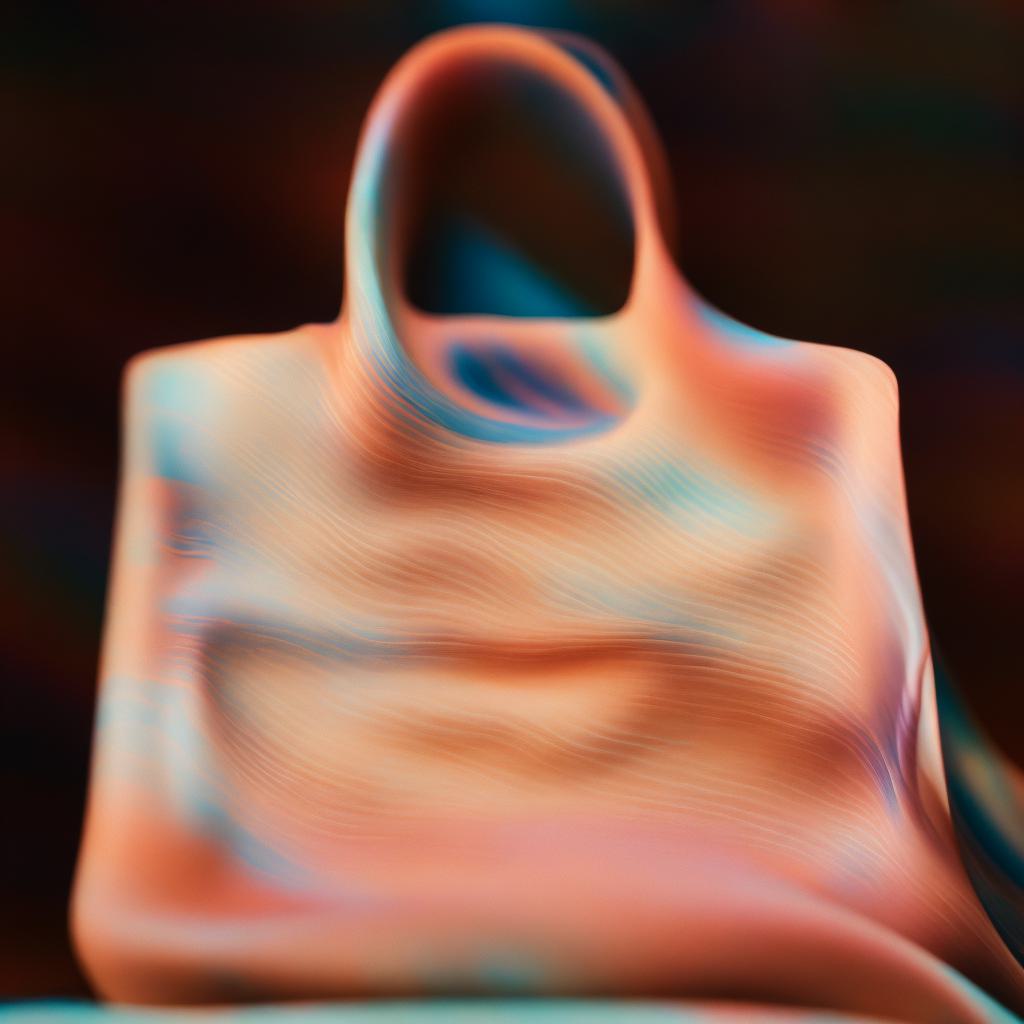} &
        \includegraphics[width=0.11\textwidth]{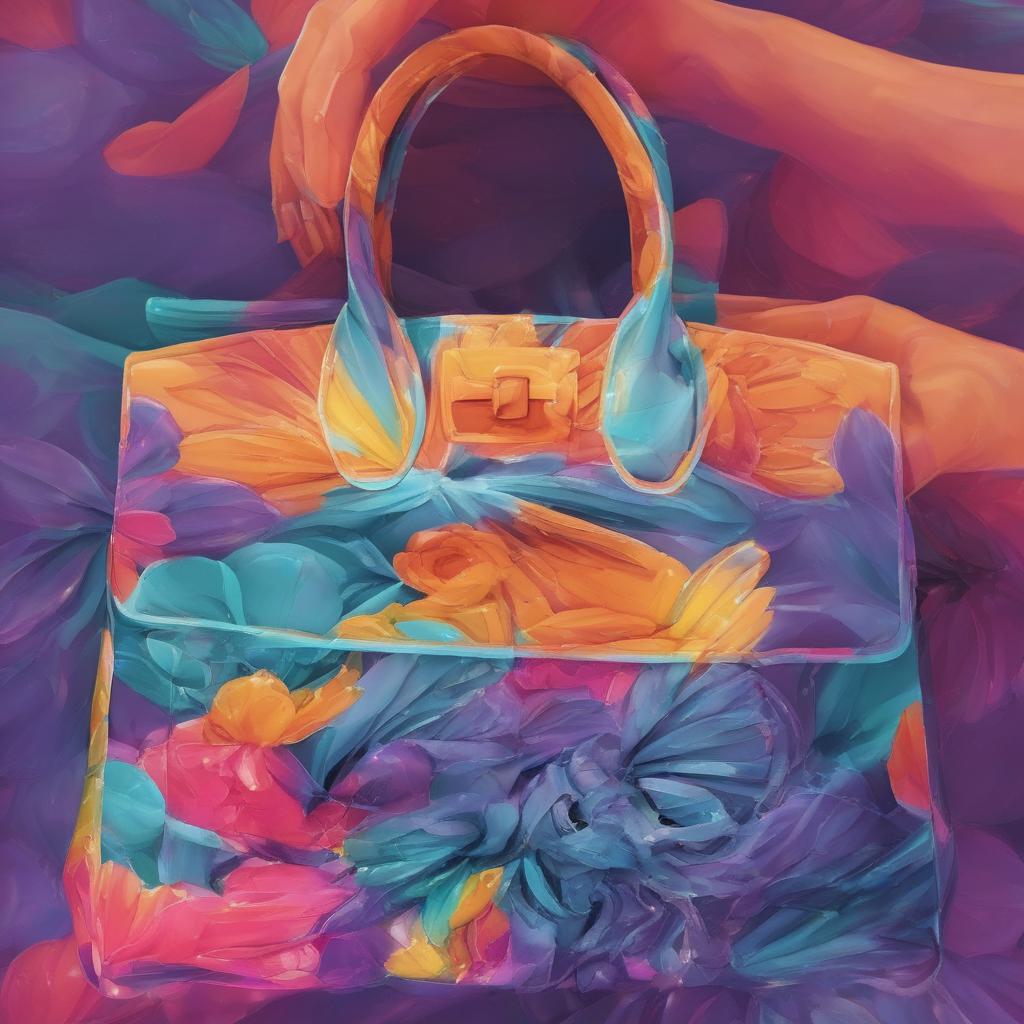} &
        \includegraphics[width=0.11\textwidth]{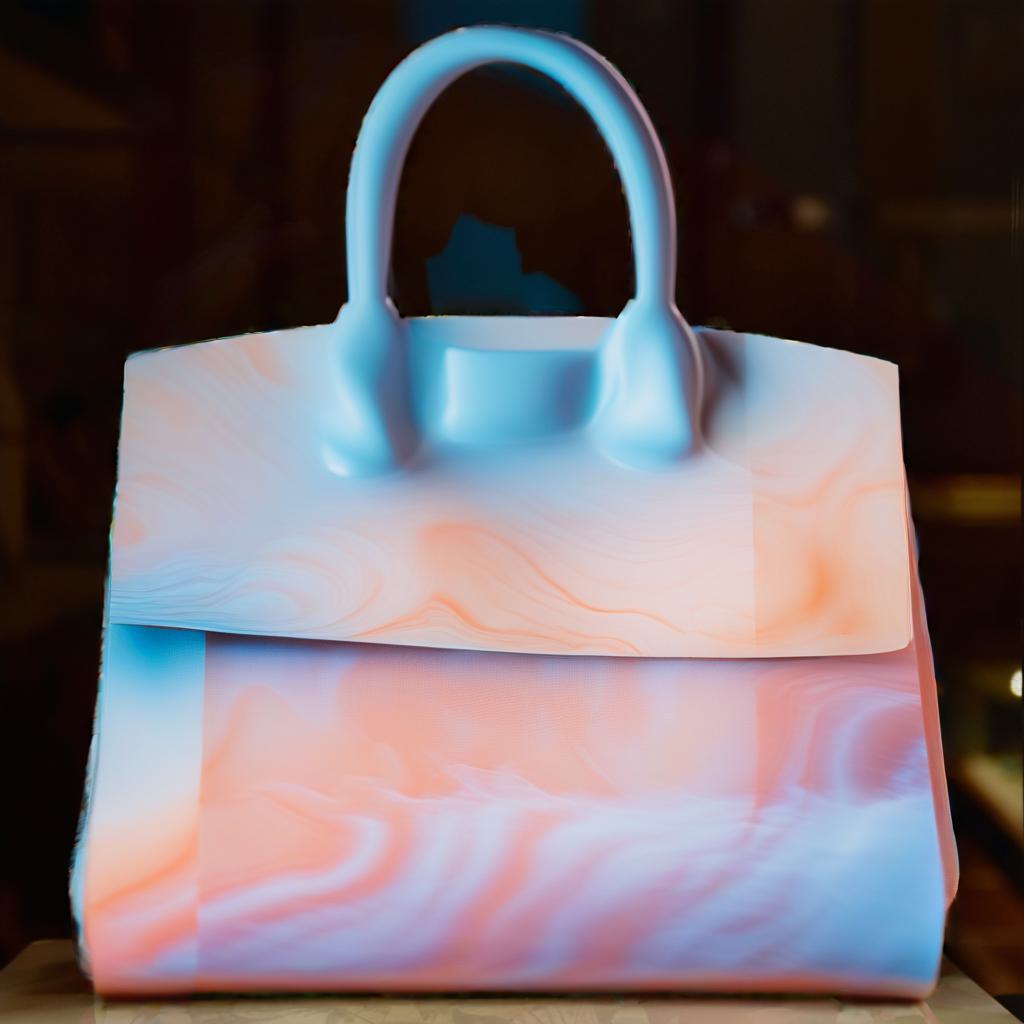} &
        \includegraphics[width=0.11\textwidth]{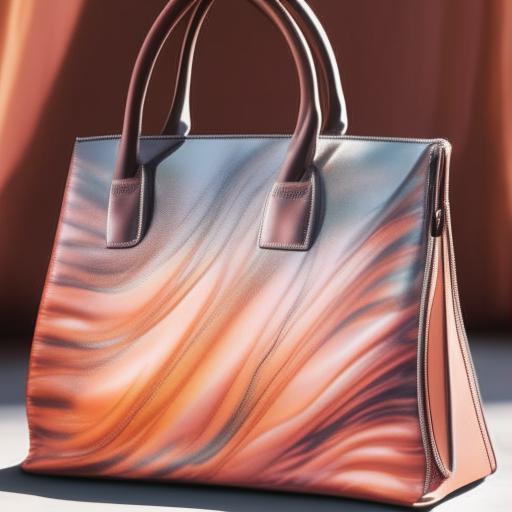} \\

        Object & Texture & CIA & IP-Adapter & Style Prompting & ZeST & pOps \\
\end{tabular}
    }
    \caption{Additional qualitative comparison for the \textit{pOps} texturing operator to alternative texturing approaches: Cross-Image Attention~\cite{alaluf2023cross}, IP-Adapter~\cite{ye2023ip-adapter}, Visual Style Prompting~\cite{jeong2024visual}, and ZeST~\cite{cheng2024zest}.}
    \label{fig:supp_texturing_comparisons}
\end{figure*}

\begin{figure*}
    \centering
    \setlength{\tabcolsep}{0.5pt}
    \addtolength{\belowcaptionskip}{-10pt}
    {
    \begin{tabular}{c c c c c c c c}

        \\

        \raisebox{0.2in}{\rotatebox{90}{ Input }} &
        \includegraphics[width=0.1\textwidth]{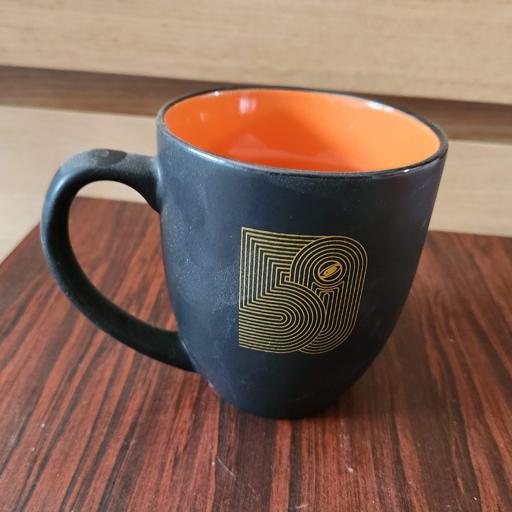} &
        \includegraphics[width=0.1\textwidth]{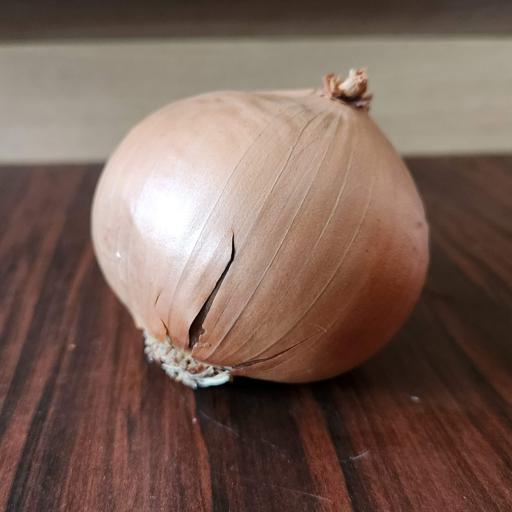} &
        \includegraphics[width=0.1\textwidth]{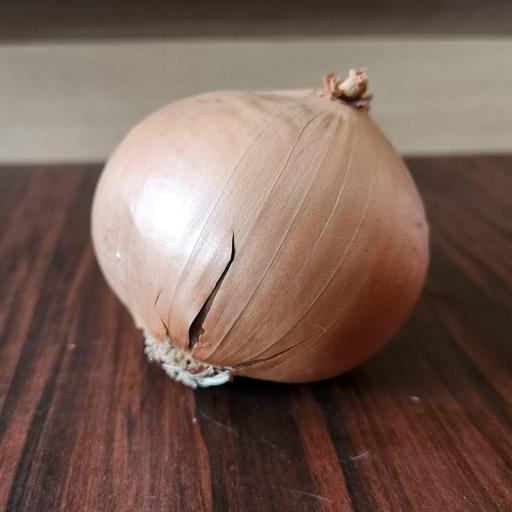} &
        \includegraphics[width=0.1\textwidth]{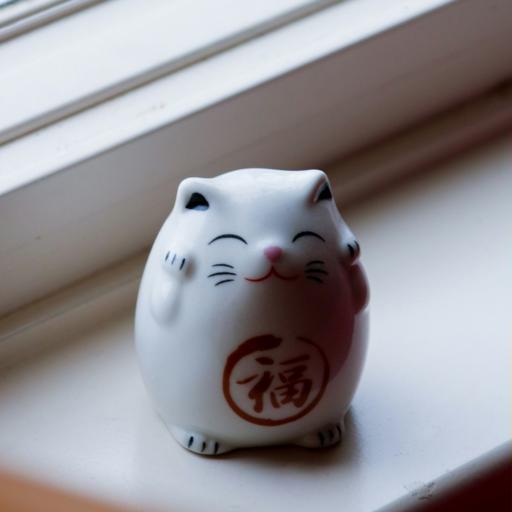} &
        \includegraphics[width=0.1\textwidth]{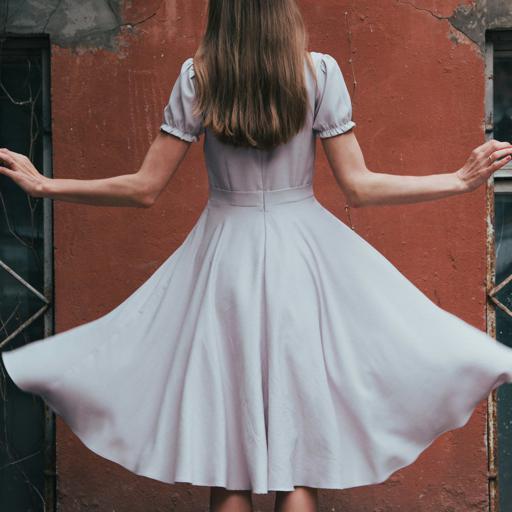} &
        \includegraphics[width=0.1\textwidth]{figures/instruct_comparisons/images/claire-abdo-_-635EI3nV8-unsplash_shattered/claire-abdo-_-635EI3nV8-unsplash.jpg} &
        \includegraphics[width=0.1\textwidth]{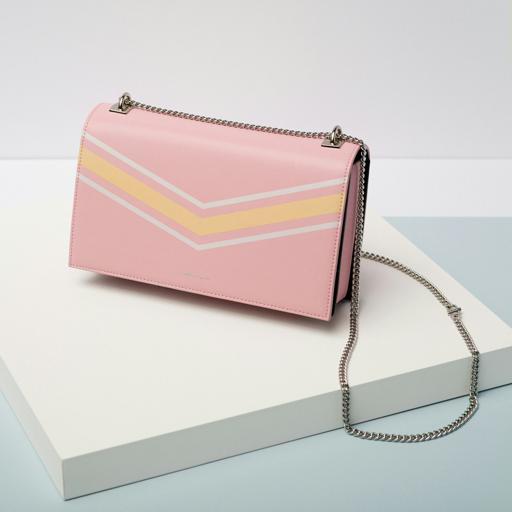} \\

        & ``futuristic'' & ``many'' & ``sliced'' & ``colorful'' & ``gothic'' & ``shattered'' & ``colorful'' \\
\end{tabular} 

    \begin{tabular}{c c c c c c c c}

        \raisebox{0.2in}{\rotatebox{90}{ Input }} &
        \includegraphics[width=0.1\textwidth]{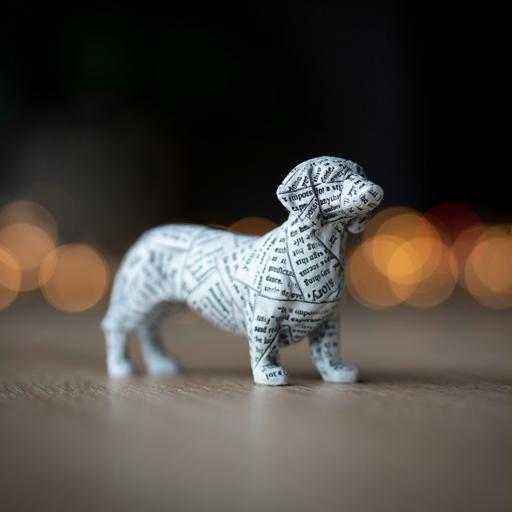} &
        \includegraphics[width=0.1\textwidth]{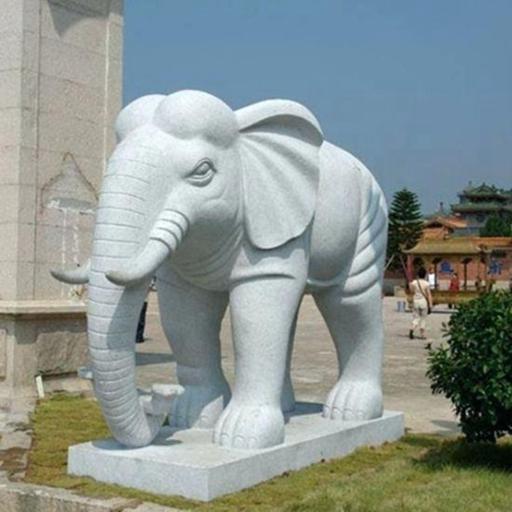} &
        \includegraphics[width=0.1\textwidth]{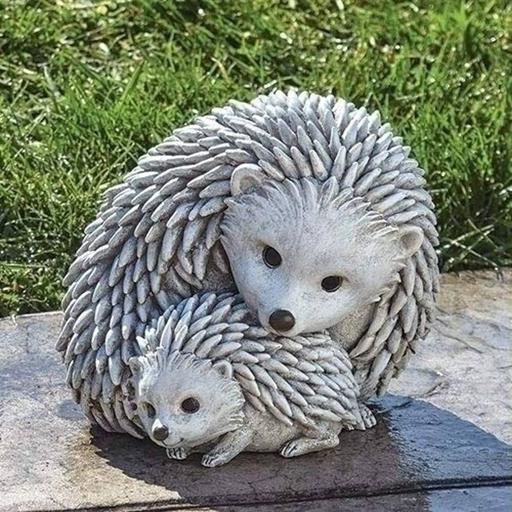} &
        \includegraphics[width=0.1\textwidth]{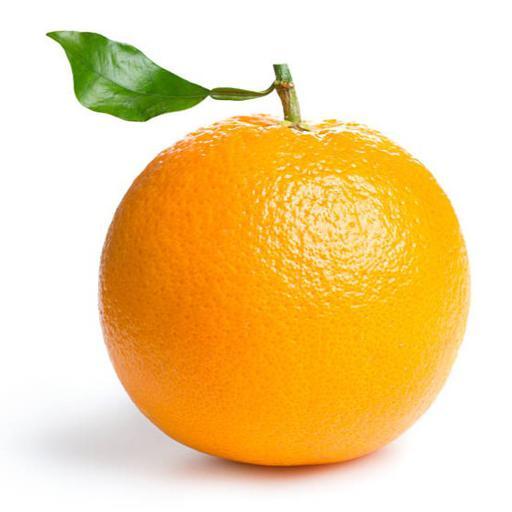} &
        \includegraphics[width=0.1\textwidth]{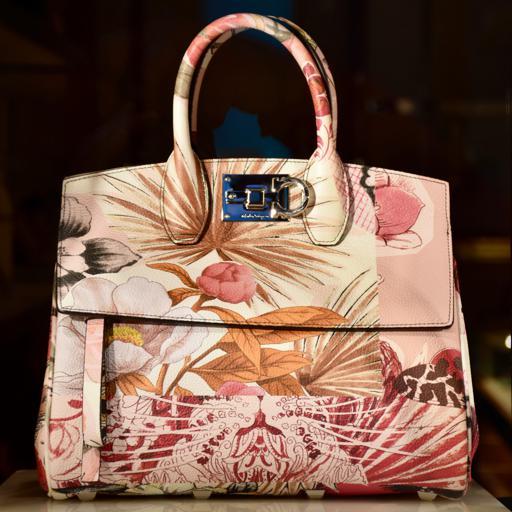} &
        \includegraphics[width=0.1\textwidth]{figures/instruct_comparisons/images/mario-losereit-mTZyJeR1Rnc-unsplash_many/mario-losereit-mTZyJeR1Rnc-unsplash.jpg} &
        \includegraphics[width=0.1\textwidth]{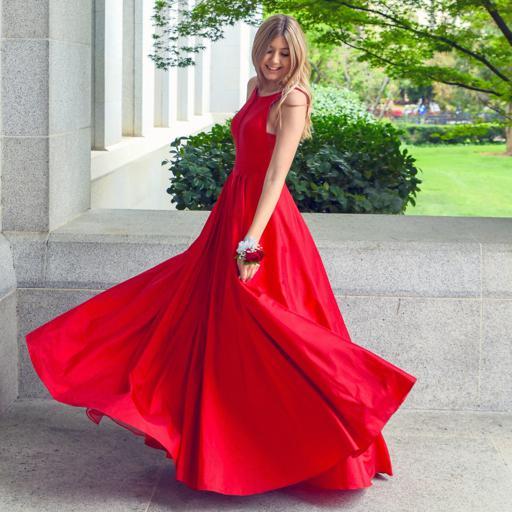} \\

        & ``skeletal'' & ``enormous'' & ``futuristic'' & ``sliced'' & ``transparent'' & ``many'' & ``soggy'' \\
\end{tabular}

    }
    \caption{Additional qualitative comparison for the \textit{pOps} instruct operator to alternative instruction-based editing approaches: InstructPix2Pix~\cite{brooks2022instructpix2pix} and two variants of IP-Adapter~\cite{ye2023ip-adapter}.}
    \label{fig:supp_instruct_comparisons}
\end{figure*}

\begin{figure*}
    \centering
    \setlength{\tabcolsep}{0.5pt}
    {
    \begin{tabular}{c c c c @{\hspace{0.4cm}} c c c @{\hspace{0.4cm}} c c c}

        \\ \\ \\ \\ \\ \\ \\ \\
\end{tabular}
    \begin{tabular}{c c c c @{\hspace{0.4cm}} c c c}

        \raisebox{0.15in}{\rotatebox{90}{ Input }} &
        \includegraphics[width=0.095\textwidth]{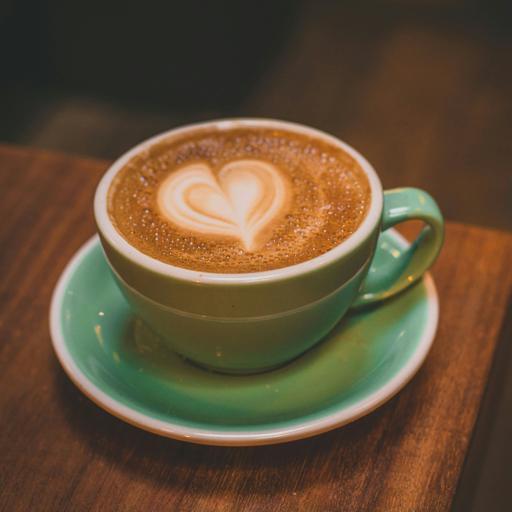} &
        \includegraphics[width=0.095\textwidth]{figures/instruct_comparisons/neti_comparison/latte_art/pablo-merchan-montes-_Tw4vCs9C-8-unsplash.jpg} &
        \includegraphics[width=0.095\textwidth]{figures/instruct_comparisons/neti_comparison/latte_art/pablo-merchan-montes-_Tw4vCs9C-8-unsplash.jpg} &
        
        \includegraphics[width=0.095\textwidth]{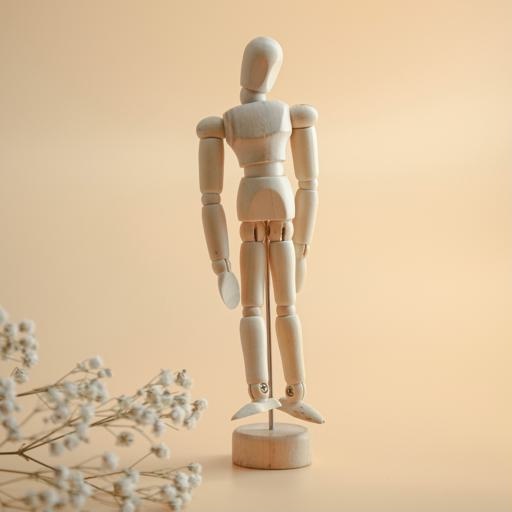} &
        \includegraphics[width=0.095\textwidth]{figures/instruct_comparisons/neti_comparison/stick_man/thoa-ngo-AZr6AOMu3l8-unsplas.jpg} &
        \includegraphics[width=0.095\textwidth]{figures/instruct_comparisons/neti_comparison/stick_man/thoa-ngo-AZr6AOMu3l8-unsplas.jpg} \\

        & ``charcoal'' & ``frosted'' & ``two'' & ``muddy'' & ``melting'' & ``comic''  \\

        \raisebox{0.175in}{\rotatebox{90}{ NeTI }} &
        \includegraphics[width=0.095\textwidth]{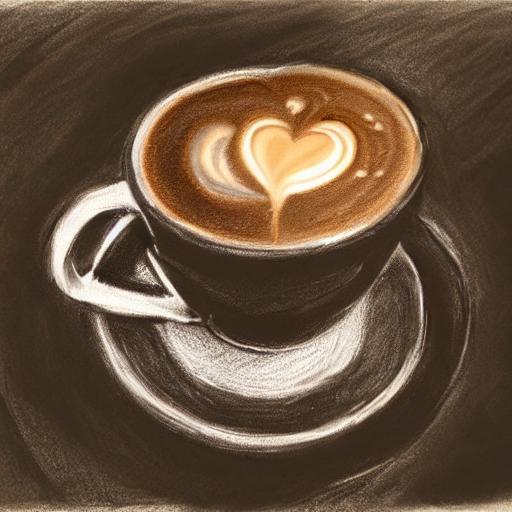} &
        \includegraphics[width=0.095\textwidth]{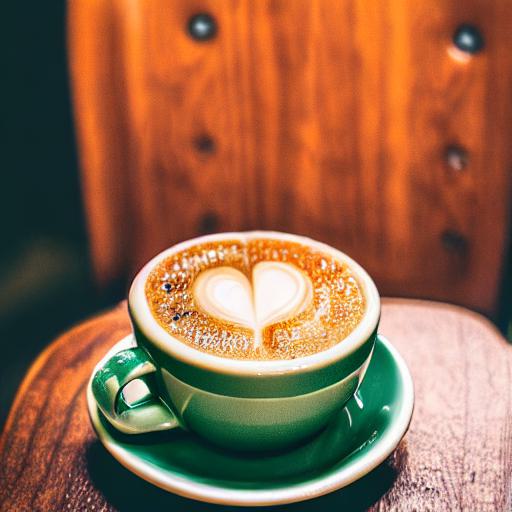} &
        \includegraphics[width=0.095\textwidth]{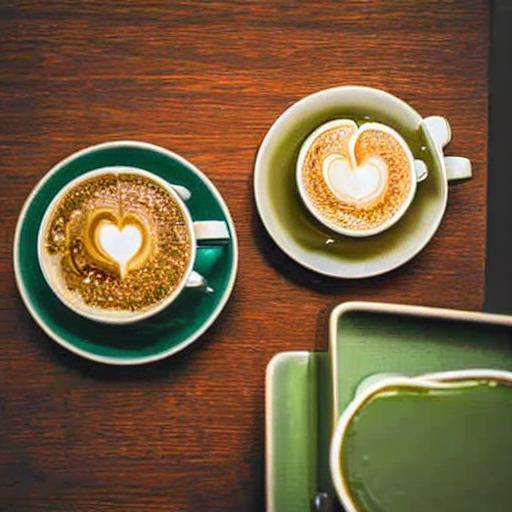} &
        
        \includegraphics[width=0.095\textwidth]{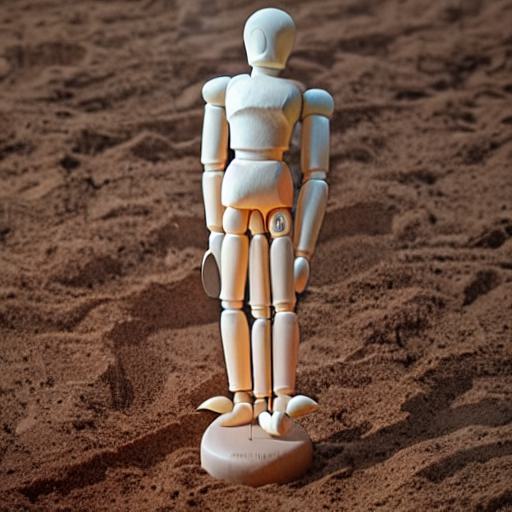} &
        \includegraphics[width=0.095\textwidth]{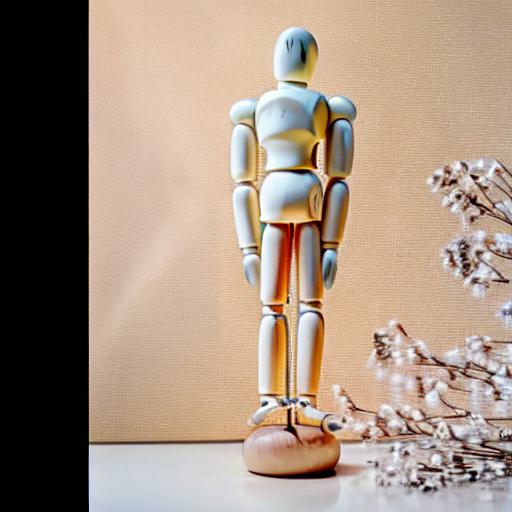} &
        \includegraphics[width=0.095\textwidth]{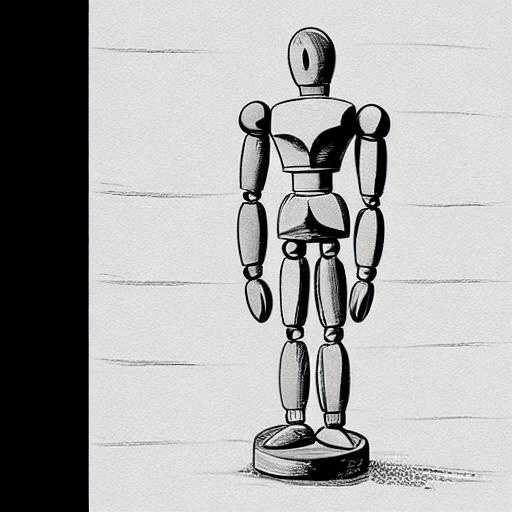} \\
\end{tabular}

    }
    \caption{Qualitative comparison for the \textit{pOps} instruct operator compared to NeTI~\cite{alaluf2023neural}, an optimization-based personalization technique.}
    \label{fig:instruct_comparisons_neti}
\end{figure*}

\begin{figure*}
    \centering
    \setlength{\tabcolsep}{0.5pt}
    \addtolength{\belowcaptionskip}{-10pt}
    {
    \begin{tabular}{c @{\hspace{0.2cm}} c c c c}

        \\ \\

        \includegraphics[width=0.15\textwidth]{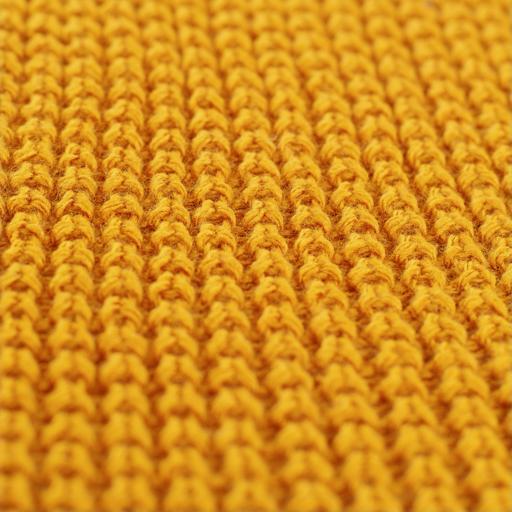} &
        \includegraphics[width=0.15\textwidth]{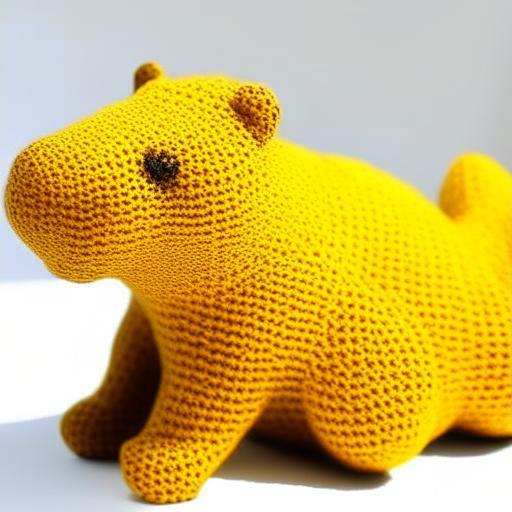} &
        \includegraphics[width=0.15\textwidth]{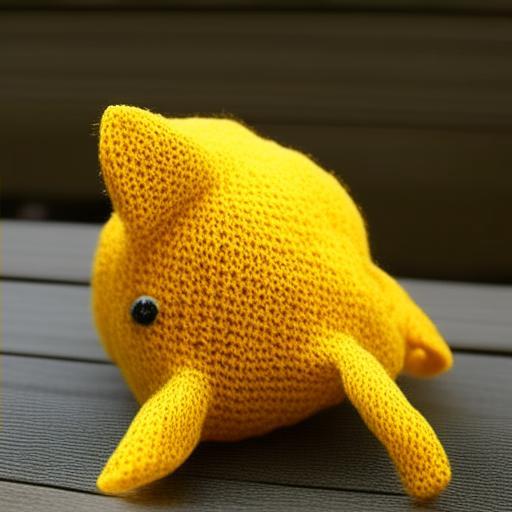} &
        \includegraphics[width=0.15\textwidth]{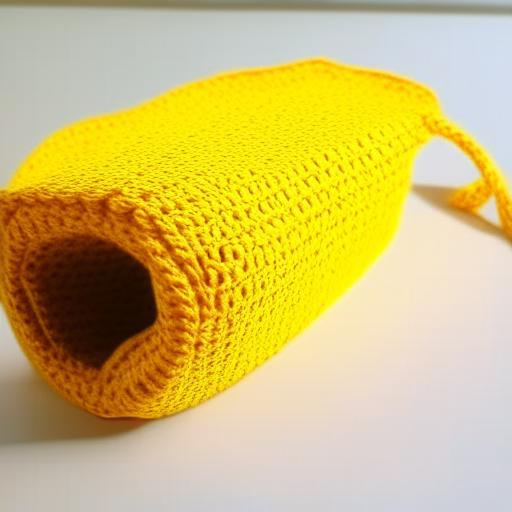} &
        \includegraphics[width=0.15\textwidth]{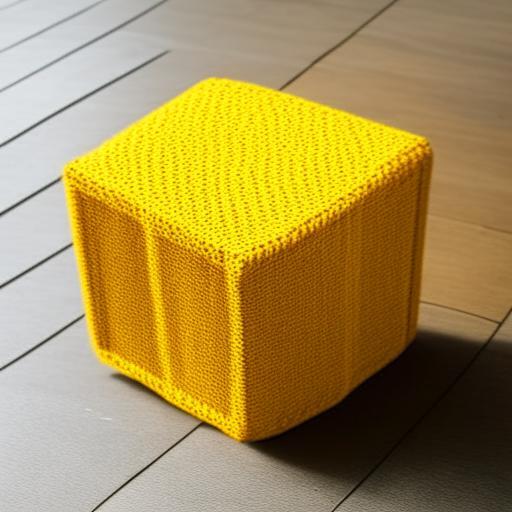} \\

        \includegraphics[width=0.15\textwidth]{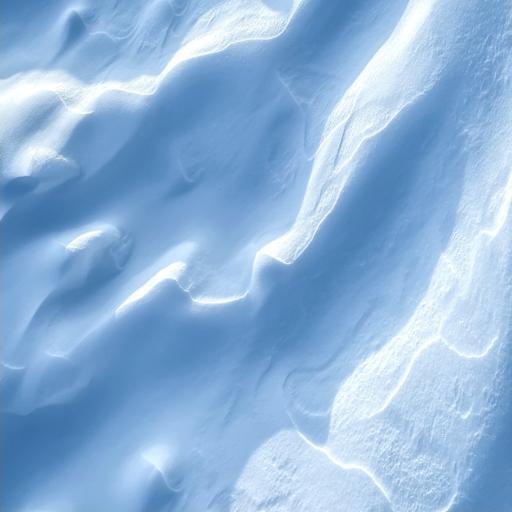} &
        \includegraphics[width=0.15\textwidth]{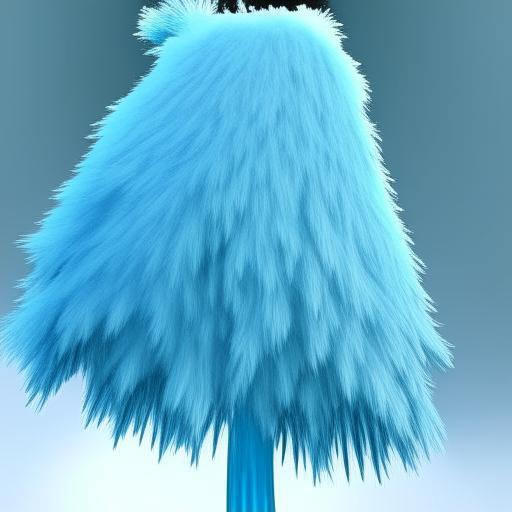} &
        \includegraphics[width=0.15\textwidth]{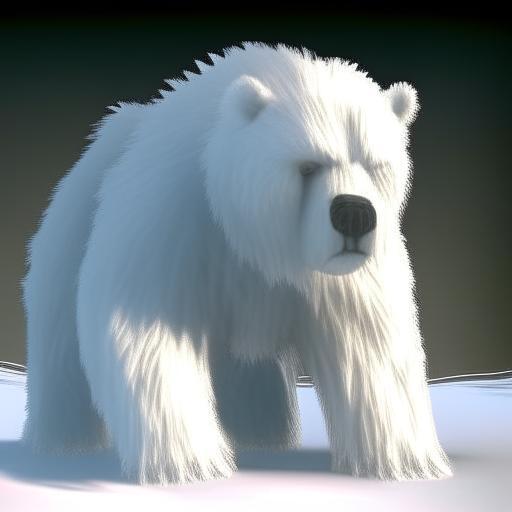} &
        \includegraphics[width=0.15\textwidth]{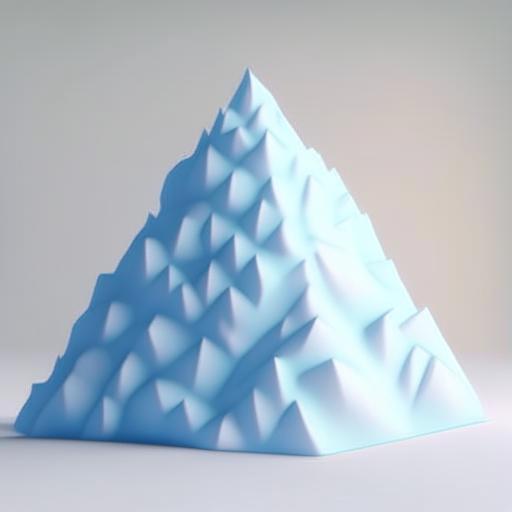} &
        \includegraphics[width=0.15\textwidth]{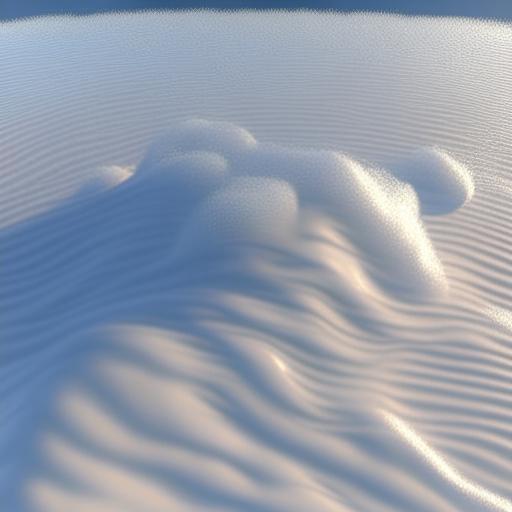} \\

        \includegraphics[width=0.15\textwidth]{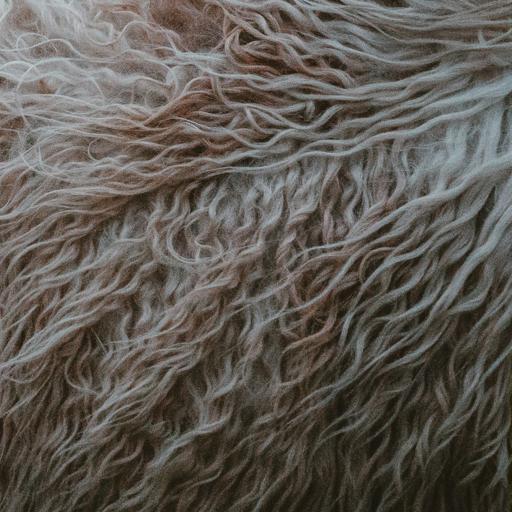} &
        \includegraphics[width=0.15\textwidth]{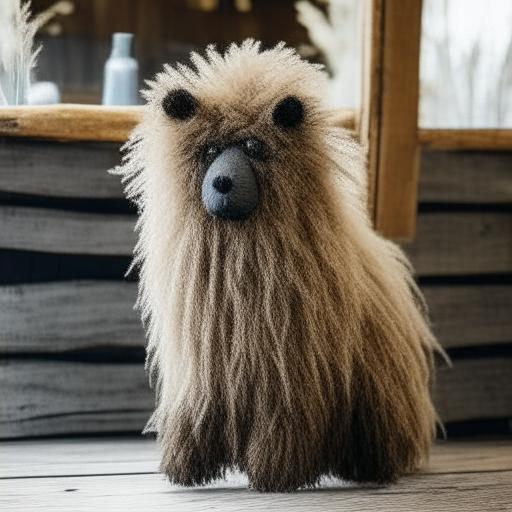} &
        \includegraphics[width=0.15\textwidth]{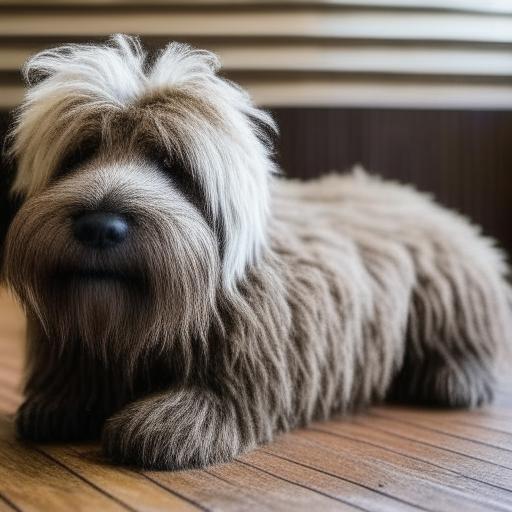} &
        \includegraphics[width=0.15\textwidth]{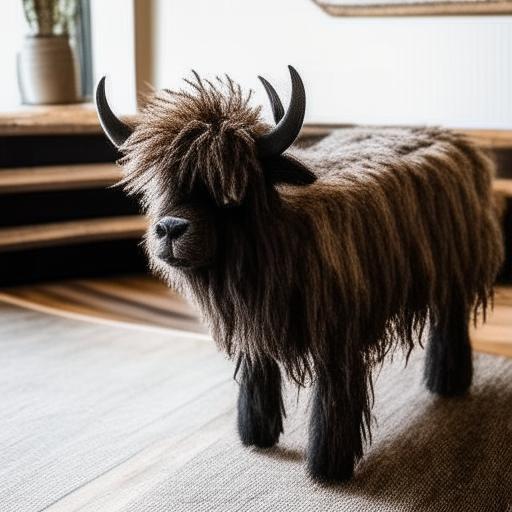} &
        \includegraphics[width=0.15\textwidth]{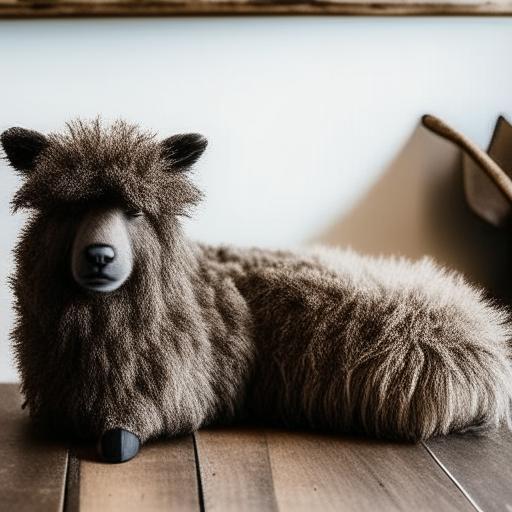} \\

        \includegraphics[width=0.15\textwidth]{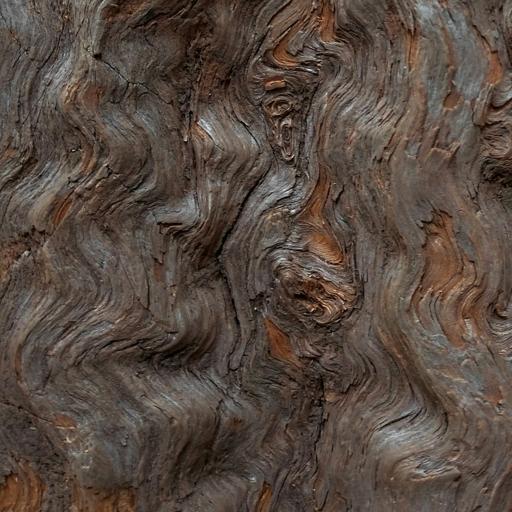} &
        \includegraphics[width=0.15\textwidth]{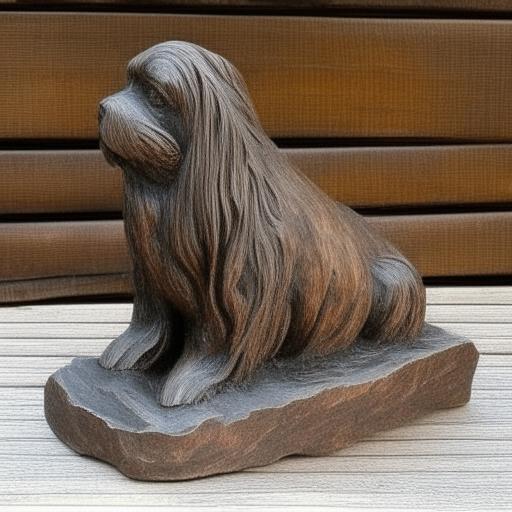} &
        \includegraphics[width=0.15\textwidth]{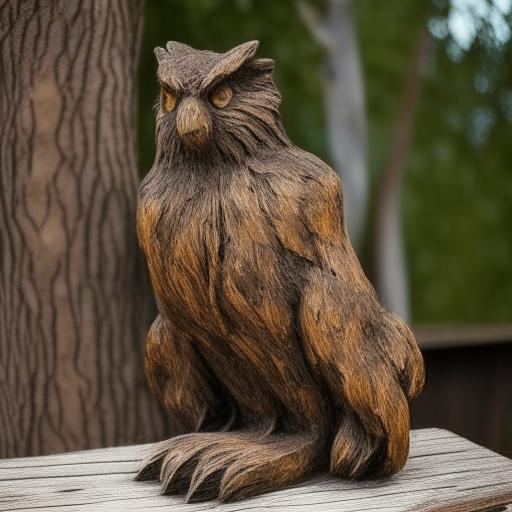} &
        \includegraphics[width=0.15\textwidth]{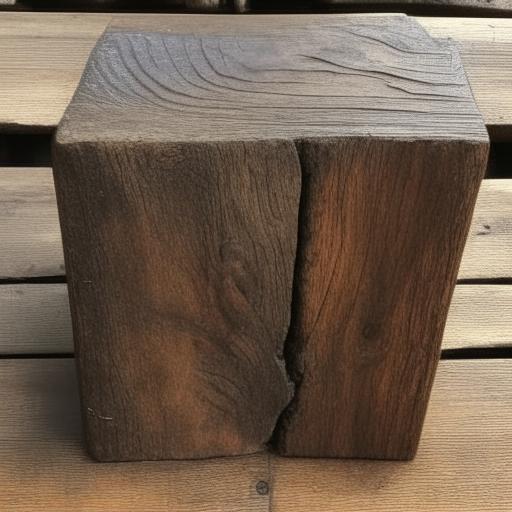} &
        \includegraphics[width=0.15\textwidth]{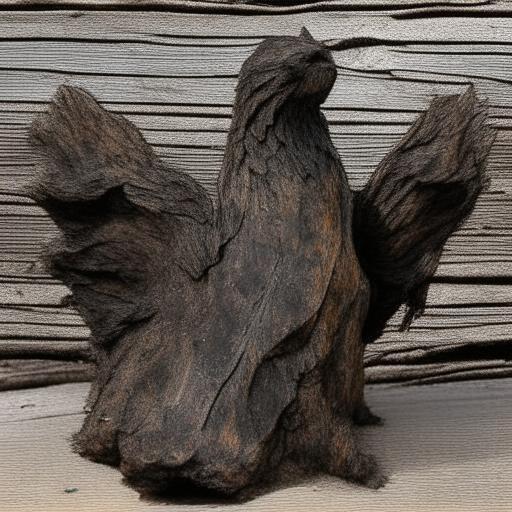} \\
\end{tabular}
    }
    \caption{Results obtained by our texturing model when null inputs are passed in place of the object input.}
    \label{fig:null_object}
\end{figure*}

\begin{figure*}
    \centering
    \setlength{\tabcolsep}{0.5pt}
    \addtolength{\belowcaptionskip}{-10pt}
    {
    \begin{tabular}{c @{\hspace{0.2cm}} c c c c}

        \\ \\

        \includegraphics[width=0.15\textwidth]{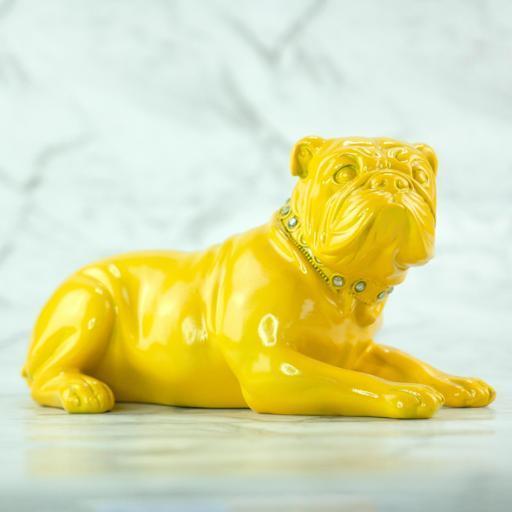} &
        \includegraphics[width=0.15\textwidth]{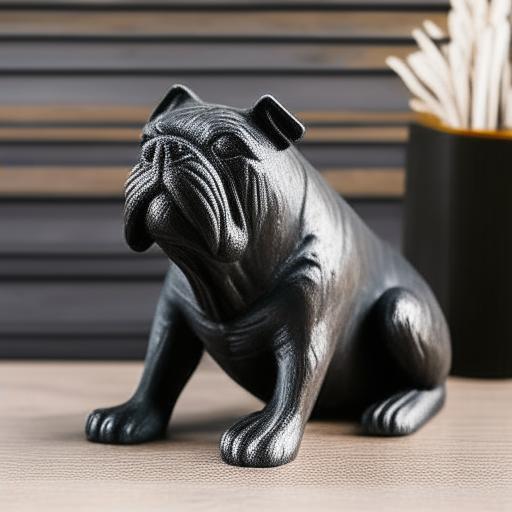} &
        \includegraphics[width=0.15\textwidth]{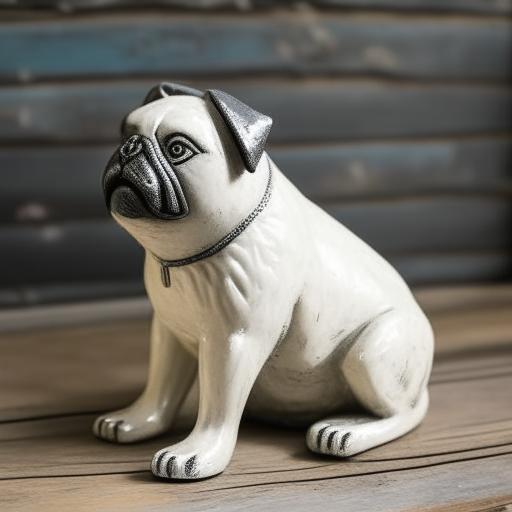} &
        \includegraphics[width=0.15\textwidth]{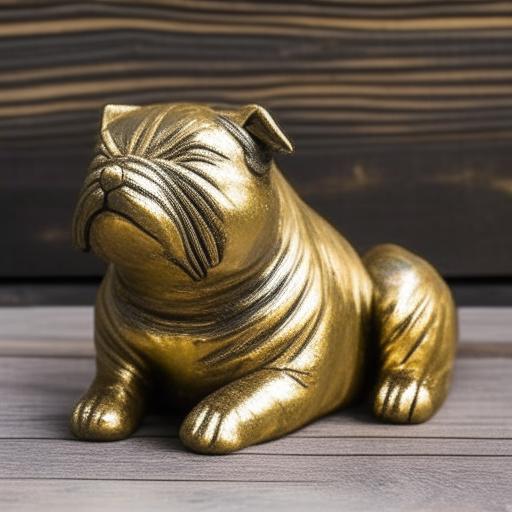} &
        \includegraphics[width=0.15\textwidth]{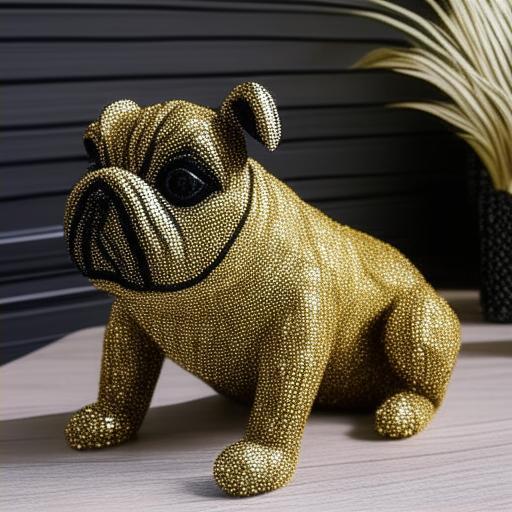} \\

        \includegraphics[width=0.15\textwidth]{figures/base_results/images/inputs/camel_in.jpg} &
        \includegraphics[width=0.15\textwidth]{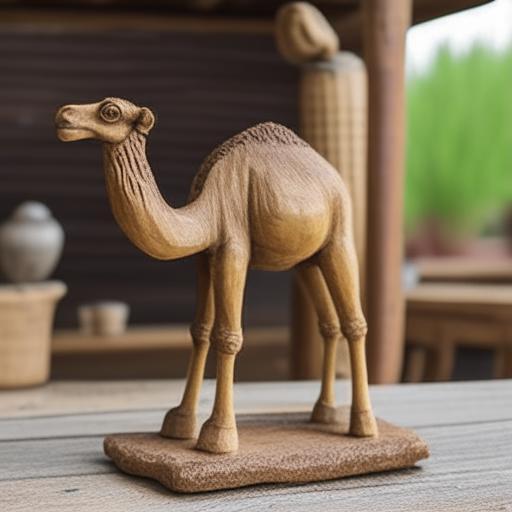} &
        \includegraphics[width=0.15\textwidth]{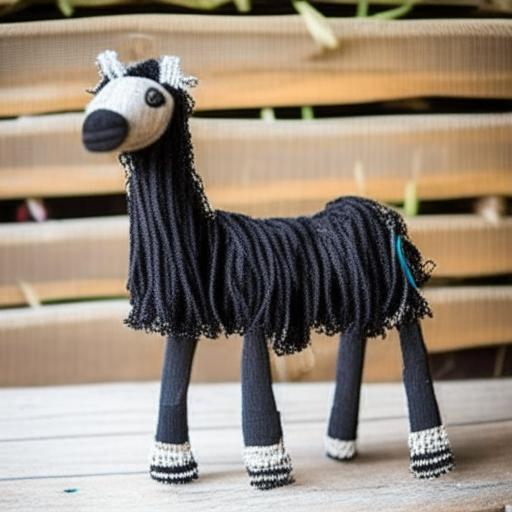} &
        \includegraphics[width=0.15\textwidth]{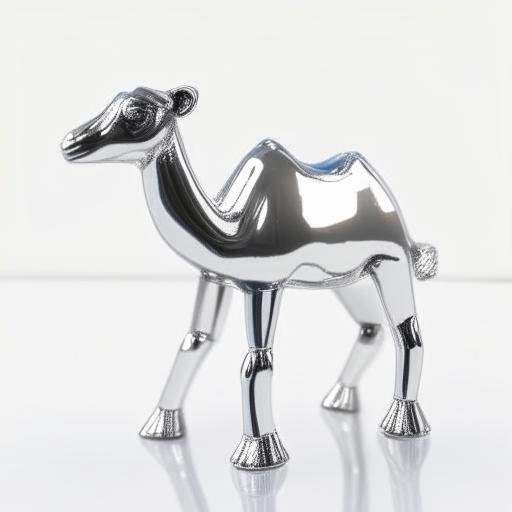} &
        \includegraphics[width=0.15\textwidth]{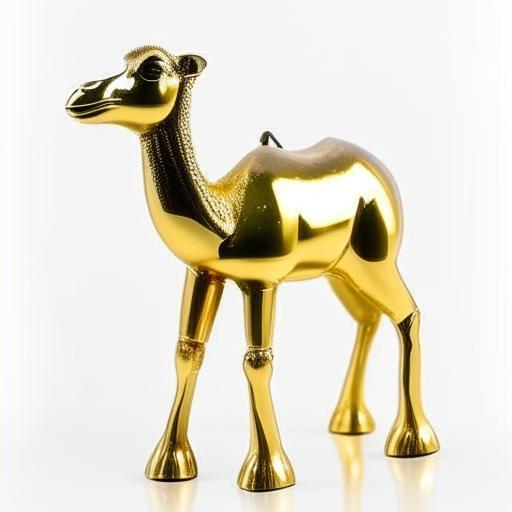} \\

        \includegraphics[width=0.15\textwidth]{figures/base_results/images/inputs/dog_words_in.jpg} &
        \includegraphics[width=0.15\textwidth]{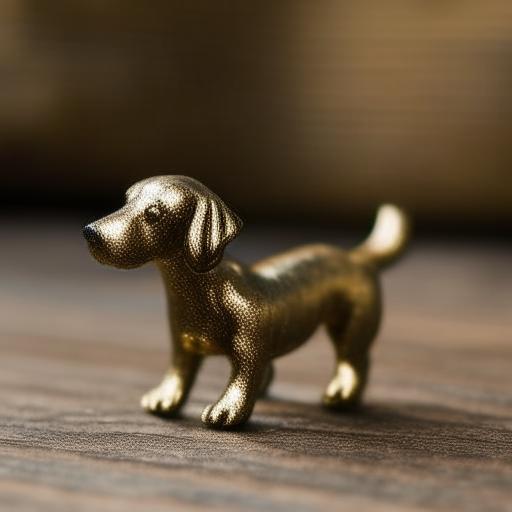} &
        \includegraphics[width=0.15\textwidth]{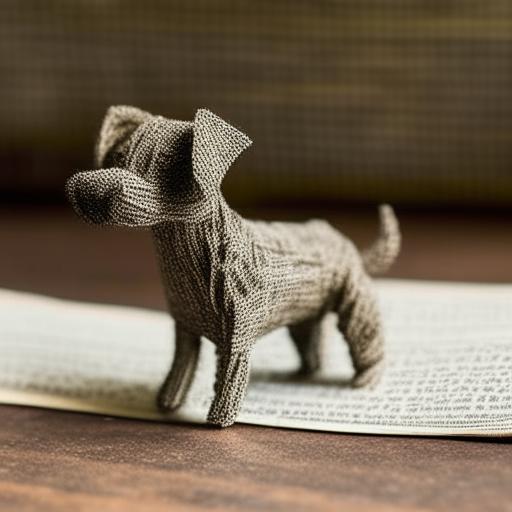} &
        \includegraphics[width=0.15\textwidth]{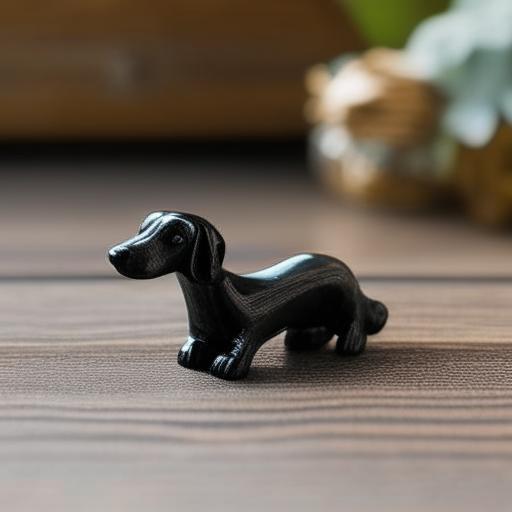} &
        \includegraphics[width=0.15\textwidth]{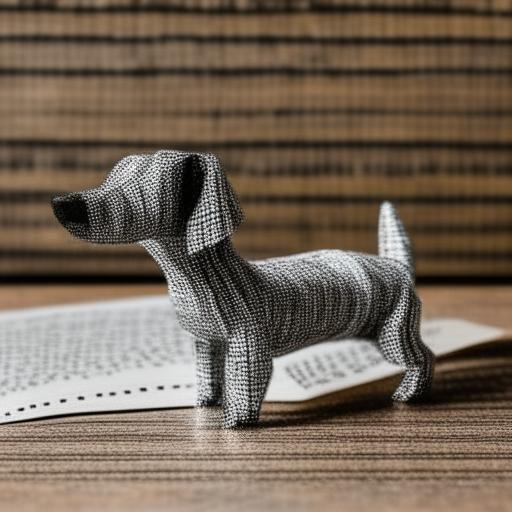} \\

       \includegraphics[width=0.15\textwidth]{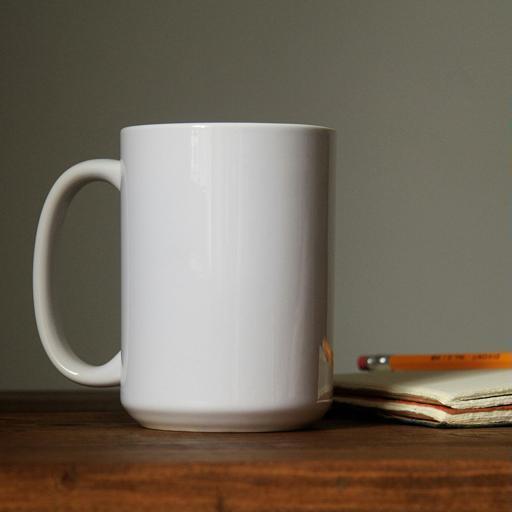} &
        \includegraphics[width=0.15\textwidth]{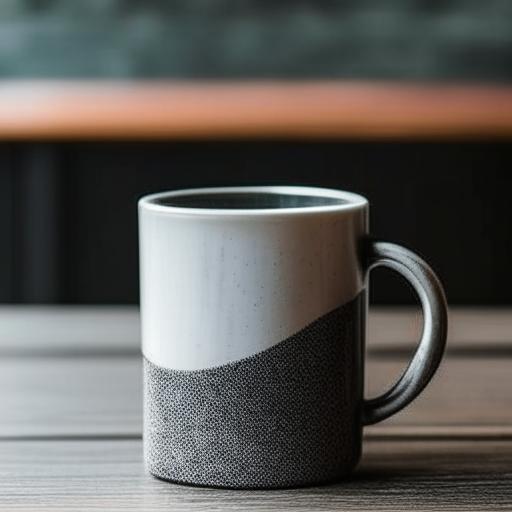} &
        \includegraphics[width=0.15\textwidth]{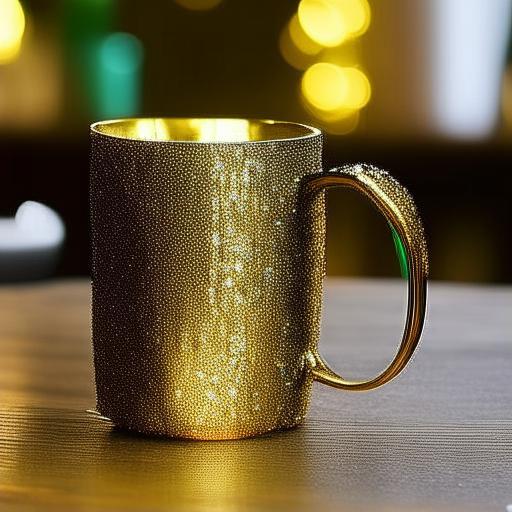} &
        \includegraphics[width=0.15\textwidth]{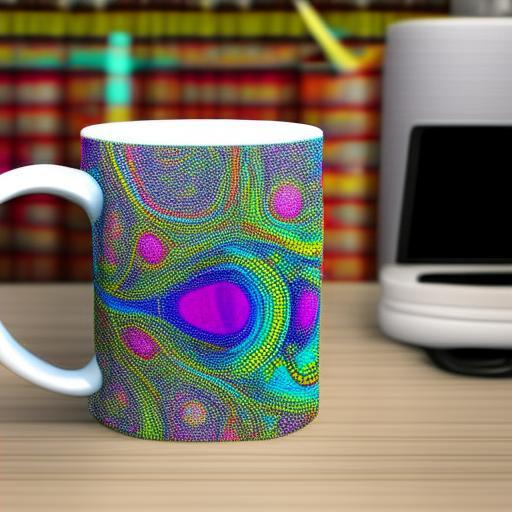} &
        \includegraphics[width=0.15\textwidth]{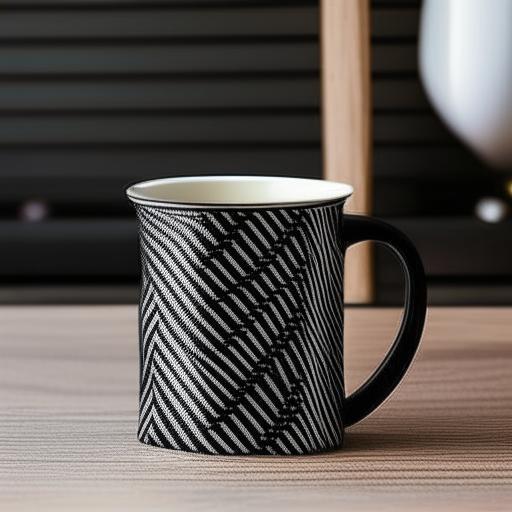} \\
\end{tabular}
    }
    \caption{Results obtained by our texturing model when null inputs are passed in place of the texture input.}
    \label{fig:null}
\end{figure*}

\begin{figure*}
    \centering
    \setlength{\tabcolsep}{0.5pt}
    \addtolength{\belowcaptionskip}{-10pt}
    {\small
    \begin{tabular}{c c @{\hspace{0.4cm}} c c @{\hspace{0.4cm}} c c}

    \end{tabular}
    \begin{tabular}{c c c @{\hspace{0.4cm}} c c c}

        \\ \\ \\ \\
        
        \includegraphics[width=0.15\textwidth]{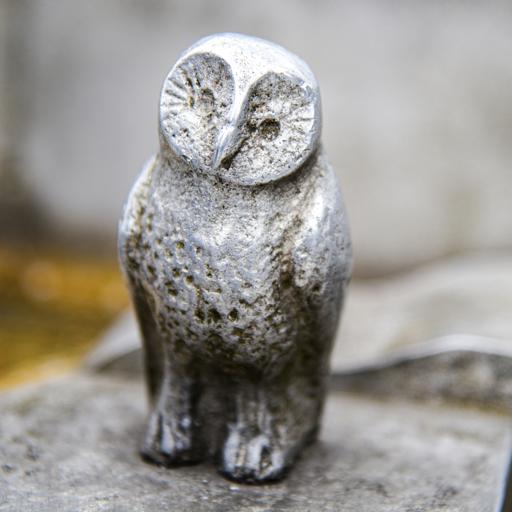} &
        \includegraphics[width=0.15\textwidth]{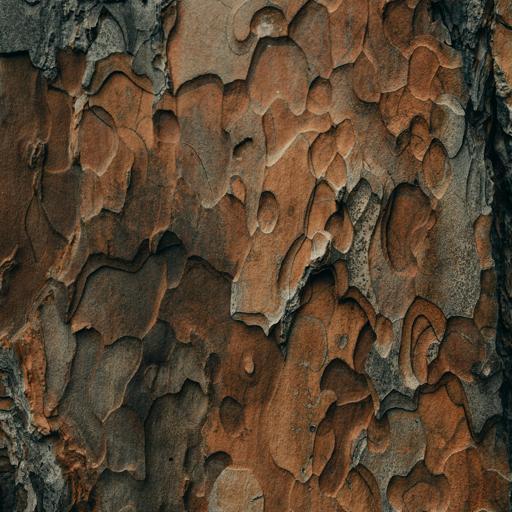} &
        \includegraphics[width=0.15\textwidth]{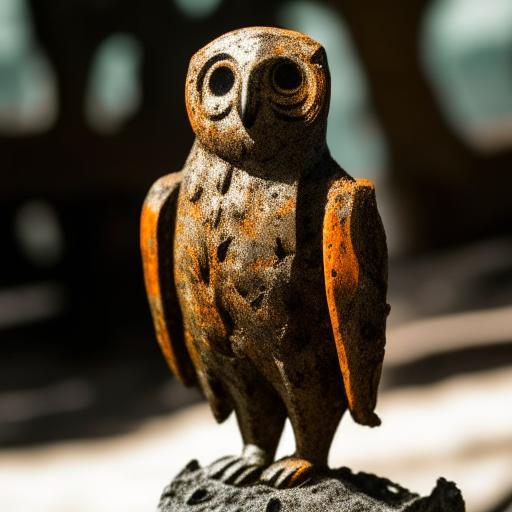} &
        \includegraphics[width=0.15\textwidth]{figures/base_results/images/inputs/owl_statue_in.jpg} &
        \includegraphics[width=0.15\textwidth]{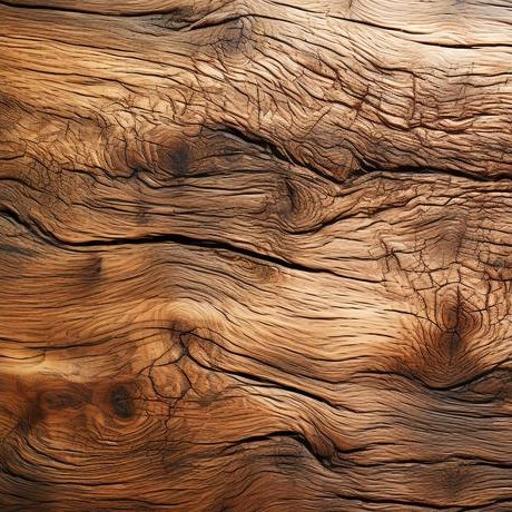} &
        \includegraphics[width=0.15\textwidth]{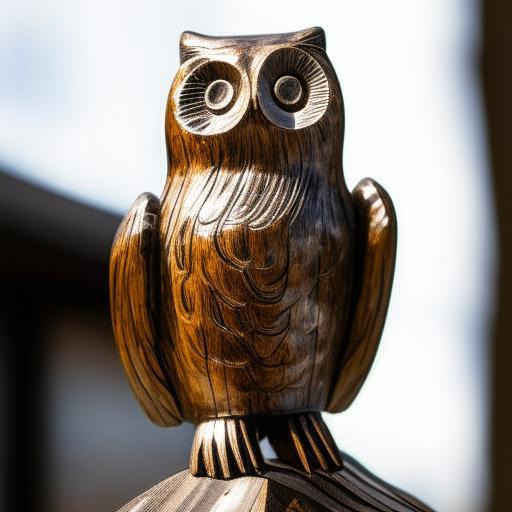}
        
        \\
        Object & Texture & Result & Object & Texture & Result \\

        \includegraphics[width=0.15\textwidth]{figures/base_results/images/inputs/owl_statue_in.jpg} &
        \includegraphics[width=0.15\textwidth]{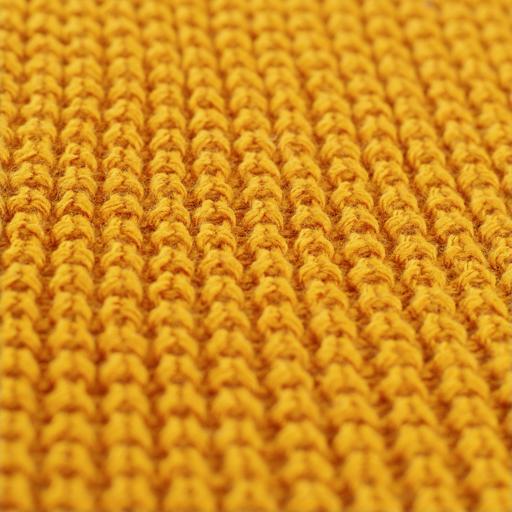} &
        \includegraphics[width=0.15\textwidth]{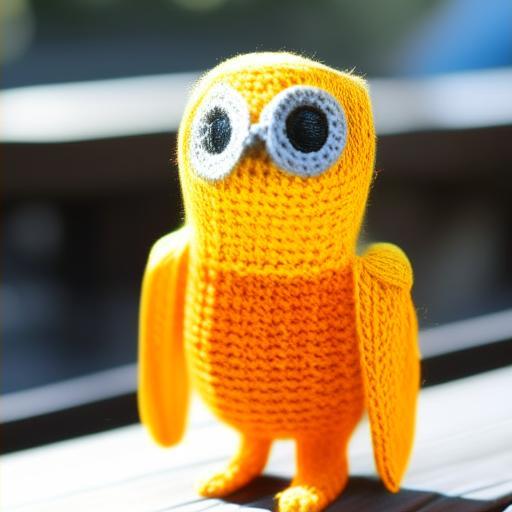} &
        \includegraphics[width=0.15\textwidth]{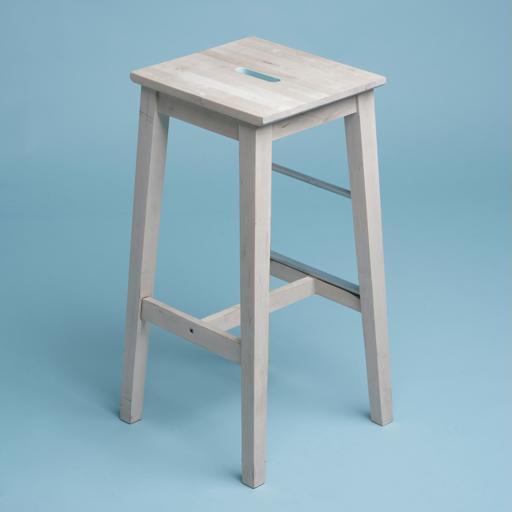} &
        \includegraphics[width=0.15\textwidth]{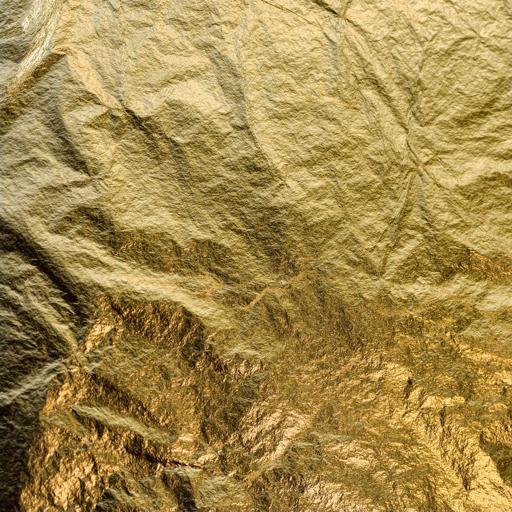} &
        \includegraphics[width=0.15\textwidth]{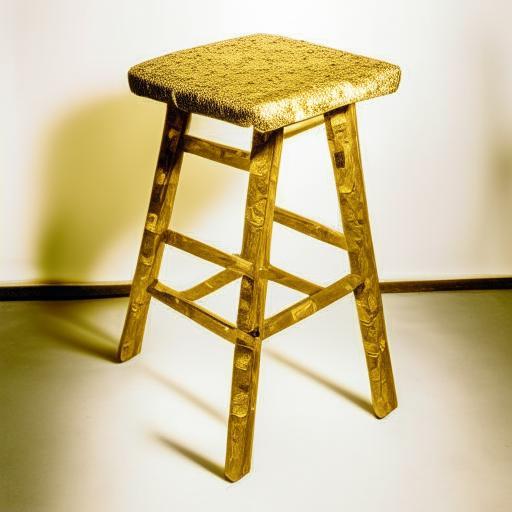} 
            
        \\
        Object & Texture & Result & Object & Texture & Result \\
\end{tabular}
    }
    \caption{Additional texturing results obtained by our \textit{pOps} method.}
    \label{fig:texturing_operator_1}
\end{figure*}

\begin{figure*}
    \centering
    \setlength{\tabcolsep}{0.5pt}
    {\small
    \begin{tabular}{c c @{\hspace{0.4cm}} c c @{\hspace{0.4cm}} c c}

    \end{tabular}
    \begin{tabular}{c c c @{\hspace{0.4cm}} c c c}

         \\ \\ \\ \\ \\ \\
        
        \includegraphics[width=0.15\textwidth]{figures/base_results/images/inputs/moose.jpg} &
        \includegraphics[width=0.15\textwidth]{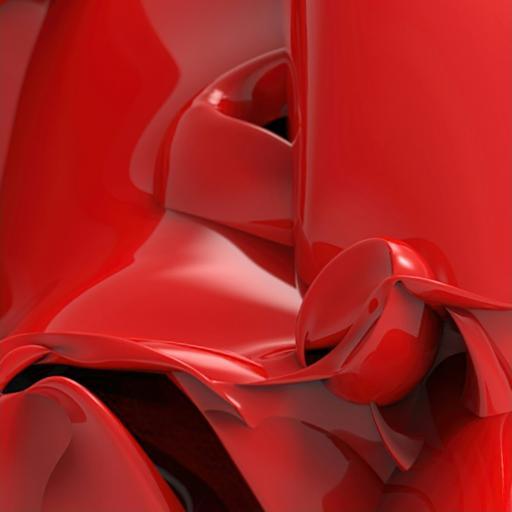} &
        \includegraphics[width=0.15\textwidth]{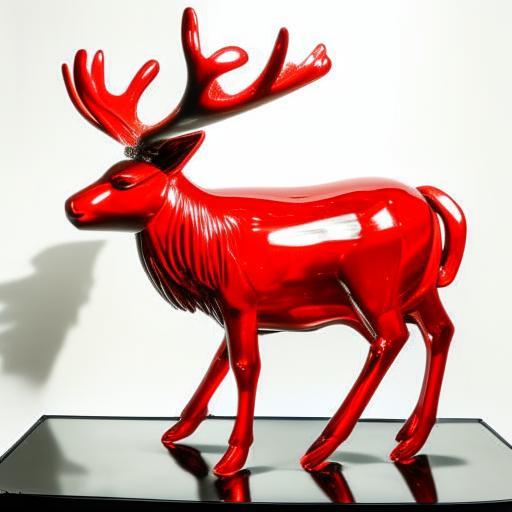} &
        \includegraphics[width=0.15\textwidth]{figures/base_results/images/inputs/moose.jpg} &
        \includegraphics[width=0.15\textwidth]{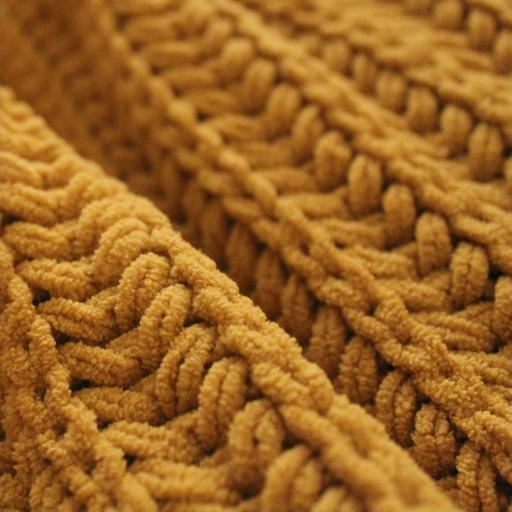} &
        \includegraphics[width=0.15\textwidth]{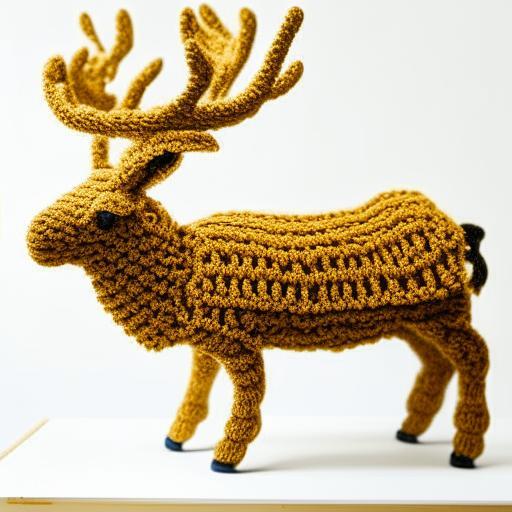}
        
        \\
        Object & Texture & Result & Object & Texture & Result \\
\end{tabular}
       \begin{tabular}{c @{\hspace{0.4cm}} c c c c c c c}

        \\

        \includegraphics[width=0.12\textwidth]{figures/base_results/images/inputs/red_dress_in_back.jpg} &
        \includegraphics[width=0.12\textwidth]{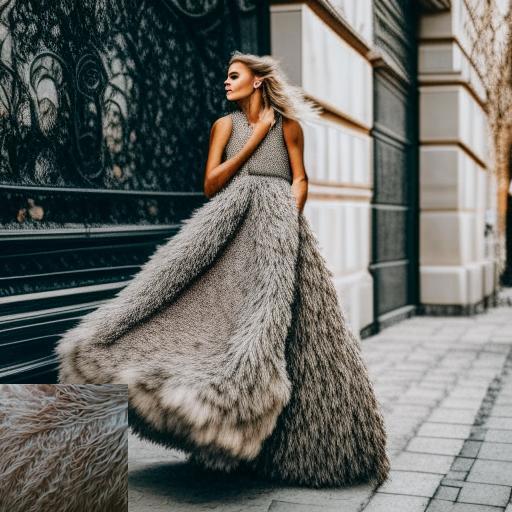} &
        \includegraphics[width=0.12\textwidth]{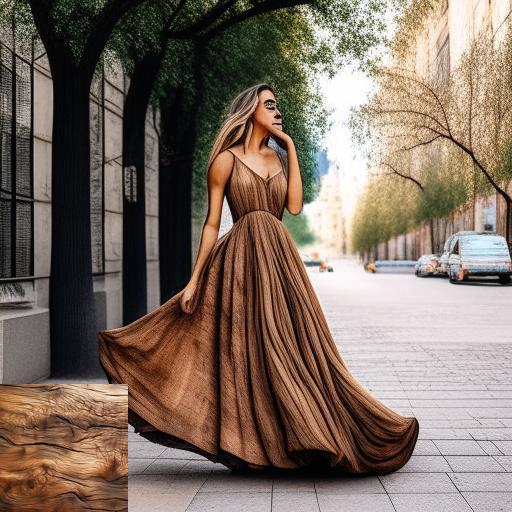} &
        \includegraphics[width=0.12\textwidth]{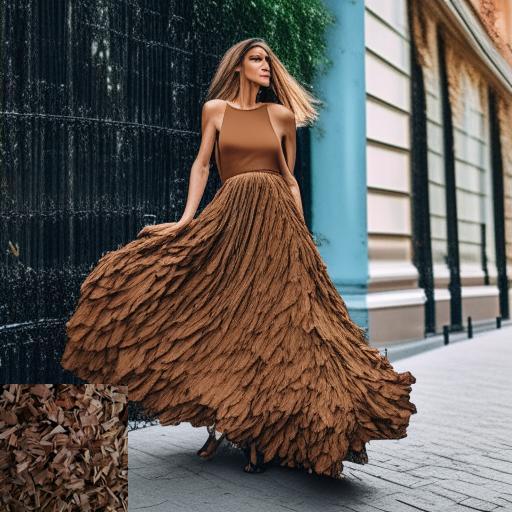} &
        \includegraphics[width=0.12\textwidth]{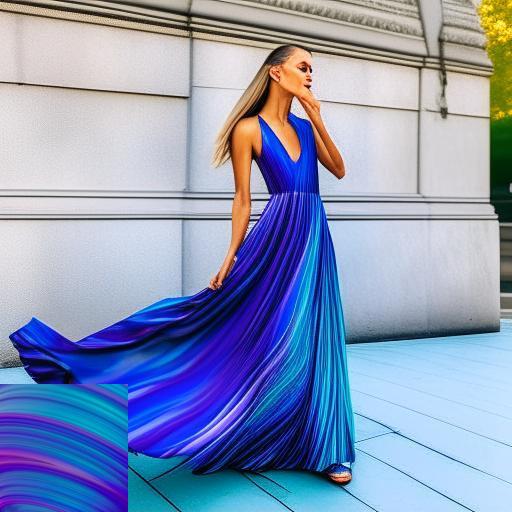} &
        \includegraphics[width=0.12\textwidth]{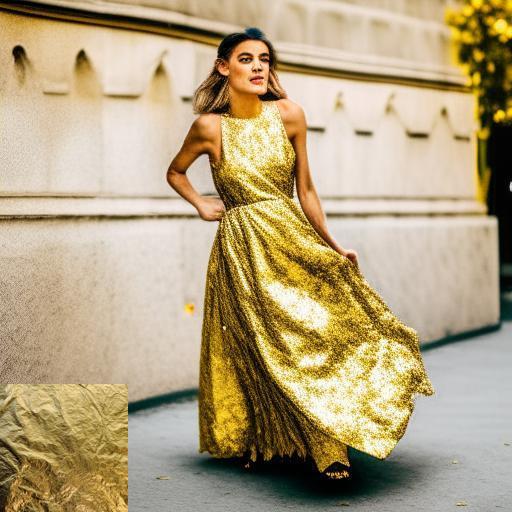} &
        \includegraphics[width=0.12\textwidth]{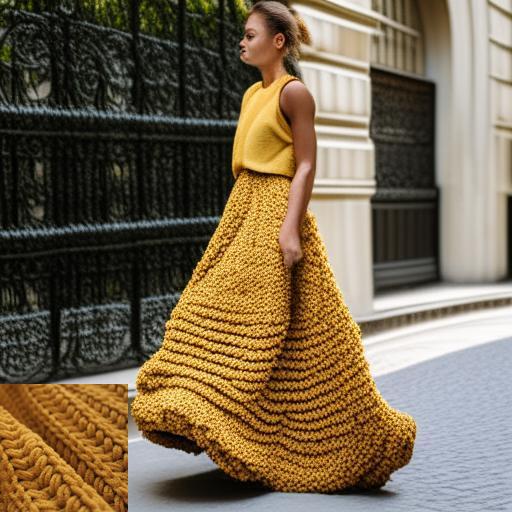} &
        \includegraphics[width=0.12\textwidth]{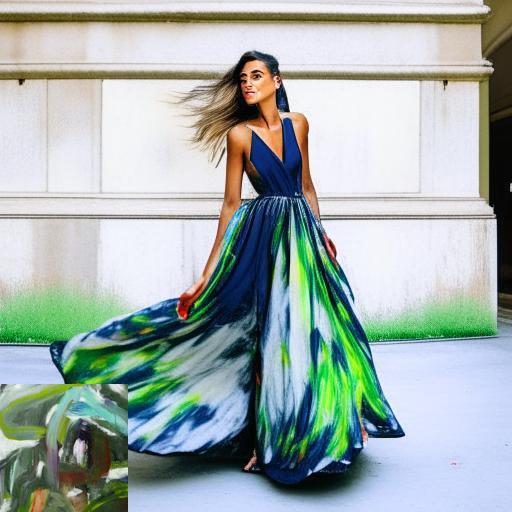} \\

        \includegraphics[width=0.12\textwidth]{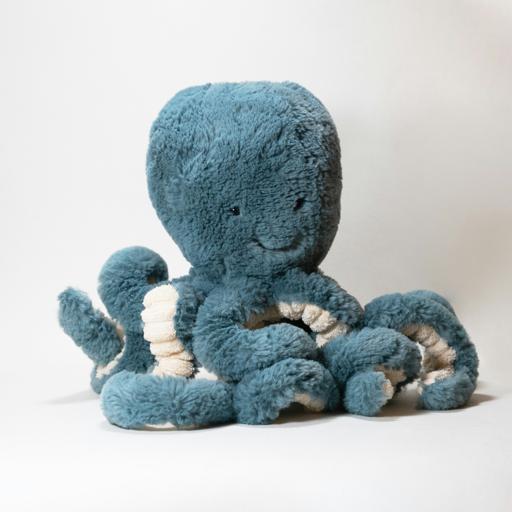} &
        \includegraphics[width=0.12\textwidth]{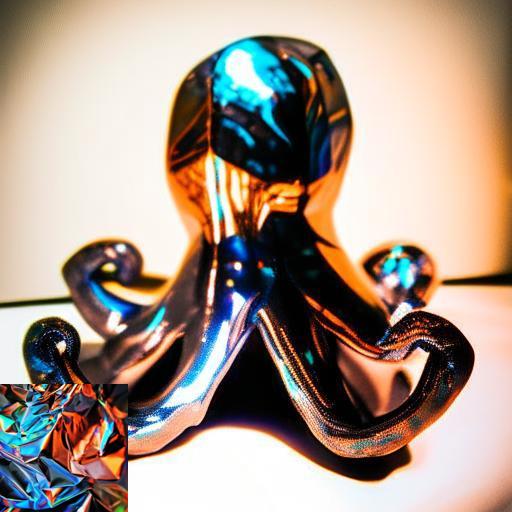} &
        \includegraphics[width=0.12\textwidth]{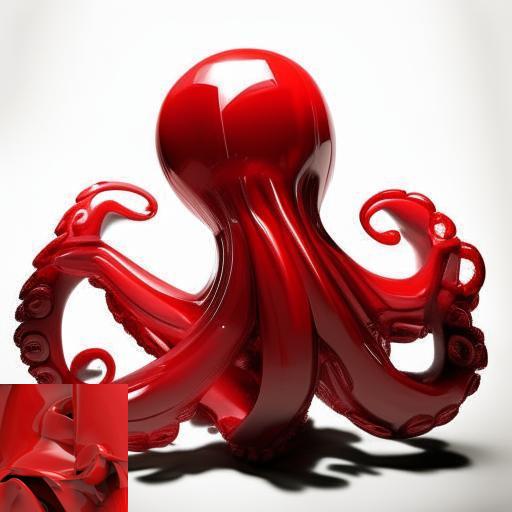} &
        \includegraphics[width=0.12\textwidth]{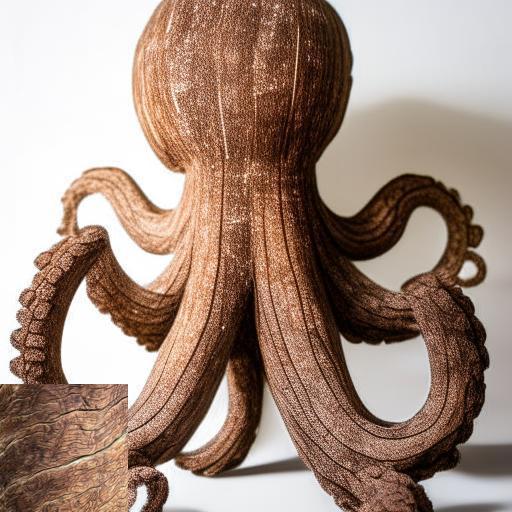} &
        \includegraphics[width=0.12\textwidth]{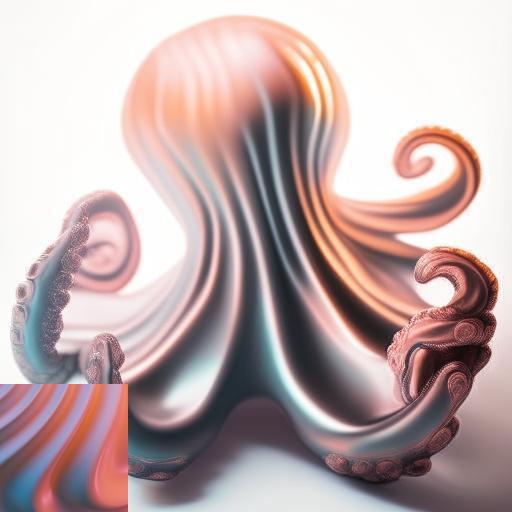} &
        \includegraphics[width=0.12\textwidth]{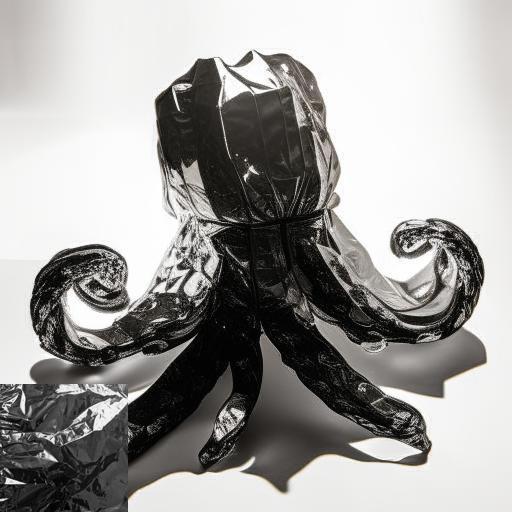} &
        \includegraphics[width=0.12\textwidth]{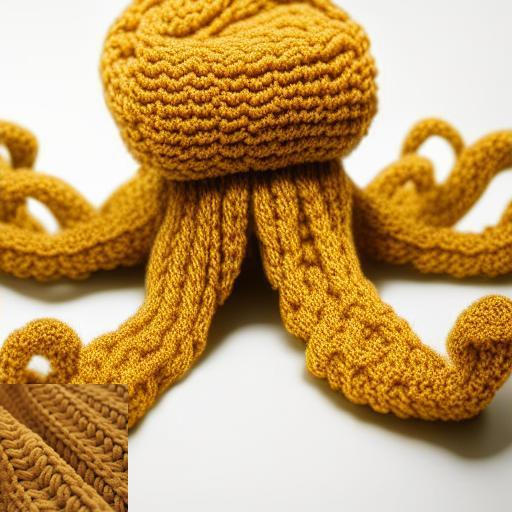} &
        \includegraphics[width=0.12\textwidth]{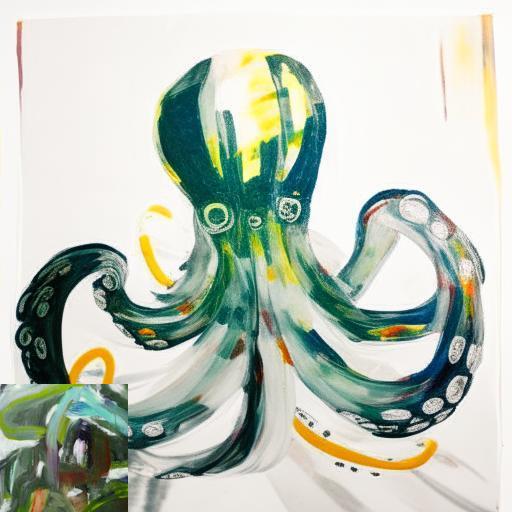} \\

    \end{tabular}
    }
    \caption{Additional texturing results obtained by our \textit{pOps} method.}
    \label{fig:texturing_operator_2}
\end{figure*}

\begin{figure*}
    \centering
    \setlength{\tabcolsep}{0.5pt}
    \addtolength{\belowcaptionskip}{-10pt}
    {
    \begin{tabular}{c @{\hspace{0.8cm}} c c c c c}

        \\[0.05cm]

        \includegraphics[width=0.12\textwidth]{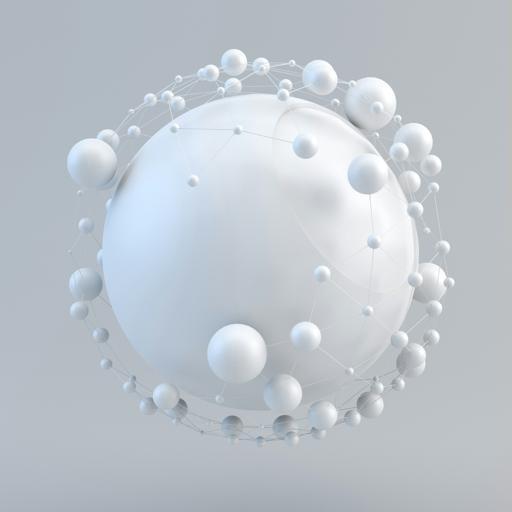} &
        \includegraphics[width=0.12\textwidth]{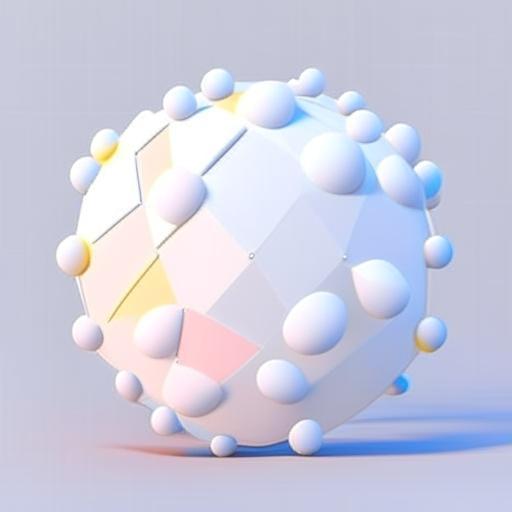} &
        \includegraphics[width=0.12\textwidth]{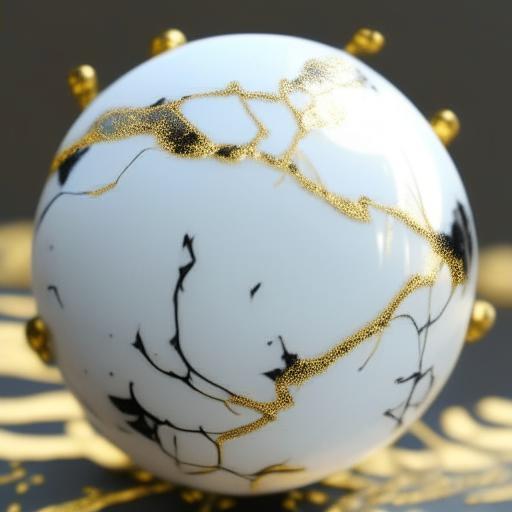} &
        \includegraphics[width=0.12\textwidth]{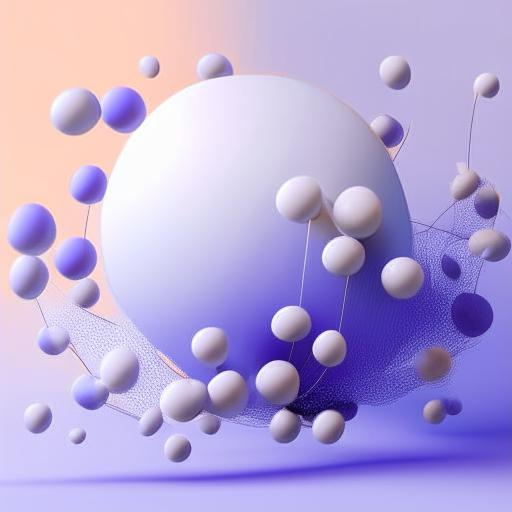} &
        \includegraphics[width=0.12\textwidth]{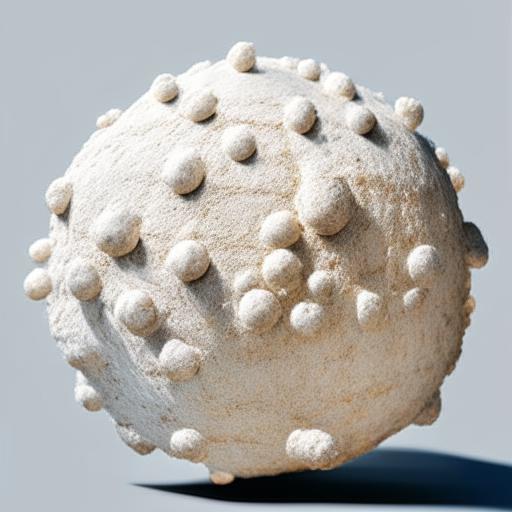} &
        \includegraphics[width=0.12\textwidth]{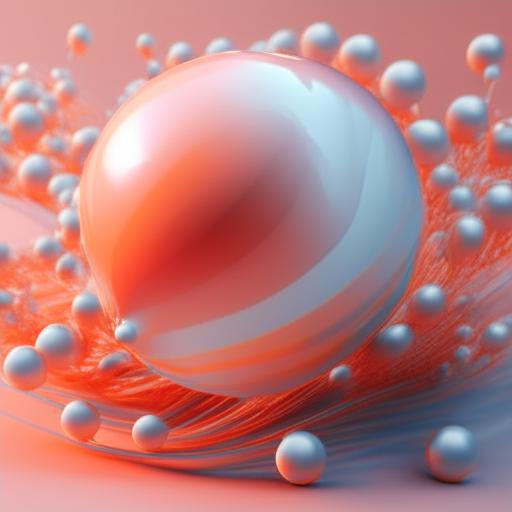} \\

        Object & \multicolumn{5}{c}{Results} \\
        
        \includegraphics[width=0.12\textwidth]{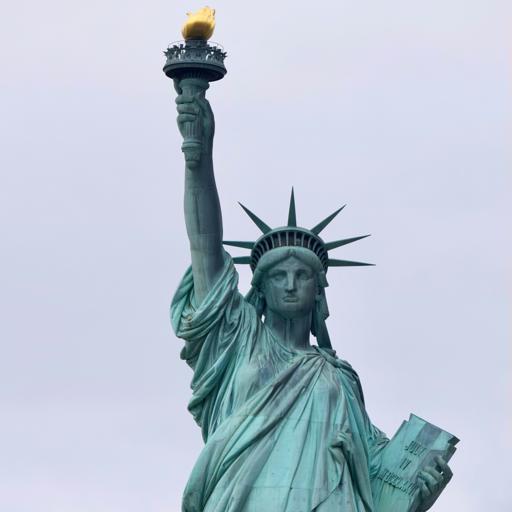} &
        \includegraphics[width=0.12\textwidth]{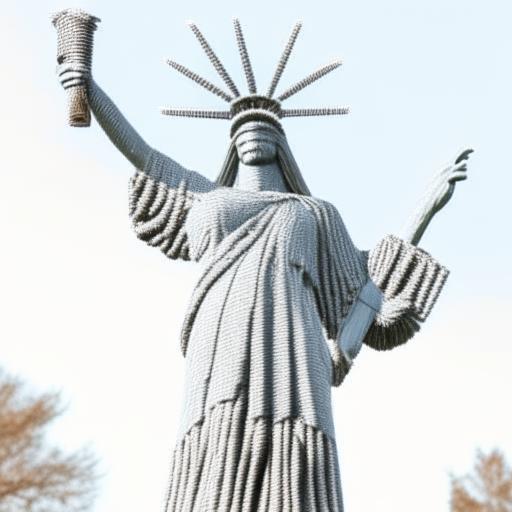} &
        \includegraphics[width=0.12\textwidth]{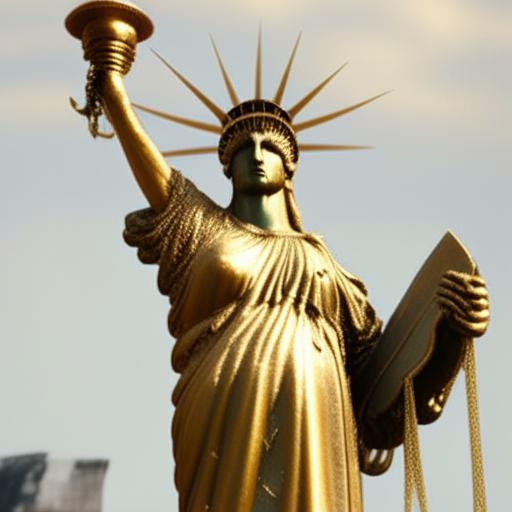} &
        \includegraphics[width=0.12\textwidth]{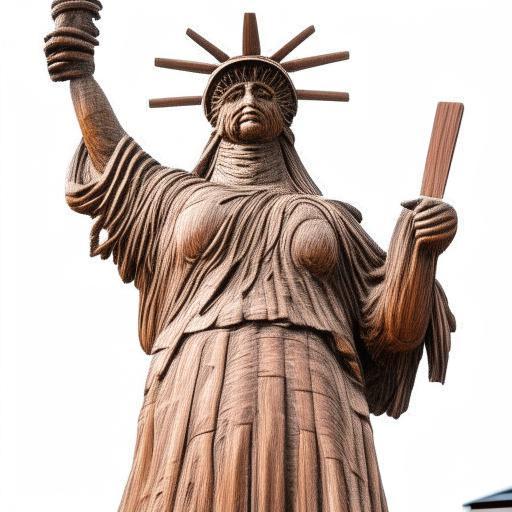} &
        \includegraphics[width=0.12\textwidth]{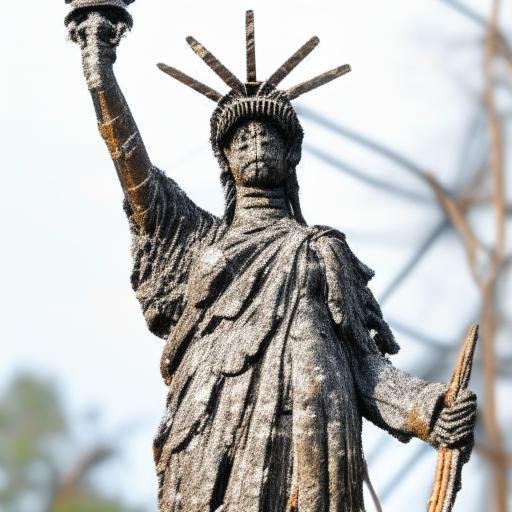} &
        \includegraphics[width=0.12\textwidth]{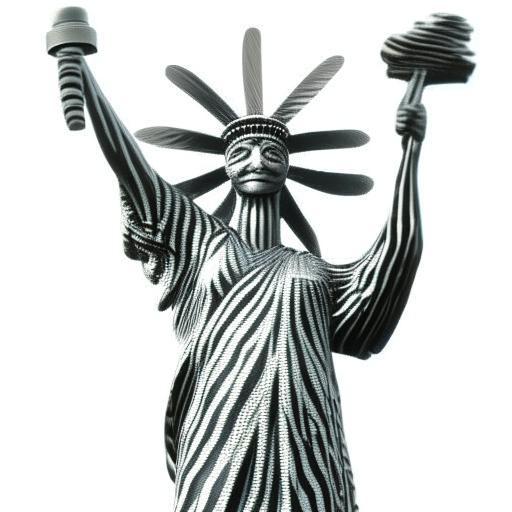} \\

        Object & \multicolumn{5}{c}{Results} \\
\end{tabular}
    }
    \caption{Additional texturing operator results obtained by our \textit{pOps} method.}
    \label{fig:texturing_ours_multi_textures}
\end{figure*}

\begin{figure*}
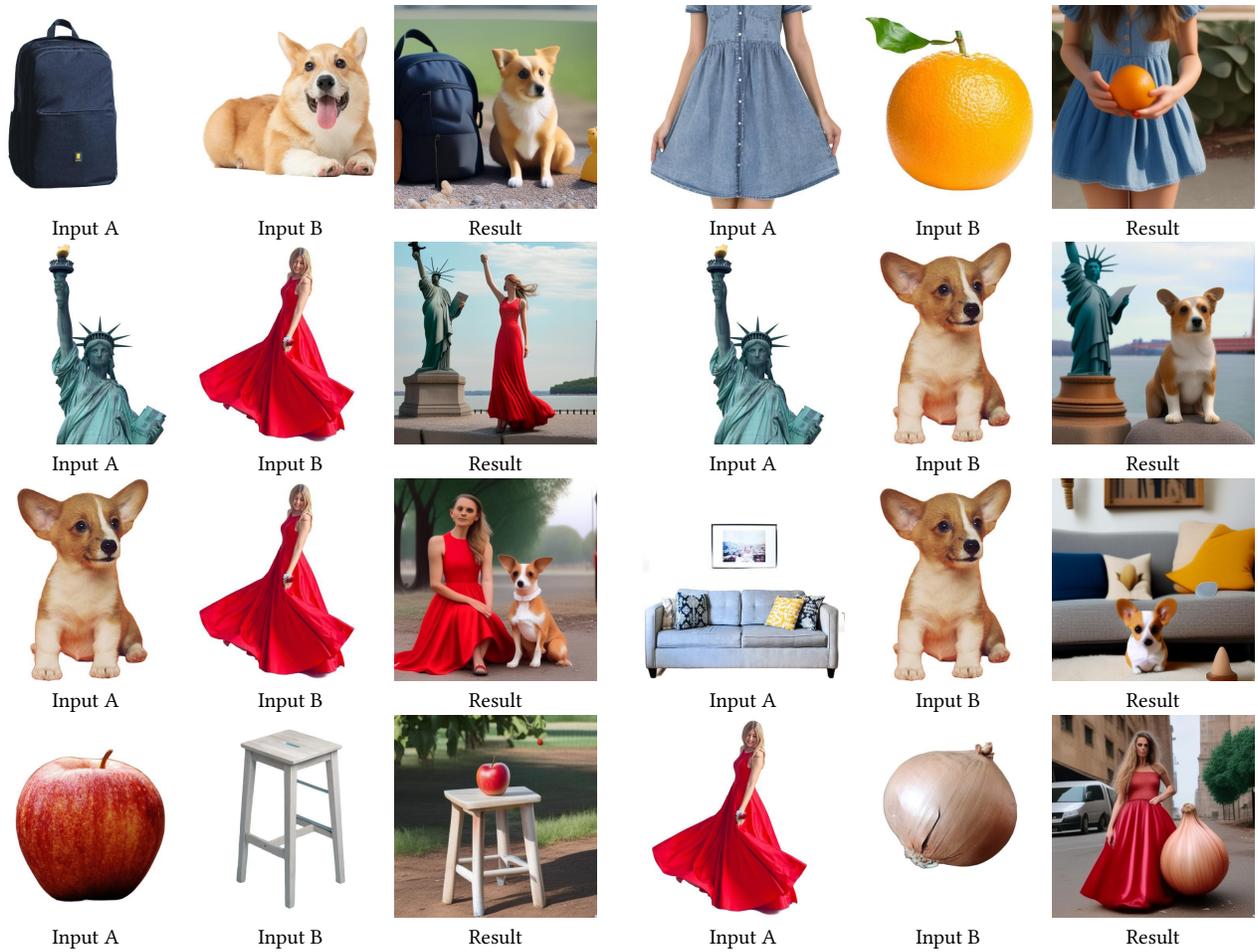

    \centering
    \setlength{\tabcolsep}{0.5pt}
    \addtolength{\belowcaptionskip}{-10pt}
    {\small
    \begin{tabular}{c c c @{\hspace{0.6cm}} c c c}

        \\ \\ \\ \\ \\ \\ \\ \\
\end{tabular}
    }
    \caption{Additional union results obtained by our \textit{pOps} method.}
    \label{fig:union_operator_1}
\end{figure*}

\begin{figure*}
    \centering
    \setlength{\tabcolsep}{0.5pt}
    \addtolength{\belowcaptionskip}{-10pt}
    {\small
    \begin{tabular}{c c c@{\hspace{0.4cm}} c c c@{\hspace{0.4cm}}}

        \\ \\ \\

        \includegraphics[width=0.15\textwidth]{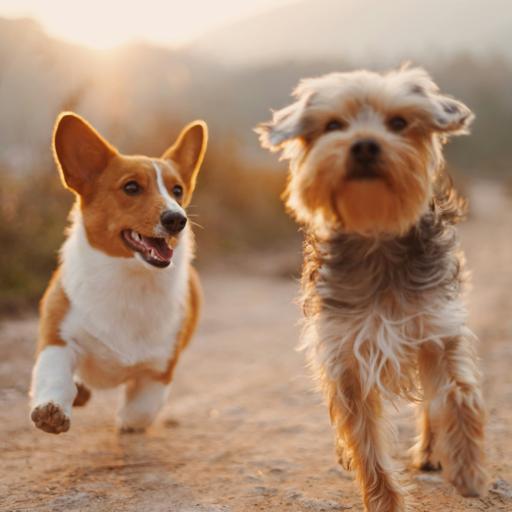} &
        \includegraphics[width=0.15\textwidth]{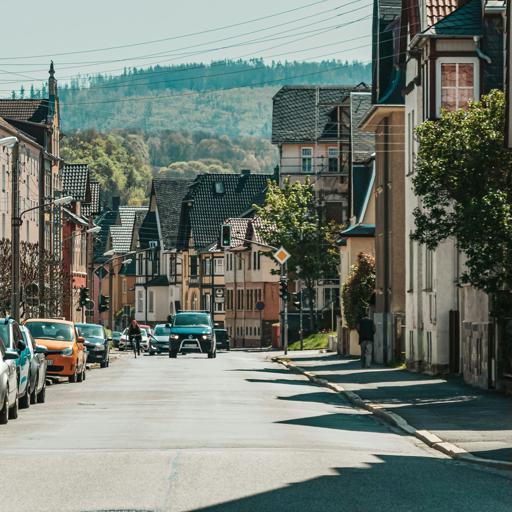} &
        \includegraphics[width=0.15\textwidth]{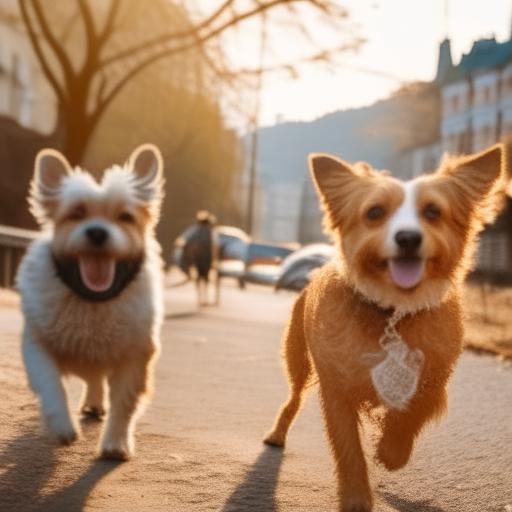} &

        \includegraphics[width=0.15\textwidth]{figures/scene_operator/images/cici-hung-nV3v8ZMRLNc-unsplash_tj-holowaychuk-1EYMue_AwDw-unsplash/cici-hung-nV3v8ZMRLNc-unsplash.jpg} &
        \includegraphics[width=0.15\textwidth]{figures/scene_operator/images/cici-hung-nV3v8ZMRLNc-unsplash_tj-holowaychuk-1EYMue_AwDw-unsplash/tj-holowaychuk-1EYMue_AwDw-unsplash.jpg} &
        \includegraphics[width=0.15\textwidth]{figures/scene_operator/images/cici-hung-nV3v8ZMRLNc-unsplash_tj-holowaychuk-1EYMue_AwDw-unsplash/1.jpg} \\

        Object & Background & Output & Object & Background & Output \\ \\
\end{tabular}
    }
    \caption{Additional scene operator results obtained by our \textit{pOps} method.}
    \label{fig:scene_operator_results_1}
\end{figure*}

\begin{figure*}
    \centering
    \setlength{\tabcolsep}{0.5pt}
    \addtolength{\belowcaptionskip}{-10pt}
    {\small
    \begin{tabular}{c c c@{\hspace{0.4cm}} c c c@{\hspace{0.4cm}}}

        \\ \\ \\

        \includegraphics[width=0.15\textwidth]{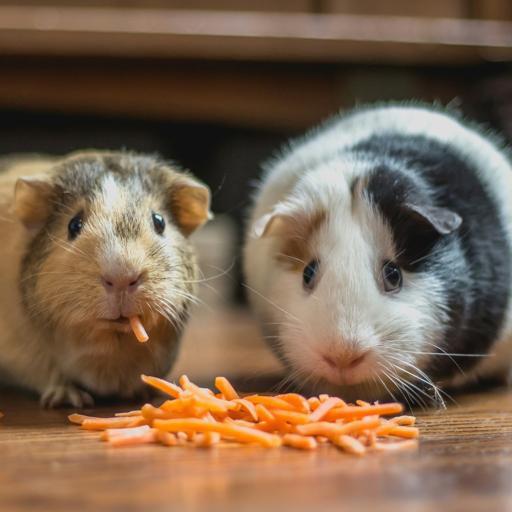} &
        \includegraphics[width=0.15\textwidth]{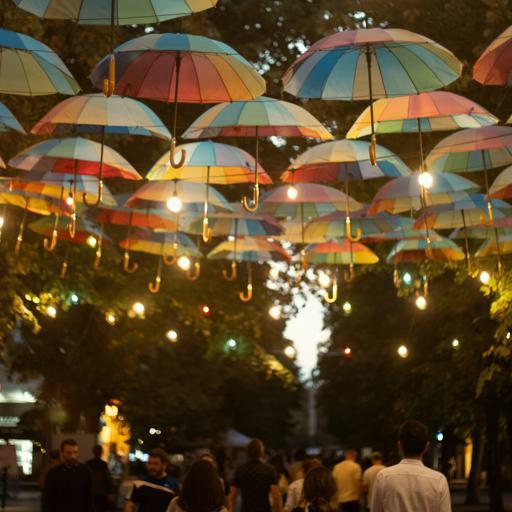} &
        \includegraphics[width=0.15\textwidth]{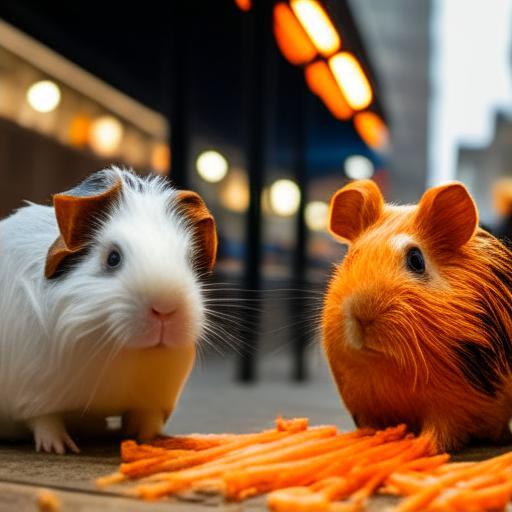} &

        \includegraphics[width=0.15\textwidth]{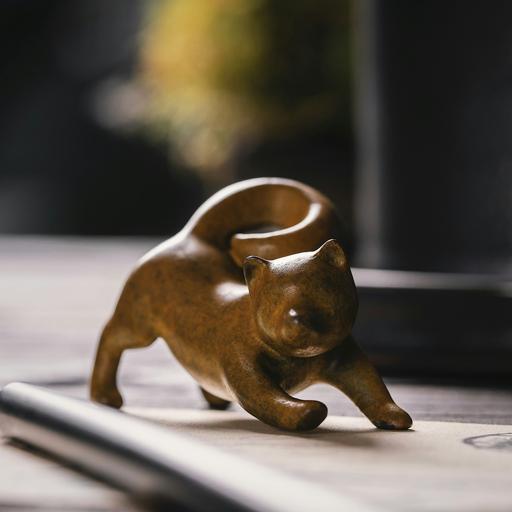} &
        \includegraphics[width=0.15\textwidth]{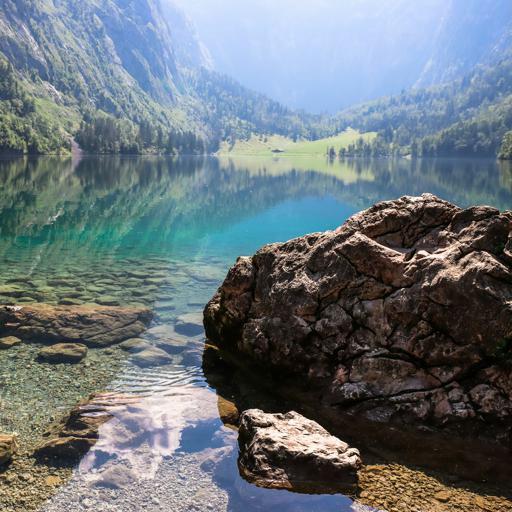} &
        \includegraphics[width=0.15\textwidth]{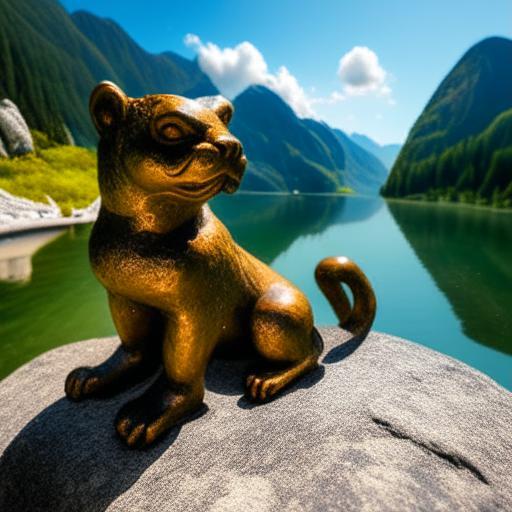} \\

        Object & Background & Output & Object & Background & Output \\ \\
\end{tabular}
    }
    \caption{Additional scene operator results obtained by our \textit{pOps} method.}
    \label{fig:scene_operator_results_2}
\end{figure*}

\begin{figure*}
    \centering
    \setlength{\tabcolsep}{0.5pt}
    \addtolength{\belowcaptionskip}{-10pt}
    {\small
    \begin{tabular}{c c c@{\hspace{0.4cm}} c c c@{\hspace{0.4cm}}}
 
        \\ \\ \\ \\ \\ \\ \\ \\
\end{tabular}
    }
    \caption{Additional scene operator results obtained by our \textit{pOps} method.}
    \label{fig:scene_operator_results_3}
\end{figure*}

\begin{figure*}
    \centering
    \setlength{\tabcolsep}{0.5pt}
    \addtolength{\belowcaptionskip}{-10pt}
    {\small
    \begin{tabular}{c c @{\hspace{0.4cm}} c c @{\hspace{0.4cm}} c c}

        \\ \\ \\ \\ \\

        \includegraphics[width=0.135\textwidth]{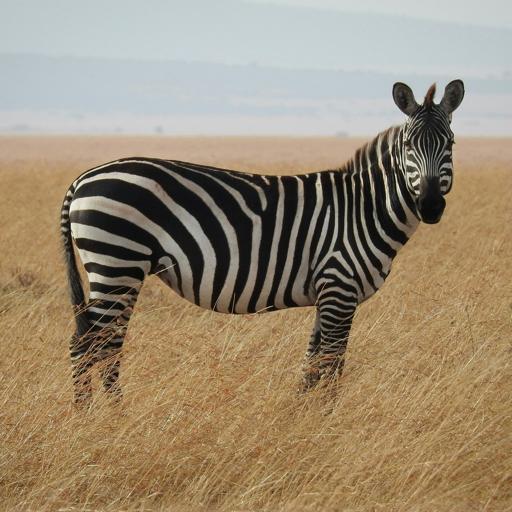} &
        \includegraphics[width=0.135\textwidth]{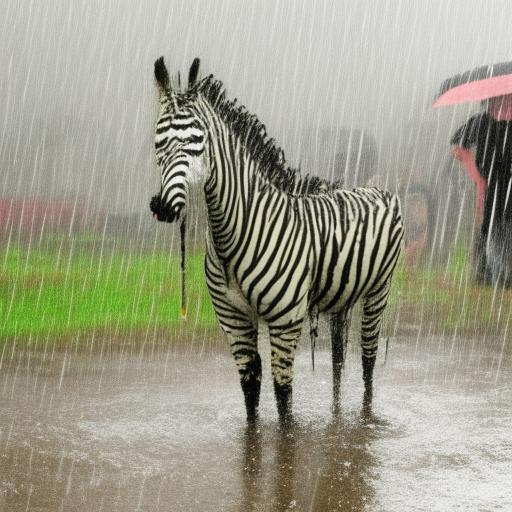} &
        \includegraphics[width=0.135\textwidth]{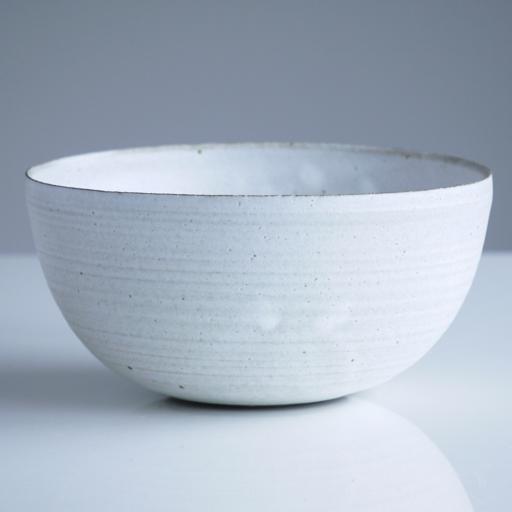} &
        \includegraphics[width=0.135\textwidth]{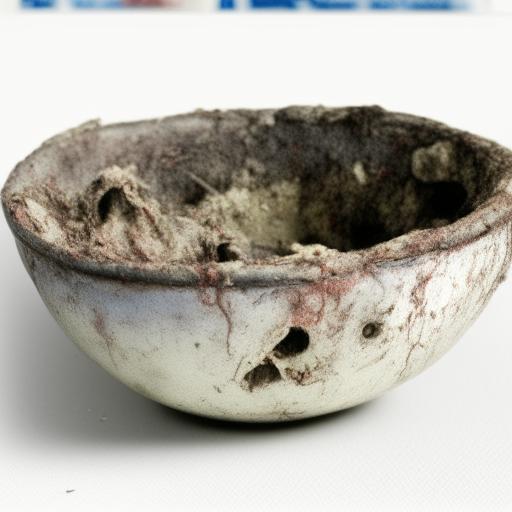} &
        \includegraphics[width=0.135\textwidth]{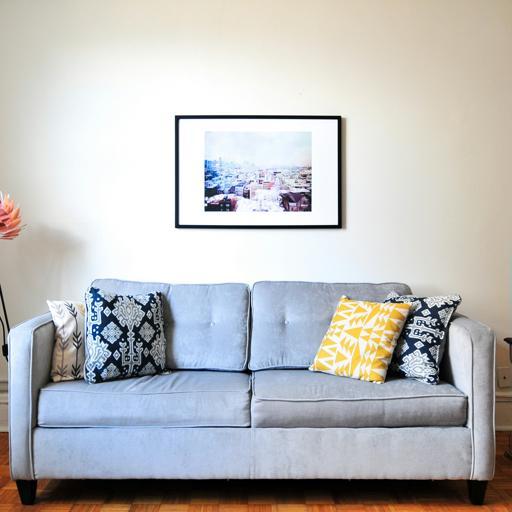} &
        \includegraphics[width=0.135\textwidth]{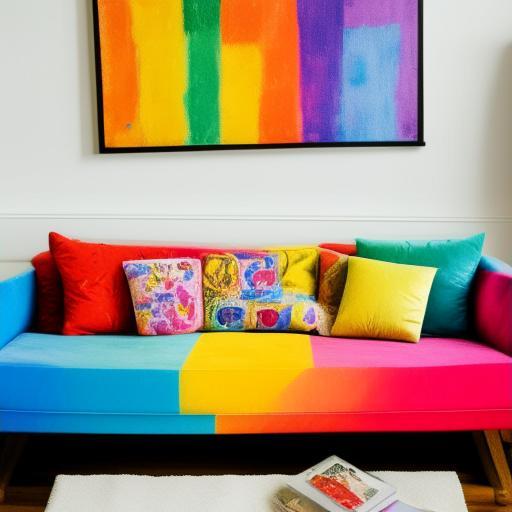} \\
        
        Input & ``soggy'' & Input & ``rotten'' & Input & ``colorful' \\

        \includegraphics[width=0.135\textwidth]{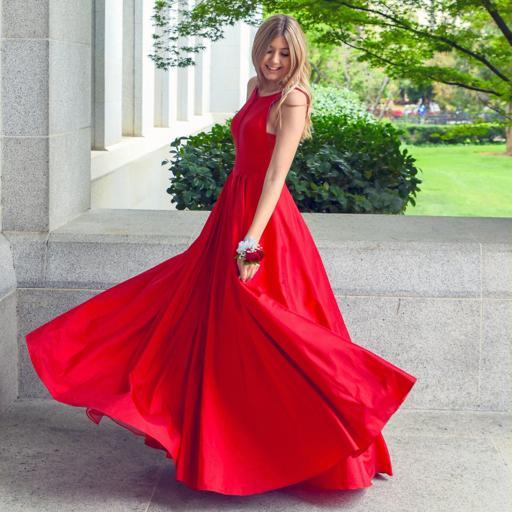} &
        \includegraphics[width=0.135\textwidth]{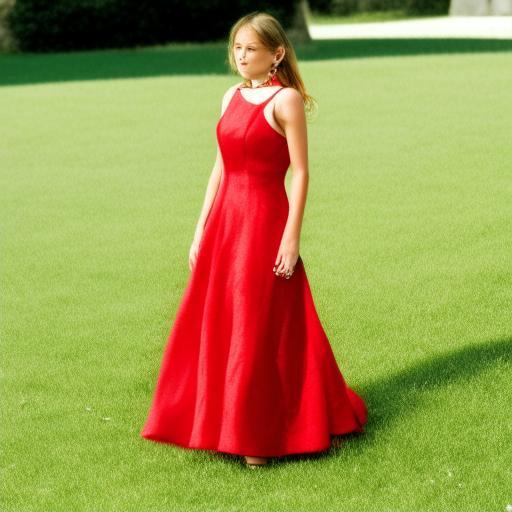} &
        \includegraphics[width=0.135\textwidth]{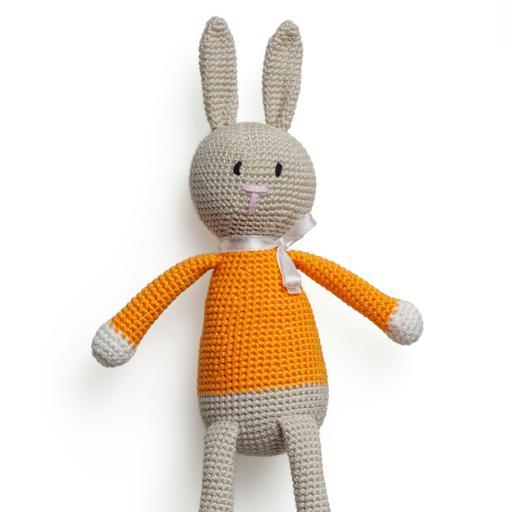} &
        \includegraphics[width=0.135\textwidth]{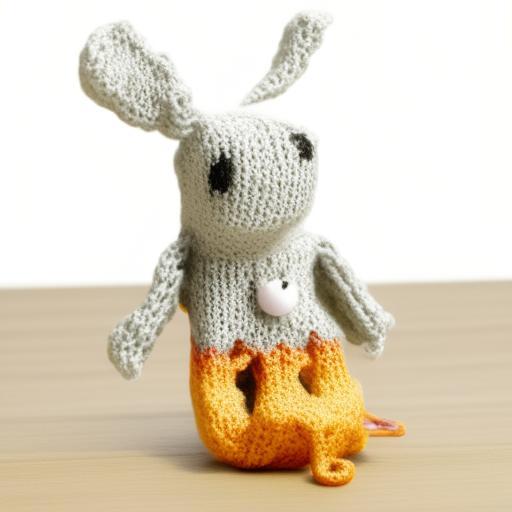} &
        \includegraphics[width=0.135\textwidth]{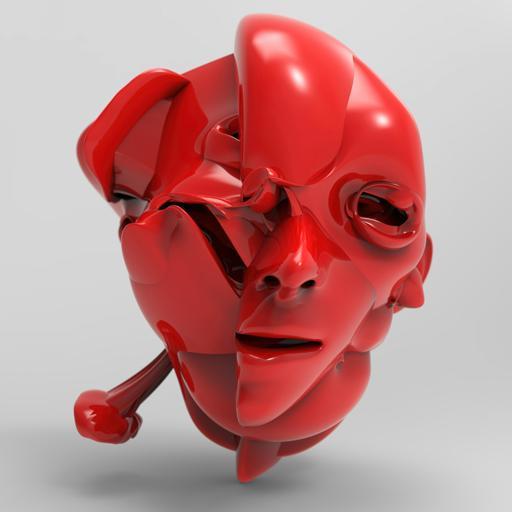} &
        \includegraphics[width=0.135\textwidth]{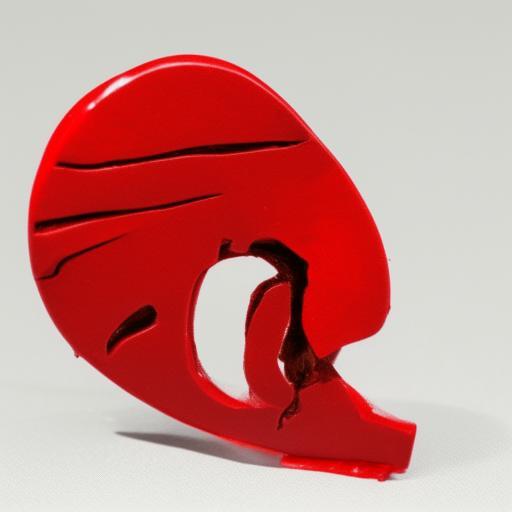} 
        
        \\
\end{tabular}
    \begin{tabular}{c c c @{\hspace{1cm}} c c c}

        \includegraphics[width=0.135\textwidth]{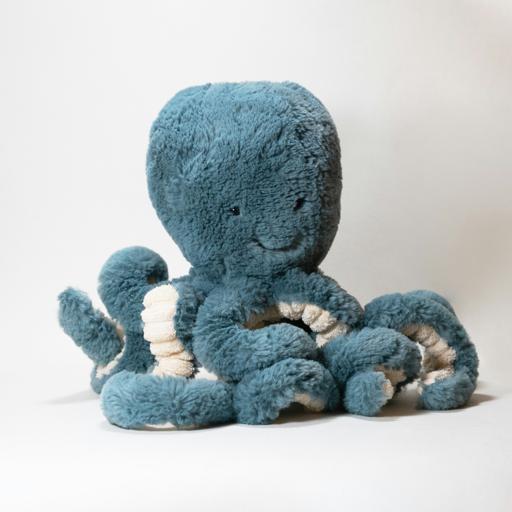} &
        \includegraphics[width=0.135\textwidth]{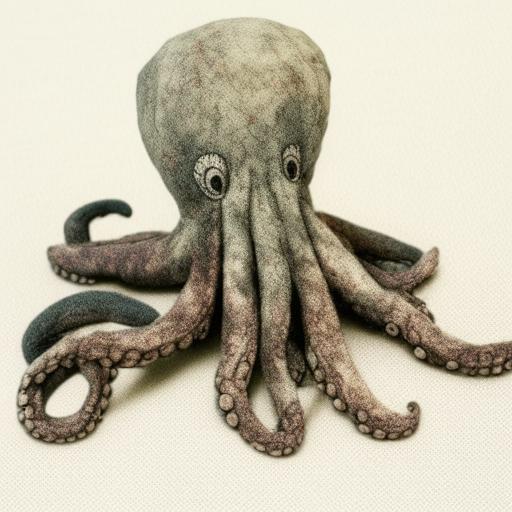} &
        \includegraphics[width=0.135\textwidth]{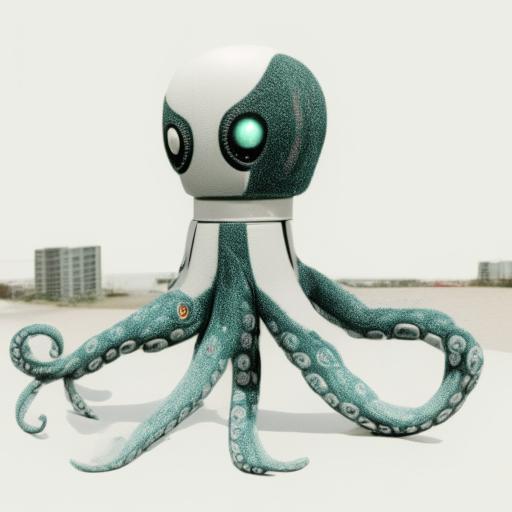} &
        \includegraphics[width=0.135\textwidth]{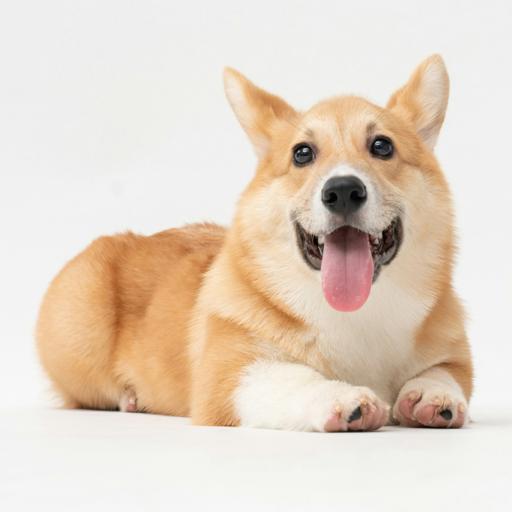} &
        \includegraphics[width=0.135\textwidth]{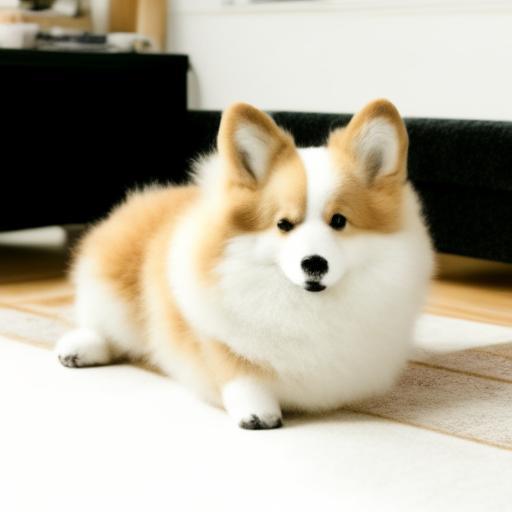} &
        \includegraphics[width=0.135\textwidth]{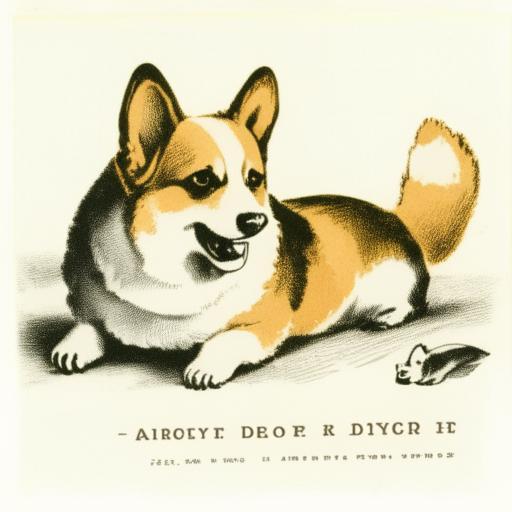} \\
        
        Input & ``aged'' & ``futuristic'' & Input & ``fluffy'' & ``litograph'' \\

        \includegraphics[width=0.135\textwidth]{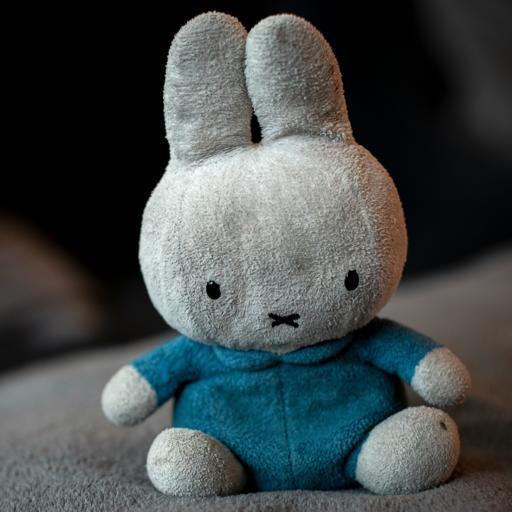} &
        \includegraphics[width=0.135\textwidth]{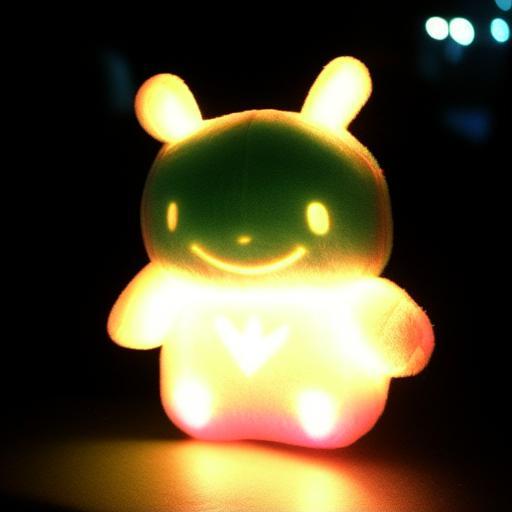} &
        \includegraphics[width=0.135\textwidth]{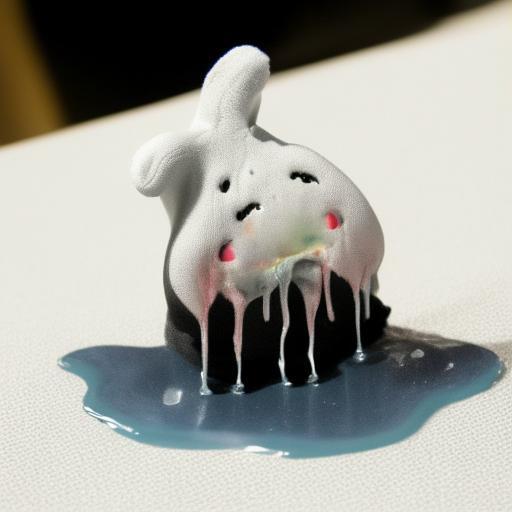} &
        \includegraphics[width=0.135\textwidth]{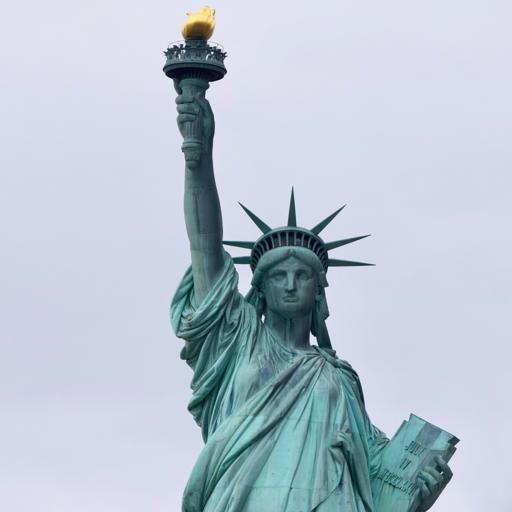} &
        \includegraphics[width=0.135\textwidth]{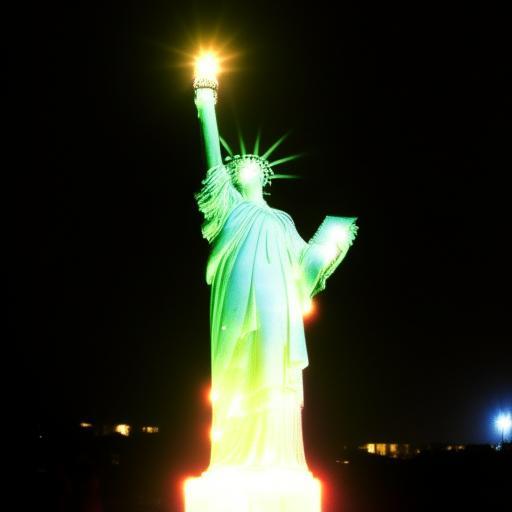} &
        \includegraphics[width=0.135\textwidth]{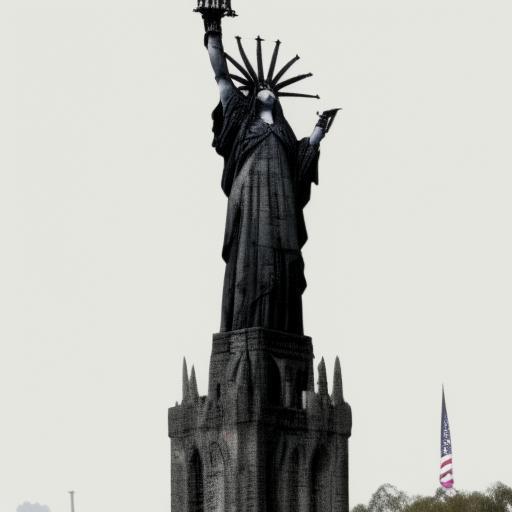} \\

        Input & ``glowing'' & ``melting'' & Input & ``glowing'' & ``gothic'' \\
\end{tabular}
    }
    \caption{Additional instruct operator results obtained by our \textit{pOps} method.}
    \label{fig:instruct_operator_1}
\end{figure*}

\begin{figure*}
    \centering
    \setlength{\tabcolsep}{0.5pt}
    \addtolength{\belowcaptionskip}{-10pt}
    {\small
    \begin{tabular}{c c c c @{\hspace{0.4cm}} c}

        \\ \\ \\ \\ \\ \\ \\
        
        \includegraphics[width=0.13\textwidth]{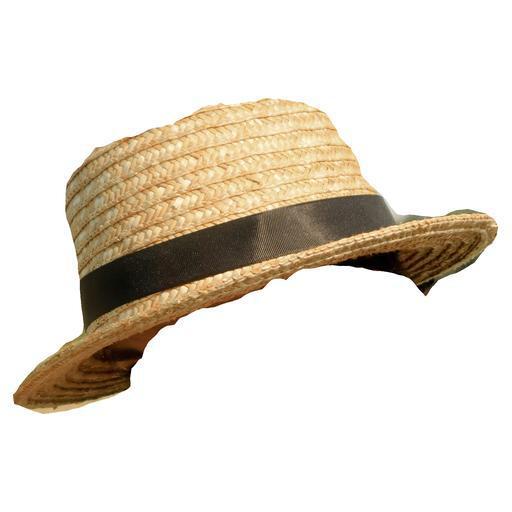} &
        \includegraphics[width=0.13\textwidth]{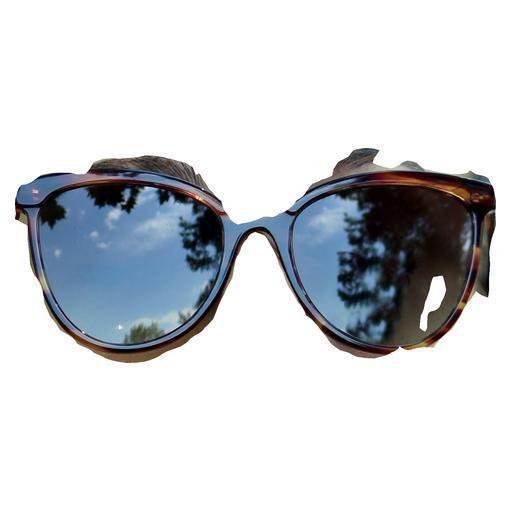} & & &
        \includegraphics[width=0.13\textwidth]{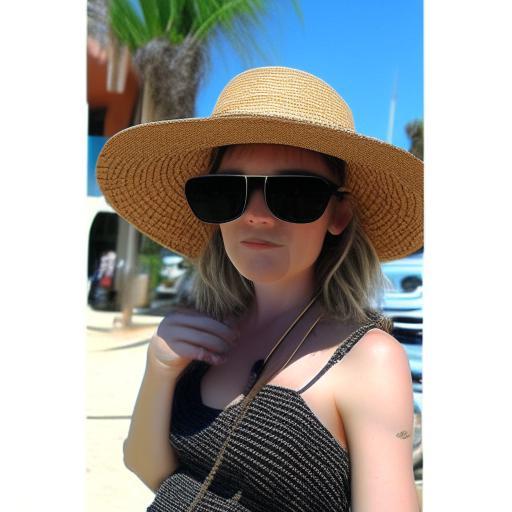} \\
\end{tabular}
    }
    \caption{Additional multi-image clothing composition results obtained by our \textit{pOps} method.}
    \label{fig:clothes_compose_1}
\end{figure*}

\begin{figure*}
    \centering
    \setlength{\tabcolsep}{0.5pt}
    \addtolength{\belowcaptionskip}{-10pt}
    {\small
    \begin{tabular}{c c c c @{\hspace{0.4cm}} c}

        \\ \\ \\ \\ \\ \\ \\
        
        \includegraphics[width=0.13\textwidth]{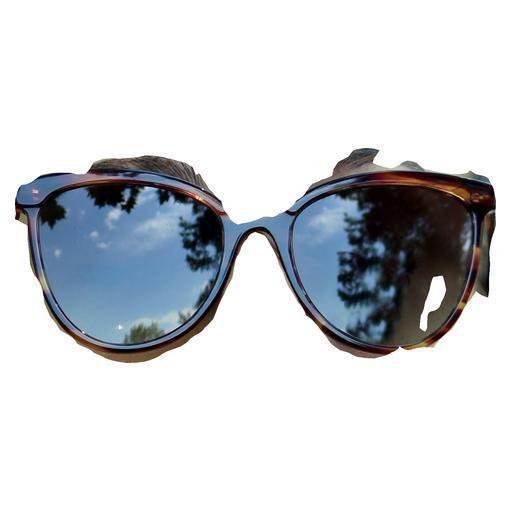} &
        \includegraphics[width=0.13\textwidth]{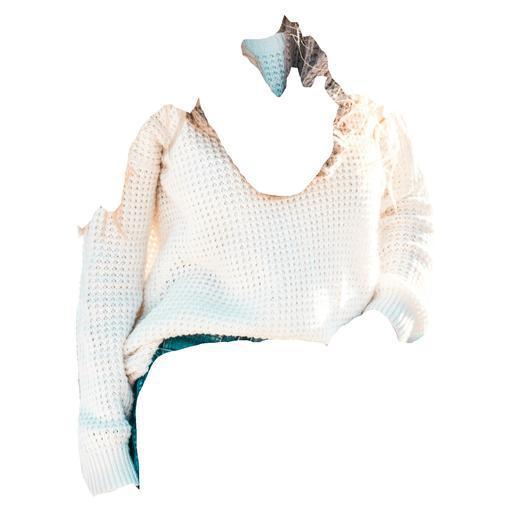} &
        \includegraphics[width=0.13\textwidth]{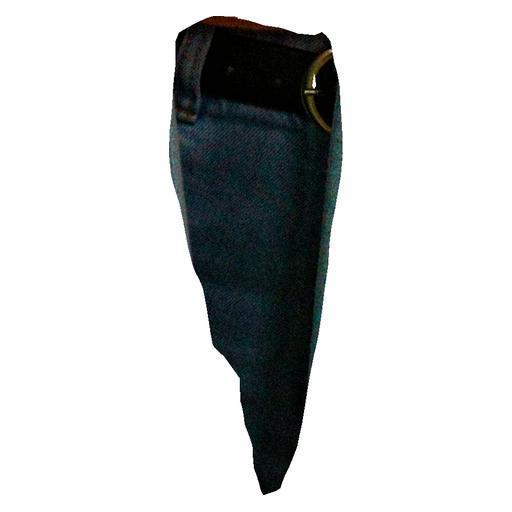} & &
        \includegraphics[width=0.13\textwidth]{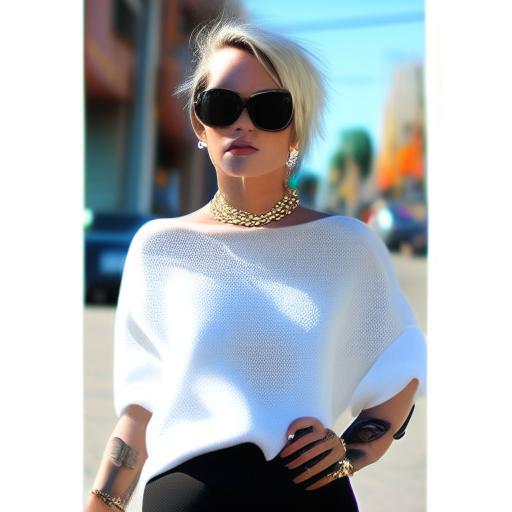} \\
\end{tabular}
    }
    \caption{Additional multi-image composition results obtained by our \textit{pOps} method.}
    \label{fig:clothes_compose_2}
\end{figure*}

\begin{figure*}
    \centering
    \setlength{\tabcolsep}{0.5pt}
    \addtolength{\belowcaptionskip}{-5pt}
    {
    \begin{tabular}{c c @{\hspace{0.4cm}} c c c}

        \includegraphics[width=0.16\textwidth]{figures/base_results/images/inputs/moose.jpg} &
        \includegraphics[width=0.16\textwidth]{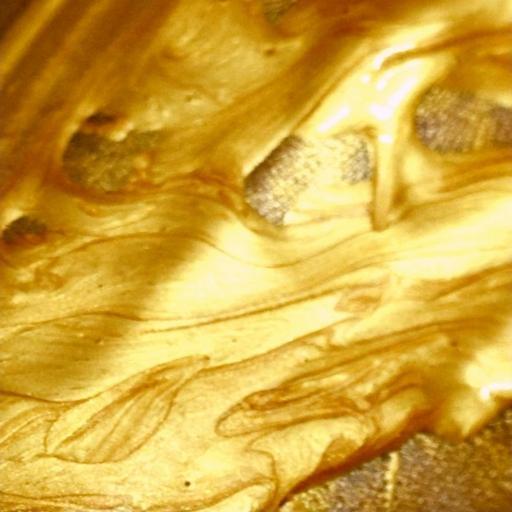} &
        \includegraphics[width=0.16\textwidth]{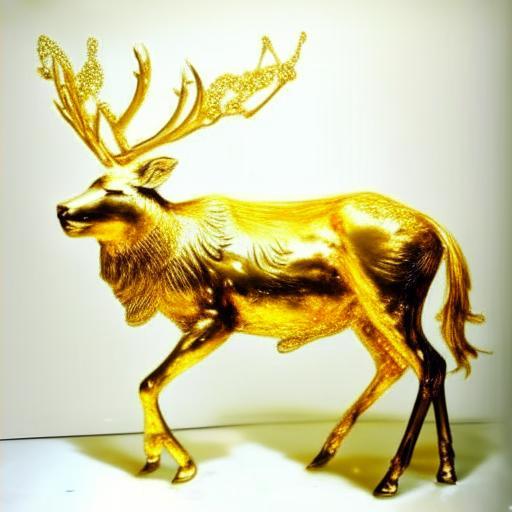} &
        \includegraphics[width=0.16\textwidth]{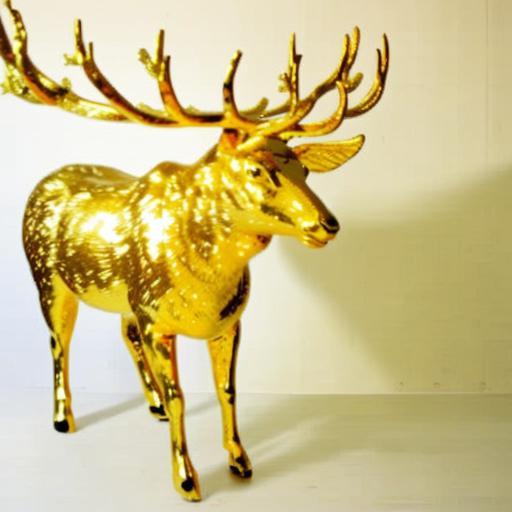} &
        \includegraphics[width=0.16\textwidth]{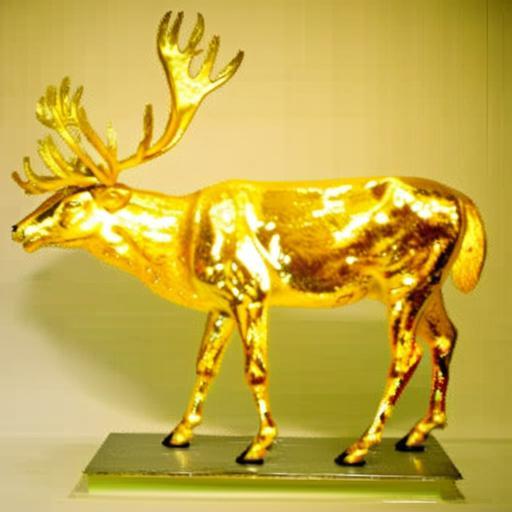} \\

        \includegraphics[width=0.16\textwidth]{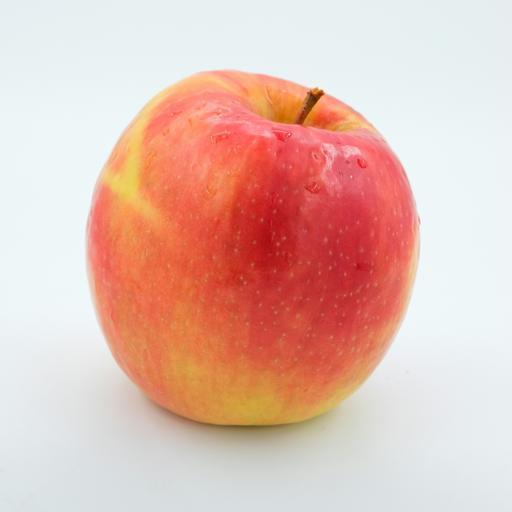} &
        \includegraphics[width=0.16\textwidth]{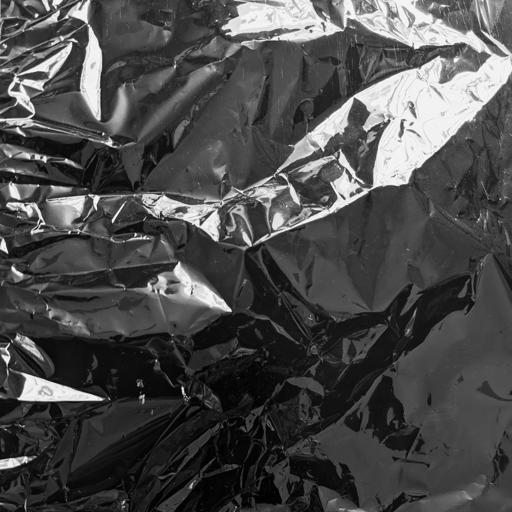} &
        \includegraphics[width=0.16\textwidth]{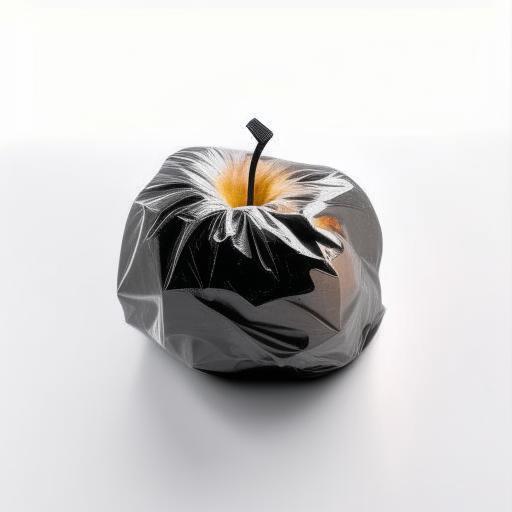} &
        \includegraphics[width=0.16\textwidth]{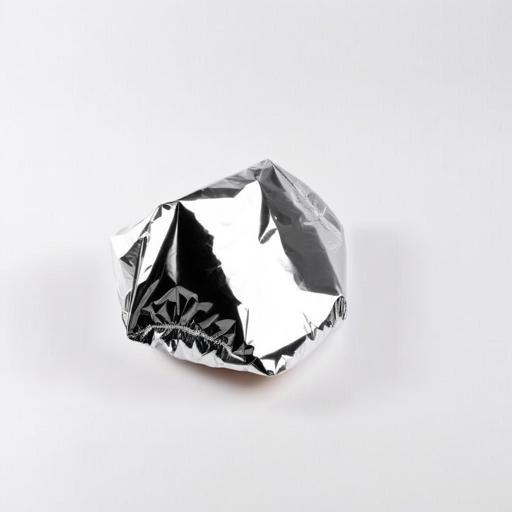} &
        \includegraphics[width=0.16\textwidth]{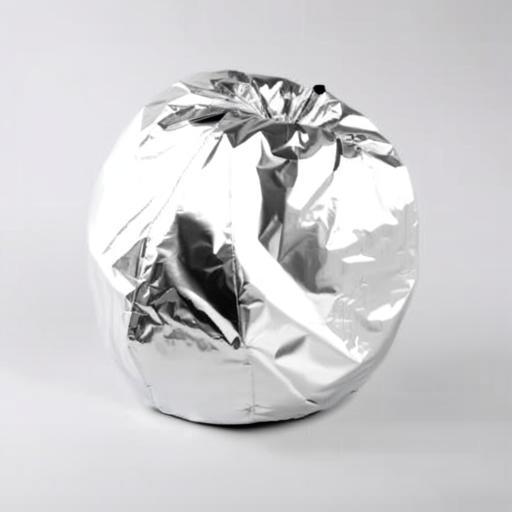} \\

        \includegraphics[width=0.16\textwidth]{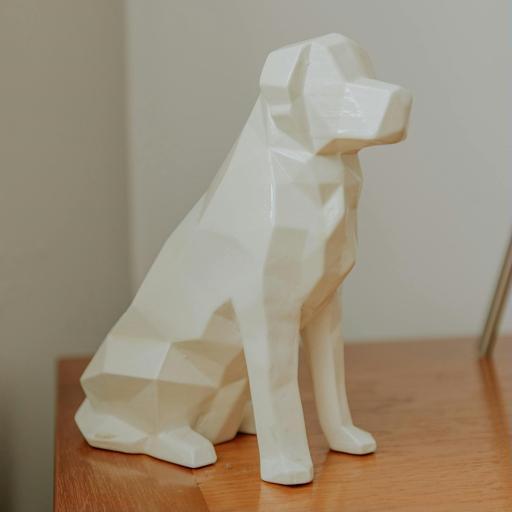} &
        \includegraphics[width=0.16\textwidth]{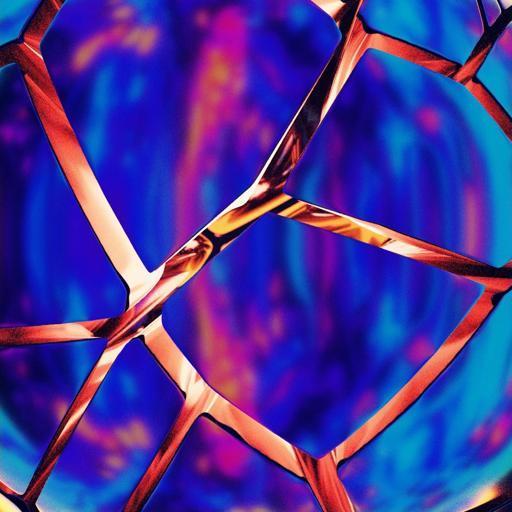} &
        \includegraphics[width=0.16\textwidth]{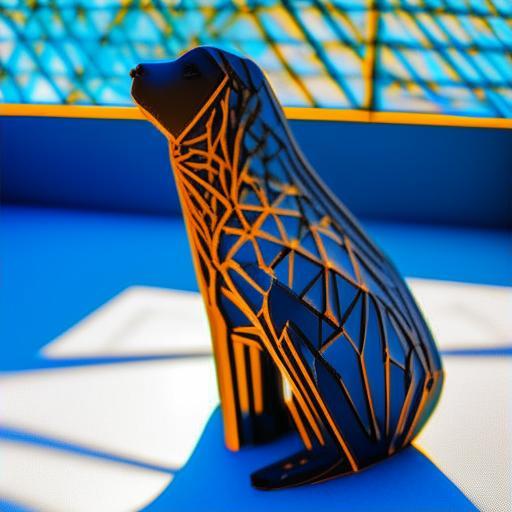} &
        \includegraphics[width=0.16\textwidth]{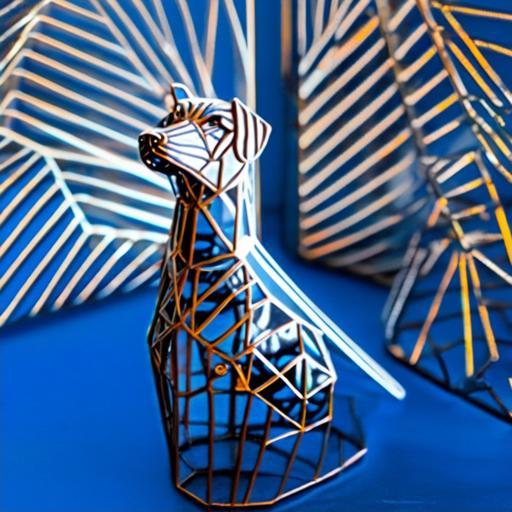} &
        \includegraphics[width=0.16\textwidth]{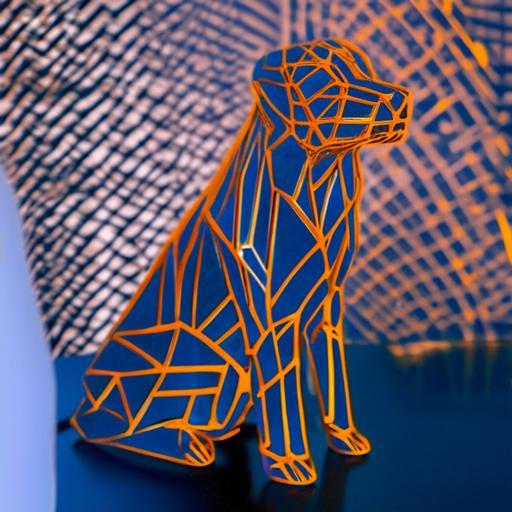} \\

        \includegraphics[width=0.16\textwidth]{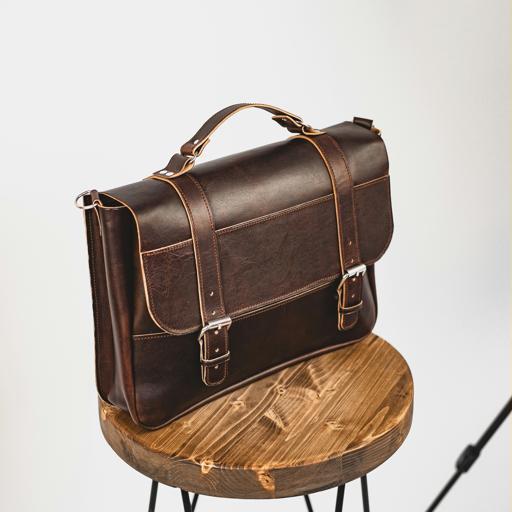} &
        \includegraphics[width=0.16\textwidth]{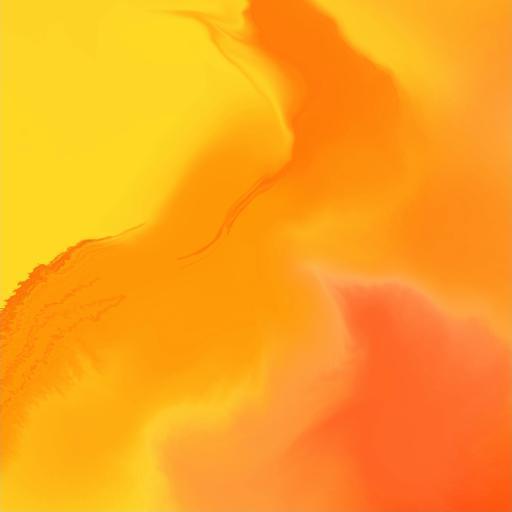} &
        \includegraphics[width=0.16\textwidth]{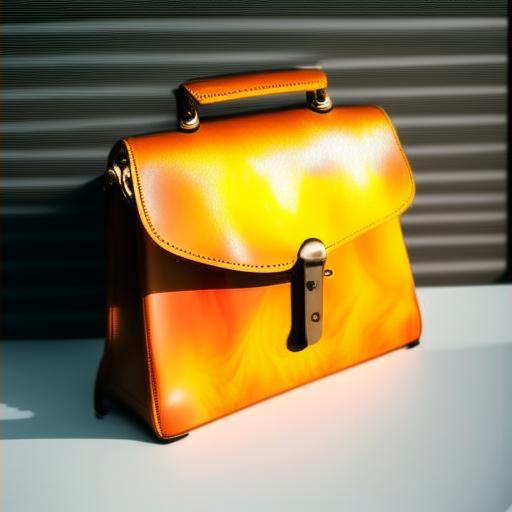} &
        \includegraphics[width=0.16\textwidth]{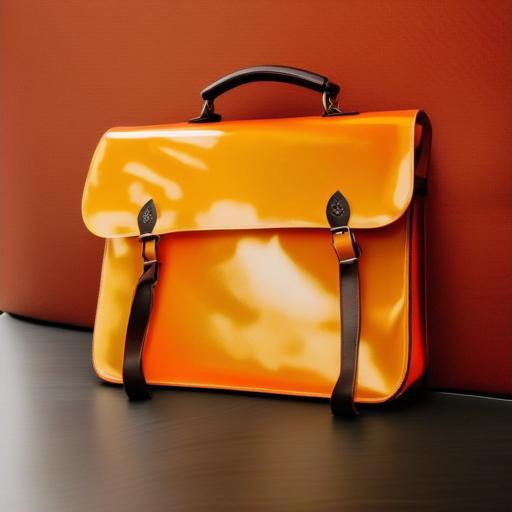} &
        \includegraphics[width=0.16\textwidth]{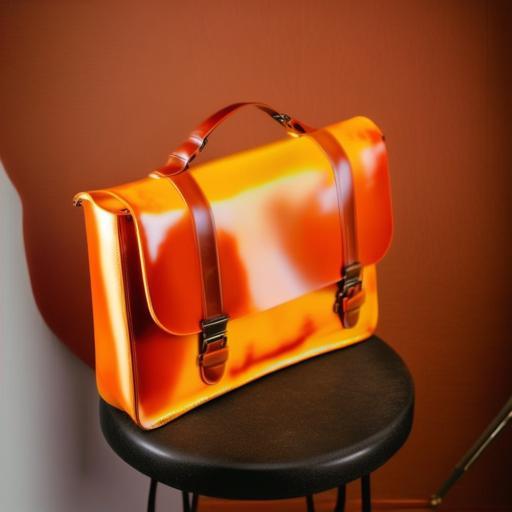} \\
\end{tabular}
    }
    \vspace{-0.2cm}
        \caption{\textbf{Different Renderers}. \textit{pOps} outputs can be directly fed to either Kandinsky or IP-Adapter and incorporated alongside spatial conditions using ControlNet~\cite{zhang2023adding}.
    }
    \label{fig:ip_adapter_supp}
\end{figure*}

\begin{figure*}
    \centering
    \setlength{\tabcolsep}{0.5pt}
    \addtolength{\belowcaptionskip}{-10pt}
    {
    \begin{tabular}{c @{\hspace{0.2cm}} c @{\hspace{0.2cm}} c @{\hspace{0.2cm}} c @{\hspace{0.2cm}} c}

        \\ \\ \\ \\ \\

        \includegraphics[width=0.15\textwidth]{figures/instruct_comparisons/images/alvan-nee-brFsZ7qszSY-unsplash_muddy/alvan-nee-brFsZ7qszSY-unsplash.jpg} &
        {\raisebox{0.375in}{\scalebox{1.2}{$\pmb{\bigcirc}$ \; ``plush'' \; $\pmb{\bigcirc}$}}} &
        \includegraphics[width=0.15\textwidth]{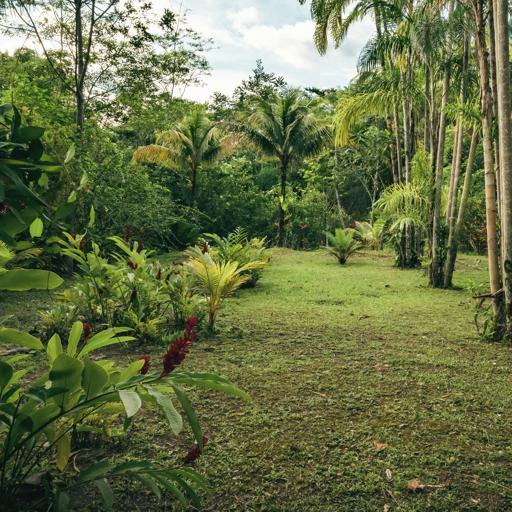} &
        {\raisebox{0.375in}{\scalebox{1.2}{$\pmb{=}$}}} &
        \includegraphics[width=0.15\textwidth]{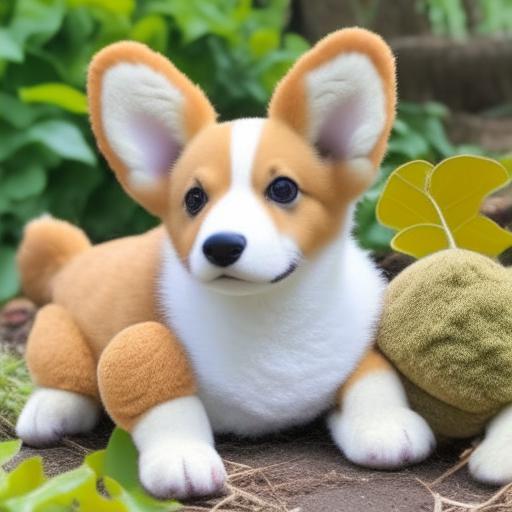} \\

        \includegraphics[width=0.15\textwidth]{figures/instruct_comparisons/images/alvan-nee-brFsZ7qszSY-unsplash_muddy/alvan-nee-brFsZ7qszSY-unsplash.jpg} &
        {\raisebox{0.375in}{\scalebox{1.2}{$\pmb{\bigcirc}$ \; ``winged'' \; $\pmb{\bigcirc}$}}} &
        \includegraphics[width=0.15\textwidth]{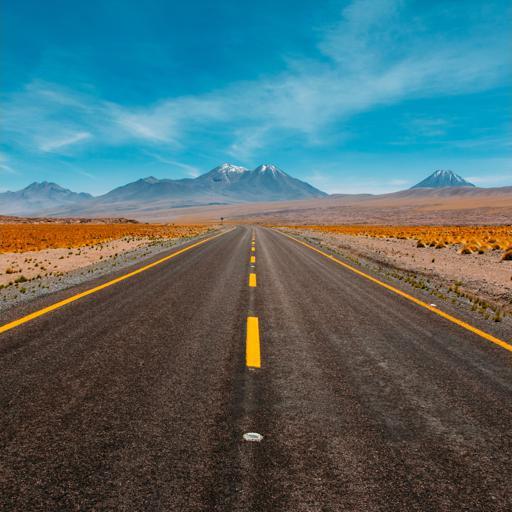} &
        {\raisebox{0.375in}{\scalebox{1.2}{$\pmb{=}$}}} &
        \includegraphics[width=0.15\textwidth]{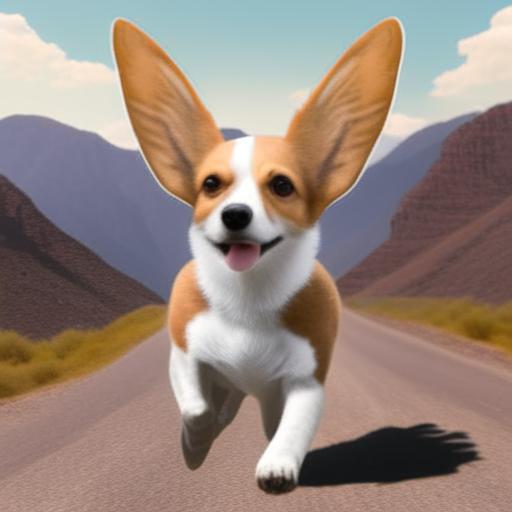} \\
\end{tabular}
    }
    \caption{Compositions of instruct and scene operators obtained by our pOps method.}
    \label{fig:instruct_scene_compositions}
\end{figure*}

\begin{figure*}
    \centering
    \setlength{\tabcolsep}{0.5pt}
    \addtolength{\belowcaptionskip}{-10pt}
    {
    \begin{tabular}{c @{\hspace{0.2cm}} c @{\hspace{0.2cm}} c @{\hspace{0.2cm}} c @{\hspace{0.2cm}} c}

        \includegraphics[width=0.11\textwidth]{figures/instruct_comparisons/images/alvan-nee-brFsZ7qszSY-unsplash_muddy/alvan-nee-brFsZ7qszSY-unsplash.jpg} &
        {\raisebox{0.25in}{\scalebox{1.2}{$\pmb{\bigcirc}$ \; ``hairless'' \; $\pmb{\bigcirc}$}}} &
        \includegraphics[width=0.11\textwidth]{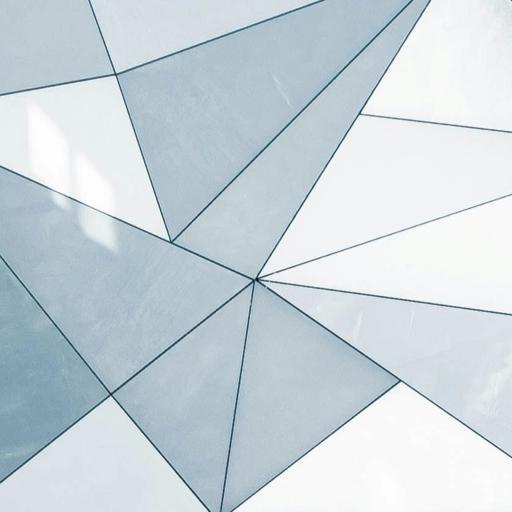} &
        {\raisebox{0.25in}{\scalebox{1.2}{$\pmb{=}$}}} &
        \includegraphics[width=0.11\textwidth]{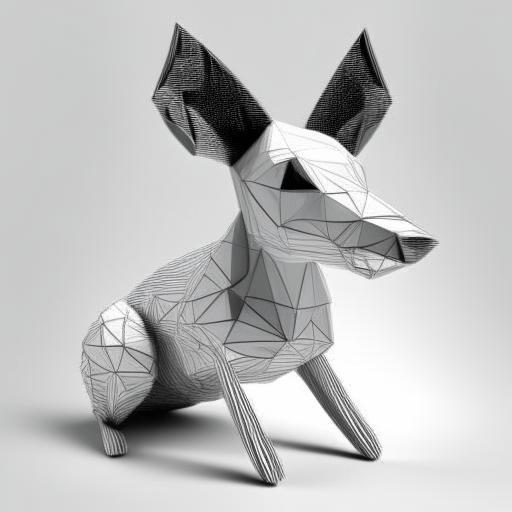} \\

        \includegraphics[width=0.11\textwidth]{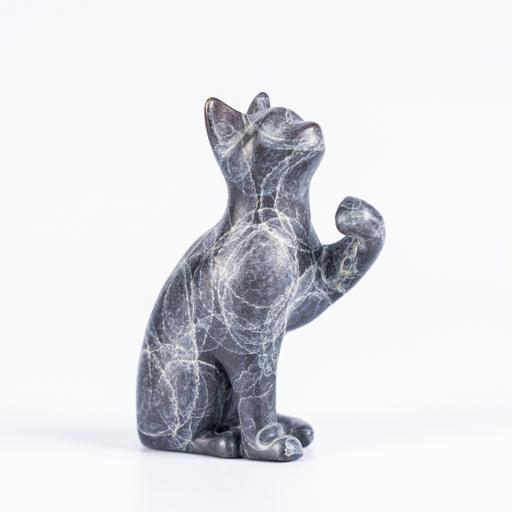} &
        {\raisebox{0.25in}{\scalebox{1.2}{$\pmb{\bigcirc}$ \; ``cartoon'' \; $\pmb{\bigcirc}$}}} &
        \includegraphics[width=0.11\textwidth]{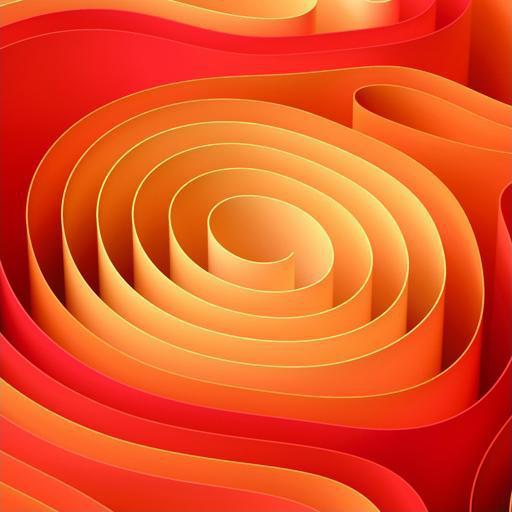} &
        {\raisebox{0.25in}{\scalebox{1.2}{$\pmb{=}$}}} &
        \includegraphics[width=0.11\textwidth]{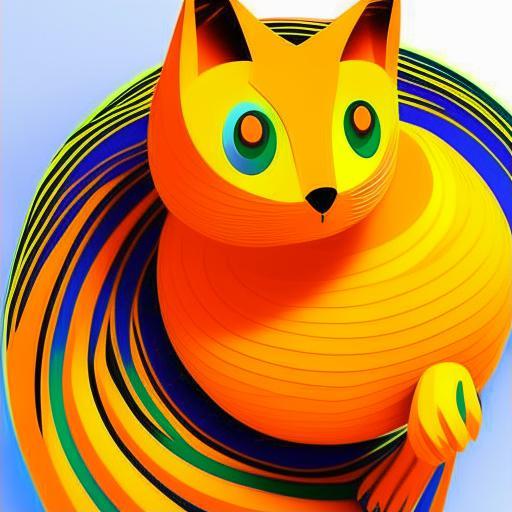} \\
\end{tabular}
    }
    \caption{Compositions of instruct and texturing operators obtained by our pOps method.}
    \label{fig:instruct_texturing_compositions}
\end{figure*}

\begin{figure*}
    \centering
    \setlength{\tabcolsep}{0.5pt}
    \addtolength{\belowcaptionskip}{-10pt}
    {
    \begin{tabular}{c @{\hspace{0.2cm}} c c @{\hspace{0.2cm}} c @{\hspace{0.2cm}} c @{\hspace{0.2cm}} c}

        \includegraphics[width=0.09\textwidth]{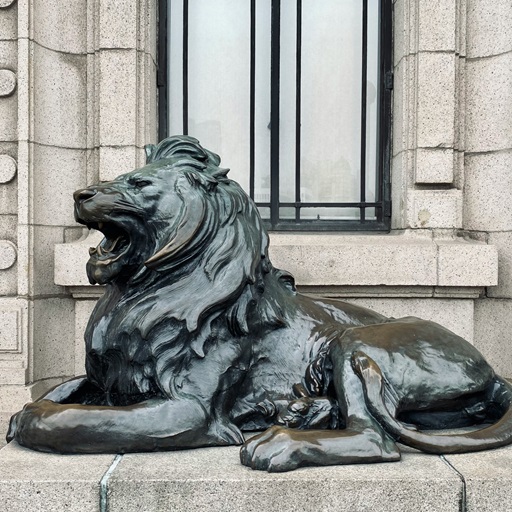} &
        {\raisebox{0.25in}{\scalebox{1.2}{$\pmb{\bigcirc}$ \; }}} &
        \includegraphics[width=0.09\textwidth]{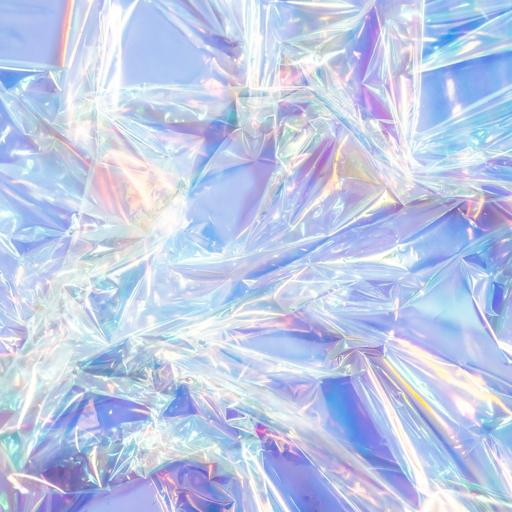} &
        {\raisebox{0.25in}{\scalebox{1.2}{$\pmb{\bigcirc}$ \; ``drawing'' \; }}} &
        {\raisebox{0.25in}{\scalebox{1.2}{$\pmb{=}$}}} &
        \includegraphics[width=0.09\textwidth]{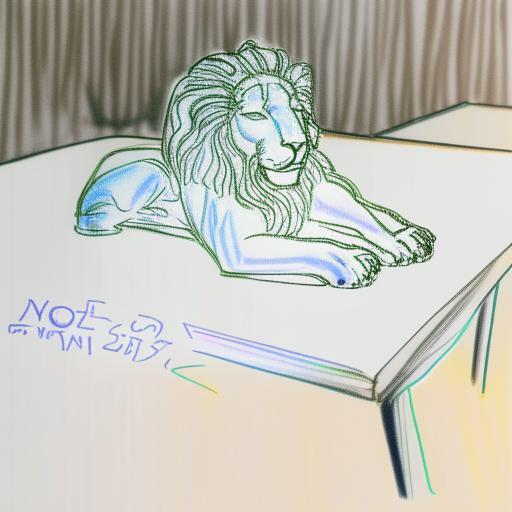} \\

        \includegraphics[width=0.09\textwidth]{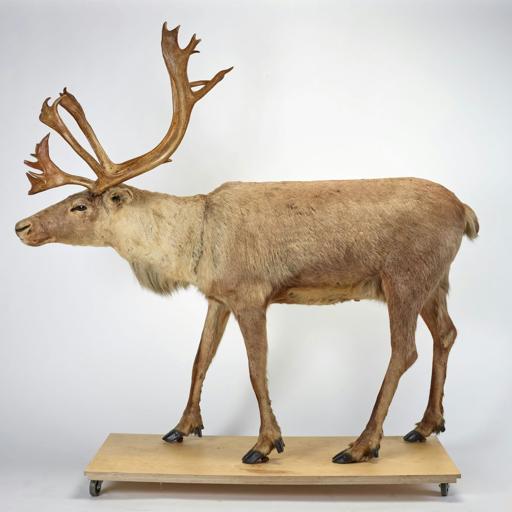} &
        {\raisebox{0.25in}{\scalebox{1.2}{$\pmb{\bigcirc}$ \; }}} &
        \includegraphics[width=0.09\textwidth]{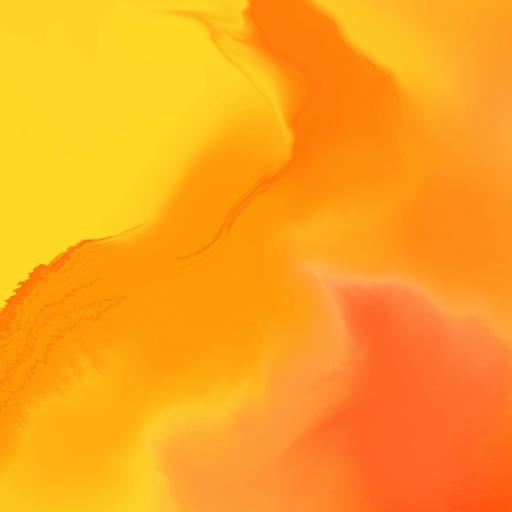} &
        {\raisebox{0.25in}{\scalebox{1.2}{$\pmb{\bigcirc}$ \; ``abstract art'' \; }}} &
        {\raisebox{0.25in}{\scalebox{1.2}{$\pmb{=}$}}} &
        \includegraphics[width=0.09\textwidth]{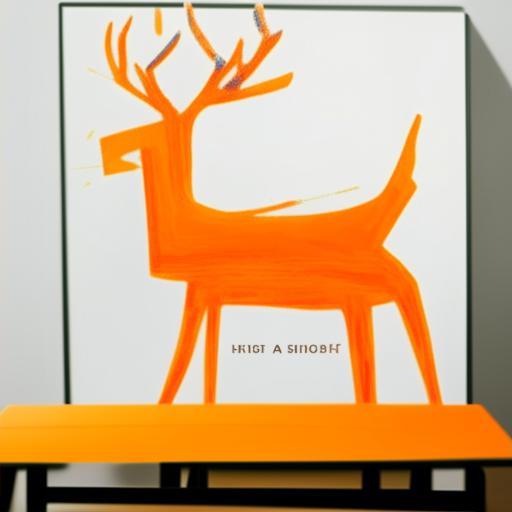} \\
\end{tabular}
    }
    \caption{Compositions of texturing and instruct operators obtained by our pOps method.}
    \label{fig:texturing_instruct_compositions}
\end{figure*}

\end{document}